%% file: main.tex
\documentclass[10pt,twocolumn,letterpaper]{article}

\usepackage[pagenumbers]{cvpr}   
\graphicspath{{figures/}}

\usepackage[T1]{fontenc}
\usepackage{amsthm}
\usepackage{placeins}
\usepackage{float}
\usepackage{array}
\usepackage{wrapfig}
\usepackage{adjustbox}
\usepackage{tikz}
\usetikzlibrary{arrows.meta}
\usepackage{algorithm}
\usepackage{algpseudocode}
\usepackage{pgfplots}
\usepackage{mathtools}
\pgfplotsset{compat=1.18}

\definecolor{cvprblue}{rgb}{0.21,0.49,0.74}
\definecolor{latencyorange}{RGB}{210,112,47}
\newcommand{\cubcmp}[3]{%
  \begingroup
  \pgfmathtruncatemacro{\cubA}{min(62,max(0,round(0.62*(#1))))}%
  \pgfmathtruncatemacro{\cubB}{min(62,max(0,round(0.62*(#2))))}%
  \begin{tikzpicture}[baseline=(cubcell.base)]%
    \node[rectangle,shade,left color=white!\cubA!cvprblue,right color=white!\cubB!cvprblue,inner xsep=1.2pt,inner ysep=0.8pt] (cubcell) {\shortstack{$#1\to#2$\\{\scriptsize\ifdim #2pt>#1pt \color{green!45!black}\else\ifdim #2pt<#1pt \color{red!75!black}\else\color{black}\fi\fi #3}}};%
  \end{tikzpicture}%
  \endgroup
}
\newcommand{\cubmain}[2]{%
  \pgfmathsetmacro{\cubdiff}{(#2)-(#1)}%
  \pgfmathtruncatemacro{\cubbaseint}{round(#1)}%
  \pgfmathtruncatemacro{\cubshade}{min(38,round(38*abs(\cubdiff)/15.5))}%
  \ifdim #2pt>#1pt
    \def\cubchange{+\pgfmathprintnumber[fixed,precision=1,zerofill]{\cubdiff}}%
  \else
    \def\cubchange{\pgfmathprintnumber[fixed,precision=1,zerofill]{\cubdiff}}%
  \fi
  \ifdim #2pt>#1pt
    \colorbox{green!\cubshade!white}{\strut\color{black}\cubbaseint\,\cubchange}%
  \else\ifdim #2pt<#1pt
    \colorbox{red!\cubshade!white}{\strut\color{black}\cubbaseint\,\cubchange}%
  \else
    \colorbox{white}{\strut\color{black}\cubbaseint\,0}%
  \fi\fi
}
\newcommand{\cubcmpfull}[3]{%
  \pgfmathtruncatemacro{\cubfullshade}{min(38,max(0,round(38*abs(#3)/15.5)))}%
  \ifdim #3pt>0pt
    \colorbox{green!\cubfullshade!white}{\shortstack{\color{black}$#1\to#2$\\{\scriptsize\color{black}#3}}}%
  \else\ifdim #3pt<0pt
    \colorbox{red!\cubfullshade!white}{\shortstack{\color{black}$#1\to#2$\\{\scriptsize\color{black}#3}}}%
  \else
    \colorbox{white}{\shortstack{\color{black}$#1\to#2$\\{\scriptsize\color{black}0.0}}}%
  \fi\fi
}
\usepackage{xcolor}
\usepackage[normalem]{ulem}
\usepackage{environ}
\usepackage{soul}
\usepackage[pagebackref,breaklinks,colorlinks,allcolors=cvprblue]{hyperref}
\input{appendix_refs}
\ifdefined\linenumbersep\fi

\def\confName{CVPR}
\def\confYear{2027}

\newcommand{\colheads}[3]{\noindent\makebox[#1][c]{\setlength{\tabcolsep}{0pt}\scriptsize\begin{tabular}{@{}*{#2}{>{\centering\arraybackslash}p{\dimexpr#1/#2\relax}}@{}}#3\end{tabular}}\par\vspace{1pt}}

\newtheorem{proposition}{Proposition}

\newcommand{\Nref}{N_{\mathrm{ref}}}
\newcommand{\dsh}{d}
\newcommand{\our}{SeeFine\xspace}
\newcommand{\vl}{\our}
\DeclareRobustCommand{\SF}[1]{\textup{SeeFine@}\allowbreak\textup{#1}}

\title{Vision Transformers Through the Looking Glass:\\
       Seeing at Pixel Resolution with Frozen Weights}

\author{
Piotr Kubaty$^{1,2}$\quad
Hanna Blazhko$^{1}$\quad
Adam Wr\'{o}bel$^{1,2}$\\
\L{}ukasz Struski$^{1}$\quad
Marek \'{S}mieja$^{1}$\quad
Jacek Tabor$^{1,3}$\\[0.5em]
{\normalsize $^{1}$Faculty of Mathematics and Computer Science, Jagiellonian University, Krak\'{o}w, Poland}\\
{\normalsize $^{2}$Doctoral School of Exact and Natural Sciences, Jagiellonian University, Krak\'{o}w, Poland}\\
{\normalsize $^{3}$Centre for Credible Artificial Intelligence, Warsaw University of Technology, Poland}
}

\begin{document}
\twocolumn[{%
\renewcommand\twocolumn[1][]{#1}%
\maketitle
\begin{center}
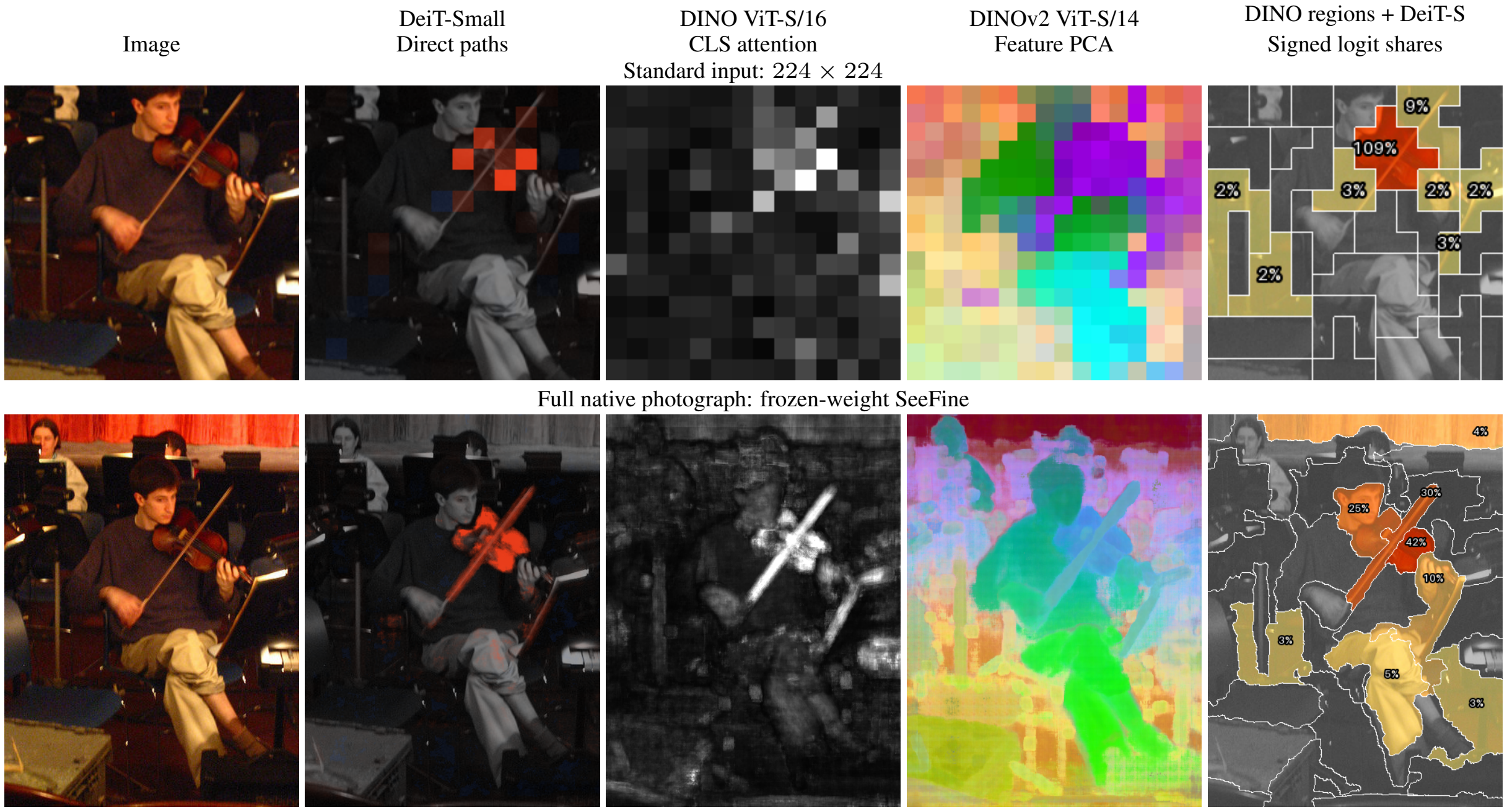

\captionsetup{type=figure}
\input{cvpr5_revision/hero_figure.tex}
\captionof{figure}{
\textbf{Native-resolution visual analysis with SeeFine.}
We compare representations extracted from standard ViT resolution ($224\times 224$) inputs (top) with those obtained using SeeFine at native image resolution (bottom).
Visualizations include direct paths, [CLS] attention maps, patch-feature PCA, and signed logit contributions across different ViTs.
Using frozen pretrained models, SeeFine enables efficient native-resolution processing, showing fine-grained spatial structures and more precise localization of visual features.
See more examples in \cref{sec:vizattn}.
}
\label{fig:hero}
\end{center}%
}]

\begin{abstract}
Vision Transformers see an image through a coarse grid of patches, so much of the fine structure their weights have learned stays hidden in their attention, features and attributions. We introduce SeeFine, a training-free way to read a frozen ViT at pixel resolution by changing only how the network is executed. It applies the unchanged patch embedding at every pixel and connects the resulting shifted copies of the patch grid through one shared class token, whose share of attention is preserved by an analytic correction. The result remains a single model with one prediction and a complete gradient graph, and needs no training data, learned upsampler or per-image optimization. SeeFine applies to supervised and self-supervised ViTs alike. With their original heads, six ImageNet classifiers gain up to $1.13$ top-1 points, whereas simply reducing the patch stride degrades most of them. Beyond classification, the same frozen checkpoints consistently improve text-guided segmentation, semantic correspondence and landmark matching in 3D CT, compare favourably with training-free alternatives, and on enlarged inputs rival an upsampler trained for dense features. Looking closer does not require retraining: the detail is already in the checkpoint at hand.
Our code is available at: \url{https://github.com/gmum/seefine}.
\end{abstract}

\begin{figure*}[t]
\centering
\input{cvpr5_revision/tikz_n2_method.tex}
\caption{
\textbf{From classical ViTs to SeeFine.}
SeeFine enables pixel-level processing with frozen pretrained ViTs by applying the original patch embedding kernel with stride one, producing one token per pixel.
To enable efficient attention computation, each token attends only to tokens within its \emph{phase} $G_{ab}$ (see~\cref{eq:phase}) and to a shared class token, which aggregates information across all phases.
All pretrained weights remain unchanged.
}
\label{fig:method}
\end{figure*}
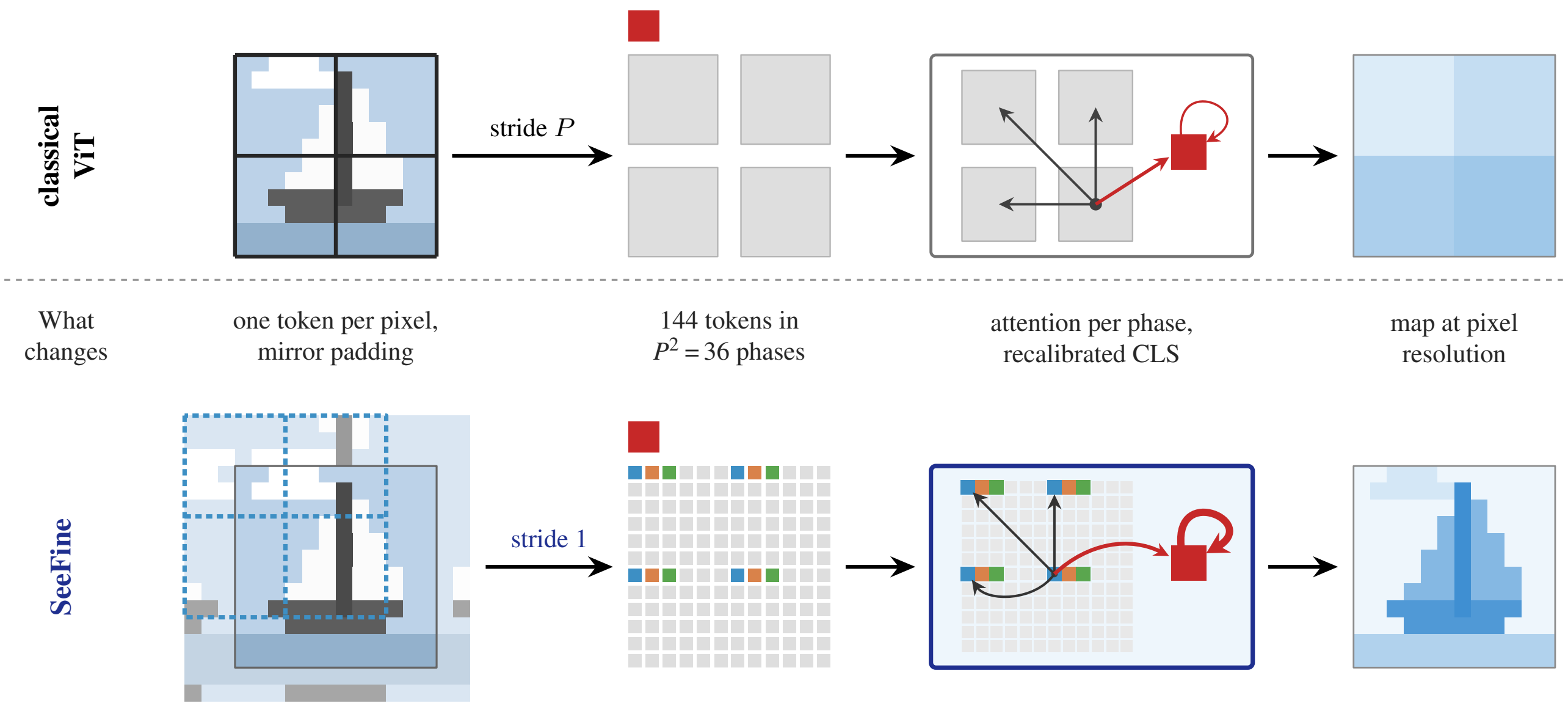

\section{Introduction}\label{sec:intro}

Pretrained Vision Transformers~\cite{dosovitskiy2021vit}, such as DINO, DINOv2 and DINOv3~\cite{caron2021dino,oquab2023dinov2,simeoni2025dinov3}, are general-purpose frozen backbones for classification, segmentation and correspondence, and their attention maps and attributions are our main window into what they have learned. All of these views live on the patch grid: a $224\times224$ image with $16\times16$ patches becomes only $14\times14$ tokens. Recognizing a violinist is one thing but seeing how a model responds to the bow, fingers and instrument is another: in \cref{fig:hero} (top), much of that structure disappears into blocks. The pretrained checkpoint has seen the bow, and its patch grid simply does not let us look closely enough.

We ask how much finer a view the checkpoint at hand can give using only its architecture and weights, with no original training data, additional learned component or per-image optimization. The finer model should also remain a single predictor, with one class-token prediction and one gradient graph, so that the supplied classifier keeps working and gradient-based explanations such as Grad-CAM or gradient-weighted rollout have one target~\cite{selvaraju2020gradcam,abnar2020quantifying,chefer2021transformer,wrobel2026dave}. Existing approaches either learn new components, such as FeatUp, JAFAR, AnyUp, ViT-Up and SPAR~\cite{fu2024featup,couairon2025jafar,wimmer2026anyup,wandel2026vitupfaithfulfeatureupsampling,kombol2026spar}, or run the frozen network more densely by reducing the patch stride~\cite{amir2022deep}, averaging shifted and flipped inputs~\cite{docherty2024upsampling} or fitting a dense map to each image~\cite{seo2026upsample}. None of them turns the dense samples into a single predictor: a smaller stride lowers classification accuracy on most models we test (\cref{tab:donors}), and averaged feature maps need an extra rule to define a prediction. Related work is discussed in \cref{sec:related_main}.

We propose SeeFine, which keeps every learned tensor and changes only how the network is executed (\cref{sec:method}). Applying the unchanged patch kernel at stride one gives one token per pixel~\cite{amir2022deep}. These tokens split into interleaved phases (\cref{fig:method}), each a shifted copy of the original patch grid; attending only within its own phase, a token sees a grid like the one the network was trained on, and the quadratic attention cost drops by $P^2$ for patch size $P$. One recurrent class token reads all phases and links them into a single model, and an analytic logit shift, following the proportional-attention principle of ToMe~\cite{bolya2023token}, keeps its share of attention despite the many more keys. \Cref{fig:hero} (bottom) and \cref{fig:ship_intro} show the result: the same frozen backbones resolve the bow and hands, and individual attention heads pick out eyes, fur and objects at native resolution. We apply SeeFine to supervised ViT and DeiT classifiers and to self-supervised DINO, DINOv2 and DINOv3 encoders with rotary positions and registers~\cite{darcet2024vision}, always keeping their existing heads. Six ImageNet classifiers gain up to $1.13$ top-1 points, whereas stride reduction alone falls below the original on most of them (\cref{tab:donors}). The frozen Talk2DINOv3 text head segments VOC2012 more accurately than with its released readout or the training-free alternatives we test (\cref{tab:talk2dinov3_native}). In semantic correspondence, SeeFine improves every backbone we test, and frozen DINOv3-B remains competitive with ViT-Up, an upsampler trained for dense features (\cref{tab:spair_main,tab:vitup_main}). The construction also extends to frozen 3D CT encoders, where it markedly reduces landmark-matching error (\cref{tab:ct3d_main}).

\noindent
Our contributions are:
\begin{itemize}\setlength{\itemsep}{2pt}\setlength{\parskip}{0pt}
\item \textbf{Native-resolution ViT processing.}
We introduce SeeFine, a training-free method that gives pretrained ViTs pixel-level representations by decomposing dense patch embeddings into interleaved phases, without modifying any weight.
\item \textbf{Efficient phase-restricted attention.}
Attention restricted to each phase and linked by a shared class token enables efficient single-pass processing, with analytic logit calibration preserving the nominal class-token attention share.
\item \textbf{Evaluation across downstream tasks.}
We evaluate quality and efficiency against existing dense-feature methods on classification, text-guided segmentation and 2D/3D correspondence.
\end{itemize}

\section{\vl}\label{sec:method}
\input{cvpr5_revision/method_story.tex}

\section{Experiments}\label{sec:experiments}\label{sec:viz}\label{sec:numbers}

We first evaluate existing classification and text-guided segmentation heads. Semantic correspondence then tests the spatial features directly, with controls for descriptor choice and the quality--cost trade-off with trained upsamplers. Ablations distinguish dense sampling from global integration, before a final transfer test in 3D CT.

\begin{table}[t]
\caption{\textbf{Frozen classifiers benefit most from the complete dense execution.} ImageNet-1k top-1 (\%) on the full validation set, with the original checkpoint, head and $224\times224$ crop. Top: donor and training-free readouts; middle: SeeFine at stride one with full or phase attention (attn.), with or without calibration (calib.). DINO: 20-NN on CLS features. III-S, III-B: DeiT-III Small and Base. The dash marks DeiT-III checkpoints that the released DVF extractor does not support because of LayerScale (\cref{app:dvf_scope}). Nearest and Bilinear leave CLS unchanged and equal the donor; HR-Dv2 returns no CLS. SeeFine is the best readout for every model. Details: \cref{sec:acc}.}
\label{tab:donors}
\centering\scriptsize\setlength{\tabcolsep}{2.5pt}
\begin{tabular}{@{}lrrrrrrr@{}}
\toprule
Readout & DeiT-T & DeiT-S & DeiT-B & ViT-S & III-S & III-B & DINO\\
\midrule
Donor & 72.20 & 79.87 & 81.97 & 81.42 & 81.36 & 83.79 & 66.41\\
Strided & 72.11 & 79.05 & 82.02 & 82.06 & 80.05 & 82.92 & 65.71\\
DVF & 70.75 & 76.66 & 81.76 & 81.34 & --- & --- & 64.86\\
\midrule
Full attn., no calib. & 69.29 & 70.03 & 80.97 & 78.63 & 73.18 & 79.77 & 61.87\\
Full attn., calib. & 71.57 & 79.97 & 82.15 & 81.72 & 81.59 & 83.80 & 65.90\\
Phase attn., no calib. & 70.97 & 77.27 & 81.79 & 81.57 & 79.78 & 82.59 & 62.31\\
\midrule
\SF{M} & \textbf{73.33} & \textbf{80.60} & \textbf{82.32} & \textbf{82.31} & \textbf{82.24} & \textbf{84.13} & \textbf{66.49}\\
\bottomrule
\end{tabular}
\end{table}

\begin{figure*}[t]
\centering
\includegraphics[width=\textwidth]{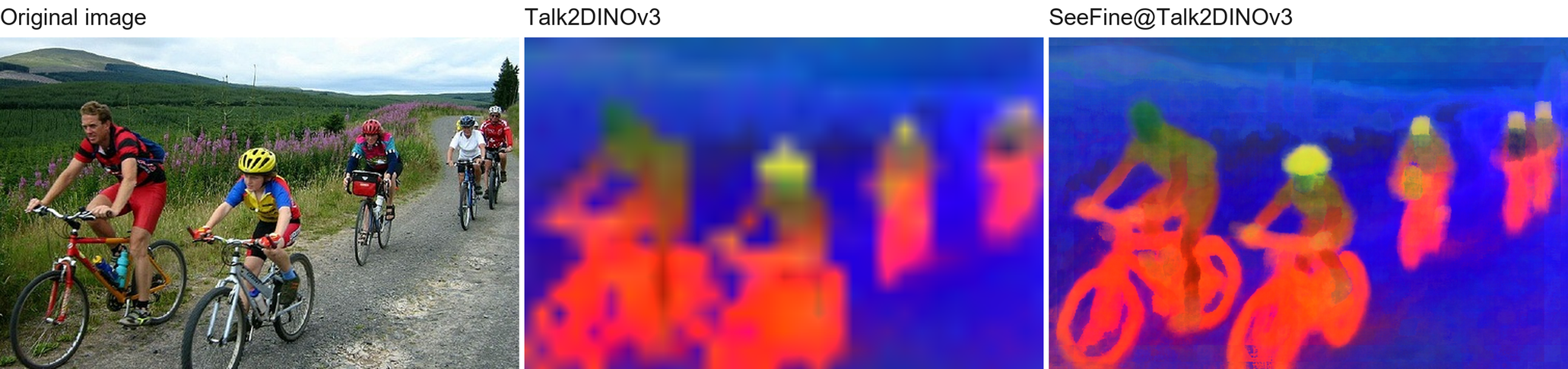}
\caption{\textbf{Three text queries come into focus in one RGB view.} Each colour channel shows similarity to one phrase: red ``bicycle'', green ``helmet'', blue ``road''; both maps share one colour scale. Compared with the $19\times31$ patch grid of Talk2DINOv3, SeeFine traces the wheels and frame and localizes the helmet. The maps show raw similarity scores.}
\label{fig:talk2dinov3_main}
\end{figure*}

\paragraph{Experimental setup.}
All backbones are released checkpoints and stay frozen: supervised ViT, DeiT and DeiT-III classifiers, self-supervised DINO, DINOv2 and DINOv3 encoders, and the TAP-CT 3D encoder. The donor is the ordinary execution of the same checkpoint on the same input. Unless labelled otherwise, SeeFine uses stride-one embedding, RGB reflection for 2D images and the analytic logit shift, with recurrent global tokens shared across the dense phases. The recorded-context control instead supplies global states from an ordinary donor pass.

Nearest and Bilinear interpolate donor features. Strided uses overlapping patch embeddings at stride 6 ($P=14$) or 7 ($P=16$) with full attention; Deep ViT Features (DVF)~\cite{amir2022deep} use stride 4; HR-Dv2~\cite{docherty2024upsampling} combines 68 shifted and flipped passes. These baselines keep the weights fixed. We use the released DVF extractor on the backbones it supports. On SPair-71k, the Strided implementation at the same stride differs from DVF by at most $0.20$ PCK@0.05 points across the three supported backbones. For DINOv2 and DINOv3, we use Strided to test dense execution with overlapping patches and full attention (\cref{app:dvf_scope}). The 2D studies use complete validation or test splits, with identical $224\times224$ crops for ImageNet and CUB and uncropped native images for segmentation and correspondence unless stated. Reported 95\% intervals use paired bootstrap resampling; each study specifies its averaging and resampling units. Runtime tables identify the GPU and measurement scope. Full protocols and supplementary evaluations are indexed in \cref{tab:appendix_protocol_index_v4}.

\paragraph{Classification with the original head.}
Classification tests whether the original head remains useful when the frozen model is evaluated densely. On the full ImageNet validation set, SeeFine improves all six supervised classifiers while keeping the checkpoint, head and input crop fixed (\cref{tab:donors}). Gains reach $1.13$ top-1 points, and SeeFine is the best evaluated readout for each model. Stride reduction alone is less reliable: Strided lowers accuracy for most classifiers, and DVF falls below the donor wherever supported. The separate DINO nearest-neighbour evaluation shows only a marginal gain and uses no supervised classification head.

The component ablation shows why denser sampling alone is not enough for classification. Without calibration, stride-one full attention substantially degrades accuracy. Calibration recovers much of this loss; combining it with phase attention gives the best results across the supervised models. In the single-image DINO benchmark, phase attention is also roughly $14\times$ faster than full attention (\cref{tab:phase_cost}).

\begin{figure*}[t]
\centering
\includegraphics[width=\textwidth]{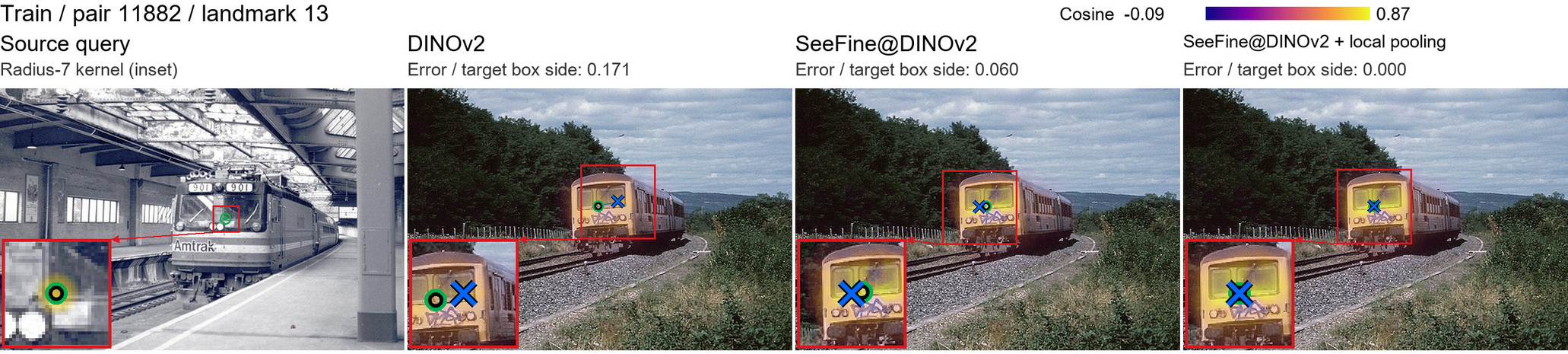}
\caption{\textbf{Finer features bring the match onto the right point.} A landmark on the source train (left) is searched for in another train photograph by cosine similarity of frozen DINOv2 features. Overlays show the similarity map near the target (bright: high), enlarged in the insets; green circle: annotated location, blue cross: best match. The coarse donor map misses the annotated point; SeeFine moves the prediction closer, and local pooling further improves this example. Over all SPair-71k test points, SeeFine is closer than DINOv2 in $53.4\%$ of cases (\cref{tab:spair_winrates}).}
\label{fig:spair_bus_main}
\end{figure*}

\begin{table}[t]\centering
\caption{\textbf{Better text-guided regions with the supplied frozen model.} VOC2012 pooled mIoU (\%) on the full validation set at native resolution, with the same frozen Talk2DINOv3 checkpoint, text projector and mask readout. Released/aligned: original/geometrically corrected donor upsampling. $\Delta$: row minus SeeFine; 95\% CI: paired image-bootstrap interval of $\Delta$. Time: median visual-encoding time per image on one GH200.  HR-Dv2 lacks the CLS attention required by the background cleaner. Details: \cref{app:talk2dinov3}.}
\label{tab:talk2dinov3_native}\footnotesize\setlength{\tabcolsep}{4pt}
\begin{tabular}{@{}lrrcr@{}}\toprule
Readout & mIoU (\%) & $\Delta$ & 95\% CI & Time (s)\\\midrule
Donor, released & 61.19 & $-5.37$ & $[-5.67,-5.11]$ & 0.021\\
Donor, aligned & 63.67 & $-2.89$ & $[-3.12,-2.71]$ & 0.021\\
Nearest & 63.88 & $-2.68$ & $[-2.87,-2.54]$ & 0.018\\
Bilinear & 63.43 & $-3.13$ & $[-3.37,-2.92]$ & 0.018\\
Strided & 65.46 & $-1.10$ & $[-1.35,-0.85]$ & 0.018\\
\midrule
SeeFine & \textbf{66.56} & --- & --- & 1.253\\
\bottomrule\end{tabular}
\end{table}

\paragraph{Text-guided segmentation.}
We next test an existing head that reads spatial features. Talk2DINO~\cite{giambellutti2025talk2dino} maps text into a self-supervised ViT's feature space, allowing each image location to be scored against a phrase. We use its released Talk2DINOv3 ViT-L checkpoint with the visual encoder, text encoder and projector frozen. Prompts, background cleaner and threshold are fixed on 100 training images and shared by every readout; no segmentation head is trained. On full native VOC2012, SeeFine gains $2.89$ mIoU points over the aligned donor and $1.10$ over Strided (\cref{tab:talk2dinov3_native}). Alignment correction accounts for part of the $5.37$-point gap to the released donor readout. The gain over that readout persists without the cleaner, and recorded donor context performs almost identically to recurrent SeeFine, so the gain does not require dense-to-global feedback (\cref{app:talk2dinov3}).

In the cyclist example in \Cref{fig:talk2dinov3_main}, the donor’s similarity maps are coarse. SeeFine@Talk2DINOv3 shows the bicycle wheels and frame more clearly and produces a more localized response to the helmet query. Both sets of maps show raw similarity scores on a shared color scale.

\begin{table}[t]\centering
\caption{\textbf{SPair-71k across frozen backbones.} Pair-mean PCK@0.05 / PCK@0.10 (\%) on all native-resolution test pairs, using final-LayerNorm last-block features. DINOv3 uses recorded donor context. Bilinear equals the donor.   SeeFine leads on DINOv2 and DINOv3; DVF and Strided are slightly higher on DINO, and with the DVF descriptor for every readout SeeFine leads (\cref{tab:descriptor_main}). The DINOv2 PCK@0.05 difference from Strided is not statistically significant. More backbones and pooling: \cref{tab:base_spair}.}\label{tab:spair_main}
\footnotesize\setlength{\tabcolsep}{4pt}
\begin{tabular}{@{}lccc@{}}\toprule
 & DINO & DINOv2 & DINOv3\\
Method & ViT-S/16 & ViT-S/14 & ViT-L/16\\\midrule
Donor & 15.74 / 28.34 & 31.89 / 49.03 & 32.92 / 51.44\\
Nearest & 11.90 / 25.53 & 25.52 / 45.89 & 24.94 / 48.10\\
Strided & 18.90 / \textbf{30.75} & 34.96 / 48.38 & 40.10 / 55.51\\
HR-Dv2 & 17.13 / 28.37 & 31.33 / 44.47 & 38.12 / 54.45\\
DVF & \textbf{18.92} / 30.61 & --- & ---\\
\midrule
SeeFine & 18.61 / 30.32 & \textbf{35.13} / \textbf{50.14} & \textbf{41.37} / \textbf{57.45}\\
\bottomrule\end{tabular}
\end{table}

\begin{table}[t]\centering\footnotesize
\caption{\textbf{The gain persists with a shared descriptor.} SPair-71k pair-mean PCK@0.05 (\%), full test split. All methods use block-9 keys with logarithmic binning on the three backbones supported by DVF. SeeFine uses bins at the same stride-4 offsets. Full descriptor controls and paired intervals: \cref{app:abl_descriptor}.}\label{tab:descriptor_main}
\setlength{\tabcolsep}{3.5pt}
\begin{tabular}{@{}lrrr@{}}\toprule
Method & DINO ViT-S/16 & ViT-S/16 & DeiT-Small\\\midrule
Donor & 13.37 & 14.62 & 11.18\\
Strided & 20.06 & 21.63 & 16.94\\
DVF & 21.50 & 22.29 & 17.72\\
SeeFine & \textbf{22.26} & \textbf{23.81} & \textbf{20.84}\\
\bottomrule\end{tabular}
\end{table}

\paragraph{Semantic correspondence.}
Semantic correspondence tests spatial features directly by matching the same part across different object instances, without a prediction head. On the full SPair-71k test split, SeeFine improves on the donor for all five tested backbones. \Cref{tab:spair_main} shows the DINO family: Strided is comparable on DINOv2, slightly ahead on DINO and behind on DINOv3. SeeFine also exceeds Strided on the supervised ViT-S and DeiT-S backbones. With the same last-block features, DVF slightly outperforms SeeFine on DINO but scores lower on ViT-S and DeiT-S (\cref{tab:base_spair}). \Cref{fig:spair_bus_main} illustrates improved localization, with a further gain from local pooling.

\paragraph{Controlling the descriptor.}
To isolate the effect of the readout, we compare methods using the same feature representation. On DINO, ViT-S and DeiT-S, block-9 keys with logarithmic binning improve both SeeFine and DVF over their respective last-block results. With this descriptor shared across methods, SeeFine achieves the highest PCK@0.05 on all three backbones (\cref{tab:descriptor_main}). Last-block features remain more effective on DINOv2 and DINOv3 (\cref{app:abl_descriptor}).

\begin{table}[t]\centering
\caption{\textbf{SeeFine and trained upsamplers at three input sizes.} SPair-71k category-macro PCK (\%) with DINOv3-B on all test pairs. ViT-Up adds LoRA and a trained decoder; AnyUp adds a trained upsampler; the other methods keep the backbone frozen. Each block shares an RGB input; the $448\times448$ block follows the ViT-Up protocol. Time: median feature-extraction latency; memory: peak allocated GPU memory; both per image on one GH200 over 24 fixed images. Bilinear equals DINOv3-B; Strided uses stride 7.   $^\ddagger$Recorded donor context, stride 4; the other SeeFine rows are recurrent. All controls: \cref{tab:vitup_comparison}.}\label{tab:vitup_main}
\footnotesize\setlength{\tabcolsep}{3pt}
\begin{tabular}{@{}lrrrrr@{}}\toprule
Method & @0.10 & @0.05 & @0.01 & Time (s) & Mem.\ (GiB)\\\midrule
\multicolumn{6}{@{}l}{\emph{Input $448\times448$}}\\
DINOv3-B & 50.03 & 34.06 & 3.90 & 0.012 & 1.51\\
Nearest & 45.69 & 22.64 & 1.09 & 0.012 & 2.08\\
Strided & 51.90 & 37.96 & 6.55 & 0.013 & 2.09\\
HR-Dv2 & 52.45 & 36.26 & 5.17 & 1.903 & 1.92\\
ViT-Up & \textbf{57.97} & \textbf{41.84} & \textbf{7.93} & 0.320 & 1.34\\
SeeFine & 55.92 & 40.00 & 5.43 & 0.501 & 9.52\\
\midrule
\multicolumn{6}{@{}l}{\emph{Native input $W\times H$}}\\
DINOv3-B & 51.72 & 35.84 & 4.62 & 0.011 & 1.59\\
Nearest & 48.24 & 25.27 & 1.34 & 0.013 & 2.20\\
Strided & 52.05 & 38.68 & \textbf{7.46} & 0.014 & 2.21\\
HR-Dv2 & 52.86 & 36.79 & 5.46 & 1.575 & 2.04\\
AnyUp & 36.51 & 19.04 & 1.92 & 1.600 & 1.50\\
SeeFine & \textbf{56.95} & 41.48 & 6.37 & 0.439 & 10.41\\
SeeFine, stride 4$^\ddagger$ & 56.92 & \textbf{41.50} & 6.60 & 0.039 & 1.62\\
\midrule
\multicolumn{6}{@{}l}{\emph{Doubled native input $2W\times2H$}}\\
DINOv3-B & 54.78 & 41.09 & 8.41 & 0.019 & 5.29\\
Nearest & 54.92 & 37.85 & 4.13 & 0.020 & 7.74\\
Strided & 57.08 & 41.95 & 9.91 & 0.045 & 7.77\\
HR-Dv2 & 49.31 & 33.30 & 6.02 & 14.912 & 7.57\\
SeeFine & \textbf{58.04} & \textbf{44.10} & \textbf{10.60} & 2.248 & 40.00\\

\bottomrule\end{tabular}
\end{table}

\input{cvpr5_revision/components_main.tex}

\begin{table}[t]\centering
\caption{\textbf{Training-free correspondence with a frozen 3D CT encoder.} Frozen TAP-CT-B-3D features; SeeFine uses recorded donor context. Within patient: median landmark error (mm) between inhale and exhale scans; across patients: points matched to the correct organ (\%). Strided: overlapping $2\times4\times4$ embedding with full attention; MIND-SSC: handcrafted descriptor; DVF and HR-Dv2 have 2D-only code. ---: not evaluated. SeeFine is the most accurate encoder readout in all four settings; MIND-SSC is more precise within a patient but has lower correct-organ accuracy across patients. Baseline definitions: \cref{app:ct3d}; details: \cref{tab:ct3d}.}\label{tab:ct3d_main}%
\footnotesize\setlength{\tabcolsep}{2.5pt}
\begin{tabular}{@{}lrrrr@{}}\toprule
 & \multicolumn{3}{c}{Within patient (mm $\downarrow$)} & Across\\\cmidrule(lr){2-4}
Readout & Lung250M & 4DCT & COPD & organ (\%) $\uparrow$\\\midrule
Same position (no matching) & 12.6 & 7.0 & 22.3 & 25.4\\
Nearest & 19.7 & 12.3 & 28.7 & 39.8\\
Trilinear upsampling & 10.1 & 7.7 & 15.0 & 50.9\\
Trilinear, $512^2$ input & 10.3 & 9.7 & 16.8 & ---\\
3D joint bilateral upsampling & 9.9 & 9.0 & 12.7 & ---\\
AnyUp, per slice & 11.3 & 12.0 & 15.2 & ---\\
Strided & 7.3 & 5.9 & 11.6 & 58.1\\
\SF{TAP-CT}, stride 1 & \textbf{5.5} & \textbf{3.0} & \textbf{10.3} & \textbf{59.4}\\\midrule
MIND-SSC (handcrafted) & 2.0 & 1.2 & 1.8 & 14.4\\
\bottomrule\end{tabular}
\end{table}

\paragraph{Trained upsamplers and the quality--cost trade-off.}
We compare DINOv3-B readouts with the released ViT-Up and AnyUp models on the same SPair split (\cref{tab:vitup_main}). This comparison follows ViT-Up's category-macro averaging, whereas the cross-backbone table reports pair means. ViT-Up~\cite{wandel2026vitupfaithfulfeatureupsampling} learns LoRA updates and a decoder; AnyUp~\cite{wimmer2026anyup} supplies an upsampler trained once for multiple backbones. At the shared $448\times448$ input, ViT-Up is more accurate and uses less time and memory. On native images, SeeFine outperforms both its donor and AnyUp under the tested protocol. Strided comes closest among the training-free baselines ($38.68$ against $41.48$ PCK@0.05) and scores higher at PCK@0.01 ($7.46$ versus $6.37$). At doubled-native input, SeeFine gains $3.00$ PCK@0.05 points over the donor on the same input, reaching $44.10$. This exceeds ViT-Up's $41.84$ at its standard $448\times448$ setting, at substantially higher time and memory (\cref{tab:vitup_main}); ViT-Up leads at the shared $448\times448$ input.

Within the recorded-context variant, stride 4 retains native-input PCK@0.05 ($41.50$ versus $41.52$) while reducing extraction time from $0.370$ to $0.039$\,s and peak memory from $10.42$ to $1.62$\,GiB (\cref{tab:current_operating_cost}).

\paragraph{Understanding the gains.}
We separate dense sampling from global integration using DINO ViT-S/16 on CUB landmark retrieval and SPair correspondence (\cref{tab:components_main}). Independent phases and shared CLS give nearly identical calibrated local-task scores, both above the donor, indicating that dense sampling accounts for most of the gain. Calibration helps the supervised classifiers and CUB but lowers SPair PCK in this DINO ablation. Sharing the CLS provides an internal prediction target and a coupled backward graph; these experiments establish neither an accuracy advantage from sharing nor more faithful explanations.

\begin{figure}[t]
\centering
\def\ctqualw{\columnwidth}%
\includegraphics[trim=7.15bp 7.15bp 7.15bp 7.15bp,clip,width=\ctqualw]{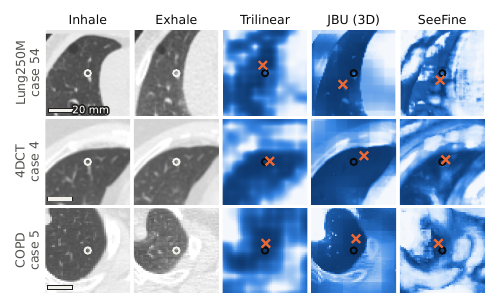}\caption{\textbf{Finer similarity maps in 3D CT.} Frozen TAP-CT-B-3D; one inhale landmark per lung dataset (fixed seed), matched against the exhale scan. Maps show feature similarity on the exhale slice through the true match; ring: true location, cross: most similar voxel  in this slice; $80\times80$\,mm crops, each with its own colour scale. Trilinear upsampling stays blocky and 3D joint bilateral upsampling (JBU) spreads similarity along tissue boundaries, whereas SeeFine's maps follow the vessels. For all readouts, including AnyUp and the handcrafted MIND-SSC see Appendix, \cref{fig:ct3d_qual_full}.}
\label{fig:ct3d_qual}
\end{figure}

\paragraph{Correspondence in 3D CT.} Finally, we test whether the same dense-readout principle transfers to volumetric encoders. We read frozen TAP-CT-B-3D at stride one along all three axes using recorded donor context, and match points by nearest-neighbour feature search without registration or training.  The datasets, baselines and full results are detailed in \cref{app:ct3d}. SeeFine reduces median landmark error relative to trilinear upsampling on all three lung datasets (e.g.\ $10.1\to5.5$\,mm on Lung250M), with lower mean error in every scan pair (\cref{tab:ct3d_main}). Larger inputs, 3D joint bilateral upsampling (JBU) and slice-wise AnyUp are less accurate. \Cref{fig:ct3d_qual} shows why: trilinear upsampling keeps the coarse token grid and JBU spreads similarity along tissue boundaries, whereas SeeFine's similarity maps follow the vessel structure. Across patients, correct-organ matching improves by $8.5$ percentage points over trilinear upsampling, with the largest gains for small or thin organs (\cref{fig:ct3d_organs}). SeeFine also outperforms Strided, the strongest other encoder readout, on all four metrics; in a paired rerun on identical query points (SeeFine $59.6\%$, Strided $58.1\%$), its correct-organ advantage is $1.5$ percentage points (95\% interval $[0.2,2.9]$). MIND-SSC remains more precise within patients but transfers poorly across patients. These results extend the benefit of dense frozen features to 3D anatomical correspondence.

\section{Related work}\label{sec:related_main}

\paragraph{Dense features from frozen ViTs.}
Learned upsamplers refine coarse ViT features with a trained module, as in FeatUp, LiFT, LoftUp, JAFAR, AnyUp and FeatSharp~\cite{fu2024featup,suri2024lift,huang2025loftup,couairon2025jafar,wimmer2026anyup,ranzinger2025featsharp}.
ViT-Up additionally adapts the backbone with LoRA and distils multi-scale features~\cite{wandel2026vitupfaithfulfeatureupsampling}, and SPAR distils sliding-window inference into the network~\cite{kombol2026spar}.
Training-free methods reduce the patch-embedding stride~\cite{amir2022deep}, average feature maps over shifted and flipped inputs~\cite{docherty2024upsampling} or optimize a dense map for each image~\cite{seo2026upsample}.
SeeFine belongs to the training-free family but integrates the dense samples inside the transformer, so the result keeps a single prediction and gradient graph.

\paragraph{Token count and attention.}
Changing the number of tokens changes attention statistics.
FlexiViT, ResFormer and NaViT address this by training for several patch sizes or resolutions~\cite{beyer2023flexivit,tian2023resformer,dehghani2023navit}, and Win-Win tunes the softmax temperature of high-resolution ViTs~\cite{leroy2024win}.
Token merging adds the log mass of merged tokens to their logits~\cite{bolya2023token}.
SeeFine applies the same principle in reverse to a class token whose relative mass would otherwise shrink, and it shifts only the class-key logits instead of rescaling all of them as a temperature does~\cite{chiang2022overcoming,nakanishi2025scalable}.
Sparse attention with a fixed spacing is common in trained architectures~\cite{child2019generating,tu2022maxvit,hassani2022dilated}, while SeeFine imposes it on frozen weights so that each phase matches the original patch grid.

\paragraph{Shifts and interpretability.}
Shift-and-stitch recovers dense outputs from shifted inputs~\cite{sermanet2014overfeat,long2015fully}, and polyphase sampling and anti-aliasing make networks more robust to small shifts~\cite{zhang2019making,chaman2021truly,ding2023reviving,rojasgomez2024making}.
SeeFine is related to these polyphase ideas but couples the phases through one class token.
Its dense readout complements explanation tools such as class-token attention~\cite{caron2021dino}, rollout~\cite{abnar2020quantifying}, relevance propagation~\cite{chefer2021transformer}, Grad-CAM~\cite{selvaraju2020gradcam} and DAVE~\cite{wrobel2026dave}, which are usually applied on the coarse patch grid.
An extended discussion is given in \cref{app:related}.

\section{Conclusion}\label{sec:conclusion}

A frozen checkpoint can offer a much finer view than its ordinary patch maps suggest. SeeFine opens that view through three simple changes to inference, giving the canonical model one recurrent CLS, one prediction and a complete gradient graph. The same idea exposes object details across backbones and, with existing heads, improves classification, text-guided segmentation, correspondence and 3D medical matching. 
SeeFine lets users choose the density of the feature grid at inference time to suit the spatial demands of the task.

\paragraph{Limitations.} First, SeeFine increases spatial resolution while keeping the checkpoint fixed. The resulting representations inherit the checkpoint's limitations. Second, tokens within half a patch of the image border also need context from outside the image, which any padding fills with synthetic content; we use RGB reflection, which improves border landmark recall over pixel replication (\cref{sec:padding-ablation}). Finally, compute and memory costs rise with resolution; in the recorded-context correspondence controls, strides 2 and 4 retain most of the gain at lower cost (\cref{app:native_controls_v4}). Full RGB-gradient inspection has a narrower tested range than forward visualization: FP32 forward and backward run up to $500\times319$ pixels, while at $2040\times1356$ only the BF16 forward fits in memory (\cref{app:gradient_cost_v4}). We offer SeeFine as a practical tool for exploring fine detail with the weights already available. Looking closer need not begin with another training run.

\section*{LLM Usage Statement}
OpenAI Codex and Anthropic Claude assisted with manuscript drafting and revision, code implementation and debugging, experiment planning, result analysis, and an internal pre-submission review. The authors are responsible for the final manuscript, software, experimental protocols and reported results.

{
    \small
    \bibliographystyle{ieeenat_fullname}
    \bibliography{references}
}

\clearpage

\appendix
\appendixrefs
\raggedbottom
\makeatletter
\setlength{\@dblfptop}{0pt}
\setlength{\@fptop}{0pt}
\makeatother
\input{cvpr5_revision/appendix_v5.tex}

\end{document}

%% file: appendix_refs.tex
\newcommand{\appxtag}{App.}
\newcommand{\appxtags}{Apps.}
\newcommand{\Appxtag}{Appendix}
\newcommand{\Appxtags}{Appendices}
\newcommand{\maintag}{Main}
\newcommand{\Maintag}{Main}
\newcommand{\appxsecstyle}{\arabic}
\newif\ifappxSnumbers
\appxSnumbersfalse

\newif\ifinappendix
\protected\def\toappx{\ifinappendix\else\appxtag~\fi}
\protected\def\Toappx{\ifinappendix\else\Appxtag~\fi}
\protected\def\tomain{\ifinappendix\maintag~\fi}
\protected\def\Tomain{\ifinappendix\Maintag~\fi}
\protected\def\appxsec{\ifinappendix Sec.\else\appxtag\fi}
\protected\def\appxsecs{\ifinappendix Secs.\else\appxtags\fi}
\protected\def\Appxsec{\ifinappendix Section\else\Appxtag\fi}
\protected\def\Appxsecs{\ifinappendix Sections\else\Appxtags\fi}

\makeatletter
\newcommand{\appendixrefs}{%
  \global\inappendixtrue
  \renewcommand{\thesection}{\appxsecstyle{section}}%
  \renewcommand{\thepage}{\texorpdfstring{\arabic{page}}{S\arabic{page}}}%
  \ifappxSnumbers\gdef\appx@numprefix{S}\else\global\let\appx@numprefix\@empty\fi
  \appx@renumber{figure}\appx@renumber{table}\appx@renumber{equation}\appx@renumber{algocf}%
  \appx@renumber{theorem}\appx@renumber{proposition}\appx@renumber{lemma}\appx@renumber{corollary}%
  \crefalias{section}{appendix}%
  \crefalias{subsection}{subappendix}%
  \crefalias{subsubsection}{subsubappendix}%
  \crefalias{figure}{appfigure}%
  \crefalias{table}{apptable}%
  \crefalias{equation}{appequation}%
  \crefalias{algocf}{appalgorithm}%
  \crefalias{algorithm}{appalgorithm}%
  \crefalias{theorem}{apptheorem}%
  \crefalias{proposition}{appproposition}%
  \crefalias{lemma}{applemma}%
  \crefalias{corollary}{appcorollary}}
\newcommand{\appx@renumber}[1]{%
  \@ifundefined{c@#1}{}{%
    \setcounter{#1}{0}%
    \@namedef{the#1}{\appx@numprefix\arabic{#1}}%
    \@namedef{theH#1}{S\arabic{#1}}}}

\newcommand{\appx@names}[6]{%
  \crefname{#1}{\tomain #3}{\tomain #4}\Crefname{#1}{\Tomain #5}{\Tomain #6}%
  \crefname{#2}{\toappx #3}{\toappx #4}\Crefname{#2}{\Toappx #5}{\Toappx #6}}
\AtEndPreamble{%
  \appx@names{section}{appendix}{Sec.}{Secs.}{Section}{Sections}%
  \appx@names{figure}{appfigure}{Fig.}{Figs.}{Figure}{Figures}%
  \appx@names{table}{apptable}{Tab.}{Tabs.}{Table}{Tables}%
  \appx@names{equation}{appequation}{Eq.}{Eqs.}{Equation}{Equations}%
  \appx@names{algorithm}{appalgorithm}{Alg.}{Algs.}{Algorithm}{Algorithms}%
  \appx@names{theorem}{apptheorem}{Thm.}{Thms.}{Theorem}{Theorems}%
  \appx@names{proposition}{appproposition}{Prop.}{Props.}{Proposition}{Propositions}%
  \appx@names{lemma}{applemma}{Lem.}{Lems.}{Lemma}{Lemmas}%
  \appx@names{corollary}{appcorollary}{Cor.}{Cors.}{Corollary}{Corollaries}%
  \crefname{appendix}{\appxsec}{\appxsecs}\Crefname{appendix}{\Appxsec}{\Appxsecs}%
  \creflabelformat{appequation}{\textup{(#2#1#3)}}%
}
\makeatother

%% file: cvpr5_revision/hero_figure.tex
{\scriptsize\setlength{\tabcolsep}{1pt}\begin{tabular}{@{}*{5}{>{\centering\arraybackslash}p{.18\textwidth}}@{}}
Image & \shortstack{DeiT-Small\\Direct paths} & \shortstack{DINO ViT-S/16\\CLS attention} & \shortstack{DINOv2 ViT-S/14\\Feature PCA} & \shortstack{DINO regions + DeiT-S\\Signed logit shares}\\
\multicolumn{5}{c}{Standard input: $224\times224$}\\
\includegraphics[width=.18\textwidth]{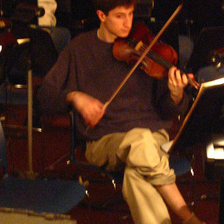}&\includegraphics[width=.18\textwidth]{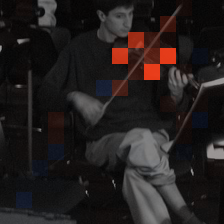}&\includegraphics[width=.18\textwidth]{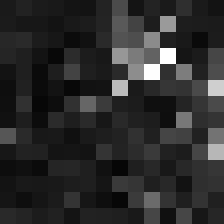}&\includegraphics[width=.18\textwidth]{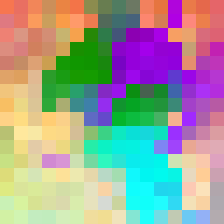}&\includegraphics[width=.18\textwidth]{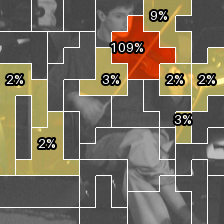}\\
\multicolumn{5}{c}{Full native photograph: frozen-weight SeeFine}\\
\includegraphics[width=.18\textwidth]{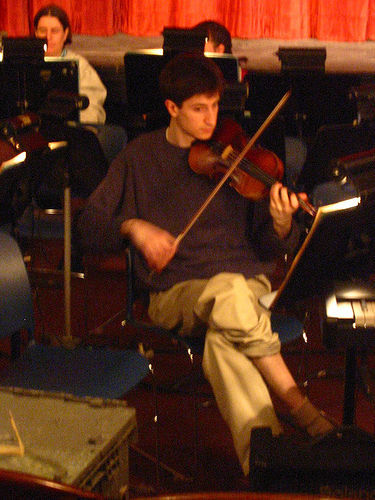}&\includegraphics[width=.18\textwidth]{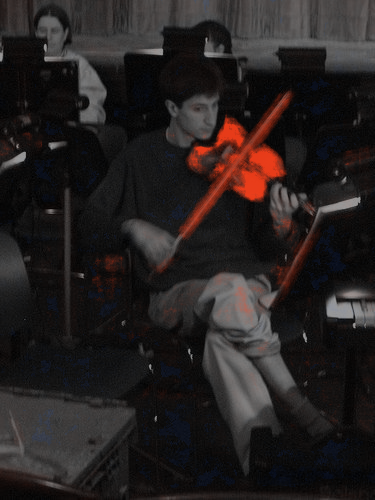}&\includegraphics[width=.18\textwidth]{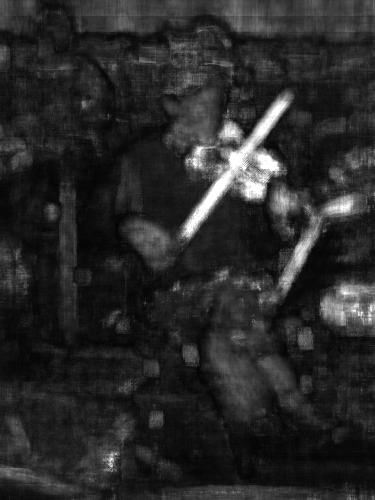}&\includegraphics[width=.18\textwidth]{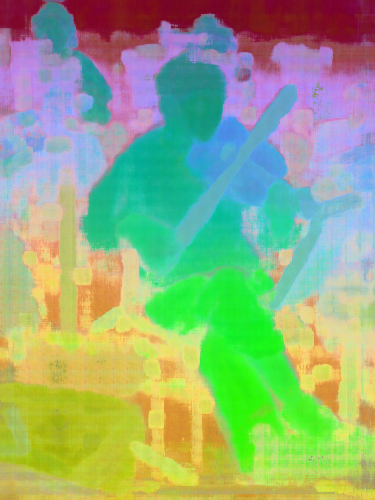}&\includegraphics[width=.18\textwidth]{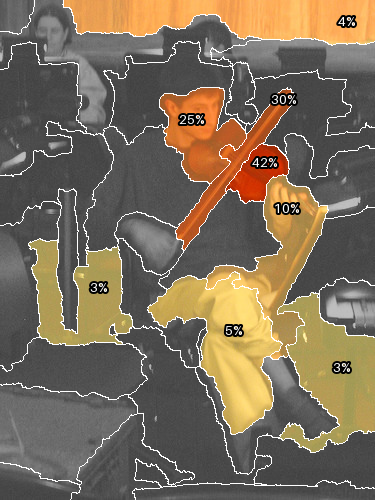}\\
\end{tabular}}

%% file: cvpr5_revision/tikz_n2_method.tex
\definecolor{nRed}{RGB}{198,40,40}%
\definecolor{nRedL}{RGB}{251,227,227}%
\definecolor{nNavy}{RGB}{31,47,143}%
\definecolor{nBlue}{RGB}{61,143,196}%
\definecolor{nBlueL}{RGB}{207,230,246}%
\definecolor{nOrange}{RGB}{217,130,74}%
\definecolor{nOrangeL}{RGB}{251,224,205}%
\definecolor{nGreen}{RGB}{90,166,78}%
\definecolor{nGreenL}{RGB}{223,240,219}%
\definecolor{nPink}{RGB}{199,107,181}%
\definecolor{nPinkL}{RGB}{243,220,238}%
\definecolor{nTok}{RGB}{222,222,222}%
\definecolor{nTokB}{RGB}{181,181,181}%
\definecolor{nSnow}{RGB}{91,127,166}%
\definecolor{nGrid}{RGB}{154,167,182}%
\begin{tikzpicture}[x=2.1mm,y=2.1mm,font=\footnotesize,line join=round]
\node[rotate=90,font=\small\bfseries,align=center,inner sep=0] at (2.2,0) {classical\\ViT};
\node[rotate=90,font=\small\bfseries,text=nNavy,inner sep=0] at (1.9,-21.2) {SeeFine};
\fill[fill={rgb,255:red,188;green,210;blue,232}] (10.9,5.2) rectangle (12.66,4.3);
\fill[fill={rgb,255:red,255;green,255;blue,255}] (12.63,5.2) rectangle (15.26,4.3);
\fill[fill={rgb,255:red,188;green,210;blue,232}] (15.23,5.2) rectangle (21.3,4.3);
\fill[fill={rgb,255:red,188;green,210;blue,232}] (10.9,4.33) rectangle (11.8,3.44);
\fill[fill={rgb,255:red,255;green,255;blue,255}] (11.77,4.33) rectangle (16.13,3.44);
\fill[fill={rgb,255:red,74;green,74;blue,74}] (16.1,4.33) rectangle (17,3.44);
\fill[fill={rgb,255:red,188;green,210;blue,232}] (16.97,4.33) rectangle (21.3,3.44);
\fill[fill={rgb,255:red,188;green,210;blue,232}] (10.9,3.47) rectangle (16.13,2.57);
\fill[fill={rgb,255:red,74;green,74;blue,74}] (16.1,3.47) rectangle (17,2.57);
\fill[fill={rgb,255:red,251;green,251;blue,251}] (16.97,3.47) rectangle (17.86,2.57);
\fill[fill={rgb,255:red,188;green,210;blue,232}] (17.83,3.47) rectangle (21.3,2.57);
\fill[fill={rgb,255:red,188;green,210;blue,232}] (10.9,2.6) rectangle (15.26,1.7);
\fill[fill={rgb,255:red,251;green,251;blue,251}] (15.23,2.6) rectangle (16.13,1.7);
\fill[fill={rgb,255:red,74;green,74;blue,74}] (16.1,2.6) rectangle (17,1.7);
\fill[fill={rgb,255:red,251;green,251;blue,251}] (16.97,2.6) rectangle (17.86,1.7);
\fill[fill={rgb,255:red,188;green,210;blue,232}] (17.83,2.6) rectangle (21.3,1.7);
\fill[fill={rgb,255:red,188;green,210;blue,232}] (10.9,1.73) rectangle (15.26,0.84);
\fill[fill={rgb,255:red,251;green,251;blue,251}] (15.23,1.73) rectangle (16.13,0.84);
\fill[fill={rgb,255:red,74;green,74;blue,74}] (16.1,1.73) rectangle (17,0.84);
\fill[fill={rgb,255:red,251;green,251;blue,251}] (16.97,1.73) rectangle (18.73,0.84);
\fill[fill={rgb,255:red,188;green,210;blue,232}] (18.7,1.73) rectangle (21.3,0.84);
\fill[fill={rgb,255:red,188;green,210;blue,232}] (10.9,0.87) rectangle (14.4,-0.03);
\fill[fill={rgb,255:red,251;green,251;blue,251}] (14.37,0.87) rectangle (16.13,-0.03);
\fill[fill={rgb,255:red,74;green,74;blue,74}] (16.1,0.87) rectangle (17,-0.03);
\fill[fill={rgb,255:red,251;green,251;blue,251}] (16.97,0.87) rectangle (18.73,-0.03);
\fill[fill={rgb,255:red,188;green,210;blue,232}] (18.7,0.87) rectangle (21.3,-0.03);
\fill[fill={rgb,255:red,188;green,210;blue,232}] (10.9,0) rectangle (14.4,-0.9);
\fill[fill={rgb,255:red,251;green,251;blue,251}] (14.37,0) rectangle (16.13,-0.9);
\fill[fill={rgb,255:red,74;green,74;blue,74}] (16.1,0) rectangle (17,-0.9);
\fill[fill={rgb,255:red,251;green,251;blue,251}] (16.97,0) rectangle (19.6,-0.9);
\fill[fill={rgb,255:red,188;green,210;blue,232}] (19.57,0) rectangle (21.3,-0.9);
\fill[fill={rgb,255:red,188;green,210;blue,232}] (10.9,-0.87) rectangle (13.53,-1.76);
\fill[fill={rgb,255:red,251;green,251;blue,251}] (13.5,-0.87) rectangle (16.13,-1.76);
\fill[fill={rgb,255:red,74;green,74;blue,74}] (16.1,-0.87) rectangle (17,-1.76);
\fill[fill={rgb,255:red,251;green,251;blue,251}] (16.97,-0.87) rectangle (19.6,-1.76);
\fill[fill={rgb,255:red,188;green,210;blue,232}] (19.57,-0.87) rectangle (21.3,-1.76);
\fill[fill={rgb,255:red,188;green,210;blue,232}] (10.9,-1.73) rectangle (12.66,-2.63);
\fill[fill={rgb,255:red,93;green,93;blue,93}] (12.63,-1.73) rectangle (16.13,-2.63);
\fill[fill={rgb,255:red,74;green,74;blue,74}] (16.1,-1.73) rectangle (17,-2.63);
\fill[fill={rgb,255:red,93;green,93;blue,93}] (16.97,-1.73) rectangle (19.6,-2.63);
\fill[fill={rgb,255:red,188;green,210;blue,232}] (19.57,-1.73) rectangle (21.3,-2.63);
\fill[fill={rgb,255:red,188;green,210;blue,232}] (10.9,-2.6) rectangle (13.53,-3.5);
\fill[fill={rgb,255:red,93;green,93;blue,93}] (13.5,-2.6) rectangle (18.73,-3.5);
\fill[fill={rgb,255:red,188;green,210;blue,232}] (18.7,-2.6) rectangle (21.3,-3.5);
\fill[fill={rgb,255:red,147;green,177;blue,204}] (10.9,-3.47) rectangle (21.3,-4.36);
\fill[fill={rgb,255:red,147;green,177;blue,204}] (10.9,-4.33) rectangle (21.3,-5.2);
\draw[line width=1.2pt,black!85] (10.9,5.2) -- (10.9,-5.2);
\draw[line width=1.2pt,black!85] (10.9,5.2) -- (21.3,5.2);
\draw[line width=1.2pt,black!85] (16.1,5.2) -- (16.1,-5.2);
\draw[line width=1.2pt,black!85] (10.9,0) -- (21.3,0);
\draw[line width=1.2pt,black!85] (21.3,5.2) -- (21.3,-5.2);
\draw[line width=1.2pt,black!85] (10.9,-5.2) -- (21.3,-5.2);
\fill[fill={rgb,255:red,218;green,230;blue,242}] (8.3,-13.4) rectangle (15.26,-14.3);
\fill[fill={rgb,255:red,253;green,253;blue,253}] (15.23,-13.4) rectangle (16.13,-14.3);
\fill[fill={rgb,255:red,155;green,155;blue,155}] (16.1,-13.4) rectangle (17,-14.3);
\fill[fill={rgb,255:red,253;green,253;blue,253}] (16.97,-13.4) rectangle (17.86,-14.3);
\fill[fill={rgb,255:red,218;green,230;blue,242}] (17.83,-13.4) rectangle (23.03,-14.3);
\fill[fill={rgb,255:red,218;green,230;blue,242}] (8.3,-14.27) rectangle (16.13,-15.16);
\fill[fill={rgb,255:red,155;green,155;blue,155}] (16.1,-14.27) rectangle (17,-15.16);
\fill[fill={rgb,255:red,253;green,253;blue,253}] (16.97,-14.27) rectangle (17.86,-15.16);
\fill[fill={rgb,255:red,218;green,230;blue,242}] (17.83,-14.27) rectangle (23.03,-15.16);
\fill[fill={rgb,255:red,255;green,255;blue,255}] (8.3,-15.13) rectangle (10.93,-16.03);
\fill[fill={rgb,255:red,218;green,230;blue,242}] (10.9,-15.13) rectangle (11.8,-16.03);
\fill[fill={rgb,255:red,255;green,255;blue,255}] (11.77,-15.13) rectangle (16.13,-16.03);
\fill[fill={rgb,255:red,155;green,155;blue,155}] (16.1,-15.13) rectangle (17,-16.03);
\fill[fill={rgb,255:red,218;green,230;blue,242}] (16.97,-15.13) rectangle (23.03,-16.03);
\fill[fill={rgb,255:red,255;green,255;blue,255}] (8.3,-16) rectangle (10.06,-16.9);
\fill[fill={rgb,255:red,218;green,230;blue,242}] (10.03,-16) rectangle (10.93,-16.9);
\fill[fill={rgb,255:red,188;green,210;blue,232}] (10.9,-16) rectangle (12.66,-16.9);
\fill[fill={rgb,255:red,255;green,255;blue,255}] (12.63,-16) rectangle (15.26,-16.9);
\fill[fill={rgb,255:red,188;green,210;blue,232}] (15.23,-16) rectangle (21.33,-16.9);
\fill[fill={rgb,255:red,218;green,230;blue,242}] (21.3,-16) rectangle (23.03,-16.9);
\fill[fill={rgb,255:red,255;green,255;blue,255}] (8.3,-16.87) rectangle (10.93,-17.76);
\fill[fill={rgb,255:red,188;green,210;blue,232}] (10.9,-16.87) rectangle (11.8,-17.76);
\fill[fill={rgb,255:red,255;green,255;blue,255}] (11.77,-16.87) rectangle (16.13,-17.76);
\fill[fill={rgb,255:red,74;green,74;blue,74}] (16.1,-16.87) rectangle (17,-17.76);
\fill[fill={rgb,255:red,188;green,210;blue,232}] (16.97,-16.87) rectangle (21.33,-17.76);
\fill[fill={rgb,255:red,218;green,230;blue,242}] (21.3,-16.87) rectangle (23.03,-17.76);
\fill[fill={rgb,255:red,218;green,230;blue,242}] (8.3,-17.73) rectangle (10.93,-18.63);
\fill[fill={rgb,255:red,188;green,210;blue,232}] (10.9,-17.73) rectangle (16.13,-18.63);
\fill[fill={rgb,255:red,74;green,74;blue,74}] (16.1,-17.73) rectangle (17,-18.63);
\fill[fill={rgb,255:red,251;green,251;blue,251}] (16.97,-17.73) rectangle (17.86,-18.63);
\fill[fill={rgb,255:red,188;green,210;blue,232}] (17.83,-17.73) rectangle (21.33,-18.63);
\fill[fill={rgb,255:red,218;green,230;blue,242}] (21.3,-17.73) rectangle (23.03,-18.63);
\fill[fill={rgb,255:red,218;green,230;blue,242}] (8.3,-18.6) rectangle (10.93,-19.5);
\fill[fill={rgb,255:red,188;green,210;blue,232}] (10.9,-18.6) rectangle (15.26,-19.5);
\fill[fill={rgb,255:red,251;green,251;blue,251}] (15.23,-18.6) rectangle (16.13,-19.5);
\fill[fill={rgb,255:red,74;green,74;blue,74}] (16.1,-18.6) rectangle (17,-19.5);
\fill[fill={rgb,255:red,251;green,251;blue,251}] (16.97,-18.6) rectangle (17.86,-19.5);
\fill[fill={rgb,255:red,188;green,210;blue,232}] (17.83,-18.6) rectangle (21.33,-19.5);
\fill[fill={rgb,255:red,218;green,230;blue,242}] (21.3,-18.6) rectangle (23.03,-19.5);
\fill[fill={rgb,255:red,218;green,230;blue,242}] (8.3,-19.47) rectangle (10.93,-20.36);
\fill[fill={rgb,255:red,188;green,210;blue,232}] (10.9,-19.47) rectangle (15.26,-20.36);
\fill[fill={rgb,255:red,251;green,251;blue,251}] (15.23,-19.47) rectangle (16.13,-20.36);
\fill[fill={rgb,255:red,74;green,74;blue,74}] (16.1,-19.47) rectangle (17,-20.36);
\fill[fill={rgb,255:red,251;green,251;blue,251}] (16.97,-19.47) rectangle (18.73,-20.36);
\fill[fill={rgb,255:red,188;green,210;blue,232}] (18.7,-19.47) rectangle (21.33,-20.36);
\fill[fill={rgb,255:red,218;green,230;blue,242}] (21.3,-19.47) rectangle (23.03,-20.36);
\fill[fill={rgb,255:red,218;green,230;blue,242}] (8.3,-20.33) rectangle (10.93,-21.23);
\fill[fill={rgb,255:red,188;green,210;blue,232}] (10.9,-20.33) rectangle (14.4,-21.23);
\fill[fill={rgb,255:red,251;green,251;blue,251}] (14.37,-20.33) rectangle (16.13,-21.23);
\fill[fill={rgb,255:red,74;green,74;blue,74}] (16.1,-20.33) rectangle (17,-21.23);
\fill[fill={rgb,255:red,251;green,251;blue,251}] (16.97,-20.33) rectangle (18.73,-21.23);
\fill[fill={rgb,255:red,188;green,210;blue,232}] (18.7,-20.33) rectangle (21.33,-21.23);
\fill[fill={rgb,255:red,218;green,230;blue,242}] (21.3,-20.33) rectangle (23.03,-21.23);
\fill[fill={rgb,255:red,218;green,230;blue,242}] (8.3,-21.2) rectangle (10.93,-22.1);
\fill[fill={rgb,255:red,188;green,210;blue,232}] (10.9,-21.2) rectangle (14.4,-22.1);
\fill[fill={rgb,255:red,251;green,251;blue,251}] (14.37,-21.2) rectangle (16.13,-22.1);
\fill[fill={rgb,255:red,74;green,74;blue,74}] (16.1,-21.2) rectangle (17,-22.1);
\fill[fill={rgb,255:red,251;green,251;blue,251}] (16.97,-21.2) rectangle (19.6,-22.1);
\fill[fill={rgb,255:red,188;green,210;blue,232}] (19.57,-21.2) rectangle (21.33,-22.1);
\fill[fill={rgb,255:red,218;green,230;blue,242}] (21.3,-21.2) rectangle (22.2,-22.1);
\fill[fill={rgb,255:red,253;green,253;blue,253}] (22.17,-21.2) rectangle (23.03,-22.1);
\fill[fill={rgb,255:red,253;green,253;blue,253}] (8.3,-22.07) rectangle (9.2,-22.96);
\fill[fill={rgb,255:red,218;green,230;blue,242}] (9.17,-22.07) rectangle (10.93,-22.96);
\fill[fill={rgb,255:red,188;green,210;blue,232}] (10.9,-22.07) rectangle (13.53,-22.96);
\fill[fill={rgb,255:red,251;green,251;blue,251}] (13.5,-22.07) rectangle (16.13,-22.96);
\fill[fill={rgb,255:red,74;green,74;blue,74}] (16.1,-22.07) rectangle (17,-22.96);
\fill[fill={rgb,255:red,251;green,251;blue,251}] (16.97,-22.07) rectangle (19.6,-22.96);
\fill[fill={rgb,255:red,188;green,210;blue,232}] (19.57,-22.07) rectangle (21.33,-22.96);
\fill[fill={rgb,255:red,218;green,230;blue,242}] (21.3,-22.07) rectangle (22.2,-22.96);
\fill[fill={rgb,255:red,253;green,253;blue,253}] (22.17,-22.07) rectangle (23.03,-22.96);
\fill[fill={rgb,255:red,166;green,166;blue,166}] (8.3,-22.93) rectangle (10.06,-23.83);
\fill[fill={rgb,255:red,218;green,230;blue,242}] (10.03,-22.93) rectangle (10.93,-23.83);
\fill[fill={rgb,255:red,188;green,210;blue,232}] (10.9,-22.93) rectangle (12.66,-23.83);
\fill[fill={rgb,255:red,93;green,93;blue,93}] (12.63,-22.93) rectangle (16.13,-23.83);
\fill[fill={rgb,255:red,74;green,74;blue,74}] (16.1,-22.93) rectangle (17,-23.83);
\fill[fill={rgb,255:red,93;green,93;blue,93}] (16.97,-22.93) rectangle (19.6,-23.83);
\fill[fill={rgb,255:red,188;green,210;blue,232}] (19.57,-22.93) rectangle (21.33,-23.83);
\fill[fill={rgb,255:red,218;green,230;blue,242}] (21.3,-22.93) rectangle (22.2,-23.83);
\fill[fill={rgb,255:red,166;green,166;blue,166}] (22.17,-22.93) rectangle (23.03,-23.83);
\fill[fill={rgb,255:red,166;green,166;blue,166}] (8.3,-23.8) rectangle (9.2,-24.7);
\fill[fill={rgb,255:red,218;green,230;blue,242}] (9.17,-23.8) rectangle (10.93,-24.7);
\fill[fill={rgb,255:red,188;green,210;blue,232}] (10.9,-23.8) rectangle (13.53,-24.7);
\fill[fill={rgb,255:red,93;green,93;blue,93}] (13.5,-23.8) rectangle (18.73,-24.7);
\fill[fill={rgb,255:red,188;green,210;blue,232}] (18.7,-23.8) rectangle (21.33,-24.7);
\fill[fill={rgb,255:red,218;green,230;blue,242}] (21.3,-23.8) rectangle (23.03,-24.7);
\fill[fill={rgb,255:red,196;green,212;blue,227}] (8.3,-24.67) rectangle (10.93,-25.56);
\fill[fill={rgb,255:red,147;green,177;blue,204}] (10.9,-24.67) rectangle (21.33,-25.56);
\fill[fill={rgb,255:red,196;green,212;blue,227}] (21.3,-24.67) rectangle (23.03,-25.56);
\fill[fill={rgb,255:red,196;green,212;blue,227}] (8.3,-25.53) rectangle (10.93,-26.43);
\fill[fill={rgb,255:red,147;green,177;blue,204}] (10.9,-25.53) rectangle (21.33,-26.43);
\fill[fill={rgb,255:red,196;green,212;blue,227}] (21.3,-25.53) rectangle (23.03,-26.43);
\fill[fill={rgb,255:red,196;green,212;blue,227}] (8.3,-26.4) rectangle (23.03,-27.3);
\fill[fill={rgb,255:red,166;green,166;blue,166}] (8.3,-27.27) rectangle (9.2,-28.13);
\fill[fill={rgb,255:red,218;green,230;blue,242}] (9.17,-27.27) rectangle (13.53,-28.13);
\fill[fill={rgb,255:red,166;green,166;blue,166}] (13.5,-27.27) rectangle (18.73,-28.13);
\fill[fill={rgb,255:red,218;green,230;blue,242}] (18.7,-27.27) rectangle (23.03,-28.13);
\draw[black!60,line width=0.6pt] (10.9,-16) rectangle (21.3,-26.4);
\draw[nBlue,line width=1.35pt,dash pattern=on 1.95pt off 1.35pt] (8.3,-13.4) rectangle (18.7,-23.8) (13.5,-13.4) -- (13.5,-23.8) (8.3,-18.6) -- (18.7,-18.6);
\draw[-{Stealth[length=2.7mm,width=2.25mm]},line width=1.2pt] (22.1,0) -- (30.4,0);
\draw[-{Stealth[length=2.7mm,width=2.25mm]},line width=1.2pt] (23.83,-21.2) -- (30.4,-21.2);
\node[anchor=south,font=\footnotesize] at (26.25,0.4) {stride $P$};
\node[anchor=south,text=nNavy,font=\footnotesize] at (27.12,-20.8) {stride 1};
\fill[nRed] (31.2,5.9) rectangle ++(1.6,1.6);
\fill[nRed] (31.2,-15.3) rectangle ++(1.6,1.6);
\filldraw[fill=nTok,draw=nTokB,line width=0.45pt] (31.2,5.2) rectangle ++(4.6,-4.6);
\filldraw[fill=nTok,draw=nTokB,line width=0.45pt] (37,5.2) rectangle ++(4.6,-4.6);
\filldraw[fill=nTok,draw=nTokB,line width=0.45pt] (31.2,-0.6) rectangle ++(4.6,-4.6);
\filldraw[fill=nTok,draw=nTokB,line width=0.45pt] (37,-0.6) rectangle ++(4.6,-4.6);
\fill[nBlue] (31.2,-16) rectangle ++(0.68,-0.68);
\fill[nOrange] (32.08,-16) rectangle ++(0.68,-0.68);
\fill[nGreen] (32.97,-16) rectangle ++(0.68,-0.68);
\fill[nTok] (33.85,-16) rectangle ++(0.68,-0.68);
\fill[nTok] (34.73,-16) rectangle ++(0.68,-0.68);
\fill[nTok] (35.62,-16) rectangle ++(0.68,-0.68);
\fill[nBlue] (36.5,-16) rectangle ++(0.68,-0.68);
\fill[nOrange] (37.38,-16) rectangle ++(0.68,-0.68);
\fill[nGreen] (38.27,-16) rectangle ++(0.68,-0.68);
\fill[nTok] (39.15,-16) rectangle ++(0.68,-0.68);
\fill[nTok] (40.03,-16) rectangle ++(0.68,-0.68);
\fill[nTok] (40.92,-16) rectangle ++(0.68,-0.68);
\fill[nTok] (31.2,-16.88) rectangle ++(0.68,-0.68);
\fill[nTok] (32.08,-16.88) rectangle ++(0.68,-0.68);
\fill[nTok] (32.97,-16.88) rectangle ++(0.68,-0.68);
\fill[nTok] (33.85,-16.88) rectangle ++(0.68,-0.68);
\fill[nTok] (34.73,-16.88) rectangle ++(0.68,-0.68);
\fill[nTok] (35.62,-16.88) rectangle ++(0.68,-0.68);
\fill[nTok] (36.5,-16.88) rectangle ++(0.68,-0.68);
\fill[nTok] (37.38,-16.88) rectangle ++(0.68,-0.68);
\fill[nTok] (38.27,-16.88) rectangle ++(0.68,-0.68);
\fill[nTok] (39.15,-16.88) rectangle ++(0.68,-0.68);
\fill[nTok] (40.03,-16.88) rectangle ++(0.68,-0.68);
\fill[nTok] (40.92,-16.88) rectangle ++(0.68,-0.68);
\fill[nTok] (31.2,-17.77) rectangle ++(0.68,-0.68);
\fill[nTok] (32.08,-17.77) rectangle ++(0.68,-0.68);
\fill[nTok] (32.97,-17.77) rectangle ++(0.68,-0.68);
\fill[nTok] (33.85,-17.77) rectangle ++(0.68,-0.68);
\fill[nTok] (34.73,-17.77) rectangle ++(0.68,-0.68);
\fill[nTok] (35.62,-17.77) rectangle ++(0.68,-0.68);
\fill[nTok] (36.5,-17.77) rectangle ++(0.68,-0.68);
\fill[nTok] (37.38,-17.77) rectangle ++(0.68,-0.68);
\fill[nTok] (38.27,-17.77) rectangle ++(0.68,-0.68);
\fill[nTok] (39.15,-17.77) rectangle ++(0.68,-0.68);
\fill[nTok] (40.03,-17.77) rectangle ++(0.68,-0.68);
\fill[nTok] (40.92,-17.77) rectangle ++(0.68,-0.68);
\fill[nTok] (31.2,-18.65) rectangle ++(0.68,-0.68);
\fill[nTok] (32.08,-18.65) rectangle ++(0.68,-0.68);
\fill[nTok] (32.97,-18.65) rectangle ++(0.68,-0.68);
\fill[nTok] (33.85,-18.65) rectangle ++(0.68,-0.68);
\fill[nTok] (34.73,-18.65) rectangle ++(0.68,-0.68);
\fill[nTok] (35.62,-18.65) rectangle ++(0.68,-0.68);
\fill[nTok] (36.5,-18.65) rectangle ++(0.68,-0.68);
\fill[nTok] (37.38,-18.65) rectangle ++(0.68,-0.68);
\fill[nTok] (38.27,-18.65) rectangle ++(0.68,-0.68);
\fill[nTok] (39.15,-18.65) rectangle ++(0.68,-0.68);
\fill[nTok] (40.03,-18.65) rectangle ++(0.68,-0.68);
\fill[nTok] (40.92,-18.65) rectangle ++(0.68,-0.68);
\fill[nTok] (31.2,-19.53) rectangle ++(0.68,-0.68);
\fill[nTok] (32.08,-19.53) rectangle ++(0.68,-0.68);
\fill[nTok] (32.97,-19.53) rectangle ++(0.68,-0.68);
\fill[nTok] (33.85,-19.53) rectangle ++(0.68,-0.68);
\fill[nTok] (34.73,-19.53) rectangle ++(0.68,-0.68);
\fill[nTok] (35.62,-19.53) rectangle ++(0.68,-0.68);
\fill[nTok] (36.5,-19.53) rectangle ++(0.68,-0.68);
\fill[nTok] (37.38,-19.53) rectangle ++(0.68,-0.68);
\fill[nTok] (38.27,-19.53) rectangle ++(0.68,-0.68);
\fill[nTok] (39.15,-19.53) rectangle ++(0.68,-0.68);
\fill[nTok] (40.03,-19.53) rectangle ++(0.68,-0.68);
\fill[nTok] (40.92,-19.53) rectangle ++(0.68,-0.68);
\fill[nTok] (31.2,-20.42) rectangle ++(0.68,-0.68);
\fill[nTok] (32.08,-20.42) rectangle ++(0.68,-0.68);
\fill[nTok] (32.97,-20.42) rectangle ++(0.68,-0.68);
\fill[nTok] (33.85,-20.42) rectangle ++(0.68,-0.68);
\fill[nTok] (34.73,-20.42) rectangle ++(0.68,-0.68);
\fill[nTok] (35.62,-20.42) rectangle ++(0.68,-0.68);
\fill[nTok] (36.5,-20.42) rectangle ++(0.68,-0.68);
\fill[nTok] (37.38,-20.42) rectangle ++(0.68,-0.68);
\fill[nTok] (38.27,-20.42) rectangle ++(0.68,-0.68);
\fill[nTok] (39.15,-20.42) rectangle ++(0.68,-0.68);
\fill[nTok] (40.03,-20.42) rectangle ++(0.68,-0.68);
\fill[nTok] (40.92,-20.42) rectangle ++(0.68,-0.68);
\fill[nBlue] (31.2,-21.3) rectangle ++(0.68,-0.68);
\fill[nOrange] (32.08,-21.3) rectangle ++(0.68,-0.68);
\fill[nGreen] (32.97,-21.3) rectangle ++(0.68,-0.68);
\fill[nTok] (33.85,-21.3) rectangle ++(0.68,-0.68);
\fill[nTok] (34.73,-21.3) rectangle ++(0.68,-0.68);
\fill[nTok] (35.62,-21.3) rectangle ++(0.68,-0.68);
\fill[nBlue] (36.5,-21.3) rectangle ++(0.68,-0.68);
\fill[nOrange] (37.38,-21.3) rectangle ++(0.68,-0.68);
\fill[nGreen] (38.27,-21.3) rectangle ++(0.68,-0.68);
\fill[nTok] (39.15,-21.3) rectangle ++(0.68,-0.68);
\fill[nTok] (40.03,-21.3) rectangle ++(0.68,-0.68);
\fill[nTok] (40.92,-21.3) rectangle ++(0.68,-0.68);
\fill[nTok] (31.2,-22.18) rectangle ++(0.68,-0.68);
\fill[nTok] (32.08,-22.18) rectangle ++(0.68,-0.68);
\fill[nTok] (32.97,-22.18) rectangle ++(0.68,-0.68);
\fill[nTok] (33.85,-22.18) rectangle ++(0.68,-0.68);
\fill[nTok] (34.73,-22.18) rectangle ++(0.68,-0.68);
\fill[nTok] (35.62,-22.18) rectangle ++(0.68,-0.68);
\fill[nTok] (36.5,-22.18) rectangle ++(0.68,-0.68);
\fill[nTok] (37.38,-22.18) rectangle ++(0.68,-0.68);
\fill[nTok] (38.27,-22.18) rectangle ++(0.68,-0.68);
\fill[nTok] (39.15,-22.18) rectangle ++(0.68,-0.68);
\fill[nTok] (40.03,-22.18) rectangle ++(0.68,-0.68);
\fill[nTok] (40.92,-22.18) rectangle ++(0.68,-0.68);
\fill[nTok] (31.2,-23.07) rectangle ++(0.68,-0.68);
\fill[nTok] (32.08,-23.07) rectangle ++(0.68,-0.68);
\fill[nTok] (32.97,-23.07) rectangle ++(0.68,-0.68);
\fill[nTok] (33.85,-23.07) rectangle ++(0.68,-0.68);
\fill[nTok] (34.73,-23.07) rectangle ++(0.68,-0.68);
\fill[nTok] (35.62,-23.07) rectangle ++(0.68,-0.68);
\fill[nTok] (36.5,-23.07) rectangle ++(0.68,-0.68);
\fill[nTok] (37.38,-23.07) rectangle ++(0.68,-0.68);
\fill[nTok] (38.27,-23.07) rectangle ++(0.68,-0.68);
\fill[nTok] (39.15,-23.07) rectangle ++(0.68,-0.68);
\fill[nTok] (40.03,-23.07) rectangle ++(0.68,-0.68);
\fill[nTok] (40.92,-23.07) rectangle ++(0.68,-0.68);
\fill[nTok] (31.2,-23.95) rectangle ++(0.68,-0.68);
\fill[nTok] (32.08,-23.95) rectangle ++(0.68,-0.68);
\fill[nTok] (32.97,-23.95) rectangle ++(0.68,-0.68);
\fill[nTok] (33.85,-23.95) rectangle ++(0.68,-0.68);
\fill[nTok] (34.73,-23.95) rectangle ++(0.68,-0.68);
\fill[nTok] (35.62,-23.95) rectangle ++(0.68,-0.68);
\fill[nTok] (36.5,-23.95) rectangle ++(0.68,-0.68);
\fill[nTok] (37.38,-23.95) rectangle ++(0.68,-0.68);
\fill[nTok] (38.27,-23.95) rectangle ++(0.68,-0.68);
\fill[nTok] (39.15,-23.95) rectangle ++(0.68,-0.68);
\fill[nTok] (40.03,-23.95) rectangle ++(0.68,-0.68);
\fill[nTok] (40.92,-23.95) rectangle ++(0.68,-0.68);
\fill[nTok] (31.2,-24.83) rectangle ++(0.68,-0.68);
\fill[nTok] (32.08,-24.83) rectangle ++(0.68,-0.68);
\fill[nTok] (32.97,-24.83) rectangle ++(0.68,-0.68);
\fill[nTok] (33.85,-24.83) rectangle ++(0.68,-0.68);
\fill[nTok] (34.73,-24.83) rectangle ++(0.68,-0.68);
\fill[nTok] (35.62,-24.83) rectangle ++(0.68,-0.68);
\fill[nTok] (36.5,-24.83) rectangle ++(0.68,-0.68);
\fill[nTok] (37.38,-24.83) rectangle ++(0.68,-0.68);
\fill[nTok] (38.27,-24.83) rectangle ++(0.68,-0.68);
\fill[nTok] (39.15,-24.83) rectangle ++(0.68,-0.68);
\fill[nTok] (40.03,-24.83) rectangle ++(0.68,-0.68);
\fill[nTok] (40.92,-24.83) rectangle ++(0.68,-0.68);
\fill[nTok] (31.2,-25.72) rectangle ++(0.68,-0.68);
\fill[nTok] (32.08,-25.72) rectangle ++(0.68,-0.68);
\fill[nTok] (32.97,-25.72) rectangle ++(0.68,-0.68);
\fill[nTok] (33.85,-25.72) rectangle ++(0.68,-0.68);
\fill[nTok] (34.73,-25.72) rectangle ++(0.68,-0.68);
\fill[nTok] (35.62,-25.72) rectangle ++(0.68,-0.68);
\fill[nTok] (36.5,-25.72) rectangle ++(0.68,-0.68);
\fill[nTok] (37.38,-25.72) rectangle ++(0.68,-0.68);
\fill[nTok] (38.27,-25.72) rectangle ++(0.68,-0.68);
\fill[nTok] (39.15,-25.72) rectangle ++(0.68,-0.68);
\fill[nTok] (40.03,-25.72) rectangle ++(0.68,-0.68);
\fill[nTok] (40.92,-25.72) rectangle ++(0.68,-0.68);
\draw[-{Stealth[length=2.7mm,width=2.25mm]},line width=1.2pt] (42.4,0) -- (46,0);
\draw[-{Stealth[length=2.7mm,width=2.25mm]},line width=1.2pt] (64.2,0) -- (67.8,0);
\draw[-{Stealth[length=2.7mm,width=2.25mm]},line width=1.2pt] (42.4,-21.2) -- (46,-21.2);
\draw[-{Stealth[length=2.7mm,width=2.25mm]},line width=1.2pt] (64.2,-21.2) -- (67.8,-21.2);
\draw[black!55,line width=0.9pt,rounded corners=1.5pt] (46.8,5.2) rectangle (63.4,-5.2);
\filldraw[fill=nBlueL!35,draw=nNavy,line width=1.35pt,rounded corners=1.5pt] (46.8,-16) rectangle (63.4,-26.4);
\filldraw[fill=nTok,draw=nTokB,line width=0.45pt] (48.4,4.4) rectangle ++(3.8,-3.8);
\filldraw[fill=nTok,draw=nTokB,line width=0.45pt] (53.4,4.4) rectangle ++(3.8,-3.8);
\filldraw[fill=nTok,draw=nTokB,line width=0.45pt] (48.4,-0.6) rectangle ++(3.8,-3.8);
\filldraw[fill=nTok,draw=nTokB,line width=0.45pt] (53.4,-0.6) rectangle ++(3.8,-3.8);
\fill[black!75] (55.3,-2.5) circle (.32);
\draw[-{Stealth[length=1.62mm,width=1.53mm]},black!75,line width=0.75pt] (55.3,-2.5) -- (50.3,2.5);
\draw[-{Stealth[length=1.62mm,width=1.53mm]},black!75,line width=0.75pt] (55.3,-2.5) -- (55.3,2.5);
\draw[-{Stealth[length=1.62mm,width=1.53mm]},black!75,line width=0.75pt] (55.3,-2.5) -- (50.3,-2.5);
\fill[nRed] (59.2,-0.7) rectangle ++(1.8,1.8);
\draw[-{Stealth[length=1.62mm,width=1.53mm]},nRed,line width=1.05pt,shorten >=0.27mm] (55.3,-2.5) -- (59.2,0);
\draw[nRed,line width=0.68pt,-{Stealth[length=1.44mm,width=1.44mm]}] (59.7,1.1) to[out=95,in=30,looseness=6] (61,0.7);
\fill[nTok!70] (48.4,-16.8) rectangle ++(0.6,-0.6);
\fill[nTok!70] (49.15,-16.8) rectangle ++(0.6,-0.6);
\fill[nTok!70] (49.89,-16.8) rectangle ++(0.6,-0.6);
\fill[nTok!70] (50.64,-16.8) rectangle ++(0.6,-0.6);
\fill[nTok!70] (51.38,-16.8) rectangle ++(0.6,-0.6);
\fill[nTok!70] (52.13,-16.8) rectangle ++(0.6,-0.6);
\fill[nTok!70] (52.87,-16.8) rectangle ++(0.6,-0.6);
\fill[nTok!70] (53.62,-16.8) rectangle ++(0.6,-0.6);
\fill[nTok!70] (54.36,-16.8) rectangle ++(0.6,-0.6);
\fill[nTok!70] (55.11,-16.8) rectangle ++(0.6,-0.6);
\fill[nTok!70] (55.85,-16.8) rectangle ++(0.6,-0.6);
\fill[nTok!70] (56.6,-16.8) rectangle ++(0.6,-0.6);
\fill[nTok!70] (48.4,-17.55) rectangle ++(0.6,-0.6);
\fill[nTok!70] (49.15,-17.55) rectangle ++(0.6,-0.6);
\fill[nTok!70] (49.89,-17.55) rectangle ++(0.6,-0.6);
\fill[nTok!70] (50.64,-17.55) rectangle ++(0.6,-0.6);
\fill[nTok!70] (51.38,-17.55) rectangle ++(0.6,-0.6);
\fill[nTok!70] (52.13,-17.55) rectangle ++(0.6,-0.6);
\fill[nTok!70] (52.87,-17.55) rectangle ++(0.6,-0.6);
\fill[nTok!70] (53.62,-17.55) rectangle ++(0.6,-0.6);
\fill[nTok!70] (54.36,-17.55) rectangle ++(0.6,-0.6);
\fill[nTok!70] (55.11,-17.55) rectangle ++(0.6,-0.6);
\fill[nTok!70] (55.85,-17.55) rectangle ++(0.6,-0.6);
\fill[nTok!70] (56.6,-17.55) rectangle ++(0.6,-0.6);
\fill[nTok!70] (48.4,-18.29) rectangle ++(0.6,-0.6);
\fill[nTok!70] (49.15,-18.29) rectangle ++(0.6,-0.6);
\fill[nTok!70] (49.89,-18.29) rectangle ++(0.6,-0.6);
\fill[nTok!70] (50.64,-18.29) rectangle ++(0.6,-0.6);
\fill[nTok!70] (51.38,-18.29) rectangle ++(0.6,-0.6);
\fill[nTok!70] (52.13,-18.29) rectangle ++(0.6,-0.6);
\fill[nTok!70] (52.87,-18.29) rectangle ++(0.6,-0.6);
\fill[nTok!70] (53.62,-18.29) rectangle ++(0.6,-0.6);
\fill[nTok!70] (54.36,-18.29) rectangle ++(0.6,-0.6);
\fill[nTok!70] (55.11,-18.29) rectangle ++(0.6,-0.6);
\fill[nTok!70] (55.85,-18.29) rectangle ++(0.6,-0.6);
\fill[nTok!70] (56.6,-18.29) rectangle ++(0.6,-0.6);
\fill[nTok!70] (48.4,-19.04) rectangle ++(0.6,-0.6);
\fill[nTok!70] (49.15,-19.04) rectangle ++(0.6,-0.6);
\fill[nTok!70] (49.89,-19.04) rectangle ++(0.6,-0.6);
\fill[nTok!70] (50.64,-19.04) rectangle ++(0.6,-0.6);
\fill[nTok!70] (51.38,-19.04) rectangle ++(0.6,-0.6);
\fill[nTok!70] (52.13,-19.04) rectangle ++(0.6,-0.6);
\fill[nTok!70] (52.87,-19.04) rectangle ++(0.6,-0.6);
\fill[nTok!70] (53.62,-19.04) rectangle ++(0.6,-0.6);
\fill[nTok!70] (54.36,-19.04) rectangle ++(0.6,-0.6);
\fill[nTok!70] (55.11,-19.04) rectangle ++(0.6,-0.6);
\fill[nTok!70] (55.85,-19.04) rectangle ++(0.6,-0.6);
\fill[nTok!70] (56.6,-19.04) rectangle ++(0.6,-0.6);
\fill[nTok!70] (48.4,-19.78) rectangle ++(0.6,-0.6);
\fill[nTok!70] (49.15,-19.78) rectangle ++(0.6,-0.6);
\fill[nTok!70] (49.89,-19.78) rectangle ++(0.6,-0.6);
\fill[nTok!70] (50.64,-19.78) rectangle ++(0.6,-0.6);
\fill[nTok!70] (51.38,-19.78) rectangle ++(0.6,-0.6);
\fill[nTok!70] (52.13,-19.78) rectangle ++(0.6,-0.6);
\fill[nTok!70] (52.87,-19.78) rectangle ++(0.6,-0.6);
\fill[nTok!70] (53.62,-19.78) rectangle ++(0.6,-0.6);
\fill[nTok!70] (54.36,-19.78) rectangle ++(0.6,-0.6);
\fill[nTok!70] (55.11,-19.78) rectangle ++(0.6,-0.6);
\fill[nTok!70] (55.85,-19.78) rectangle ++(0.6,-0.6);
\fill[nTok!70] (56.6,-19.78) rectangle ++(0.6,-0.6);
\fill[nTok!70] (48.4,-20.53) rectangle ++(0.6,-0.6);
\fill[nTok!70] (49.15,-20.53) rectangle ++(0.6,-0.6);
\fill[nTok!70] (49.89,-20.53) rectangle ++(0.6,-0.6);
\fill[nTok!70] (50.64,-20.53) rectangle ++(0.6,-0.6);
\fill[nTok!70] (51.38,-20.53) rectangle ++(0.6,-0.6);
\fill[nTok!70] (52.13,-20.53) rectangle ++(0.6,-0.6);
\fill[nTok!70] (52.87,-20.53) rectangle ++(0.6,-0.6);
\fill[nTok!70] (53.62,-20.53) rectangle ++(0.6,-0.6);
\fill[nTok!70] (54.36,-20.53) rectangle ++(0.6,-0.6);
\fill[nTok!70] (55.11,-20.53) rectangle ++(0.6,-0.6);
\fill[nTok!70] (55.85,-20.53) rectangle ++(0.6,-0.6);
\fill[nTok!70] (56.6,-20.53) rectangle ++(0.6,-0.6);
\fill[nTok!70] (48.4,-21.27) rectangle ++(0.6,-0.6);
\fill[nTok!70] (49.15,-21.27) rectangle ++(0.6,-0.6);
\fill[nTok!70] (49.89,-21.27) rectangle ++(0.6,-0.6);
\fill[nTok!70] (50.64,-21.27) rectangle ++(0.6,-0.6);
\fill[nTok!70] (51.38,-21.27) rectangle ++(0.6,-0.6);
\fill[nTok!70] (52.13,-21.27) rectangle ++(0.6,-0.6);
\fill[nTok!70] (52.87,-21.27) rectangle ++(0.6,-0.6);
\fill[nTok!70] (53.62,-21.27) rectangle ++(0.6,-0.6);
\fill[nTok!70] (54.36,-21.27) rectangle ++(0.6,-0.6);
\fill[nTok!70] (55.11,-21.27) rectangle ++(0.6,-0.6);
\fill[nTok!70] (55.85,-21.27) rectangle ++(0.6,-0.6);
\fill[nTok!70] (56.6,-21.27) rectangle ++(0.6,-0.6);
\fill[nTok!70] (48.4,-22.02) rectangle ++(0.6,-0.6);
\fill[nTok!70] (49.15,-22.02) rectangle ++(0.6,-0.6);
\fill[nTok!70] (49.89,-22.02) rectangle ++(0.6,-0.6);
\fill[nTok!70] (50.64,-22.02) rectangle ++(0.6,-0.6);
\fill[nTok!70] (51.38,-22.02) rectangle ++(0.6,-0.6);
\fill[nTok!70] (52.13,-22.02) rectangle ++(0.6,-0.6);
\fill[nTok!70] (52.87,-22.02) rectangle ++(0.6,-0.6);
\fill[nTok!70] (53.62,-22.02) rectangle ++(0.6,-0.6);
\fill[nTok!70] (54.36,-22.02) rectangle ++(0.6,-0.6);
\fill[nTok!70] (55.11,-22.02) rectangle ++(0.6,-0.6);
\fill[nTok!70] (55.85,-22.02) rectangle ++(0.6,-0.6);
\fill[nTok!70] (56.6,-22.02) rectangle ++(0.6,-0.6);
\fill[nTok!70] (48.4,-22.76) rectangle ++(0.6,-0.6);
\fill[nTok!70] (49.15,-22.76) rectangle ++(0.6,-0.6);
\fill[nTok!70] (49.89,-22.76) rectangle ++(0.6,-0.6);
\fill[nTok!70] (50.64,-22.76) rectangle ++(0.6,-0.6);
\fill[nTok!70] (51.38,-22.76) rectangle ++(0.6,-0.6);
\fill[nTok!70] (52.13,-22.76) rectangle ++(0.6,-0.6);
\fill[nTok!70] (52.87,-22.76) rectangle ++(0.6,-0.6);
\fill[nTok!70] (53.62,-22.76) rectangle ++(0.6,-0.6);
\fill[nTok!70] (54.36,-22.76) rectangle ++(0.6,-0.6);
\fill[nTok!70] (55.11,-22.76) rectangle ++(0.6,-0.6);
\fill[nTok!70] (55.85,-22.76) rectangle ++(0.6,-0.6);
\fill[nTok!70] (56.6,-22.76) rectangle ++(0.6,-0.6);
\fill[nTok!70] (48.4,-23.51) rectangle ++(0.6,-0.6);
\fill[nTok!70] (49.15,-23.51) rectangle ++(0.6,-0.6);
\fill[nTok!70] (49.89,-23.51) rectangle ++(0.6,-0.6);
\fill[nTok!70] (50.64,-23.51) rectangle ++(0.6,-0.6);
\fill[nTok!70] (51.38,-23.51) rectangle ++(0.6,-0.6);
\fill[nTok!70] (52.13,-23.51) rectangle ++(0.6,-0.6);
\fill[nTok!70] (52.87,-23.51) rectangle ++(0.6,-0.6);
\fill[nTok!70] (53.62,-23.51) rectangle ++(0.6,-0.6);
\fill[nTok!70] (54.36,-23.51) rectangle ++(0.6,-0.6);
\fill[nTok!70] (55.11,-23.51) rectangle ++(0.6,-0.6);
\fill[nTok!70] (55.85,-23.51) rectangle ++(0.6,-0.6);
\fill[nTok!70] (56.6,-23.51) rectangle ++(0.6,-0.6);
\fill[nTok!70] (48.4,-24.25) rectangle ++(0.6,-0.6);
\fill[nTok!70] (49.15,-24.25) rectangle ++(0.6,-0.6);
\fill[nTok!70] (49.89,-24.25) rectangle ++(0.6,-0.6);
\fill[nTok!70] (50.64,-24.25) rectangle ++(0.6,-0.6);
\fill[nTok!70] (51.38,-24.25) rectangle ++(0.6,-0.6);
\fill[nTok!70] (52.13,-24.25) rectangle ++(0.6,-0.6);
\fill[nTok!70] (52.87,-24.25) rectangle ++(0.6,-0.6);
\fill[nTok!70] (53.62,-24.25) rectangle ++(0.6,-0.6);
\fill[nTok!70] (54.36,-24.25) rectangle ++(0.6,-0.6);
\fill[nTok!70] (55.11,-24.25) rectangle ++(0.6,-0.6);
\fill[nTok!70] (55.85,-24.25) rectangle ++(0.6,-0.6);
\fill[nTok!70] (56.6,-24.25) rectangle ++(0.6,-0.6);
\fill[nTok!70] (48.4,-25) rectangle ++(0.6,-0.6);
\fill[nTok!70] (49.15,-25) rectangle ++(0.6,-0.6);
\fill[nTok!70] (49.89,-25) rectangle ++(0.6,-0.6);
\fill[nTok!70] (50.64,-25) rectangle ++(0.6,-0.6);
\fill[nTok!70] (51.38,-25) rectangle ++(0.6,-0.6);
\fill[nTok!70] (52.13,-25) rectangle ++(0.6,-0.6);
\fill[nTok!70] (52.87,-25) rectangle ++(0.6,-0.6);
\fill[nTok!70] (53.62,-25) rectangle ++(0.6,-0.6);
\fill[nTok!70] (54.36,-25) rectangle ++(0.6,-0.6);
\fill[nTok!70] (55.11,-25) rectangle ++(0.6,-0.6);
\fill[nTok!70] (55.85,-25) rectangle ++(0.6,-0.6);
\fill[nTok!70] (56.6,-25) rectangle ++(0.6,-0.6);
\fill[nBlue] (48.33,-17.47) rectangle (49.07,-16.73);
\fill[nOrange] (49.08,-17.47) rectangle (49.82,-16.73);
\fill[nGreen] (49.82,-17.47) rectangle (50.56,-16.73);
\fill[nBlue] (52.8,-17.47) rectangle (53.54,-16.73);
\fill[nOrange] (53.55,-17.47) rectangle (54.29,-16.73);
\fill[nGreen] (54.29,-17.47) rectangle (55.03,-16.73);
\fill[nBlue] (48.33,-21.94) rectangle (49.07,-21.2);
\fill[nOrange] (49.08,-21.94) rectangle (49.82,-21.2);
\fill[nGreen] (49.82,-21.94) rectangle (50.56,-21.2);
\fill[nBlue] (52.8,-21.94) rectangle (53.54,-21.2);
\fill[nOrange] (53.55,-21.94) rectangle (54.29,-21.2);
\fill[nGreen] (54.29,-21.94) rectangle (55.03,-21.2);
\fill[black!75] (53.17,-21.57) circle (.17);
\fill[nRed] (59.2,-21.9) rectangle ++(1.8,1.8);
\draw[-{Stealth[length=1.62mm,width=1.53mm]},black!80,line width=0.75pt,shorten >=0.76mm] (53.17,-21.57) to (53.17,-17.1);
\draw[-{Stealth[length=1.62mm,width=1.53mm]},black!80,line width=0.75pt,shorten >=0.76mm] (53.17,-21.57) to (48.7,-17.1);
\draw[-{Stealth[length=1.62mm,width=1.53mm]},black!80,line width=0.75pt,shorten >=0.76mm] (53.17,-21.57) to[bend left=55] (48.7,-21.57);
\draw[-{Stealth[length=1.62mm,width=1.53mm]},nRed,line width=1.2pt,shorten >=0.27mm] (53.17,-21.57) to[bend left=35] (59.2,-20.7);
\draw[nRed,line width=2.25pt,-{Stealth[length=2.34mm,width=2.52mm]}] (59.7,-20.1) to[out=95,in=30,looseness=6] (61,-20.5);
\fill[fill={rgb,255:red,220;green,236;blue,248}] (68.6,5.2) rectangle (73.83,-0.03);
\fill[fill={rgb,255:red,196;green,221;blue,242}] (73.8,5.2) rectangle (79,-0.03);
\fill[fill={rgb,255:red,172;green,207;blue,236}] (68.6,0) rectangle (73.83,-5.2);
\fill[fill={rgb,255:red,159;green,200;blue,233}] (73.8,0) rectangle (79,-5.2);
\fill[fill={rgb,255:red,238;green,246;blue,252}] (68.6,-16) rectangle (70.36,-16.9);
\fill[fill={rgb,255:red,212;green,231;blue,246}] (70.33,-16) rectangle (72.96,-16.9);
\fill[fill={rgb,255:red,238;green,246;blue,252}] (72.93,-16) rectangle (79,-16.9);
\fill[fill={rgb,255:red,238;green,246;blue,252}] (68.6,-16.87) rectangle (69.5,-17.76);
\fill[fill={rgb,255:red,212;green,231;blue,246}] (69.47,-16.87) rectangle (73.83,-17.76);
\fill[fill={rgb,255:red,63;green,143;blue,210}] (73.8,-16.87) rectangle (74.7,-17.76);
\fill[fill={rgb,255:red,238;green,246;blue,252}] (74.67,-16.87) rectangle (79,-17.76);
\fill[fill={rgb,255:red,238;green,246;blue,252}] (68.6,-17.73) rectangle (73.83,-18.63);
\fill[fill={rgb,255:red,63;green,143;blue,210}] (73.8,-17.73) rectangle (74.7,-18.63);
\fill[fill={rgb,255:red,133;green,184;blue,227}] (74.67,-17.73) rectangle (75.56,-18.63);
\fill[fill={rgb,255:red,238;green,246;blue,252}] (75.53,-17.73) rectangle (79,-18.63);
\fill[fill={rgb,255:red,238;green,246;blue,252}] (68.6,-18.6) rectangle (72.96,-19.5);
\fill[fill={rgb,255:red,133;green,184;blue,227}] (72.93,-18.6) rectangle (73.83,-19.5);
\fill[fill={rgb,255:red,63;green,143;blue,210}] (73.8,-18.6) rectangle (74.7,-19.5);
\fill[fill={rgb,255:red,133;green,184;blue,227}] (74.67,-18.6) rectangle (75.56,-19.5);
\fill[fill={rgb,255:red,238;green,246;blue,252}] (75.53,-18.6) rectangle (79,-19.5);
\fill[fill={rgb,255:red,238;green,246;blue,252}] (68.6,-19.47) rectangle (72.96,-20.36);
\fill[fill={rgb,255:red,133;green,184;blue,227}] (72.93,-19.47) rectangle (73.83,-20.36);
\fill[fill={rgb,255:red,63;green,143;blue,210}] (73.8,-19.47) rectangle (74.7,-20.36);
\fill[fill={rgb,255:red,133;green,184;blue,227}] (74.67,-19.47) rectangle (76.43,-20.36);
\fill[fill={rgb,255:red,238;green,246;blue,252}] (76.4,-19.47) rectangle (79,-20.36);
\fill[fill={rgb,255:red,238;green,246;blue,252}] (68.6,-20.33) rectangle (72.1,-21.23);
\fill[fill={rgb,255:red,133;green,184;blue,227}] (72.07,-20.33) rectangle (73.83,-21.23);
\fill[fill={rgb,255:red,63;green,143;blue,210}] (73.8,-20.33) rectangle (74.7,-21.23);
\fill[fill={rgb,255:red,133;green,184;blue,227}] (74.67,-20.33) rectangle (76.43,-21.23);
\fill[fill={rgb,255:red,238;green,246;blue,252}] (76.4,-20.33) rectangle (79,-21.23);
\fill[fill={rgb,255:red,238;green,246;blue,252}] (68.6,-21.2) rectangle (72.1,-22.1);
\fill[fill={rgb,255:red,133;green,184;blue,227}] (72.07,-21.2) rectangle (73.83,-22.1);
\fill[fill={rgb,255:red,63;green,143;blue,210}] (73.8,-21.2) rectangle (74.7,-22.1);
\fill[fill={rgb,255:red,133;green,184;blue,227}] (74.67,-21.2) rectangle (77.3,-22.1);
\fill[fill={rgb,255:red,238;green,246;blue,252}] (77.27,-21.2) rectangle (79,-22.1);
\fill[fill={rgb,255:red,238;green,246;blue,252}] (68.6,-22.07) rectangle (71.23,-22.96);
\fill[fill={rgb,255:red,133;green,184;blue,227}] (71.2,-22.07) rectangle (73.83,-22.96);
\fill[fill={rgb,255:red,63;green,143;blue,210}] (73.8,-22.07) rectangle (74.7,-22.96);
\fill[fill={rgb,255:red,133;green,184;blue,227}] (74.67,-22.07) rectangle (77.3,-22.96);
\fill[fill={rgb,255:red,238;green,246;blue,252}] (77.27,-22.07) rectangle (79,-22.96);
\fill[fill={rgb,255:red,238;green,246;blue,252}] (68.6,-22.93) rectangle (70.36,-23.83);
\fill[fill={rgb,255:red,80;green,153;blue,214}] (70.33,-22.93) rectangle (73.83,-23.83);
\fill[fill={rgb,255:red,63;green,143;blue,210}] (73.8,-22.93) rectangle (74.7,-23.83);
\fill[fill={rgb,255:red,80;green,153;blue,214}] (74.67,-22.93) rectangle (77.3,-23.83);
\fill[fill={rgb,255:red,238;green,246;blue,252}] (77.27,-22.93) rectangle (79,-23.83);
\fill[fill={rgb,255:red,238;green,246;blue,252}] (68.6,-23.8) rectangle (71.23,-24.7);
\fill[fill={rgb,255:red,80;green,153;blue,214}] (71.2,-23.8) rectangle (76.43,-24.7);
\fill[fill={rgb,255:red,238;green,246;blue,252}] (76.4,-23.8) rectangle (79,-24.7);
\fill[fill={rgb,255:red,177;green,210;blue,237}] (68.6,-24.67) rectangle (79,-25.56);
\fill[fill={rgb,255:red,177;green,210;blue,237}] (68.6,-25.53) rectangle (79,-26.4);
\draw[black!45,line width=0.52pt] (68.6,5.2) rectangle ++(10.4,-10.4);
\draw[black!45,line width=0.52pt] (68.6,-16) rectangle ++(10.4,-10.4);
\draw[black!40,line width=0.6pt,dash pattern=on 1.8pt off 1.8pt] (-1,-6.4) -- (79.4,-6.4);
\node[align=center,text=black!85,inner sep=0,font=\footnotesize] at (2.2,-9.4) {What\\changes};
\node[align=center,text=black!85,inner sep=0,font=\footnotesize] at (16.1,-9.4) {one token per pixel,\\mirror padding};
\node[align=center,text=black!85,inner sep=0,font=\footnotesize] at (36.4,-9.4) {144 tokens in\\\textit{P}\textsuperscript{2}\,=\,36 phases};
\node[align=center,text=black!85,inner sep=0,font=\footnotesize] at (55.1,-9.4) {attention per phase,\\recalibrated CLS};
\node[align=center,text=black!85,inner sep=0,font=\footnotesize] at (73.8,-9.4) {map at pixel\\resolution};
\end{tikzpicture}

%% file: cvpr5_revision/method_story.tex
Our goal is to enable pretrained ViTs to process images at native resolution efficiently, without modifying their learned weights. The main challenge is to extend patch-based processing to pixel-level resolution while avoiding the computational cost of full self-attention over the resulting dense token grid. 
We address this by adapting the patch embedding and attention computation, while reusing all pretrained weights. The overall approach is illustrated in~\cref{fig:method} and developed in three steps below.

\begin{figure}[!t]
\centering
\colheads{\columnwidth}{3}{Image&Heads 1 / 2 / 3&Heads 4 / 5 / 6}
\setlength{\tabcolsep}{0pt}\renewcommand{\arraystretch}{0}%
\begin{tabular}{@{}c@{\hspace{1pt}}c@{\hspace{1pt}}c@{}}
\includegraphics[width=.329\columnwidth]{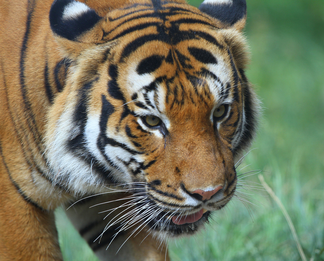} &
\includegraphics[width=.329\columnwidth]{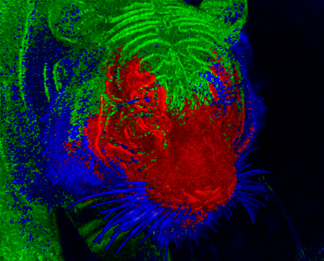} &
\includegraphics[width=.329\columnwidth]{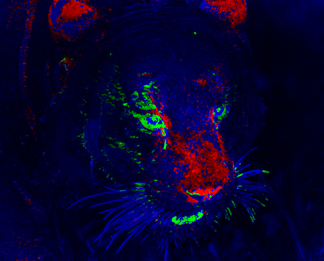}\\[1pt]
\includegraphics[width=.329\columnwidth]{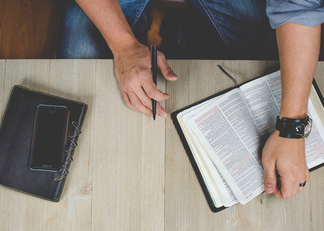} &
\includegraphics[width=.329\columnwidth]{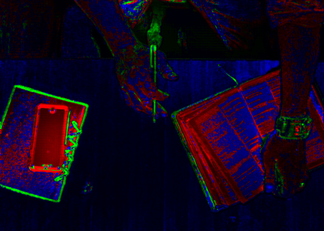} &
\includegraphics[width=.329\columnwidth]{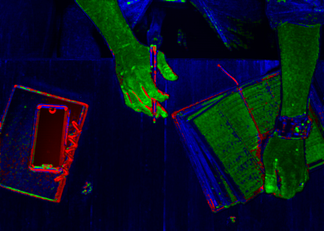}
\end{tabular}
\caption{
\textbf{Effect of SeeFine on attention maps from multiple heads.}
We visualize the final-layer [CLS] token attention maps from different heads of DINO ViT-S/16, processed by SeeFine at native image resolution.
Each color represents a different attention head.
SeeFine shows fine-grained spatial details and enables pixel-level analysis of individual attention heads.
Additional examples are provided in \cref{app:div2k}.
}
\label{fig:ship_intro}
\end{figure}

\paragraph{Step I: Patch phases.}
A standard ViT divides an image into non-overlapping patches using a patch embedding kernel of size $P\times P$ and stride $P$.
By applying the same pretrained kernel with stride one, we can produce a dense grid of overlapping patch tokens.
The resulting grid can be decomposed into $P^2$ separate grids, which we call \emph{patch phases}.
Each phase corresponds to a spatially shifted version of the original patch grid, with neighboring patch centers separated by $P$ pixels. Formally, for an image of size $H\times W$, the dense grid contains $N=HW$ tokens, which we partition into indexed phases:
\begin{equation}
G_{ab} = \{(i,j) : i \bmod P = a,\ j \bmod P = b\}
\label{eq:phase}
\end{equation}
where $ 0 \leq a,b < P$.
Since each phase $G_{ab}$ preserves the original patch spacing, it can be processed independently by the frozen ViT using the same pretrained weights.
Interleaving the resulting patch representations produces a dense feature map with one token per pixel (see~\cref{fig:method}).
However, naive processing of independent phases produces separate class tokens and predictions, without any interaction between phases.
To addresses this limitation, we connect the phases within a single ViT pass.

\begin{figure}[t]
\centering
\input{cvpr5_revision/tikz_n4_phase_attention.tex}
\caption{\textbf{Phase-restricted attention.} Three of the $P^2$ phases are shown, each with its $2\times2$ tokens, in the colours of \cref{fig:method}. Patch queries of phase $G$ attend to the keys of their own phase and to the shared class token (CLS), whose key logit is shifted by $d(|G|)$; the class query attends to all keys, with self-key shift $d(N)$. Blocks between phases are never computed, so the cost falls from $N^2$ to about $N^2/P^2$.  }
\label{fig:phase_attention}\label{fig:independent_phases}
\end{figure}

\paragraph{Step II: Connecting phases.}
Processing all patch phases $G_{ab}$ with full self-attention would be computationally expensive due to its quadratic complexity.
To address this, SeeFine introduces an efficient attention pattern that connects the phases through a shared class token $\boldsymbol{cls}$.
Specifically, each patch token attends only to tokens within its own phase $G_{ab}$ and to $\boldsymbol{cls}$, while the $\boldsymbol{cls}$ token attends to itself and all patch tokens across phases (see \cref{fig:phase_attention}):
\begin{equation}
\begin{aligned}
\operatorname{Attn}_{G_{ab}}
&\coloneqq \operatorname{Attn}
\left(Q_{G_{ab}}, K_{[\boldsymbol{cls}|G_{ab}]}, V_{[\boldsymbol{cls}|G_{ab}]}\right)\\
\operatorname{Attn}_{\boldsymbol{cls}}
&\coloneqq \operatorname{Attn}
\left(Q_{\boldsymbol{cls}}, K_{[\boldsymbol{cls}|\mathrm{all}]}, V_{[\boldsymbol{cls}|\mathrm{all}]}\right)
\end{aligned}
\end{equation}
Here, $Q$, $K$, and $V$ denote the query, key, and value projections, and $\mathrm{all}$ denotes patch tokens from all phases.
This avoids full attention over the dense token grid while allowing information to flow between phases through the shared class token.
As a result, SeeFine processes all phases in a single forward pass and produces a single prediction using the pretrained classifier.

\paragraph{Step III: Calibrating class-token attention.}
The phase-restricted attention introduced in Step II changes the number of keys available to each query, affecting the relative attention assigned to the shared class token $\boldsymbol{cls}$.
For equal attention logits, the class token receives an attention share of $1/(n+1)$, where $n$ is the number of spatial keys.
As the number of keys increases, this share decreases, potentially weakening the contribution of the class token.
To preserve its attention share, we introduce a correction to the class-token attention logit.
Following the proportional-attention principle~\cite{bolya2023token}, we account for the difference between the number of spatial keys $n$ and the reference token count $\Nref$ used during pretraining.
This gives the correction
\begin{equation}\label{eq:rule}
\dsh(n) = \log\frac{n}{\Nref},
\end{equation}
which is added to the class-token key logit $K_{\boldsymbol{cls}}$.
For equal logits, the resulting attention share becomes
\begin{equation}
s = \frac{e^{\dsh(n)}}{n+e^{\dsh(n)}}
= \frac{1}{\Nref+1},
\end{equation}
matching the reference configuration.
For patch queries, $n$ corresponds to the number of spatial keys within their phase $G_{ab}$, while the class query uses the total number of spatial keys.
The correction preserves the class-token attention share without modifying the pretrained weights. 
The actual attention distribution remains input-dependent.

\paragraph{Preserving pretrained representations.}
Although SeeFine modifies the execution of pretrained ViTs, it reuses all learned weights without modification.
We evaluate whether the resulting representations remain consistent with the original model through two complementary experiments.
First, at the original patch locations, SeeFine with recorded donor context reproduces DINOv3 features with a mean relative error of $1.4\cdot10^{-6}$ and input gradients within $0.1\%$ of the original model (\cref{app:anchor_agreement}).
Second, we evaluate spatial consistency using 100 ImageNet images under known affine transformations.
Matching features extracted with \SF{DINO} reduces the median point localization error from $4.33$ to $0.80$ pixels and increases the proportion of points localized within four pixels from $45.75\%$ to $96.12\%$ (\cref{app:spatial}).
Together, these results demonstrate numerical agreement at the original patch locations under recorded donor context and improved spatial precision at native resolution.

\paragraph{Implementation and computational efficiency.}
SeeFine implements dense patch embedding using a stride-one convolution with the original pretrained patch embedding kernel.
To efficiently compute phase-restricted attention, tokens are grouped by phase and folded into the batch dimension, enabling the use of fused attention kernels.
The shared class-token attention is computed separately, once per transformer block, while all pretrained projections, normalization layers, MLPs, and the classifier remain unchanged (\cref{app:algorithm}).
For $N$ dense patch tokens and patch size $P$, this reduces the attention complexity from $\mathcal{O}(N^2)$ to $\mathcal{O}(N^2/P^2 + N)$.
In practice, SeeFine processes a $375\times500$ image with DINO ViT-S/16 in \textit{0.16s}, compared to \textit{2.23s} for full attention over all pixel tokens (\cref{app:abl_components}).
It also processes a $2040\times1452$ image, containing approximately three million tokens, in \textit{6.3s} on a single GPU (\cref{app:runtime}).
Using a stride $S$ that divides $P$ provides a trade-off between spatial resolution and computational cost.

%% file: cvpr5_revision/tikz_n4_phase_attention.tex
\definecolor{nRed}{RGB}{198,40,40}%
\definecolor{nRedL}{RGB}{251,227,227}%
\definecolor{nNavy}{RGB}{31,47,143}%
\definecolor{nBlue}{RGB}{61,143,196}%
\definecolor{nBlueL}{RGB}{207,230,246}%
\definecolor{nOrange}{RGB}{217,130,74}%
\definecolor{nOrangeL}{RGB}{251,224,205}%
\definecolor{nGreen}{RGB}{90,166,78}%
\definecolor{nGreenL}{RGB}{223,240,219}%
\definecolor{nPink}{RGB}{199,107,181}%
\definecolor{nPinkL}{RGB}{243,220,238}%
\definecolor{nTok}{RGB}{222,222,222}%
\definecolor{nTokB}{RGB}{181,181,181}%
\definecolor{nSnow}{RGB}{91,127,166}%
\definecolor{nGrid}{RGB}{154,167,182}%
\providecommand{\phaseG}[2]{G_{#1,#2}}%
\begin{tikzpicture}[x=1.23mm,y=1.23mm,font=\footnotesize,line join=round]
\filldraw[fill=nRedL,draw=black!20,line width=.25pt] (14,0) rectangle ++(2.2,-2.2);
\filldraw[fill=nRedL,draw=black!20,line width=.25pt] (16.7,0) rectangle ++(2.2,-2.2);
\filldraw[fill=nRedL,draw=black!20,line width=.25pt] (19.4,0) rectangle ++(2.2,-2.2);
\filldraw[fill=nRedL,draw=black!20,line width=.25pt] (22.1,0) rectangle ++(2.2,-2.2);
\filldraw[fill=nRedL,draw=black!20,line width=.25pt] (24.8,0) rectangle ++(2.2,-2.2);
\filldraw[fill=nRedL,draw=black!20,line width=.25pt] (27.5,0) rectangle ++(2.2,-2.2);
\filldraw[fill=nRedL,draw=black!20,line width=.25pt] (30.2,0) rectangle ++(2.2,-2.2);
\filldraw[fill=nRedL,draw=black!20,line width=.25pt] (32.9,0) rectangle ++(2.2,-2.2);
\filldraw[fill=nRedL,draw=black!20,line width=.25pt] (35.6,0) rectangle ++(2.2,-2.2);
\filldraw[fill=nRedL,draw=black!20,line width=.25pt] (38.3,0) rectangle ++(2.2,-2.2);
\filldraw[fill=nRedL,draw=black!20,line width=.25pt] (41,0) rectangle ++(2.2,-2.2);
\filldraw[fill=nRedL,draw=black!20,line width=.25pt] (43.7,0) rectangle ++(2.2,-2.2);
\filldraw[fill=nRedL,draw=black!20,line width=.25pt] (46.4,0) rectangle ++(2.2,-2.2);
\filldraw[fill=nRedL,draw=black!20,line width=.25pt] (14,-2.7) rectangle ++(2.2,-2.2);
\filldraw[fill=nBlueL,draw=black!20,line width=.25pt] (16.7,-2.7) rectangle ++(2.2,-2.2);
\filldraw[fill=nBlueL,draw=black!20,line width=.25pt] (19.4,-2.7) rectangle ++(2.2,-2.2);
\filldraw[fill=nBlueL,draw=black!20,line width=.25pt] (22.1,-2.7) rectangle ++(2.2,-2.2);
\filldraw[fill=nBlueL,draw=black!20,line width=.25pt] (24.8,-2.7) rectangle ++(2.2,-2.2);
\draw[black!20,line width=.25pt] (27.5,-2.7) rectangle ++(2.2,-2.2);
\draw[black!25,line width=.25pt] (27.5,-3.8) -- ++(1.1,1.1) (28.6,-4.9) -- ++(1.1,1.1);
\draw[black!20,line width=.25pt] (30.2,-2.7) rectangle ++(2.2,-2.2);
\draw[black!25,line width=.25pt] (30.2,-3.8) -- ++(1.1,1.1) (31.3,-4.9) -- ++(1.1,1.1);
\draw[black!20,line width=.25pt] (32.9,-2.7) rectangle ++(2.2,-2.2);
\draw[black!25,line width=.25pt] (32.9,-3.8) -- ++(1.1,1.1) (34,-4.9) -- ++(1.1,1.1);
\draw[black!20,line width=.25pt] (35.6,-2.7) rectangle ++(2.2,-2.2);
\draw[black!25,line width=.25pt] (35.6,-3.8) -- ++(1.1,1.1) (36.7,-4.9) -- ++(1.1,1.1);
\draw[black!20,line width=.25pt] (38.3,-2.7) rectangle ++(2.2,-2.2);
\draw[black!25,line width=.25pt] (38.3,-3.8) -- ++(1.1,1.1) (39.4,-4.9) -- ++(1.1,1.1);
\draw[black!20,line width=.25pt] (41,-2.7) rectangle ++(2.2,-2.2);
\draw[black!25,line width=.25pt] (41,-3.8) -- ++(1.1,1.1) (42.1,-4.9) -- ++(1.1,1.1);
\draw[black!20,line width=.25pt] (43.7,-2.7) rectangle ++(2.2,-2.2);
\draw[black!25,line width=.25pt] (43.7,-3.8) -- ++(1.1,1.1) (44.8,-4.9) -- ++(1.1,1.1);
\draw[black!20,line width=.25pt] (46.4,-2.7) rectangle ++(2.2,-2.2);
\draw[black!25,line width=.25pt] (46.4,-3.8) -- ++(1.1,1.1) (47.5,-4.9) -- ++(1.1,1.1);
\filldraw[fill=nRedL,draw=black!20,line width=.25pt] (14,-5.4) rectangle ++(2.2,-2.2);
\filldraw[fill=nBlueL,draw=black!20,line width=.25pt] (16.7,-5.4) rectangle ++(2.2,-2.2);
\filldraw[fill=nBlueL,draw=black!20,line width=.25pt] (19.4,-5.4) rectangle ++(2.2,-2.2);
\filldraw[fill=nBlueL,draw=black!20,line width=.25pt] (22.1,-5.4) rectangle ++(2.2,-2.2);
\filldraw[fill=nBlueL,draw=black!20,line width=.25pt] (24.8,-5.4) rectangle ++(2.2,-2.2);
\draw[black!20,line width=.25pt] (27.5,-5.4) rectangle ++(2.2,-2.2);
\draw[black!25,line width=.25pt] (27.5,-6.5) -- ++(1.1,1.1) (28.6,-7.6) -- ++(1.1,1.1);
\draw[black!20,line width=.25pt] (30.2,-5.4) rectangle ++(2.2,-2.2);
\draw[black!25,line width=.25pt] (30.2,-6.5) -- ++(1.1,1.1) (31.3,-7.6) -- ++(1.1,1.1);
\draw[black!20,line width=.25pt] (32.9,-5.4) rectangle ++(2.2,-2.2);
\draw[black!25,line width=.25pt] (32.9,-6.5) -- ++(1.1,1.1) (34,-7.6) -- ++(1.1,1.1);
\draw[black!20,line width=.25pt] (35.6,-5.4) rectangle ++(2.2,-2.2);
\draw[black!25,line width=.25pt] (35.6,-6.5) -- ++(1.1,1.1) (36.7,-7.6) -- ++(1.1,1.1);
\draw[black!20,line width=.25pt] (38.3,-5.4) rectangle ++(2.2,-2.2);
\draw[black!25,line width=.25pt] (38.3,-6.5) -- ++(1.1,1.1) (39.4,-7.6) -- ++(1.1,1.1);
\draw[black!20,line width=.25pt] (41,-5.4) rectangle ++(2.2,-2.2);
\draw[black!25,line width=.25pt] (41,-6.5) -- ++(1.1,1.1) (42.1,-7.6) -- ++(1.1,1.1);
\draw[black!20,line width=.25pt] (43.7,-5.4) rectangle ++(2.2,-2.2);
\draw[black!25,line width=.25pt] (43.7,-6.5) -- ++(1.1,1.1) (44.8,-7.6) -- ++(1.1,1.1);
\draw[black!20,line width=.25pt] (46.4,-5.4) rectangle ++(2.2,-2.2);
\draw[black!25,line width=.25pt] (46.4,-6.5) -- ++(1.1,1.1) (47.5,-7.6) -- ++(1.1,1.1);
\filldraw[fill=nRedL,draw=black!20,line width=.25pt] (14,-8.1) rectangle ++(2.2,-2.2);
\filldraw[fill=nBlueL,draw=black!20,line width=.25pt] (16.7,-8.1) rectangle ++(2.2,-2.2);
\filldraw[fill=nBlueL,draw=black!20,line width=.25pt] (19.4,-8.1) rectangle ++(2.2,-2.2);
\filldraw[fill=nBlueL,draw=black!20,line width=.25pt] (22.1,-8.1) rectangle ++(2.2,-2.2);
\filldraw[fill=nBlueL,draw=black!20,line width=.25pt] (24.8,-8.1) rectangle ++(2.2,-2.2);
\draw[black!20,line width=.25pt] (27.5,-8.1) rectangle ++(2.2,-2.2);
\draw[black!25,line width=.25pt] (27.5,-9.2) -- ++(1.1,1.1) (28.6,-10.3) -- ++(1.1,1.1);
\draw[black!20,line width=.25pt] (30.2,-8.1) rectangle ++(2.2,-2.2);
\draw[black!25,line width=.25pt] (30.2,-9.2) -- ++(1.1,1.1) (31.3,-10.3) -- ++(1.1,1.1);
\draw[black!20,line width=.25pt] (32.9,-8.1) rectangle ++(2.2,-2.2);
\draw[black!25,line width=.25pt] (32.9,-9.2) -- ++(1.1,1.1) (34,-10.3) -- ++(1.1,1.1);
\draw[black!20,line width=.25pt] (35.6,-8.1) rectangle ++(2.2,-2.2);
\draw[black!25,line width=.25pt] (35.6,-9.2) -- ++(1.1,1.1) (36.7,-10.3) -- ++(1.1,1.1);
\draw[black!20,line width=.25pt] (38.3,-8.1) rectangle ++(2.2,-2.2);
\draw[black!25,line width=.25pt] (38.3,-9.2) -- ++(1.1,1.1) (39.4,-10.3) -- ++(1.1,1.1);
\draw[black!20,line width=.25pt] (41,-8.1) rectangle ++(2.2,-2.2);
\draw[black!25,line width=.25pt] (41,-9.2) -- ++(1.1,1.1) (42.1,-10.3) -- ++(1.1,1.1);
\draw[black!20,line width=.25pt] (43.7,-8.1) rectangle ++(2.2,-2.2);
\draw[black!25,line width=.25pt] (43.7,-9.2) -- ++(1.1,1.1) (44.8,-10.3) -- ++(1.1,1.1);
\draw[black!20,line width=.25pt] (46.4,-8.1) rectangle ++(2.2,-2.2);
\draw[black!25,line width=.25pt] (46.4,-9.2) -- ++(1.1,1.1) (47.5,-10.3) -- ++(1.1,1.1);
\filldraw[fill=nRedL,draw=black!20,line width=.25pt] (14,-10.8) rectangle ++(2.2,-2.2);
\filldraw[fill=nBlueL,draw=black!20,line width=.25pt] (16.7,-10.8) rectangle ++(2.2,-2.2);
\filldraw[fill=nBlueL,draw=black!20,line width=.25pt] (19.4,-10.8) rectangle ++(2.2,-2.2);
\filldraw[fill=nBlueL,draw=black!20,line width=.25pt] (22.1,-10.8) rectangle ++(2.2,-2.2);
\filldraw[fill=nBlueL,draw=black!20,line width=.25pt] (24.8,-10.8) rectangle ++(2.2,-2.2);
\draw[black!20,line width=.25pt] (27.5,-10.8) rectangle ++(2.2,-2.2);
\draw[black!25,line width=.25pt] (27.5,-11.9) -- ++(1.1,1.1) (28.6,-13) -- ++(1.1,1.1);
\draw[black!20,line width=.25pt] (30.2,-10.8) rectangle ++(2.2,-2.2);
\draw[black!25,line width=.25pt] (30.2,-11.9) -- ++(1.1,1.1) (31.3,-13) -- ++(1.1,1.1);
\draw[black!20,line width=.25pt] (32.9,-10.8) rectangle ++(2.2,-2.2);
\draw[black!25,line width=.25pt] (32.9,-11.9) -- ++(1.1,1.1) (34,-13) -- ++(1.1,1.1);
\draw[black!20,line width=.25pt] (35.6,-10.8) rectangle ++(2.2,-2.2);
\draw[black!25,line width=.25pt] (35.6,-11.9) -- ++(1.1,1.1) (36.7,-13) -- ++(1.1,1.1);
\draw[black!20,line width=.25pt] (38.3,-10.8) rectangle ++(2.2,-2.2);
\draw[black!25,line width=.25pt] (38.3,-11.9) -- ++(1.1,1.1) (39.4,-13) -- ++(1.1,1.1);
\draw[black!20,line width=.25pt] (41,-10.8) rectangle ++(2.2,-2.2);
\draw[black!25,line width=.25pt] (41,-11.9) -- ++(1.1,1.1) (42.1,-13) -- ++(1.1,1.1);
\draw[black!20,line width=.25pt] (43.7,-10.8) rectangle ++(2.2,-2.2);
\draw[black!25,line width=.25pt] (43.7,-11.9) -- ++(1.1,1.1) (44.8,-13) -- ++(1.1,1.1);
\draw[black!20,line width=.25pt] (46.4,-10.8) rectangle ++(2.2,-2.2);
\draw[black!25,line width=.25pt] (46.4,-11.9) -- ++(1.1,1.1) (47.5,-13) -- ++(1.1,1.1);
\filldraw[fill=nRedL,draw=black!20,line width=.25pt] (14,-13.5) rectangle ++(2.2,-2.2);
\draw[black!20,line width=.25pt] (16.7,-13.5) rectangle ++(2.2,-2.2);
\draw[black!25,line width=.25pt] (16.7,-14.6) -- ++(1.1,1.1) (17.8,-15.7) -- ++(1.1,1.1);
\draw[black!20,line width=.25pt] (19.4,-13.5) rectangle ++(2.2,-2.2);
\draw[black!25,line width=.25pt] (19.4,-14.6) -- ++(1.1,1.1) (20.5,-15.7) -- ++(1.1,1.1);
\draw[black!20,line width=.25pt] (22.1,-13.5) rectangle ++(2.2,-2.2);
\draw[black!25,line width=.25pt] (22.1,-14.6) -- ++(1.1,1.1) (23.2,-15.7) -- ++(1.1,1.1);
\draw[black!20,line width=.25pt] (24.8,-13.5) rectangle ++(2.2,-2.2);
\draw[black!25,line width=.25pt] (24.8,-14.6) -- ++(1.1,1.1) (25.9,-15.7) -- ++(1.1,1.1);
\filldraw[fill=nOrangeL,draw=black!20,line width=.25pt] (27.5,-13.5) rectangle ++(2.2,-2.2);
\filldraw[fill=nOrangeL,draw=black!20,line width=.25pt] (30.2,-13.5) rectangle ++(2.2,-2.2);
\filldraw[fill=nOrangeL,draw=black!20,line width=.25pt] (32.9,-13.5) rectangle ++(2.2,-2.2);
\filldraw[fill=nOrangeL,draw=black!20,line width=.25pt] (35.6,-13.5) rectangle ++(2.2,-2.2);
\draw[black!20,line width=.25pt] (38.3,-13.5) rectangle ++(2.2,-2.2);
\draw[black!25,line width=.25pt] (38.3,-14.6) -- ++(1.1,1.1) (39.4,-15.7) -- ++(1.1,1.1);
\draw[black!20,line width=.25pt] (41,-13.5) rectangle ++(2.2,-2.2);
\draw[black!25,line width=.25pt] (41,-14.6) -- ++(1.1,1.1) (42.1,-15.7) -- ++(1.1,1.1);
\draw[black!20,line width=.25pt] (43.7,-13.5) rectangle ++(2.2,-2.2);
\draw[black!25,line width=.25pt] (43.7,-14.6) -- ++(1.1,1.1) (44.8,-15.7) -- ++(1.1,1.1);
\draw[black!20,line width=.25pt] (46.4,-13.5) rectangle ++(2.2,-2.2);
\draw[black!25,line width=.25pt] (46.4,-14.6) -- ++(1.1,1.1) (47.5,-15.7) -- ++(1.1,1.1);
\filldraw[fill=nRedL,draw=black!20,line width=.25pt] (14,-16.2) rectangle ++(2.2,-2.2);
\draw[black!20,line width=.25pt] (16.7,-16.2) rectangle ++(2.2,-2.2);
\draw[black!25,line width=.25pt] (16.7,-17.3) -- ++(1.1,1.1) (17.8,-18.4) -- ++(1.1,1.1);
\draw[black!20,line width=.25pt] (19.4,-16.2) rectangle ++(2.2,-2.2);
\draw[black!25,line width=.25pt] (19.4,-17.3) -- ++(1.1,1.1) (20.5,-18.4) -- ++(1.1,1.1);
\draw[black!20,line width=.25pt] (22.1,-16.2) rectangle ++(2.2,-2.2);
\draw[black!25,line width=.25pt] (22.1,-17.3) -- ++(1.1,1.1) (23.2,-18.4) -- ++(1.1,1.1);
\draw[black!20,line width=.25pt] (24.8,-16.2) rectangle ++(2.2,-2.2);
\draw[black!25,line width=.25pt] (24.8,-17.3) -- ++(1.1,1.1) (25.9,-18.4) -- ++(1.1,1.1);
\filldraw[fill=nOrangeL,draw=black!20,line width=.25pt] (27.5,-16.2) rectangle ++(2.2,-2.2);
\filldraw[fill=nOrangeL,draw=black!20,line width=.25pt] (30.2,-16.2) rectangle ++(2.2,-2.2);
\filldraw[fill=nOrangeL,draw=black!20,line width=.25pt] (32.9,-16.2) rectangle ++(2.2,-2.2);
\filldraw[fill=nOrangeL,draw=black!20,line width=.25pt] (35.6,-16.2) rectangle ++(2.2,-2.2);
\draw[black!20,line width=.25pt] (38.3,-16.2) rectangle ++(2.2,-2.2);
\draw[black!25,line width=.25pt] (38.3,-17.3) -- ++(1.1,1.1) (39.4,-18.4) -- ++(1.1,1.1);
\draw[black!20,line width=.25pt] (41,-16.2) rectangle ++(2.2,-2.2);
\draw[black!25,line width=.25pt] (41,-17.3) -- ++(1.1,1.1) (42.1,-18.4) -- ++(1.1,1.1);
\draw[black!20,line width=.25pt] (43.7,-16.2) rectangle ++(2.2,-2.2);
\draw[black!25,line width=.25pt] (43.7,-17.3) -- ++(1.1,1.1) (44.8,-18.4) -- ++(1.1,1.1);
\draw[black!20,line width=.25pt] (46.4,-16.2) rectangle ++(2.2,-2.2);
\draw[black!25,line width=.25pt] (46.4,-17.3) -- ++(1.1,1.1) (47.5,-18.4) -- ++(1.1,1.1);
\filldraw[fill=nRedL,draw=black!20,line width=.25pt] (14,-18.9) rectangle ++(2.2,-2.2);
\draw[black!20,line width=.25pt] (16.7,-18.9) rectangle ++(2.2,-2.2);
\draw[black!25,line width=.25pt] (16.7,-20) -- ++(1.1,1.1) (17.8,-21.1) -- ++(1.1,1.1);
\draw[black!20,line width=.25pt] (19.4,-18.9) rectangle ++(2.2,-2.2);
\draw[black!25,line width=.25pt] (19.4,-20) -- ++(1.1,1.1) (20.5,-21.1) -- ++(1.1,1.1);
\draw[black!20,line width=.25pt] (22.1,-18.9) rectangle ++(2.2,-2.2);
\draw[black!25,line width=.25pt] (22.1,-20) -- ++(1.1,1.1) (23.2,-21.1) -- ++(1.1,1.1);
\draw[black!20,line width=.25pt] (24.8,-18.9) rectangle ++(2.2,-2.2);
\draw[black!25,line width=.25pt] (24.8,-20) -- ++(1.1,1.1) (25.9,-21.1) -- ++(1.1,1.1);
\filldraw[fill=nOrangeL,draw=black!20,line width=.25pt] (27.5,-18.9) rectangle ++(2.2,-2.2);
\filldraw[fill=nOrangeL,draw=black!20,line width=.25pt] (30.2,-18.9) rectangle ++(2.2,-2.2);
\filldraw[fill=nOrangeL,draw=black!20,line width=.25pt] (32.9,-18.9) rectangle ++(2.2,-2.2);
\filldraw[fill=nOrangeL,draw=black!20,line width=.25pt] (35.6,-18.9) rectangle ++(2.2,-2.2);
\draw[black!20,line width=.25pt] (38.3,-18.9) rectangle ++(2.2,-2.2);
\draw[black!25,line width=.25pt] (38.3,-20) -- ++(1.1,1.1) (39.4,-21.1) -- ++(1.1,1.1);
\draw[black!20,line width=.25pt] (41,-18.9) rectangle ++(2.2,-2.2);
\draw[black!25,line width=.25pt] (41,-20) -- ++(1.1,1.1) (42.1,-21.1) -- ++(1.1,1.1);
\draw[black!20,line width=.25pt] (43.7,-18.9) rectangle ++(2.2,-2.2);
\draw[black!25,line width=.25pt] (43.7,-20) -- ++(1.1,1.1) (44.8,-21.1) -- ++(1.1,1.1);
\draw[black!20,line width=.25pt] (46.4,-18.9) rectangle ++(2.2,-2.2);
\draw[black!25,line width=.25pt] (46.4,-20) -- ++(1.1,1.1) (47.5,-21.1) -- ++(1.1,1.1);
\filldraw[fill=nRedL,draw=black!20,line width=.25pt] (14,-21.6) rectangle ++(2.2,-2.2);
\draw[black!20,line width=.25pt] (16.7,-21.6) rectangle ++(2.2,-2.2);
\draw[black!25,line width=.25pt] (16.7,-22.7) -- ++(1.1,1.1) (17.8,-23.8) -- ++(1.1,1.1);
\draw[black!20,line width=.25pt] (19.4,-21.6) rectangle ++(2.2,-2.2);
\draw[black!25,line width=.25pt] (19.4,-22.7) -- ++(1.1,1.1) (20.5,-23.8) -- ++(1.1,1.1);
\draw[black!20,line width=.25pt] (22.1,-21.6) rectangle ++(2.2,-2.2);
\draw[black!25,line width=.25pt] (22.1,-22.7) -- ++(1.1,1.1) (23.2,-23.8) -- ++(1.1,1.1);
\draw[black!20,line width=.25pt] (24.8,-21.6) rectangle ++(2.2,-2.2);
\draw[black!25,line width=.25pt] (24.8,-22.7) -- ++(1.1,1.1) (25.9,-23.8) -- ++(1.1,1.1);
\filldraw[fill=nOrangeL,draw=black!20,line width=.25pt] (27.5,-21.6) rectangle ++(2.2,-2.2);
\filldraw[fill=nOrangeL,draw=black!20,line width=.25pt] (30.2,-21.6) rectangle ++(2.2,-2.2);
\filldraw[fill=nOrangeL,draw=black!20,line width=.25pt] (32.9,-21.6) rectangle ++(2.2,-2.2);
\filldraw[fill=nOrangeL,draw=black!20,line width=.25pt] (35.6,-21.6) rectangle ++(2.2,-2.2);
\draw[black!20,line width=.25pt] (38.3,-21.6) rectangle ++(2.2,-2.2);
\draw[black!25,line width=.25pt] (38.3,-22.7) -- ++(1.1,1.1) (39.4,-23.8) -- ++(1.1,1.1);
\draw[black!20,line width=.25pt] (41,-21.6) rectangle ++(2.2,-2.2);
\draw[black!25,line width=.25pt] (41,-22.7) -- ++(1.1,1.1) (42.1,-23.8) -- ++(1.1,1.1);
\draw[black!20,line width=.25pt] (43.7,-21.6) rectangle ++(2.2,-2.2);
\draw[black!25,line width=.25pt] (43.7,-22.7) -- ++(1.1,1.1) (44.8,-23.8) -- ++(1.1,1.1);
\draw[black!20,line width=.25pt] (46.4,-21.6) rectangle ++(2.2,-2.2);
\draw[black!25,line width=.25pt] (46.4,-22.7) -- ++(1.1,1.1) (47.5,-23.8) -- ++(1.1,1.1);
\filldraw[fill=nRedL,draw=black!20,line width=.25pt] (14,-24.3) rectangle ++(2.2,-2.2);
\draw[black!20,line width=.25pt] (16.7,-24.3) rectangle ++(2.2,-2.2);
\draw[black!25,line width=.25pt] (16.7,-25.4) -- ++(1.1,1.1) (17.8,-26.5) -- ++(1.1,1.1);
\draw[black!20,line width=.25pt] (19.4,-24.3) rectangle ++(2.2,-2.2);
\draw[black!25,line width=.25pt] (19.4,-25.4) -- ++(1.1,1.1) (20.5,-26.5) -- ++(1.1,1.1);
\draw[black!20,line width=.25pt] (22.1,-24.3) rectangle ++(2.2,-2.2);
\draw[black!25,line width=.25pt] (22.1,-25.4) -- ++(1.1,1.1) (23.2,-26.5) -- ++(1.1,1.1);
\draw[black!20,line width=.25pt] (24.8,-24.3) rectangle ++(2.2,-2.2);
\draw[black!25,line width=.25pt] (24.8,-25.4) -- ++(1.1,1.1) (25.9,-26.5) -- ++(1.1,1.1);
\draw[black!20,line width=.25pt] (27.5,-24.3) rectangle ++(2.2,-2.2);
\draw[black!25,line width=.25pt] (27.5,-25.4) -- ++(1.1,1.1) (28.6,-26.5) -- ++(1.1,1.1);
\draw[black!20,line width=.25pt] (30.2,-24.3) rectangle ++(2.2,-2.2);
\draw[black!25,line width=.25pt] (30.2,-25.4) -- ++(1.1,1.1) (31.3,-26.5) -- ++(1.1,1.1);
\draw[black!20,line width=.25pt] (32.9,-24.3) rectangle ++(2.2,-2.2);
\draw[black!25,line width=.25pt] (32.9,-25.4) -- ++(1.1,1.1) (34,-26.5) -- ++(1.1,1.1);
\draw[black!20,line width=.25pt] (35.6,-24.3) rectangle ++(2.2,-2.2);
\draw[black!25,line width=.25pt] (35.6,-25.4) -- ++(1.1,1.1) (36.7,-26.5) -- ++(1.1,1.1);
\filldraw[fill=nGreenL,draw=black!20,line width=.25pt] (38.3,-24.3) rectangle ++(2.2,-2.2);
\filldraw[fill=nGreenL,draw=black!20,line width=.25pt] (41,-24.3) rectangle ++(2.2,-2.2);
\filldraw[fill=nGreenL,draw=black!20,line width=.25pt] (43.7,-24.3) rectangle ++(2.2,-2.2);
\filldraw[fill=nGreenL,draw=black!20,line width=.25pt] (46.4,-24.3) rectangle ++(2.2,-2.2);
\filldraw[fill=nRedL,draw=black!20,line width=.25pt] (14,-27) rectangle ++(2.2,-2.2);
\draw[black!20,line width=.25pt] (16.7,-27) rectangle ++(2.2,-2.2);
\draw[black!25,line width=.25pt] (16.7,-28.1) -- ++(1.1,1.1) (17.8,-29.2) -- ++(1.1,1.1);
\draw[black!20,line width=.25pt] (19.4,-27) rectangle ++(2.2,-2.2);
\draw[black!25,line width=.25pt] (19.4,-28.1) -- ++(1.1,1.1) (20.5,-29.2) -- ++(1.1,1.1);
\draw[black!20,line width=.25pt] (22.1,-27) rectangle ++(2.2,-2.2);
\draw[black!25,line width=.25pt] (22.1,-28.1) -- ++(1.1,1.1) (23.2,-29.2) -- ++(1.1,1.1);
\draw[black!20,line width=.25pt] (24.8,-27) rectangle ++(2.2,-2.2);
\draw[black!25,line width=.25pt] (24.8,-28.1) -- ++(1.1,1.1) (25.9,-29.2) -- ++(1.1,1.1);
\draw[black!20,line width=.25pt] (27.5,-27) rectangle ++(2.2,-2.2);
\draw[black!25,line width=.25pt] (27.5,-28.1) -- ++(1.1,1.1) (28.6,-29.2) -- ++(1.1,1.1);
\draw[black!20,line width=.25pt] (30.2,-27) rectangle ++(2.2,-2.2);
\draw[black!25,line width=.25pt] (30.2,-28.1) -- ++(1.1,1.1) (31.3,-29.2) -- ++(1.1,1.1);
\draw[black!20,line width=.25pt] (32.9,-27) rectangle ++(2.2,-2.2);
\draw[black!25,line width=.25pt] (32.9,-28.1) -- ++(1.1,1.1) (34,-29.2) -- ++(1.1,1.1);
\draw[black!20,line width=.25pt] (35.6,-27) rectangle ++(2.2,-2.2);
\draw[black!25,line width=.25pt] (35.6,-28.1) -- ++(1.1,1.1) (36.7,-29.2) -- ++(1.1,1.1);
\filldraw[fill=nGreenL,draw=black!20,line width=.25pt] (38.3,-27) rectangle ++(2.2,-2.2);
\filldraw[fill=nGreenL,draw=black!20,line width=.25pt] (41,-27) rectangle ++(2.2,-2.2);
\filldraw[fill=nGreenL,draw=black!20,line width=.25pt] (43.7,-27) rectangle ++(2.2,-2.2);
\filldraw[fill=nGreenL,draw=black!20,line width=.25pt] (46.4,-27) rectangle ++(2.2,-2.2);
\filldraw[fill=nRedL,draw=black!20,line width=.25pt] (14,-29.7) rectangle ++(2.2,-2.2);
\draw[black!20,line width=.25pt] (16.7,-29.7) rectangle ++(2.2,-2.2);
\draw[black!25,line width=.25pt] (16.7,-30.8) -- ++(1.1,1.1) (17.8,-31.9) -- ++(1.1,1.1);
\draw[black!20,line width=.25pt] (19.4,-29.7) rectangle ++(2.2,-2.2);
\draw[black!25,line width=.25pt] (19.4,-30.8) -- ++(1.1,1.1) (20.5,-31.9) -- ++(1.1,1.1);
\draw[black!20,line width=.25pt] (22.1,-29.7) rectangle ++(2.2,-2.2);
\draw[black!25,line width=.25pt] (22.1,-30.8) -- ++(1.1,1.1) (23.2,-31.9) -- ++(1.1,1.1);
\draw[black!20,line width=.25pt] (24.8,-29.7) rectangle ++(2.2,-2.2);
\draw[black!25,line width=.25pt] (24.8,-30.8) -- ++(1.1,1.1) (25.9,-31.9) -- ++(1.1,1.1);
\draw[black!20,line width=.25pt] (27.5,-29.7) rectangle ++(2.2,-2.2);
\draw[black!25,line width=.25pt] (27.5,-30.8) -- ++(1.1,1.1) (28.6,-31.9) -- ++(1.1,1.1);
\draw[black!20,line width=.25pt] (30.2,-29.7) rectangle ++(2.2,-2.2);
\draw[black!25,line width=.25pt] (30.2,-30.8) -- ++(1.1,1.1) (31.3,-31.9) -- ++(1.1,1.1);
\draw[black!20,line width=.25pt] (32.9,-29.7) rectangle ++(2.2,-2.2);
\draw[black!25,line width=.25pt] (32.9,-30.8) -- ++(1.1,1.1) (34,-31.9) -- ++(1.1,1.1);
\draw[black!20,line width=.25pt] (35.6,-29.7) rectangle ++(2.2,-2.2);
\draw[black!25,line width=.25pt] (35.6,-30.8) -- ++(1.1,1.1) (36.7,-31.9) -- ++(1.1,1.1);
\filldraw[fill=nGreenL,draw=black!20,line width=.25pt] (38.3,-29.7) rectangle ++(2.2,-2.2);
\filldraw[fill=nGreenL,draw=black!20,line width=.25pt] (41,-29.7) rectangle ++(2.2,-2.2);
\filldraw[fill=nGreenL,draw=black!20,line width=.25pt] (43.7,-29.7) rectangle ++(2.2,-2.2);
\filldraw[fill=nGreenL,draw=black!20,line width=.25pt] (46.4,-29.7) rectangle ++(2.2,-2.2);
\filldraw[fill=nRedL,draw=black!20,line width=.25pt] (14,-32.4) rectangle ++(2.2,-2.2);
\draw[black!20,line width=.25pt] (16.7,-32.4) rectangle ++(2.2,-2.2);
\draw[black!25,line width=.25pt] (16.7,-33.5) -- ++(1.1,1.1) (17.8,-34.6) -- ++(1.1,1.1);
\draw[black!20,line width=.25pt] (19.4,-32.4) rectangle ++(2.2,-2.2);
\draw[black!25,line width=.25pt] (19.4,-33.5) -- ++(1.1,1.1) (20.5,-34.6) -- ++(1.1,1.1);
\draw[black!20,line width=.25pt] (22.1,-32.4) rectangle ++(2.2,-2.2);
\draw[black!25,line width=.25pt] (22.1,-33.5) -- ++(1.1,1.1) (23.2,-34.6) -- ++(1.1,1.1);
\draw[black!20,line width=.25pt] (24.8,-32.4) rectangle ++(2.2,-2.2);
\draw[black!25,line width=.25pt] (24.8,-33.5) -- ++(1.1,1.1) (25.9,-34.6) -- ++(1.1,1.1);
\draw[black!20,line width=.25pt] (27.5,-32.4) rectangle ++(2.2,-2.2);
\draw[black!25,line width=.25pt] (27.5,-33.5) -- ++(1.1,1.1) (28.6,-34.6) -- ++(1.1,1.1);
\draw[black!20,line width=.25pt] (30.2,-32.4) rectangle ++(2.2,-2.2);
\draw[black!25,line width=.25pt] (30.2,-33.5) -- ++(1.1,1.1) (31.3,-34.6) -- ++(1.1,1.1);
\draw[black!20,line width=.25pt] (32.9,-32.4) rectangle ++(2.2,-2.2);
\draw[black!25,line width=.25pt] (32.9,-33.5) -- ++(1.1,1.1) (34,-34.6) -- ++(1.1,1.1);
\draw[black!20,line width=.25pt] (35.6,-32.4) rectangle ++(2.2,-2.2);
\draw[black!25,line width=.25pt] (35.6,-33.5) -- ++(1.1,1.1) (36.7,-34.6) -- ++(1.1,1.1);
\filldraw[fill=nGreenL,draw=black!20,line width=.25pt] (38.3,-32.4) rectangle ++(2.2,-2.2);
\filldraw[fill=nGreenL,draw=black!20,line width=.25pt] (41,-32.4) rectangle ++(2.2,-2.2);
\filldraw[fill=nGreenL,draw=black!20,line width=.25pt] (43.7,-32.4) rectangle ++(2.2,-2.2);
\filldraw[fill=nGreenL,draw=black!20,line width=.25pt] (46.4,-32.4) rectangle ++(2.2,-2.2);
\fill[nRed] (14.2,0.8) rectangle ++(1.8,1.8);
\fill[nRed] (11.4,-0.2) rectangle ++(1.8,-1.8);
\fill[nBlue] (16.9,0.8) rectangle ++(1.8,1.8);
\fill[nBlue] (11.4,-2.9) rectangle ++(1.8,-1.8);
\fill[nBlue] (19.6,0.8) rectangle ++(1.8,1.8);
\fill[nBlue] (11.4,-5.6) rectangle ++(1.8,-1.8);
\fill[nBlue] (22.3,0.8) rectangle ++(1.8,1.8);
\fill[nBlue] (11.4,-8.3) rectangle ++(1.8,-1.8);
\fill[nBlue] (25,0.8) rectangle ++(1.8,1.8);
\fill[nBlue] (11.4,-11) rectangle ++(1.8,-1.8);
\fill[nOrange] (27.7,0.8) rectangle ++(1.8,1.8);
\fill[nOrange] (11.4,-13.7) rectangle ++(1.8,-1.8);
\fill[nOrange] (30.4,0.8) rectangle ++(1.8,1.8);
\fill[nOrange] (11.4,-16.4) rectangle ++(1.8,-1.8);
\fill[nOrange] (33.1,0.8) rectangle ++(1.8,1.8);
\fill[nOrange] (11.4,-19.1) rectangle ++(1.8,-1.8);
\fill[nOrange] (35.8,0.8) rectangle ++(1.8,1.8);
\fill[nOrange] (11.4,-21.8) rectangle ++(1.8,-1.8);
\fill[nGreen] (38.5,0.8) rectangle ++(1.8,1.8);
\fill[nGreen] (11.4,-24.5) rectangle ++(1.8,-1.8);
\fill[nGreen] (41.2,0.8) rectangle ++(1.8,1.8);
\fill[nGreen] (11.4,-27.2) rectangle ++(1.8,-1.8);
\fill[nGreen] (43.9,0.8) rectangle ++(1.8,1.8);
\fill[nGreen] (11.4,-29.9) rectangle ++(1.8,-1.8);
\fill[nGreen] (46.6,0.8) rectangle ++(1.8,1.8);
\fill[nGreen] (11.4,-32.6) rectangle ++(1.8,-1.8);
\node[text=nRed] at (15.1,4.3) {CLS};
\node[text=nRed,anchor=east] at (10.8,-1.1) {CLS};
\node[text=nBlue] at (21.85,4.3) {$\phaseG{0}{0}$};
\node[text=nBlue,anchor=east] at (10.8,-7.85) {$\phaseG{0}{0}$};
\node[text=nOrange] at (32.65,4.3) {$\phaseG{0}{1}$};
\node[text=nOrange,anchor=east] at (10.8,-18.65) {$\phaseG{0}{1}$};
\node[text=nGreen] at (43.45,4.3) {$\phaseG{0}{2}$};
\node[text=nGreen,anchor=east] at (10.8,-29.45) {$\phaseG{0}{2}$};
\draw[nBlue,line width=.8pt,rounded corners=.4pt] (16.4,-2.4) rectangle ++(10.9,-10.9);
\draw[nOrange,line width=.8pt,rounded corners=.4pt] (27.2,-13.2) rectangle ++(10.9,-10.9);
\draw[nGreen,line width=.8pt,rounded corners=.4pt] (38,-24) rectangle ++(10.9,-10.9);
\draw[nRed,line width=.7pt,rounded corners=.4pt] (13.7,0.3) rectangle ++(35.2,-2.8) (13.7,0.3) rectangle ++(2.8,-35.2);
\node[text=nRed] at (15.1,7.2) {$+d$};
\node[text=black!80] at (31.55,7.4) {keys};
\node[rotate=90,text=black!80] at (3.6,-17.55) {queries};
\node[text=black!70,font=\small] at (51.3,-17.3) {$\times$};
\draw[nRed,line width=.6pt] (53.7,-0.15) rectangle ++(1.9,-1.9);
\filldraw[fill=nRedL,draw=nRed,line width=.6pt] (61.7,-0.15) rectangle ++(1.9,-1.9);
\draw[nBlue,line width=.6pt] (53.7,-2.85) rectangle ++(1.9,-1.9);
\filldraw[fill=nBlueL,draw=nBlue,line width=.6pt] (61.7,-2.85) rectangle ++(1.9,-1.9);
\draw[nBlue,line width=.6pt] (53.7,-5.55) rectangle ++(1.9,-1.9);
\filldraw[fill=nBlueL,draw=nBlue,line width=.6pt] (61.7,-5.55) rectangle ++(1.9,-1.9);
\draw[nBlue,line width=.6pt] (53.7,-8.25) rectangle ++(1.9,-1.9);
\filldraw[fill=nBlueL,draw=nBlue,line width=.6pt] (61.7,-8.25) rectangle ++(1.9,-1.9);
\draw[nBlue,line width=.6pt] (53.7,-10.95) rectangle ++(1.9,-1.9);
\filldraw[fill=nBlueL,draw=nBlue,line width=.6pt] (61.7,-10.95) rectangle ++(1.9,-1.9);
\draw[nOrange,line width=.6pt] (53.7,-13.65) rectangle ++(1.9,-1.9);
\filldraw[fill=nOrangeL,draw=nOrange,line width=.6pt] (61.7,-13.65) rectangle ++(1.9,-1.9);
\draw[nOrange,line width=.6pt] (53.7,-16.35) rectangle ++(1.9,-1.9);
\filldraw[fill=nOrangeL,draw=nOrange,line width=.6pt] (61.7,-16.35) rectangle ++(1.9,-1.9);
\draw[nOrange,line width=.6pt] (53.7,-19.05) rectangle ++(1.9,-1.9);
\filldraw[fill=nOrangeL,draw=nOrange,line width=.6pt] (61.7,-19.05) rectangle ++(1.9,-1.9);
\draw[nOrange,line width=.6pt] (53.7,-21.75) rectangle ++(1.9,-1.9);
\filldraw[fill=nOrangeL,draw=nOrange,line width=.6pt] (61.7,-21.75) rectangle ++(1.9,-1.9);
\draw[nGreen,line width=.6pt] (53.7,-24.45) rectangle ++(1.9,-1.9);
\filldraw[fill=nGreenL,draw=nGreen,line width=.6pt] (61.7,-24.45) rectangle ++(1.9,-1.9);
\draw[nGreen,line width=.6pt] (53.7,-27.15) rectangle ++(1.9,-1.9);
\filldraw[fill=nGreenL,draw=nGreen,line width=.6pt] (61.7,-27.15) rectangle ++(1.9,-1.9);
\draw[nGreen,line width=.6pt] (53.7,-29.85) rectangle ++(1.9,-1.9);
\filldraw[fill=nGreenL,draw=nGreen,line width=.6pt] (61.7,-29.85) rectangle ++(1.9,-1.9);
\draw[nGreen,line width=.6pt] (53.7,-32.55) rectangle ++(1.9,-1.9);
\filldraw[fill=nGreenL,draw=nGreen,line width=.6pt] (61.7,-32.55) rectangle ++(1.9,-1.9);
\node[text=black!80,anchor=base] at (54.7,3.52) {values};
\node[text=black!80,anchor=base] at (62.7,3.52) {outputs};
\draw[-{Stealth[length=1.5mm,width=1.25mm]},line width=.8pt] (56.4,-17.3) -- (61,-17.3);
\node[anchor=north west,align=left,inner sep=0,text=black!80] at (11.4,-36.7) {\textcolor{nRed}{$+d$}: class-key shift, $d(|G|)$ for patch queries,\\$d(N)$ for the class query};
\draw[black!20,line width=.25pt] (11.4,-44.1) rectangle ++(2.2,-2.2);
\draw[black!25,line width=.25pt] (11.4,-45.2) -- ++(1.1,1.1) (12.5,-46.3) -- ++(1.1,1.1);
\node[anchor=west,text=black!80,inner sep=0] at (14.4,-45.2) {masked: never computed};
\end{tikzpicture}

%% file: cvpr5_revision/components_main.tex
\begin{table}[t]\centering\footnotesize
\caption{\textbf{Dense sampling and integration have distinct roles.} Frozen DINO ViT-S/16, full test splits. CUB: image-macro visible-landmark recall (\%) within the top $10\%$ of mean CLS attention on identical $224\times224$ crops. SPair: pair-mean PCK@0.05; this table interpolates grid features to the pixel lattice before cosine matching, whereas \cref{tab:spair_main} interpolates the cosine maps of grid readouts, so the scores of the two tables are not directly comparable. Independent and shared phases give similar local-task gains.}\label{tab:components_main}
\setlength{\tabcolsep}{4pt}
\begin{tabular}{@{}lrr@{}}\toprule
Execution & CUB & SPair\\\midrule
Ordinary donor & 63.49 & 17.71\\
Independent phases, calibrated & 70.80 & 18.62\\
Shared CLS, uncalibrated & 70.45 & 19.38\\
Shared CLS, calibrated & 70.77 & 18.63\\
\bottomrule\end{tabular}\end{table}

%% file: cvpr5_revision/appendix_v5.tex
\maketitlesupplementary
\noindent The supplement follows the experiments of \cref{sec:experiments}: implementation and preservation checks (\cref{app:algorithm}), classification (\cref{sec:acc}), text-guided segmentation (\cref{app:talk2dinov3}), semantic correspondence (\cref{app:spair}), CUB landmarks (\cref{app:cubparts}), medical images (\cref{app:medical}) and depth (\cref{app:downstream}), followed by qualitative results, runtime and related work. All backbones stay frozen; \cref{tab:appendix_protocol_index_v4} lists the protocol of every study.

\paragraph{Training-free baselines.}
\emph{Nearest} and \emph{Bilinear} interpolate the donor grid. \emph{Strided} slides the patch embedding at stride 6 for $P=14$, following JAFAR~\cite{couairon2025jafar}, and at stride 7 for $P=16$ (a similar ratio), with resampled positions, full attention and no correction. \emph{Deep ViT Features (DVF)}~\cite{amir2022deep} use overlapping patches at stride 4. We run the released DVF extractor on compatible checkpoints and use Strided for DINOv2 and DINOv3 (\cref{app:dvf_scope}). Our last-block SPair comparison applies the checkpoint's final LayerNorm to every readout, including DVF. \Cref{app:abl_descriptor} also applies DVF's original descriptor (block-9 keys with logarithmic binning) to all readouts. \emph{HR-Dv2}~\cite{docherty2024upsampling} combines 68 shifted and flipped passes at stride 4 and returns no class token, so it is omitted from tasks that read one. In the tables, --- marks a combination that the method or its released code does not support.
\input{cvpr5_revision/appendix_protocol_index.tex}

\section{Method and implementation}\label{app:algorithm}\label{app:method}
SeeFine changes how an existing ViT is evaluated: it slides the patch embedding densely, partitions tokens by phase and calibrates the shared class key. Patch weights, attention projections, normalization, residual MLPs and the prediction head are reused. The resulting feature map has one entry per input pixel.

\subsection{Image geometry and positions}\label{sec:dense}
For patch side $P$, reflect RGB values before normalization by $a=\lfloor P/2\rfloor$ pixels at top/left and $b=P-1-a$ at bottom/right, including corners. Valid stride-one convolution then yields exactly $H\times W$ tokens. For $P=16$, eight pixels of padding before the image and seven after it place each dense output at its source pixel. Reflection follows PyTorch \texttt{reflect} and excludes the edge sample itself. Very small images use repeated reflection to reach the required width.

Absolute position embeddings use PyTorch bicubic interpolation with \texttt{align\_corners=False}. For a source positional grid of height $G$ and an output lattice of height $H$, output index $i$ samples source coordinate $u=(i+\tfrac12)G/H-\tfrac12$. Boundary handling follows the interpolation operator, and the horizontal axis is treated in the same way. The sampled embeddings can therefore differ slightly from the donor's at its patch centres. The anchor-position control in \cref{app:anchor_agreement} separates this effect from phase coupling. DINOv3 RoPE uses the donor's periods over either the source-image extent or the donor's patch support (\cref{sec:practical,app:voc_context_v4}). RGB reflection supplies embedding context without enlarging the output domain.

\paragraph{Border padding.}\label{sec:padding-ablation}
For a border-centred patch, some RGB values lie outside the image. We compare mirror reflection with nearest-pixel replication on CUB landmarks (\cref{tab:padding-ablation}). We evaluate both padding rules on all $5{,}794$ CUB test images, using the same frozen DINO ViT-S/16, $224\times224$ crop, position sampling, analytic corrections and landmark metric. The patch-query correction is exactly zero for both rules because each phase contains $196$ real patch keys. We report recall over all landmarks and separately for landmarks in the outer eight-pixel band.

\begin{table}[t]\centering\footnotesize
\caption{\textbf{A controlled boundary ablation.} Image-macro landmark recall (\%) on CUB. Both rows use the same frozen \SF{DINO} model and differ only in how the RGB image is extended beyond its borders.}\label{tab:padding-ablation}
\setlength{\tabcolsep}{3pt}
\begin{tabular}{@{}lrr@{}}\toprule
RGB extension & All landmarks & Outer 8-pixel band\\\midrule
Nearest-pixel replication & 70.77 & 69.51\\
Mirror reflection (default) & 70.77 & 72.17\\\bottomrule
\end{tabular}
\end{table}

Mirror improves outer-band recall by $2.66$ points (paired 95\% image-bootstrap interval $[0.89,4.44]$), while the interior difference is $-0.10$ points ($[-0.18,-0.01]$). Aggregate recall is unchanged at the shown precision. This improvement at the border motivates reflection as the default.

\subsection{One attention rule, two key counts}\label{sec:phase}\label{sec:constants}
At stride one, define $G_{ab}=\{(i,j):i\bmod P=a,\ j\bmod P=b\}$. A patch query reads its own phase and the shared class token. The class query reads all $N=HW$ real patches and itself. Before softmax, add
\begin{equation}
d_{\mathrm{pix},G}=\log\frac{|G|}{\Nref},\qquad
d_{\mathrm{cls}}=\log\frac{HW}{\Nref}
\end{equation}
to the class-key logit of a patch query and to the class-query self logit, respectively. $\Nref$ is the donor reference patch count, excluding global tokens. It follows the donor configuration: the checkpoint-native DINOv2 ViT-S/14 positional table is $37\times37$ ($\Nref=1369$), whereas a resampled $224$-pixel configuration has $16\times16$ positions ($\Nref=256$). The native-resolution SPair evaluation preserves the former table on both sides before interpolation to each image extent. For each patch query, the numerator is the number of keys in its own phase, which can vary across a rectangular image. For the class query, it is the full patch count. Dummy keys used for batching receive $-\infty$ logits and are excluded from both counts.

\begin{algorithm}[H]
\caption{SeeFine for a frozen single-CLS ViT}\label{alg:seefine}
\footnotesize
\begin{algorithmic}[1]
\Require Frozen model $M$, RGB image $x\in\mathbb R^{H\times W\times3}$, patch side $P$, reference count $\Nref$
\Ensure Prediction and pixel-aligned features
\State $a\gets\lfloor P/2\rfloor$; $b\gets P-1-a$
\State $x^+\gets\operatorname{ReflectRGB}(x; a,b)$; normalize $x^+$
\State $Z\gets\operatorname{PatchEmbed}(x^+;\mathrm{stride}=1)$
\State Add source-aligned donor positions to $Z$; initialize donor CLS state $c$
\State Precompute phase indices, real-key masks, $\log(|G|/\Nref)$ and $\log(HW/\Nref)$
\For{each frozen block $\ell$}
  \State Compute normalized inputs and $Q,K,V$ once for $[c;Z]$
  \State In parallel, $A_G\gets\operatorname{Attn}(Q_G,[K_c;K_G],[V_c;V_G])$ for every phase
  \State In the preceding softmax, shift only $K_c$ by $\log(|G|/\Nref)$; mask dummy keys
  \State $A_c\gets\operatorname{Attn}(Q_c,[K_c;K_Z],[V_c;V_Z])$, shifting its self key by $\log(HW/\Nref)$
  \State Apply unchanged output projections and both attention residuals simultaneously
  \State Apply the block's unchanged pre-norm MLP residual to $c$ and $Z$
\EndFor
\State Apply final normalization and the unchanged prediction head; return prediction and $Z$
\end{algorithmic}
\end{algorithm}

\input{cvpr5_revision/reference_policy_v5.tex}

\subsection{Practical implementation}\label{sec:practical}
The phase axis can be folded into the batch for fused scaled-dot-product attention. Counts, masks and log shifts depend only on image geometry and are computed once, then reused across layers. Within each block, every phase and the class query read the states entering that block, and the updates are applied together, so the result does not depend on the order of the phases. Dummy query outputs are discarded when the spatial grid is restored.

At a $224\times224$ input with $P=16$, each of the $256$ phases has $196$ patches, so $d_{\mathrm{pix},G}=0$, while $d_{\mathrm{cls}}=\log256$. Both shifts can be nonzero at native resolution. \Cref{tab:components_main} compares their effects on CUB and SPair. The donor-recovery check uses the original stride, the original valid preprocessing and a single phase.

\paragraph{Registers and rotary positions.} DINOv3 keeps its original CLS and four registers. In recurrent SeeFine, each global query attends to every valid spatial key and all five global keys. Each patch query attends to the valid keys in its own phase and to the same five global keys. Each global key receives $\log(N_{\rm keys}/\Nref)$ for a global query and $\log(|J_G|/\Nref)$ for a patch query, where $J_G$ contains the phase's valid keys. All attention updates read the incoming block states. The recorded-context ablation supplies the donor's incoming global states instead and requires an additional donor pass (\cref{alg:donor_context_v4}). The current-input reference, rotary domain and complete-window key support are held fixed when comparing the two variants.

\paragraph{Geometry per model family.}\label{app:geometry_policy}
The reference count and geometry are fixed per model family before any task is scored. For a supported absolute-position checkpoint, the spatial size of its configured positional table sets $\Nref$, positions are interpolated over the source-image extent, and all reflected-window keys are kept. For DINOv3, we use the valid donor count of the given image, donor-support RoPE and complete RGB windows as keys. Border-centred windows, which contain reflected pixels, remain queries but are not used as keys. The original CLS and registers are retained and updated recurrently. Each query uses its own real-key count in the nominal-mass rule. Fixed-reference DINOv3 and recorded globals are the controls listed in \cref{tab:reference_policy_scope_v4}.

\subsection{Recorded context and agreement with the donor}\label{app:recorded_context}
\paragraph{DINOv3 global context: integration and preservation.}
To separate integration from donor preservation, we ablate how the DINOv3 global states are supplied (\cref{app:talk2dinov3}). Recurrent globals read the dense tokens and return their integrated context to subsequent blocks. The recorded-context control instead obtains the incoming globals from an ordinary donor pass and supplies them read-only to dense patch queries. Both variants set $\Nref$ to the donor's patch count on the current image, normalize RoPE coordinates to the valid donor support and use complete RGB patch windows as keys. The query lattice still includes reflected border-centred windows. The recorded control recovers donor features and tested local-feature input gradients at donor anchors, while recurrent feedback changes the integrated function (\cref{tab:global_feedback_agreement_v4}). Their full-VOC mIoUs are $66.60$ and $66.56$, respectively (\cref{tab:voc_current_globals_v4}).
\input{cvpr4_revision_20261027/donor_context_algorithm.tex}
\input{cvpr5_revision/donor_anchor_contract_v5.tex}
\input{cvpr5_revision/anchor_agreement_v5.tex}
\input{cvpr5_revision/component_v5.tex}
\FloatBarrier
\clearpage

\section{Image classification}\label{sec:acc}\label{sec:accdonors}
\Cref{tab:donors} evaluates the same $224\times224$ crop and frozen classification head in each pair. All rows use all $50{,}000$ validation images. Gains range from $0.34$ to $1.13$ percentage points on the six classifiers. Strided feeds its class token to the same classification head. It improves on the donor only for ViT-S and DeiT-B and scores below SeeFine for all six classifiers. DVF reads the last-block class token, as SeeFine does, so \cref{tab:donors} involves no descriptor difference (\cref{app:abl_descriptor}).

DINO has no supervised ImageNet classification head. Its separate 20-nearest-neighbour row uses normalized CLS features, 50 training images per class and all $50{,}000$ validation images. The bank and queries use the corresponding donor or SeeFine readout consistently. In this separate nearest-neighbour evaluation, accuracy rises from $66.41\%$ with donor features to $66.49\%$ with SeeFine.

\paragraph{Component ablation.}\label{app:abl_components}
SeeFine combines a stride-one embedding with mirror padding, phase attention and mass calibration. \Cref{tab:abl_components_imagenet} removes components on ImageNet-1k, the main quantitative result. It starts from Strided at stride one, adds mirror padding and calibration under full attention, and then switches to phase attention. \Cref{app:components_v4} tests sharing and calibration separately on DINO.

Calibration matters most. Without it, full attention at stride one falls $1.0$ to $9.8$ points below the donor on all six classifiers, and phase attention falls $0.53$ to $3.33$ points below SeeFine. With the embedding and calibration held fixed, phase attention cuts runtime about 14-fold, from $2.23$\,s to $0.156$\,s on one $375\times500$ image (DINO ViT-S/16, H100; \cref{tab:phase_cost}). SeeFine attains the highest accuracy on all six classifiers.

\begin{table}[H]\centering\scriptsize
\caption{\textbf{Component ablation on ImageNet-1k.} Top-1 (\%), full validation set, unchanged head. III-S, III-B: DeiT-III Small and Base. For the DINO 20-NN, SeeFine reaches $66.49$ and phase attention without calibration $62.31$.}\label{tab:abl_components_imagenet}
\setlength{\tabcolsep}{2.5pt}
\begin{tabular}{@{}lrrrrrr@{}}\toprule
Variant & DeiT-T & DeiT-S & DeiT-B & ViT-S & III-S & III-B\\\midrule
Donor & 72.20 & 79.87 & 81.97 & 81.42 & 81.36 & 83.79\\
Strided, stride 1 & 69.06 & 70.61 & 81.01 & 78.91 & 73.01 & 79.77\\
Stride 1 + mirror, full attention & 69.29 & 70.03 & 80.97 & 78.63 & 73.18 & 79.77\\
+ calibration, full attention & 71.57 & 79.97 & 82.15 & 81.72 & 81.59 & 83.80\\
Phase attention, no calibration & 70.97 & 77.27 & 81.79 & 81.57 & 79.78 & 82.59\\
SeeFine & \textbf{73.33} & \textbf{80.60} & \textbf{82.32} & \textbf{82.31} & \textbf{82.24} & \textbf{84.13}\\
\bottomrule\end{tabular}
\end{table}

\begin{table}[H]\centering\scriptsize\par{
\caption{\textbf{Phase attention reduces runtime.} Median runtime over five synchronized runs after one warm-up, using frozen DINO ViT-S/16 on one H100 with a single $375\times500$ image (batch size 1). The stride-one comparison holds the embedding and calibration fixed to isolate the effect of the attention pattern.}\label{tab:phase_cost}
\begin{tabular}{@{}lr@{}}\toprule
Readout & Time (s)\\\midrule
Donor (patch grid) & 0.016\\
Stride 1, full attention + calibration & 2.230\\
SeeFine (phase attention + calibration) & 0.156\\
\bottomrule\end{tabular}
}\par\end{table}

\paragraph{Reference count.}\label{app:abl_nref}
The reference count $\Nref$ sets the calibration in \cref{sec:phase}. We change it by a factor of four in each direction, which shifts the class-token logit by $\mp\log 4$ (\cref{tab:abl_nref_imagenet}).

On ImageNet-1k the default $\Nref$ is highest on all six classifiers. Using $\Nref\times\frac14$ lowers accuracy by $0.36$ to $9.12$ points and $\Nref\times4$ by $0.35$ to $1.81$ points.

\begin{table}[H]\centering\footnotesize
\caption{\textbf{Reference-count sensitivity on ImageNet-1k.} Top-1 (\%), full validation set, unchanged head. III-S, III-B: DeiT-III Small and Base.}\label{tab:abl_nref_imagenet}
\setlength{\tabcolsep}{3pt}
\begin{tabular}{@{}lrrrrrr@{}}\toprule
Variant & DeiT-T & DeiT-S & DeiT-B & ViT-S & III-S & III-B\\\midrule
$\Nref\times\frac14$ & 64.21 & 78.64 & 81.96 & 80.01 & 80.32 & 83.16\\
$\Nref\times4$ & 71.52 & 79.11 & 81.97 & 81.32 & 81.06 & 83.37\\
SeeFine (default) & \textbf{73.33} & \textbf{80.60} & \textbf{82.32} & \textbf{82.31} & \textbf{82.24} & \textbf{84.13}\\
\bottomrule\end{tabular}
\end{table}
\FloatBarrier

\section{Text-guided segmentation}\label{app:talk2dinov3}
Talk2DINO~\cite{giambellutti2025talk2dino} maps frozen CLIP text features into a self-supervised visual encoder's patch-feature space. We use the released Talk2DINOv3 ViT-L checkpoint to ask whether its pretrained text head benefits from native pixel-resolution features. The visual encoder, text encoder and text projector remain frozen; no segmentation head is trained.

\paragraph{Global tokens and geometry.}
The released visual encoder has 24 blocks, one CLS and four registers. Recurrent SeeFine (UP) updates these global tokens jointly with all dense phases. The recorded-context control (RP) instead takes the donor's incoming global states before each block. All readouts receive the same unresized photograph. The donor's stride-16 embedding omits incomplete trailing strips, while the dense embedding uses 8/7 RGB reflection and returns one feature per source pixel. UP and RP share $\Nref=\lfloor H/16\rfloor\lfloor W/16\rfloor$, donor-support RoPE and complete RGB patch windows as keys (\cref{tab:talk_policy_v4}). The main-paper figure uses UP. The qualitative figures at the end of this section use recorded context with a fixed reference of 784, source-image RoPE and reflected keys.
\input{cvpr5_revision/talk_policy_v5.tex}

\paragraph{One shared text and mask readout.}
The official templates for the 20 VOC classes pass through the frozen text encoder and projector. Normalised projected prompts are averaged and normalised again. Pixel scores are sigmoid-transformed cosine similarities. For background cleaning, post-RoPE CLS attention, conditioned on patch keys, pools the final visual features. Each head's text cosine similarity gives its softmax weight across heads. The weighted attention density is rescaled to the score range and blended with the raw score using weight $0.2$. Every feature readout uses this same mask rule.

\input{cvpr4_revision_20261027/cleaner_definition.tex}

\paragraph{Full held-out validation.}
The text ensemble, mask readout and threshold were fixed on 100 VOC2012 training images (seed 1310), disjoint from validation. Probabilities are mapped to the original label dimensions and void pixels are ignored. UP reaches $66.56\%$ mIoU, $2.89$ points above the geometrically aligned donor and $5.37$ above the donor with the released upsampling (paired 95\% image-bootstrap CI $[5.11,5.67]$, 2,000 resamples). All readouts process the full frame in one pass, whereas the released benchmark uses sliding windows. The training-free baselines are listed in \cref{tab:talk2dinov3_native}.  HR-Dv2 is omitted because it does not provide the CLS attention required by the background cleaner.

\input{cvpr5_revision/voc_context_v5.tex}

\begin{figure*}[p]
\centering
\includegraphics[width=\textwidth]{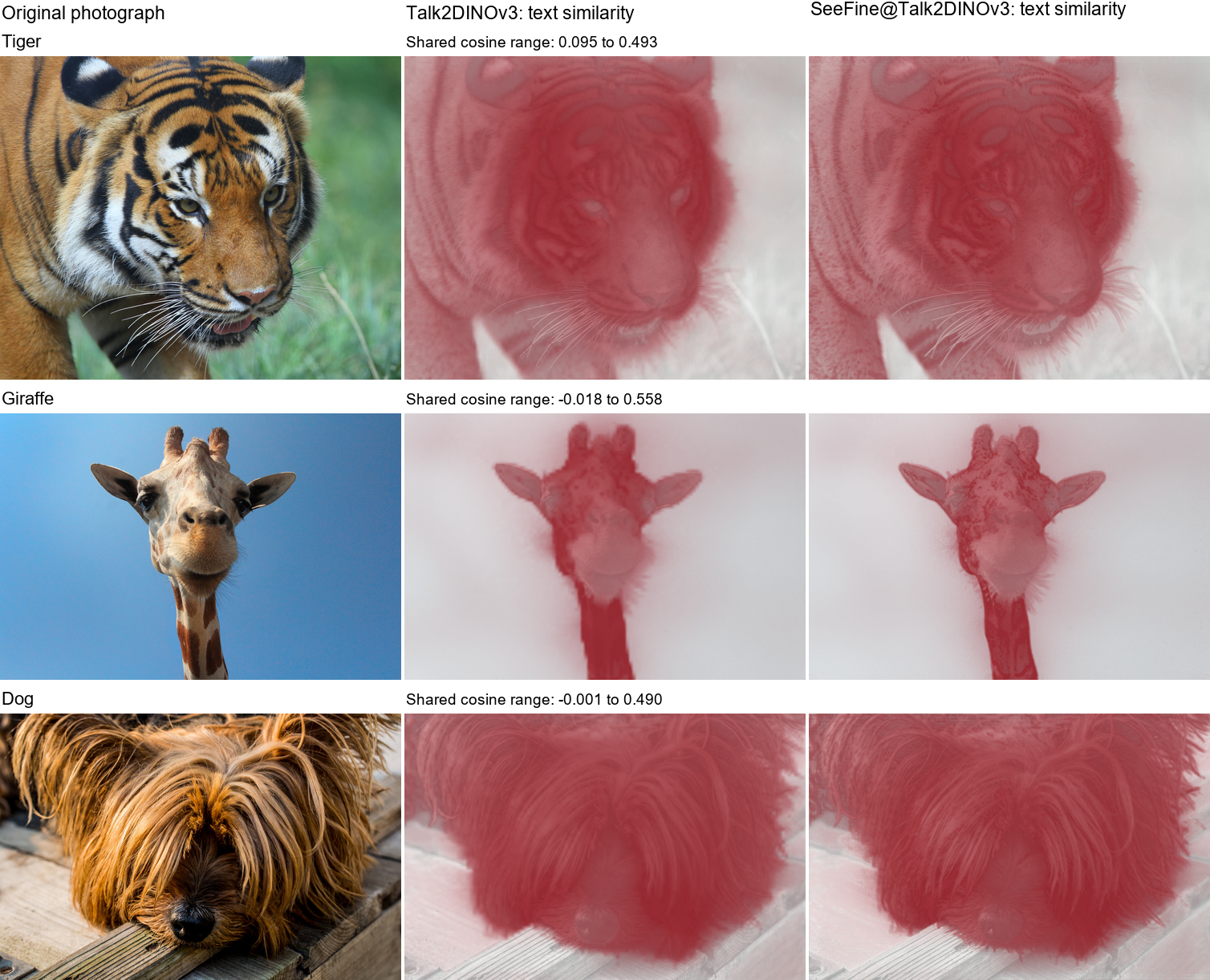}
\caption{\textbf{DIV2K responses to object text.} Frozen Talk2DINOv3 ViT-L (DINOv3 ViT-L/16 visual encoder), compared with \SF{Talk2DINOv3} using the same weights and recorded donor context. Tiger (0010), giraffe (0781) and dog (0596), using the prompts shown above their rows. Columns: original photograph, text--feature cosine similarity in Talk2DINOv3 and in \SF{Talk2DINOv3}. Larger cosine values give stronger dark-red overlays on a bright desaturated photograph. Each pair uses one shared display range: the pooled $2$nd--$98$th percentiles of both source-grid maps. The maps show continuous similarity.}
\label{fig:talk2dinov3_div2k_text_main}
\end{figure*}

\begin{figure*}[p]
\centering
\includegraphics[width=\textwidth]{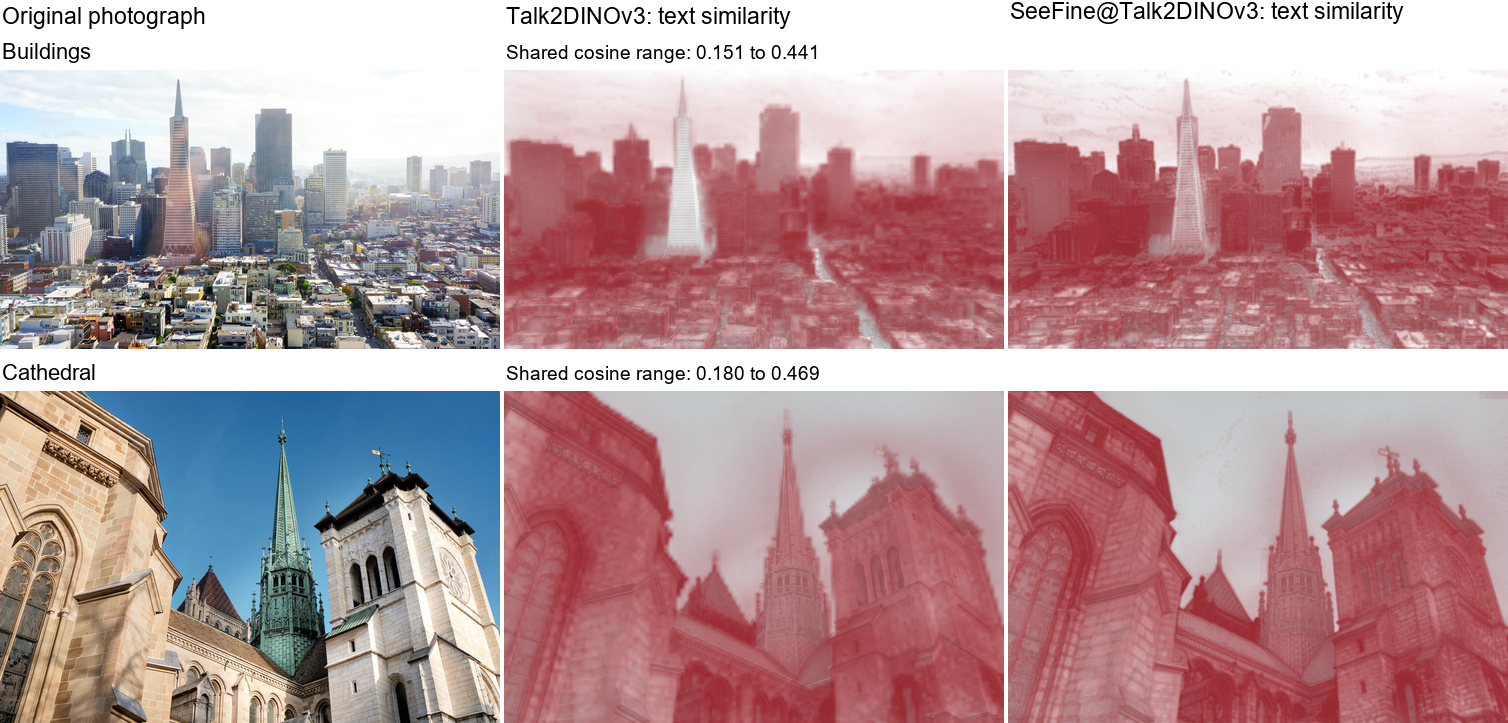}
\caption{\textbf{DIV2K architecture, queried with text.} Frozen Talk2DINOv3 ViT-L (DINOv3 ViT-L/16) versus \SF{Talk2DINOv3}. Rows: skyline 0045 (``buildings'') and cathedral 0756 (``cathedral''). Columns: source photograph, donor and SeeFine cosine responses. Each pair shares a colour scale; display as in \cref{fig:talk2dinov3_div2k_text_main}.}
\label{fig:talk2dinov3_div2k_text_architecture}
\end{figure*}

\begin{figure*}[p]
\centering
\includegraphics[width=\textwidth]{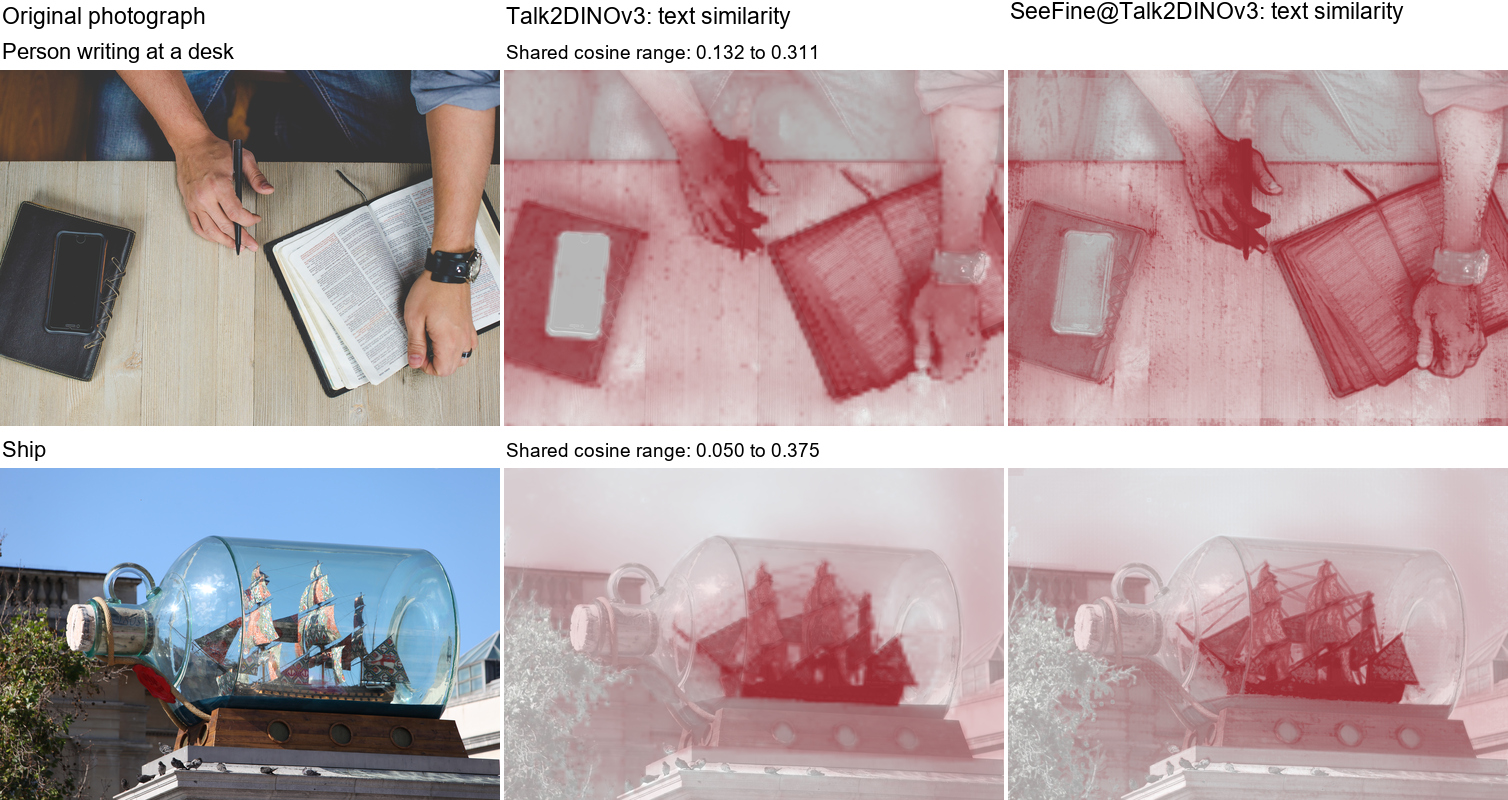}
\caption{\textbf{DIV2K objects, queried with phrases.} Same frozen Talk2DINOv3 ViT-L (DINOv3 ViT-L/16) versus \SF{Talk2DINOv3}. Rows: image 0678 (``person writing at a desk'') and 0737 (``ship''). Columns and paired colour scales follow \cref{fig:talk2dinov3_div2k_text_architecture}.}
\label{fig:talk2dinov3_div2k_text_objects}
\end{figure*}

\begin{figure*}[p]
\centering
\includegraphics[width=\textwidth]{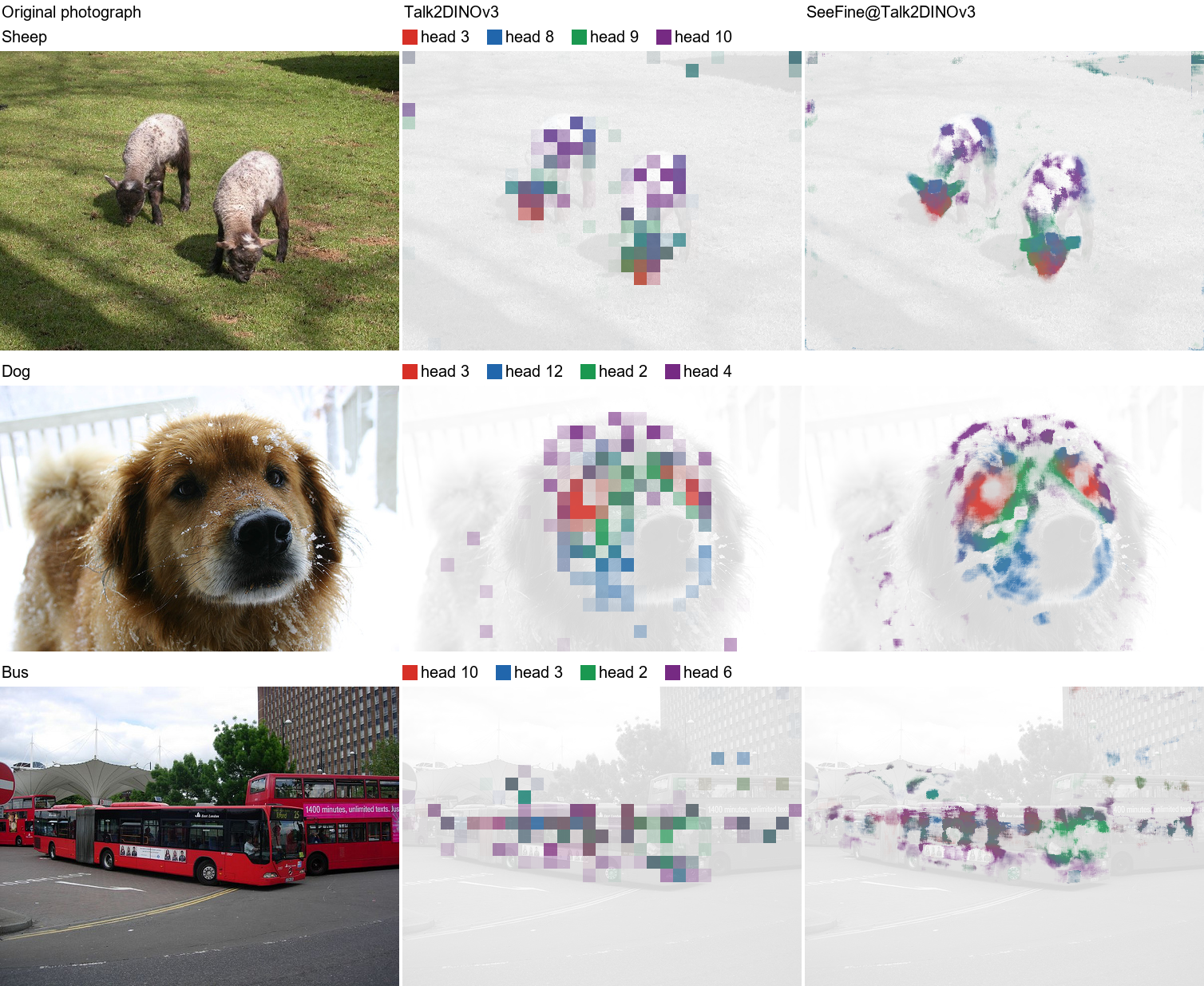}
\caption{\textbf{Complementary CLS heads for the text query.} Each row shows the photograph and the selected CLS heads of frozen Talk2DINOv3 and \SF{Talk2DINOv3} with recorded donor context. A colour identifies the same head in both panels of a row and is assigned separately for each example. The background blends $84\%$ white with $16\%$ image luminance. A head's display weight rises from zero at its $90$th attention percentile to one at its $99.5$th. Combined opacity is capped at $85\%$, and overlapping colours are mixed symmetrically. \Cref{app:talk2dinov3} describes head selection.}
\label{fig:talk2dinov3_fourheads}
\end{figure*}

\begin{figure*}[p]
\centering
\includegraphics[width=\textwidth,height=.66\textheight,keepaspectratio]{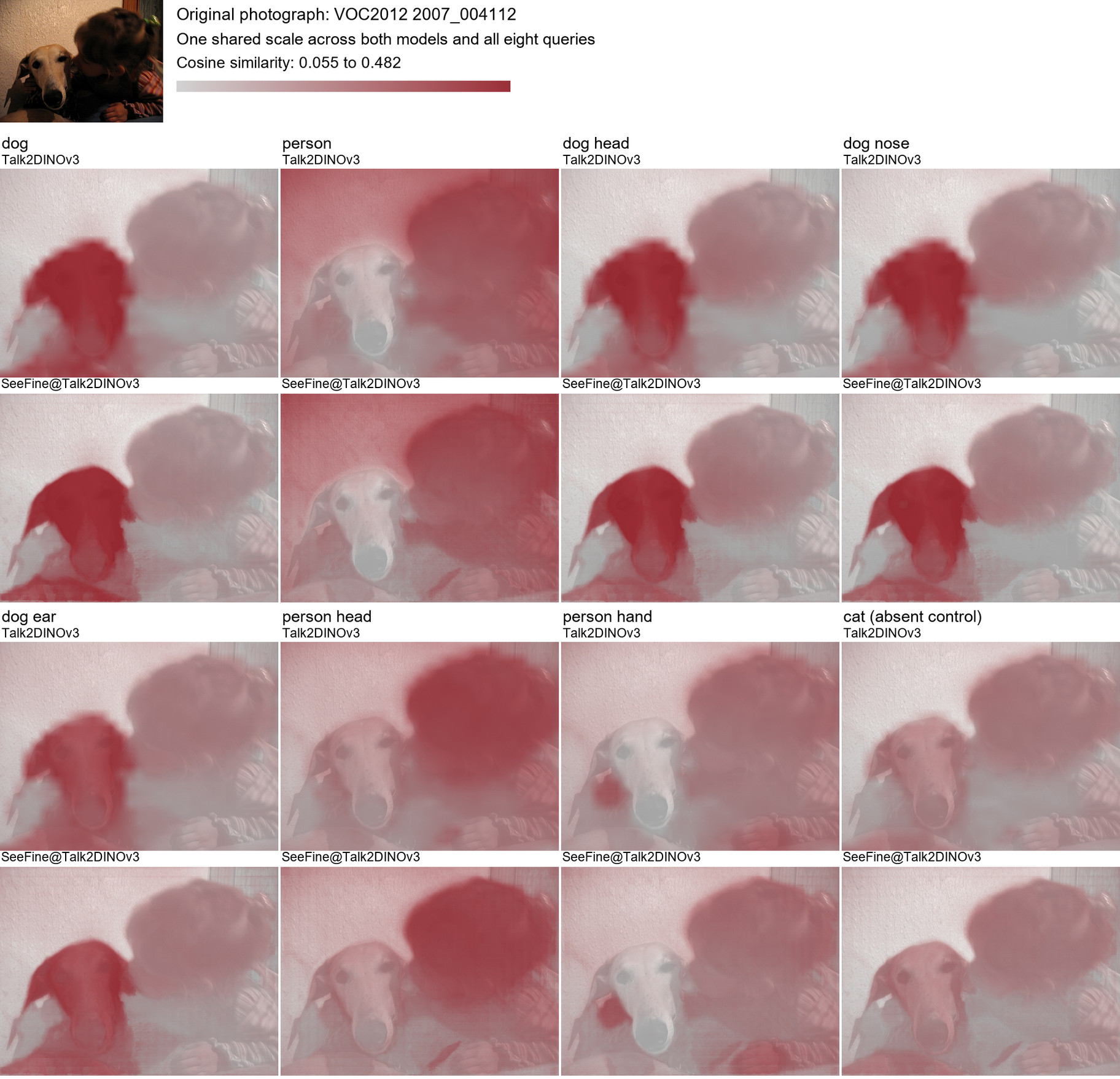}
\caption{\textbf{Object and part queries: dog and person.} The photograph appears above the query panels, and each query shows Talk2DINOv3 on top and \SF{Talk2DINOv3} below. Darker red marks higher text--feature cosine similarity on a bright desaturated image. The shared scale, shown above the panels, spans the pooled $2$nd--$98$th percentiles of both models across all eight phrases, including the absent-object control. Donor maps are expanded bilinearly for display, and \SF{Talk2DINOv3} scores each source pixel directly. Both models use the same projected text-template ensembles.}
\label{fig:talk2dinov3_parts_dog}
\end{figure*}

\begin{figure*}[p]
\centering
\includegraphics[width=\textwidth]{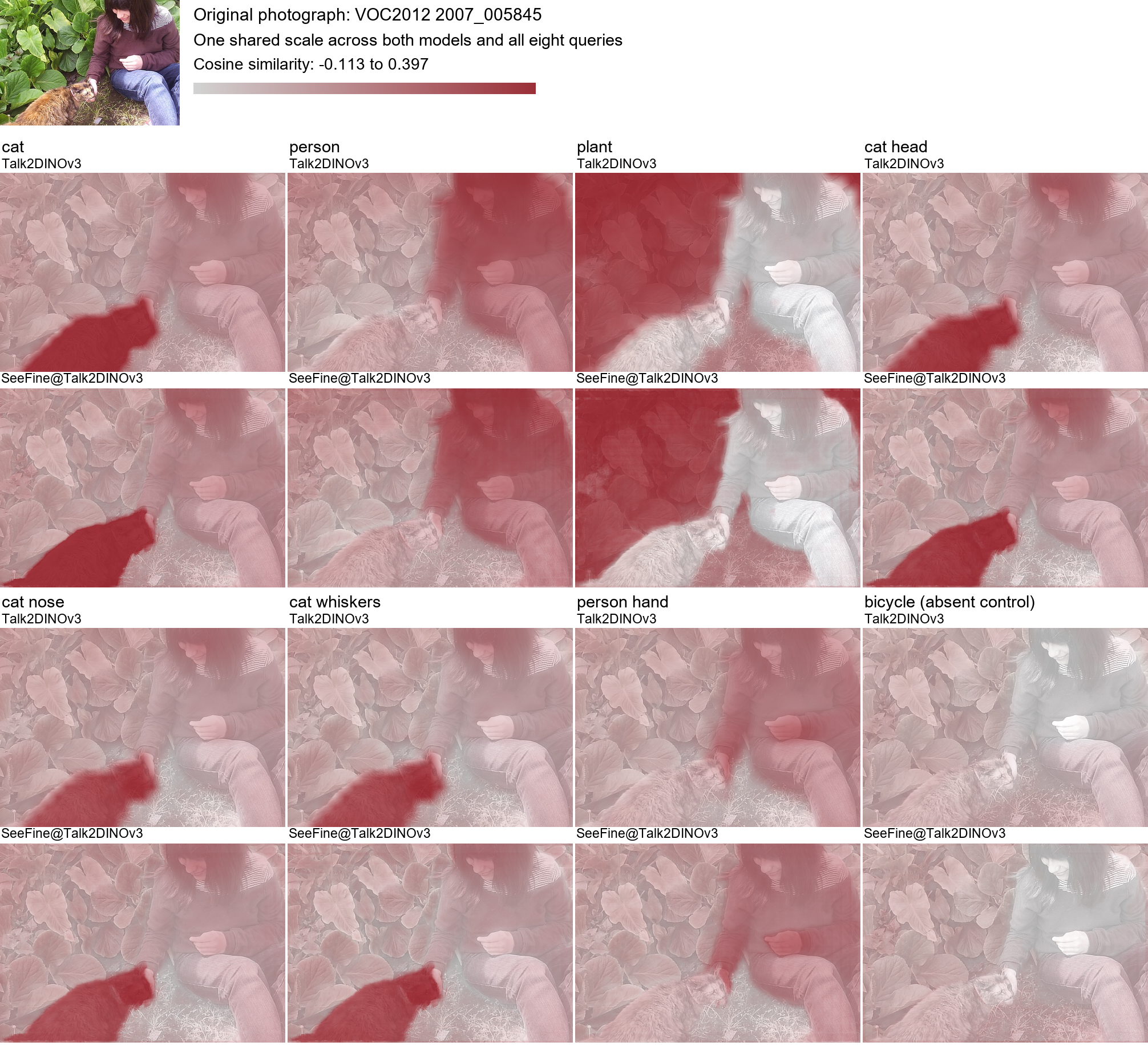}
\caption{\textbf{Object and part queries: cat and person.} Each query shows the Talk2DINOv3 response above the \SF{Talk2DINOv3} response. Display scales, text-template ensembles and the absent-object control follow \cref{fig:talk2dinov3_parts_dog}.}
\label{fig:talk2dinov3_parts_cat}
\end{figure*}

\begin{figure*}[p]
\centering
\includegraphics[width=\textwidth]{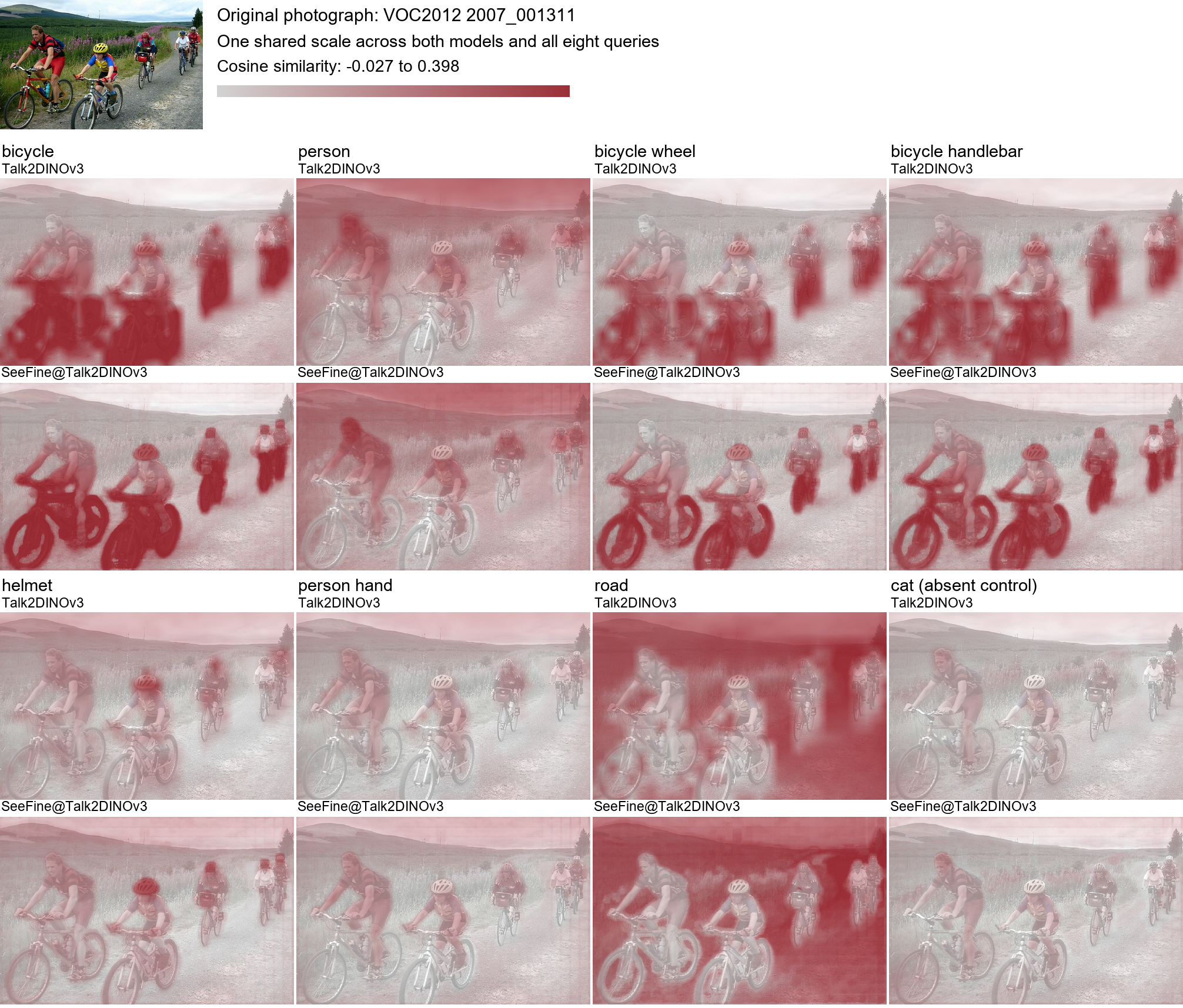}
\caption{\textbf{Object and part queries: cyclists.} Each query shows the Talk2DINOv3 response above the \SF{Talk2DINOv3} response. Display scales, text-template ensembles and the absent-object control follow \cref{fig:talk2dinov3_parts_dog}.}
\label{fig:talk2dinov3_parts_bicycle}
\end{figure*}

\paragraph{Qualitative examples.} \Cref{fig:talk2dinov3_div2k_text_main,fig:talk2dinov3_div2k_text_architecture,fig:talk2dinov3_div2k_text_objects} compare responses to object text on native DIV2K photographs (longer side $2{,}040$ pixels), \cref{fig:talk2dinov3_fourheads} shows complementary attention heads, and \cref{fig:talk2dinov3_parts_dog,fig:talk2dinov3_parts_cat,fig:talk2dinov3_parts_bicycle} show object and part queries in VOC2012 scenes. Donor maps are expanded bilinearly for display, while \SF{Talk2DINOv3} provides one value per pixel; each donor--SeeFine pair shares one colour scale. The panels visualize continuous similarity and attention responses.

\paragraph{Selecting complementary heads.} The encoder has 16 CLS heads. We rank dense heads by the text similarity of their attention-pooled features, then greedily skip a head whose above-90th-percentile attention region has IoU above $0.6$ with a previously selected head, selecting up to four. Donor and SeeFine use the same selected indices and colours. Opacity spans each map's $90$th--$99.5$th percentiles. Head selection uses only text similarity and attention maps. Each colour identifies an attention head.

\paragraph{Object and part queries in multi-object scenes.}\label{app:talk2dinov3_parts}
\Cref{fig:talk2dinov3_parts_dog,fig:talk2dinov3_parts_cat,fig:talk2dinov3_parts_bicycle} show three uncropped VOC2012 scenes, chosen by object presence: a dog and a person, a cat and a person, and cyclists. We encode each image once and score eight phrases, including an absent-object control. A single colour scale across the phrases and readouts of an image preserves differences in response strength. Part queries such as ``nose'' or ``whiskers'' respond over a larger region of the animal, with finer contours under SeeFine.

\FloatBarrier

\clearpage
\input{cvpr5_revision/spair_final.tex}

\FloatBarrier

\clearpage
\section{Landmark retrieval on CUB}\label{app:cubparts}
We use all $5{,}794$ official test images of CUB-200-2011~\cite{wah2011cub}, with $66{,}567$ visible annotations retained after the shared centre crop. Frozen DINO ViT-S/16 supplies final-block CLS-to-patch attention. All readouts see exactly the same $224\times224$ RGB image: the donor produces a $14\times14$ map and \SF{DINO} a $224\times224$ map. A visible landmark is assigned to its nearest token centre and is retrieved if that token belongs to the top $10\%$ of the native attention map. In \cref{tab:cubparts_detail}, the mean-attention column ranks the average of the six head maps.

For the aggregate score, we average recall over landmarks within each image and then over images. Each part-specific row pools the visible annotations of that part. Mean-map image-macro recall rises from $63.49$ to $70.77\%$ ($+7.28$ points). A paired image bootstrap with $2{,}000$ resamples and seed 1310 gives $[6.95,7.59]$ points. Aggregate recall rises for all six heads, and \cref{tab:cubparts_detail} shows the part-level changes.

\begin{figure}[!ht]
\centering
\colheads{\columnwidth}{3}{Image&DINO&\SF{DINO}}
\includegraphics[width=\columnwidth]{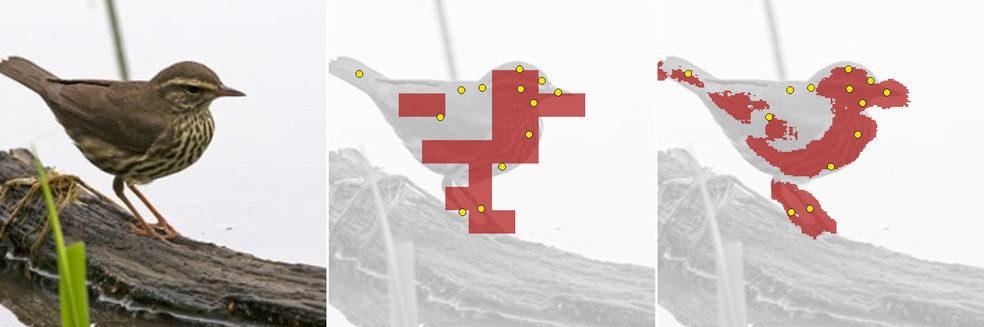}
\caption{\textbf{Fine bird parts, recovered by the same frozen attention.} The red mask marks the top $10\%$ of class-token attention, and yellow dots show the annotated landmarks. SeeFine covers $11$ of the $13$ landmarks and DINO covers $4$.}
\label{fig:cubparts_main}
\end{figure}

\Cref{tab:cubparts_main} compares the training-free readouts. SeeFine improves recall over the donor and performs similarly to Strided, and DVF is $0.52$ points higher. The comparable gains from these dense readouts point to sampling density as the main source of the improvement. \Cref{fig:cubparts_main} illustrates the finer attention around individual parts.

\begin{table}[t]\centering\footnotesize
\caption{\textbf{Fine attention improves landmark retrieval.} Image-macro CUB landmark recall (\%) on the full test set at the top $10\%$ of mean CLS attention. Strided and SeeFine have similar recall, and DVF scores $0.52$ points above SeeFine. HR-Dv2 is not listed: this task reads class-token attention, and HR-Dv2 returns no class token. \Cref{tab:base_cub} gives head-level scores and paired intervals.}
\label{tab:cubparts_main}
\setlength{\tabcolsep}{6pt}
\begin{tabular}{@{}lr@{}}\toprule
Readout & Mean-attention recall (\%)\\\midrule
Donor (= Nearest) & 63.52\\
Bilinear & 66.24\\
Strided & 70.73\\
DVF & \textbf{71.29}\\
SeeFine & 70.77\\
\bottomrule\end{tabular}
\end{table}

\begin{table}[t]
\centering
\caption{\textbf{Small bird parts become easier to retrieve.} Top-$10\%$ landmark recall on the CUB-200-2011 test set for each CLS-attention head of frozen DINO ViT-S/16. Each cell shows the donor recall (\%) and the change with SeeFine in percentage points (green: gain, red: loss). The mean-attention column ranks the average of the six head maps. Every head improves in aggregate, and losses are confined to a few part--head cells.}
\label{tab:cubparts_detail}
\scriptsize
\setlength{\fboxsep}{1pt}
\setlength{\tabcolsep}{0.6pt}
\renewcommand{\arraystretch}{1.03}
\begin{tabular}{@{}l*{7}{c}@{}}
\toprule
Part & H1 & H2 & H3 & H4 & H5 & H6 & \shortstack{Mean\\attention}\\
\midrule
\textbf{All} & \cubmain{64.5031760}{68.1414476} & \cubmain{41.2522676}{44.3672450} & \cubmain{42.1749586}{47.2290610} & \cubmain{63.1649715}{67.6968069} & \cubmain{56.2483032}{63.8143701} & \cubmain{40.7099976}{46.8064303} & \cubmain{63.4884389}{70.7663289}\\
Beak & \cubmain{92.9527040}{96.9474279} & \cubmain{19.0879970}{21.8202374} & \cubmain{30.6010929}{41.0401357} & \cubmain{96.1748634}{99.0201621} & \cubmain{80.3843980}{92.2743546} & \cubmain{63.5764085}{77.3695120} & \cubmain{90.9930281}{97.0793292}\\
Forehead & \cubmain{90.4904530}{96.8738300} & \cubmain{29.2587046}{31.9543242} & \cubmain{24.4290528}{25.2152752} & \cubmain{85.6608012}{92.4560090} & \cubmain{69.7491576}{80.8873081} & \cubmain{33.1336578}{34.4627480} & \cubmain{76.0950955}{85.9977537}\\
Crown & \cubmain{85.7461438}{93.3469615} & \cubmain{37.4279874}{43.7465155} & \cubmain{33.2651923}{39.4350492} & \cubmain{71.9754692}{80.5612340} & \cubmain{68.5560305}{80.4125627} & \cubmain{32.0200706}{37.0748931} & \cubmain{65.8241962}{77.1789630}\\
Nape & \cubmain{85.0955414}{91.1010009} & \cubmain{40.3093722}{42.4749773} & \cubmain{37.3612375}{41.2920837} & \cubmain{59.9454049}{66.7151956} & \cubmain{50.6824386}{56.1783439} & \cubmain{24.7679709}{26.3876251} & \cubmain{58.3075523}{64.8953594}\\
Left eye & \cubmain{98.4285714}{99.2857143} & \cubmain{42.3928571}{50.6428571} & \cubmain{42.4642857}{53.2500000} & \cubmain{95.1071429}{97.2857143} & \cubmain{92.4642857}{96.8571429} & \cubmain{54.8571429}{67.3571429} & \cubmain{95.6071429}{98.3571429}\\
Right eye & \cubmain{98.4221599}{98.9481066} & \cubmain{44.4249649}{51.4375877} & \cubmain{42.0056101}{52.4894811} & \cubmain{94.5301543}{97.5455820} & \cubmain{92.5315568}{96.6690042} & \cubmain{55.2244039}{68.1276297} & \cubmain{95.5820477}{98.1416550}\\
Throat & \cubmain{94.9428163}{97.9270908} & \cubmain{27.1265189}{24.2673338} & \cubmain{42.2802001}{44.8355969} & \cubmain{82.1837026}{87.4017155} & \cubmain{58.7026447}{61.8298785} & \cubmain{33.8456040}{32.6840600} & \cubmain{81.5582559}{87.7591137}\\
Breast & \cubmain{80.5024372}{84.2519685} & \cubmain{57.5928009}{61.1923510} & \cubmain{68.5789276}{74.5219348} & \cubmain{31.0648669}{32.8833896} & \cubmain{44.8631421}{47.7690289} & \cubmain{33.4833146}{38.1327334} & \cubmain{55.5868016}{60.8548931}\\
Belly & \cubmain{50.4132231}{55.3522235} & \cubmain{59.7796143}{65.0137741} & \cubmain{69.5001968}{77.9024006} & \cubmain{22.1566312}{21.2317985} & \cubmain{43.8016529}{50.0196773} & \cubmain{48.0322708}{53.3451397} & \cubmain{46.3597009}{51.2003148}\\
Back & \cubmain{30.8532423}{30.1934016} & \cubmain{52.9237770}{57.5199090} & \cubmain{32.4687144}{33.9021615} & \cubmain{18.3617747}{19.0671217} & \cubmain{31.0352673}{33.1058020} & \cubmain{29.9431172}{33.9931741} & \cubmain{28.6916951}{29.5563140}\\
Left wing & \cubmain{43.5762817}{45.0277949} & \cubmain{82.5200741}{84.1877702} & \cubmain{71.4947498}{74.9845584} & \cubmain{30.5744287}{32.0877085} & \cubmain{49.4132180}{54.0765905} & \cubmain{46.1087091}{50.5867820} & \cubmain{63.3106856}{66.0592959}\\
Right wing & \cubmain{45.1652765}{47.3277726} & \cubmain{83.5341365}{84.8625270} & \cubmain{71.7639790}{76.1198641} & \cubmain{29.9351251}{32.3138709} & \cubmain{50.4788384}{54.3713315} & \cubmain{48.0074143}{51.4365153} & \cubmain{63.8554217}{68.0877356}\\
Left leg & \cubmain{14.1832098}{16.1205453} & \cubmain{14.4702224}{12.9634059} & \cubmain{14.6376465}{15.0442478} & \cubmain{81.7029419}{90.6003348} & \cubmain{43.1714901}{56.6371681} & \cubmain{44.4630471}{55.2738579} & \cubmain{50.9208323}{65.4628079}\\
Right leg & \cubmain{14.6859903}{16.0628019} & \cubmain{14.3478261}{13.3091787} & \cubmain{15.2898551}{14.8309179} & \cubmain{82.6570048}{90.9178744} & \cubmain{42.9951691}{56.9082126} & \cubmain{44.7342995}{55.7246377} & \cubmain{51.1111111}{65.9661836}\\
Tail & \cubmain{22.9758777}{29.3049916} & \cubmain{44.6859326}{54.6930977} & \cubmain{52.7824218}{63.7926917} & \cubmain{75.2089802}{83.6398376} & \cubmain{44.3276809}{58.7294005} & \cubmain{41.2467160}{52.3764032} & \cubmain{47.0981610}{61.1416289}\\
\bottomrule
\end{tabular}
\setlength{\fboxsep}{3pt}
\end{table}

\paragraph{Training-free baselines.}\label{app:base_cub}
\Cref{tab:base_cub,tab:cubparts_main} evaluate all readouts with one implementation. This implementation gives donor recall of $63.52\%$, compared with $63.49\%$ in the part and head analysis of \cref{tab:cubparts_detail}. SeeFine rounds to $70.77\%$ in both evaluations, and each interval uses paired results from its own evaluation. Nearest equals the donor, because a landmark is assigned to the token containing it. Strided is on par with SeeFine ($-0.04$ points, $[-0.26,+0.17]$), and the stride-4 attention of DVF scores $0.52$ points higher ($[+0.31,+0.75]$).

\begin{table*}[t]\centering\footnotesize
\caption{\textbf{CUB landmark recall with training-free baselines.} Image-macro recall (\%) at the top 10\% of final-block DINO ViT-S/16 CLS attention, over all landmarks of the full test set. The $\Delta$ column subtracts SeeFine's mean-attention recall from that of each baseline, with paired 95\% image-bootstrap CIs. HR-Dv2 is not listed: this task reads class-token attention, and HR-Dv2 returns no class token.}\label{tab:base_cub}
\setlength{\tabcolsep}{2.5pt}
\begin{tabular}{@{}lrrrrrrrrc@{}}\toprule
Readout & H1 & H2 & H3 & H4 & H5 & H6 & \shortstack{Mean\\attention} & $\Delta$ & 95\% CI\\\midrule
Donor (= Nearest) & 64.6 & 41.3 & 42.3 & 63.2 & 56.3 & 40.8 & 63.52 & $-7.25$ & $[-7.58,-6.92]$\\
Bilinear & 64.8 & 38.9 & 40.7 & 62.3 & 60.5 & 41.4 & 66.24 & $-4.53$ & $[-4.82,-4.25]$\\
Strided & 67.7 & 44.3 & 48.5 & 68.5 & 67.6 & 50.8 & 70.73 & $-0.04$ & $[-0.26,+0.17]$\\
DVF & 68.4 & 46.1 & 50.1 & 69.9 & 69.0 & 52.8 & \textbf{71.29} & $+0.52$ & $[+0.31,+0.75]$\\
\midrule
SeeFine & 68.2 & 44.5 & 47.3 & 67.8 & 63.9 & 46.9 & 70.77 & --- & ---\\
\bottomrule\end{tabular}
\end{table*}

\begin{figure*}[p]\centering
\includegraphics[width=\textwidth,height=.84\textheight,keepaspectratio]{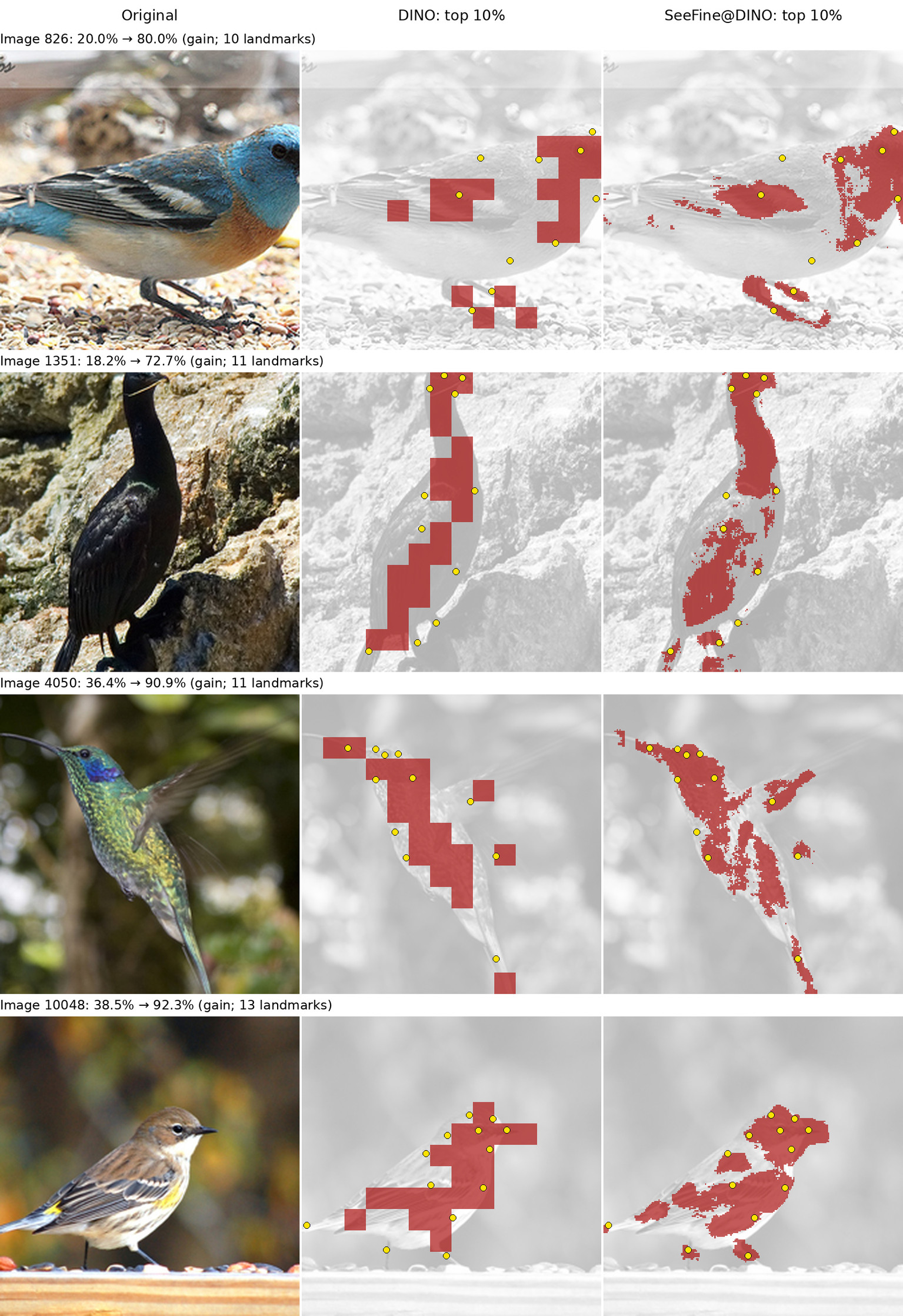}
\caption{\textbf{CUB landmark retrieval: examples with large gains.} The same $224\times224$ source crop, DINO ViT-S/16 mean CLS attention on its $14\times14$ grid, and \SF{DINO}. Red marks the top $10\%$ of each native map; yellow dots show visible in-crop landmarks. The donor mask is expanded with nearest-neighbour blocks only for display. Per-image recall and annotation counts are printed above each row. These are four of the six largest gains among test images with at least ten retained landmarks.}\label{fig:cubparts_qual_1}
\end{figure*}

\begin{figure*}[p]\centering
\includegraphics[width=\textwidth,height=.84\textheight,keepaspectratio]{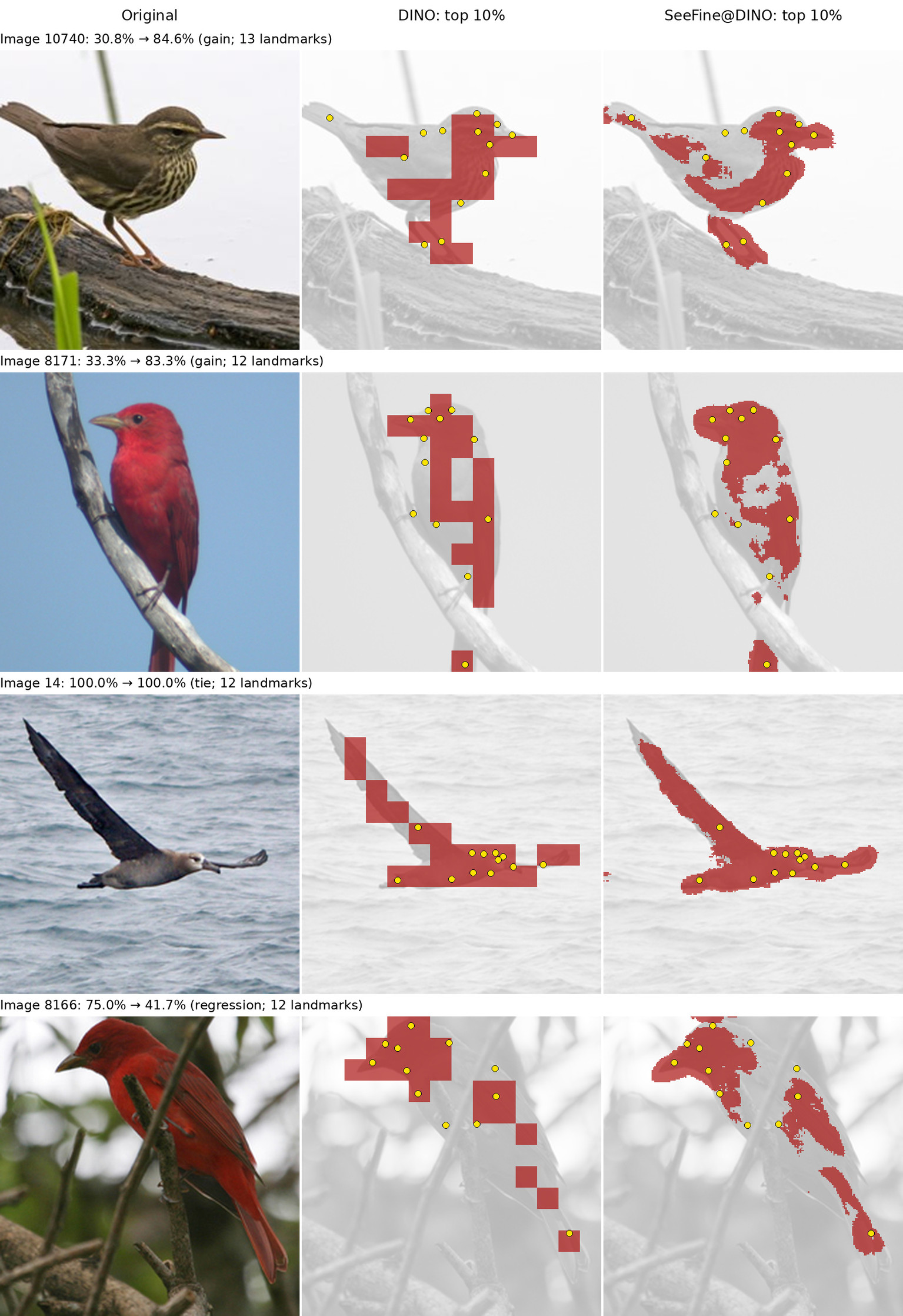}
\caption{\textbf{CUB landmark retrieval: further examples.} The first two rows show the remaining two of the six largest gains. SeeFine and DINO retrieve the same number of landmarks in the third row, and SeeFine retrieves fewer in the fourth. The original, DINO donor and \SF{DINO} use the same geometry, masks and landmarks as \cref{fig:cubparts_qual_1}.}\label{fig:cubparts_qual_2}
\end{figure*}

\paragraph{Qualitative selection.} \Cref{fig:cubparts_qual_1,fig:cubparts_qual_2} show six gains among images with at least ten visible landmarks, one tie and one regression. 


\FloatBarrier

\clearpage\onecolumn
\section{Medical images}\label{app:medical}
\subsection{Volumetric CT encoders}\label{app:ct3d}
Like 2D vision transformers, volumetric encoders embed fixed-size patches, add learned positional embeddings and mix tokens through attention, so SeeFine extends to 3D volumes. In CT, each token covers a large voxel block: $4\times8\times8=256$ voxels for TAP-CT and $16^3=4{,}096$ for 3DINO. A single token can therefore cover small structures such as vessels, airways and adrenal glands.

We evaluate whether finer feature readouts improve correspondence in measurable tasks. We use two public CT encoders, 35 pairs of lung scans with expert landmarks, and 15 pairs of abdominal scans from different patients. All experiments are training-free: the encoders remain frozen, and correspondences are computed by nearest-neighbour search.

\paragraph{3D readout.}
The 3D construction follows the 2D procedure along all three spatial axes. For patches of size $P_d\times P_h\times P_w$ and stride $S$, we pad each axis by $P-S$ voxels, using mirrored padding in-plane and neighbouring slices along the scan axis. We then apply the original patch convolution at stride $S$ and trilinearly resample the learned positional embeddings. Tokens are divided into $\prod_i P_i/S_i$ phases, each preserving the original token spacing. As in 2D, global attention keys are shifted by $\log(|G|/\Nref)$. Class and register states are supplied by an ordinary donor pass at each block, using the recorded-context variant (\cref{alg:donor_context_v4}).

\paragraph{Baselines.} Nearest assigns each voxel its containing-patch token and predicts patch centres. Trilinear upsamples the donor grid. Strided uses full attention and the checkpoint's position interpolation. Its embedding stride follows the rule of the 2D baseline: $2\times4\times4$ for TAP-CT and $7$ for 3DINO. At the original patch stride, its tokens reproduce the donor. DVF and HR-Dv2 have released implementations for 2D images only and are not evaluated here.

We evaluate two volumetric DINOv2-based encoders:
TAP-CT-B-3D~\cite{veenboer2025tapct}, a ViT-B with $4\times8\times8$ patches, pretrained on $12\times224\times224$ CT crops, and 3DINO-ViT~\cite{xu2025threedino}, a ViT-L with $16^3$ patches, pretrained on CT, MRI, and PET. TAP-CT supports 256 phases at stride one, each containing $3\times28\times28=\Nref$ tokens, requiring no global-key shift. For 3DINO, we use strides 2 and 4, corresponding to 512 and 64 phases, respectively. At the original patch stride, the 3D readout reproduces the original features (minimum cosine similarity $0.9999998$). Scans are resized in-plane to $224\times224$ following TAP-CT's preprocessing and processed in blocks of 12 slices for TAP-CT and 48 for 3DINO. No model weights are updated.

\paragraph{Evaluation tasks.}
We evaluate feature correspondence in two settings.

\emph{Within-patient correspondence.}
We test whether features identify the same anatomical point between inhale and exhale scans. The evaluation uses Lung250M-4B~\cite{falta2023lung250m} (15 scan pairs, 1{,}477 landmarks), DIR-Lab 4DCT, and COPDgene~\cite{castillo2009dirlab,castillo2013copdgene} (10 pairs each, 300 landmarks per pair). For each inhale landmark, we find the most similar feature across all voxels in the exhale lung using cosine similarity. Performance is measured by the distance to its annotated counterpart in millimetres. No registration, spatial constraints, or search windows are used.

\emph{Across-patient correspondence.}
We test whether features identify the same anatomy in different patients. Using AbdomenCTCT from Learn2Reg~\cite{xu2016abdomen,hering2023learn2reg}, we form 15 disjoint pairs from 30 labelled scans, each containing 13 organs. We sample 20 interior points per organ in the first scan and match them across the second scan, restricting the search to voxels with HU $>-500$. A match is correct if it falls inside the corresponding organ.

\begin{figure}[H]
\centering
\includegraphics[width=\textwidth]{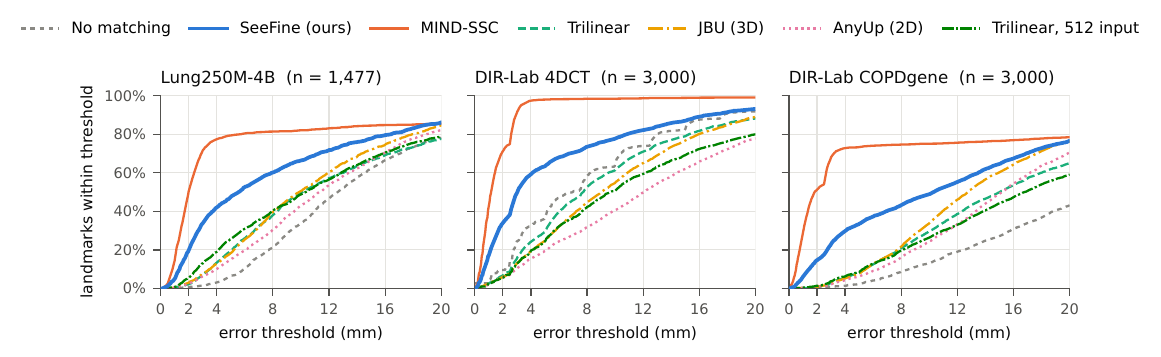}
\caption{\textbf{Landmark accuracy across error thresholds.} All feature curves use the same frozen TAP-CT-B-3D encoder; MIND-SSC is a handcrafted descriptor and ``No matching'' uses voxel positions. \SF{TAP-CT} outperforms the upsampling baselines across all three datasets. MIND-SSC achieves the highest accuracy at small errors but has a longer tail of distant matches. The staircase on 4DCT reflects its 2.5\,mm slice spacing.}
\label{fig:ct3d_curves}
\end{figure}

\paragraph{Results.}
\Cref{tab:ct3d,fig:ct3d_curves} summarize the results.

For within-patient correspondence, \SF{TAP-CT} substantially improves localization over trilinear upsampling. Median landmark error decreases from $10.1$, $7.7$, and $15.0$\,mm to $5.5$, $3.0$, and $10.3$\,mm on Lung250M-4B, DIR-Lab 4DCT, and COPDgene, respectively. Mean per-landmark error decreases by $5.1\,[3.6,6.6]$, $4.8\,[4.0,5.7]$, and $8.2\,[6.8,9.5]$\,mm (bootstrap 95\% intervals), with improvements in all 35 scan pairs.

\begin{figure}[H]
\centering
\includegraphics[width=.7\textwidth]{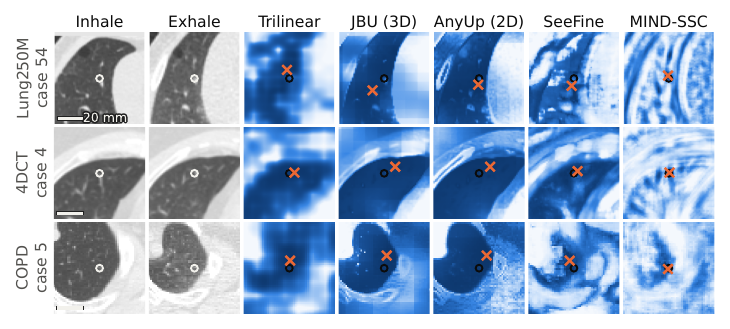}
\caption{\textbf{Similarity maps for five readouts.} The landmarks of \cref{fig:ct3d_qual} (fixed seed, one per lung dataset), frozen TAP-CT-B-3D except MIND-SSC. Maps show feature similarity on the exhale slice through the true match. The ring marks the true location, and the cross marks the most similar voxel in this slice. Each $80\times80$\,mm crop has its own colour scale. Trilinear keeps the coarse token grid, JBU and AnyUp spread similarity along tissue boundaries, and SeeFine follows the vessels. MIND-SSC localizes most precisely within a patient.}
\label{fig:ct3d_qual_full}
\end{figure}

\SF{TAP-CT} gives lower median errors than trilinear upsampling at $512^2$, 3D joint bilateral upsampling~\cite{kopf2007joint} and slice-wise AnyUp~\cite{wimmer2026anyup} on all three lung datasets (\cref{tab:ct3d}). In the examples of \cref{fig:ct3d_qual_full}, interpolation keeps the coarse token grid, the alternative upsamplers spread along tissue boundaries, and the SeeFine responses follow the vessel structure.

The improvement also transfers to 3DINO. At stride 2, \SF{3DINO} reduces mean landmark error by $1.3\,[0.8,1.8]$, $3.5\,[3.3,3.7]$, and $2.3\,[2.0,2.7]$\,mm, improving results in 32 of 35 scan pairs. The gains are smaller than for TAP-CT, consistent with its coarser $16^3$ patches and the need to use a larger stride.

\begin{figure}[H]
\centering
\includegraphics[width=\textwidth]{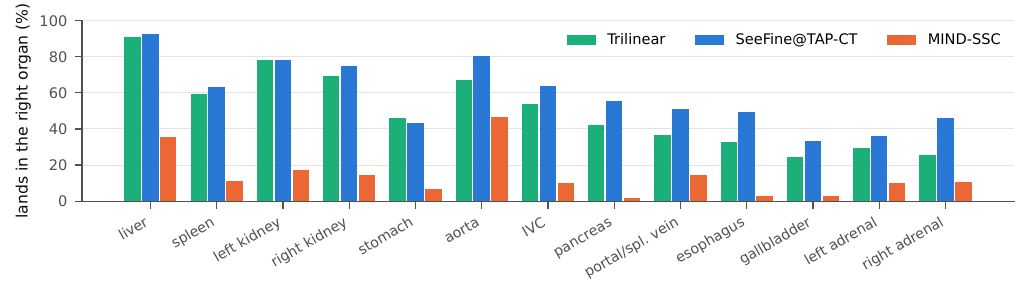}
\caption{\textbf{Cross-patient anatomical correspondence.} Percentage of points whose nearest-neighbour match falls inside the same organ in another patient (AbdomenCTCT, 15 pairs, 20 points per organ). \SF{TAP-CT} outperforms trilinear upsampling on 11 of 13 organs, with particularly large gains for small and thin structures.}
\label{fig:ct3d_organs}
\end{figure}

Across patients, \SF{TAP-CT} matches $59.4\%$ of sampled points to the correct organ, compared with $50.9\%$ for trilinear upsampling (gain $8.5$ percentage points; 95\% interval $[6.9,10.0]$). The gains are especially large for small or thin structures, including the aorta, pancreas, portal vein, oesophagus and right adrenal (\cref{fig:ct3d_organs}). Matching each point to the same voxel position in the other scan gives $25.4\%$, $34.0$ percentage points below SeeFine. The gain carries over to 3DINO, where SeeFine raises correct-organ matches from $34.0\%$ to $40.7\%$ (+$6.7$ percentage points; 95\% interval $[5.2,8.3]$) and improves all 13 organs.

\paragraph{Computational cost.}
At stride one, each $12\times224\times224$ block contains $602{,}112$ tokens, and SeeFine needs about 82\,s of GPU time per Lung250M scan on a single RTX~4090, compared with less than one second for the donor pass. At stride $1\times2\times2$, processing takes 16\,s per scan at a median error of $5.5$\,mm, and at stride $2\times4\times4$ it takes 2\,s at $6.5$\,mm. Across patients, this coarser setting reaches $59.5\%$ correct-organ matches, close to the $59.4\%$ at stride one.

\paragraph{Scope and limitations.}
For within-patient correspondence, the handcrafted MIND-SSC descriptor~\cite{heinrich2013mind}, commonly used in lung registration, is more precise, with median errors of $1.2$--$2.0$\,mm. Its local intensity-based representation suits matching the same anatomy across breathing phases. Across patients, it reaches $14.4\%$ correct-organ matches, compared with $25.4\%$ for the same-position baseline and $59.4\%$ for \SF{TAP-CT}, whose encoder features capture anatomy that transfers across patients.

These experiments evaluate frozen features through nearest-neighbour correspondence, without a registration method. A dedicated registration method could use SeeFine features as input.

\begin{table}[H]
\centering\footnotesize
\setlength{\tabcolsep}{4pt}
\begin{tabular}{@{}lccccccc@{}}
\toprule
 & \multicolumn{2}{c}{Lung250M-4B} & \multicolumn{2}{c}{DIR-Lab 4DCT} & \multicolumn{2}{c}{DIR-Lab COPDgene} & Across patients \\
\cmidrule(lr){2-3}\cmidrule(lr){4-5}\cmidrule(lr){6-7}\cmidrule(l){8-8}
Readout & median$\downarrow$ & $\le$4\,mm$\uparrow$ & median$\downarrow$ & $\le$4\,mm$\uparrow$ & median$\downarrow$ & $\le$4\,mm$\uparrow$ & right organ$\uparrow$ \\
\midrule
Same position (no matching) & 12.6 & 3\% & 7.0 & 33\% & 22.3 & 1\% & 25.4\% \\
\midrule
\multicolumn{8}{@{}l}{\emph{TAP-CT-B-3D}} \\
Nearest & 19.7 & 5\% & 12.3 & 9\% & 28.7 & 2\% & 39.8\%\\
Trilinear upsampling & 10.1 & 14\% & 7.7 & 24\% & 15.0 & 6\% & 50.9\% \\
Trilinear, $512^2$ input & 10.3 & 19\% & 9.7 & 20\% & 16.8 & 6\% & -- \\
3D joint bilateral upsampling & 9.9 & 13\% & 9.0 & 19\% & 12.7 & 5\% & -- \\
AnyUp, per slice & 11.3 & 10\% & 12.0 & 16\% & 15.2 & 4\% & -- \\
Strided & 7.3 & 28\% & 5.9 & 35\% & 11.6 & 13\% & 58.1\%\\
\SF{TAP-CT}, stride 1 & \textbf{5.5} & \textbf{42\%} & \textbf{3.0} & \textbf{59\%} & \textbf{10.3} & \textbf{30\%} & \textbf{59.4\%} \\
\multicolumn{8}{@{}l}{\emph{3DINO-ViT}} \\
Nearest & 17.9 & 1\% & 18.1 & 1\% & 22.6 & 1\% & 28.4\%\\
Trilinear upsampling & 11.3 & 8\% & 7.6 & 27\% & 13.3 & 5\% & 34.0\% \\
Strided & 10.9 & 11\% & 7.7 & 28\% & 13.5 & 7\% & 38.7\%\\
\SF{3DINO}, stride 2 & \textbf{9.3} & \textbf{24\%} & \textbf{3.3} & \textbf{56\%} & \textbf{10.4} & \textbf{18\%} & \textbf{40.7\%} \\
\midrule
\multicolumn{8}{@{}l}{\emph{Handcrafted}} \\
MIND-SSC & 2.0 & 77\% & 1.2 & 98\% & 1.8 & 73\% & 14.4\% \\
\bottomrule
\end{tabular}
\caption{\textbf{Training-free correspondence with frozen 3D encoders.} Within-patient results report median landmark error (mm) and the percentage of matches within 4\,mm. Across-patient results report the percentage of points matched to the correct organ (15 pairs, 13 organs, 20 points per organ, the same points for every method). The same-position baseline uses no feature matching. Methods using the same encoder share its frozen weights. MIND-SSC is a handcrafted descriptor. Nearest, Trilinear and Strided are defined under \emph{Baselines} above; DVF and HR-Dv2 have 2D-only code. ---: not evaluated.}
\label{tab:ct3d}
\end{table}

\subsection{Slice-wise tumour-mask propagation on brain MRI}\label{app:brats-slices}
We test whether a denser frozen representation can carry an annotation from one MRI slice to another, using released DINO ViT-S/16 on 2D FLAIR slices of BraTS 2019~\cite{menze2015brats,bakas2017tcga,balakrishnan2023brats2d}. The set contains 4,715 slices of $240\times240$ pixels from 152 cases. Each case's middle slice is its only labelled support. We score all 4,563 tumour-containing target slices from the 150 cases with at least two slices. The two single-slice cases supply no targets.

All readouts clip FLAIR to $[-0.5,8]$, replicate it to RGB and apply the same ImageNet normalization. Normalized support features are averaged inside the union of tumour labels and in a 15-pixel surrounding ring to form foreground/background prototypes. Each target pixel selects the prototype with higher cosine similarity, without fitting a threshold. Target masks are used only for scoring. The donor samples the support mask at its $15\times15$ patch centres to build prototypes, assigns one label per patch and expands the labels by nearest-neighbour interpolation. All other rows, Nearest included, pool features under the full-resolution support mask and ring and label every pixel of the $240\times240$ lattice. Nearest and Bilinear interpolate the donor grid, Strided uses stride 7, HR-Dv2 its released 68-pass recombination and DVF its extractor at stride 4. SeeFine uses stride-one features with the same frozen weights and reference 196 (\cref{tab:reference_policy_scope_v4}).

\begin{figure}[H]
\centering
\includegraphics[width=.88\textwidth,height=.43\textheight,keepaspectratio]{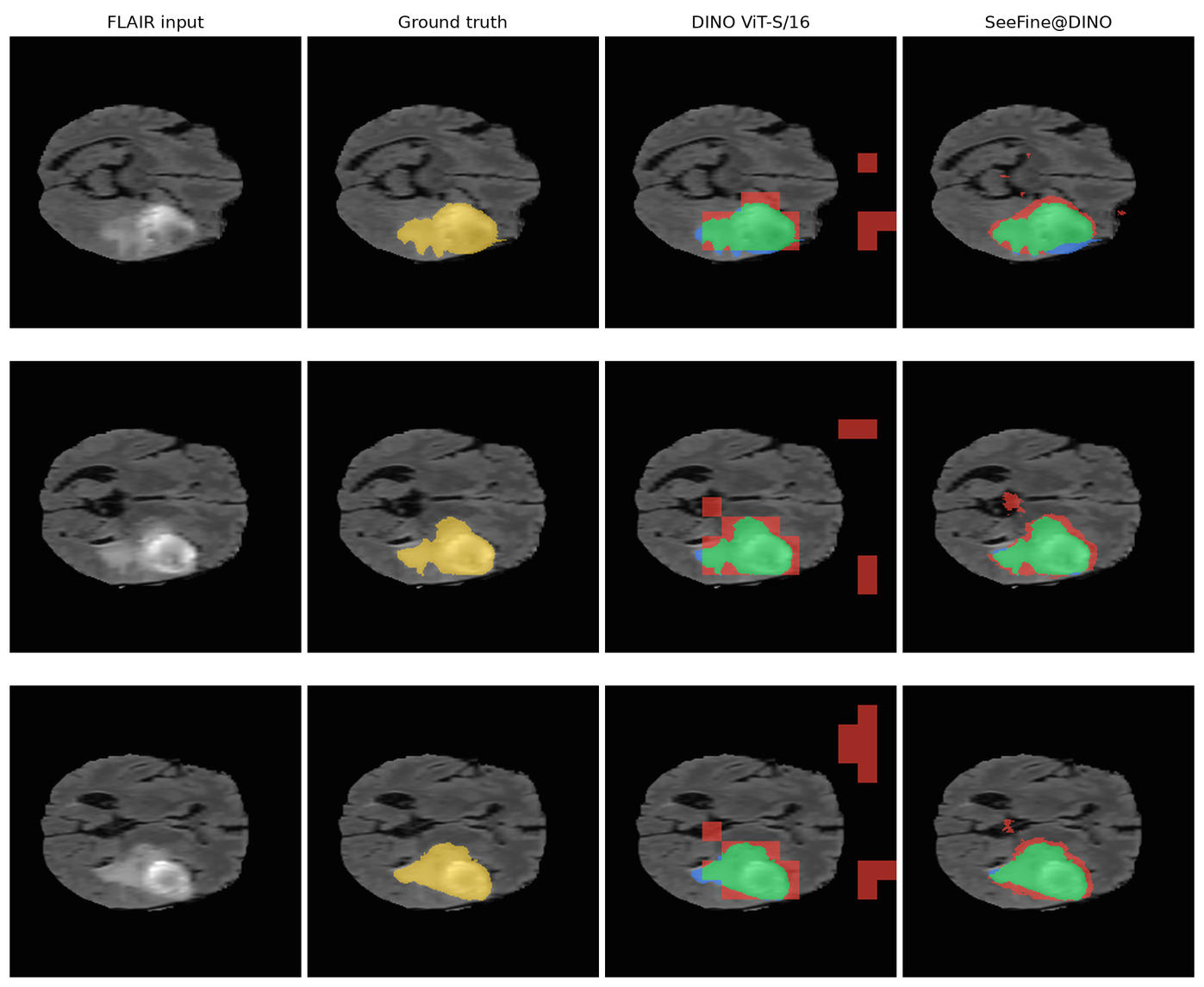}
\caption{\textbf{A support annotation carried across FLAIR slices.} Target slices from BRATS\_001, using support slice 67. Columns: input, target ground truth, frozen DINO and SeeFine@DINO. Green/red/blue mark true positives/false positives/false negatives. Target labels are used only for comparison and scoring.}
\label{fig:brats-dino-slices}
\end{figure}

Case-macro Dice averages target-slice Dice within each case, then across cases. All readouts use last-block tokens before the final LayerNorm. SeeFine raises case-macro Dice from $70.19$ to $75.98\%$, a gain of $5.79$ points (paired 95\% case-bootstrap CI $[4.54,7.09]$, 2,000 resamples). Bilinear interpolation gives the highest Dice, $77.45\%$, $1.47$ points above SeeFine, and DVF reaches $74.20\%$ (\cref{tab:brats_full}). \Cref{fig:brats-dino-slices} separately illustrates support slice 67 and target slices 63, 71 and 75 from one case.

\begin{table}[H]\centering\footnotesize
\caption{\textbf{Tumour-mask propagation on BraTS 2D FLAIR slices.} Dice (\%) of frozen DINO ViT-S/16 readouts on all tumour-containing target slices. Case-macro Dice averages the slices within each case and then weights cases equally; slice-macro Dice weights slices equally. All readouts use pre-LayerNorm last-block features, the same support annotation and cosine-prototype assignment.}
\label{tab:brats_full}
\setlength{\tabcolsep}{8pt}
\begin{tabular}{@{}lrr@{}}\toprule
Readout & Case-macro Dice (\%) & Slice-macro Dice (\%)\\\midrule
DINO ViT-S/16 & 70.19 & 70.03\\
Nearest & 72.08 & 71.69\\
Bilinear & \textbf{77.45} & \textbf{76.78}\\
Strided & 75.61 & 74.83\\
HR-Dv2 & 74.59 & 74.13\\
DVF & 74.20 & 73.48\\
\midrule
\SF{DINO} & 75.98 & 75.99\\
\bottomrule\end{tabular}
\end{table}

\clearpage
\section{Depth from frozen features}\label{app:downstream}
The NYUv2 experiment asks whether a simple task head can use the finer frozen features. We train a separate depth head for each readout of frozen DINOv2 ViT-S/14, using the same $795$ labelled training images, $654$ test images and training recipe.

The head is a $1\times1$ convolution predicting 256 depth bins over $[10^{-3},10]$ metres, followed by the Probe3D positive-probability normalization and expectation. Training uses the same DepthLoss, AdamW (learning rate $5\cdot10^{-4}$, weight decay $0.01$), batch eight, 149 warm-up steps and cosine decay over $36{,}000$ steps, seed zero. Frozen features are cached once. All readouts use the same $224\times224$ view and label lattice. Donor features are interpolated to the lattice, and SeeFine features are read on it directly.

\paragraph{Results.} SeeFine improves three of the four depth metrics, with separately trained heads and one training seed. It lowers scale-aware RMSE from $0.5456$ to $0.5388$ and raises $d_1$ from $76.86$ to $77.57\%$. With scale-and-shift alignment, $d_1$ rises from $87.61$ to $87.78\%$, and RMSE changes from $0.4137$ to $0.4175$. Nearest, Strided and HR-Dv2 score below the donor on all four metrics (\cref{tab:canonical_nyu}). HR-Dv2 uses the DINOv2 checkpoint without registers, which has the donor's weights; the separately trained checkpoint with registers gives SA RMSE $0.5515$.
\IfFileExists{cvpr5_revision/nyu_results_v5.tex}{\input{cvpr5_revision/nyu_results_v5.tex}}{}
\FloatBarrier

\clearpage
\section{Qualitative results}\label{app:gallery}\label{app:regions}\label{app:div2k}
\subsection{Native-resolution attention}
\begin{figure}[H]
\centering
\begin{minipage}{.55\textwidth}
\noindent\makebox[\linewidth][c]{%
\begin{tabular}{@{}>{\centering\arraybackslash}p{.5\linewidth}@{\hspace{.01\linewidth}}>{\centering\arraybackslash}p{.5\linewidth}@{}}
\scriptsize Image & \scriptsize Winning head (1--4)
\end{tabular}}\par\vspace{1pt}
\makebox[\linewidth][c]{\includegraphics[width=1.01\linewidth]{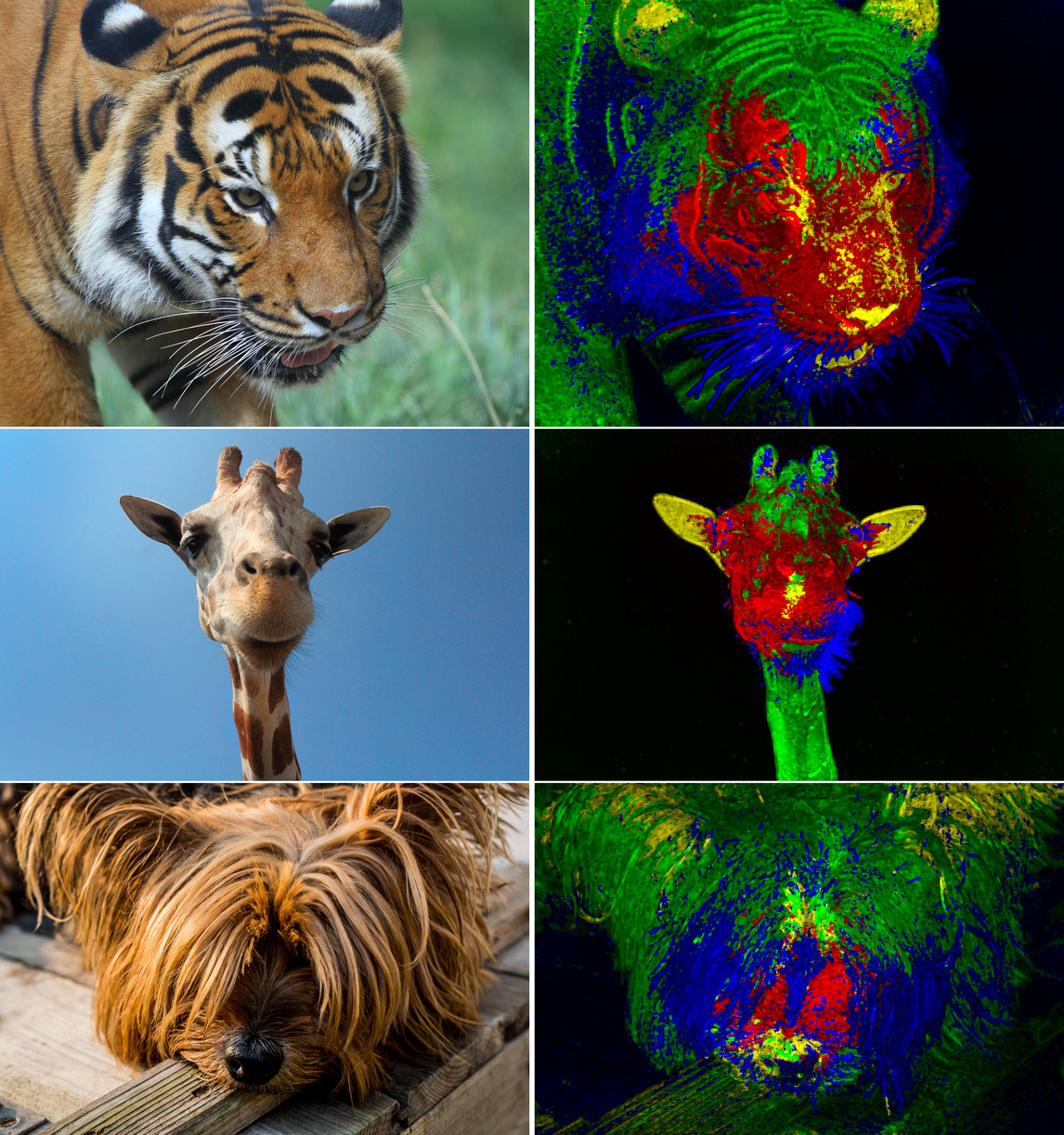}}
\end{minipage}
\caption{\textbf{Different heads reveal different details.} Frozen DINO ViT-S/16 read by \SF{DINO} on complete DIV2K photographs of a tiger, a giraffe and a Yorkshire terrier. The right column assigns each pixel the colour of the strongest normalized final-block CLS-attention head among heads $1$--$4$ (red, green, blue and yellow). Each head is normalized by its $p_{99.5}$, and brightness follows the sum of normalized attention. In the terrier, the colours appear to separate face, fur, whiskers and especially ears, without assigned part labels. \Cref{sec:vizattn} gives the display rule, and the following figures show all six heads in two triplets.}
\label{fig:div2k_desk_heads_main}
\end{figure}
The gallery shows final-block CLS attention of frozen DINO ViT-S/16 read by \SF{DINO} on complete native-resolution DIV2K photographs of animals, people, buildings and machinery. \Cref{sec:vizattn} defines the head-colour displays.
\begingroup
\setlength{\intextsep}{2pt}
\captionsetup{skip=2pt}
\begin{figure}[H]
\centering
\IfFileExists{appendix_canonical_20261014/div2k_heads_1.jpg}{\includegraphics[width=\textwidth]{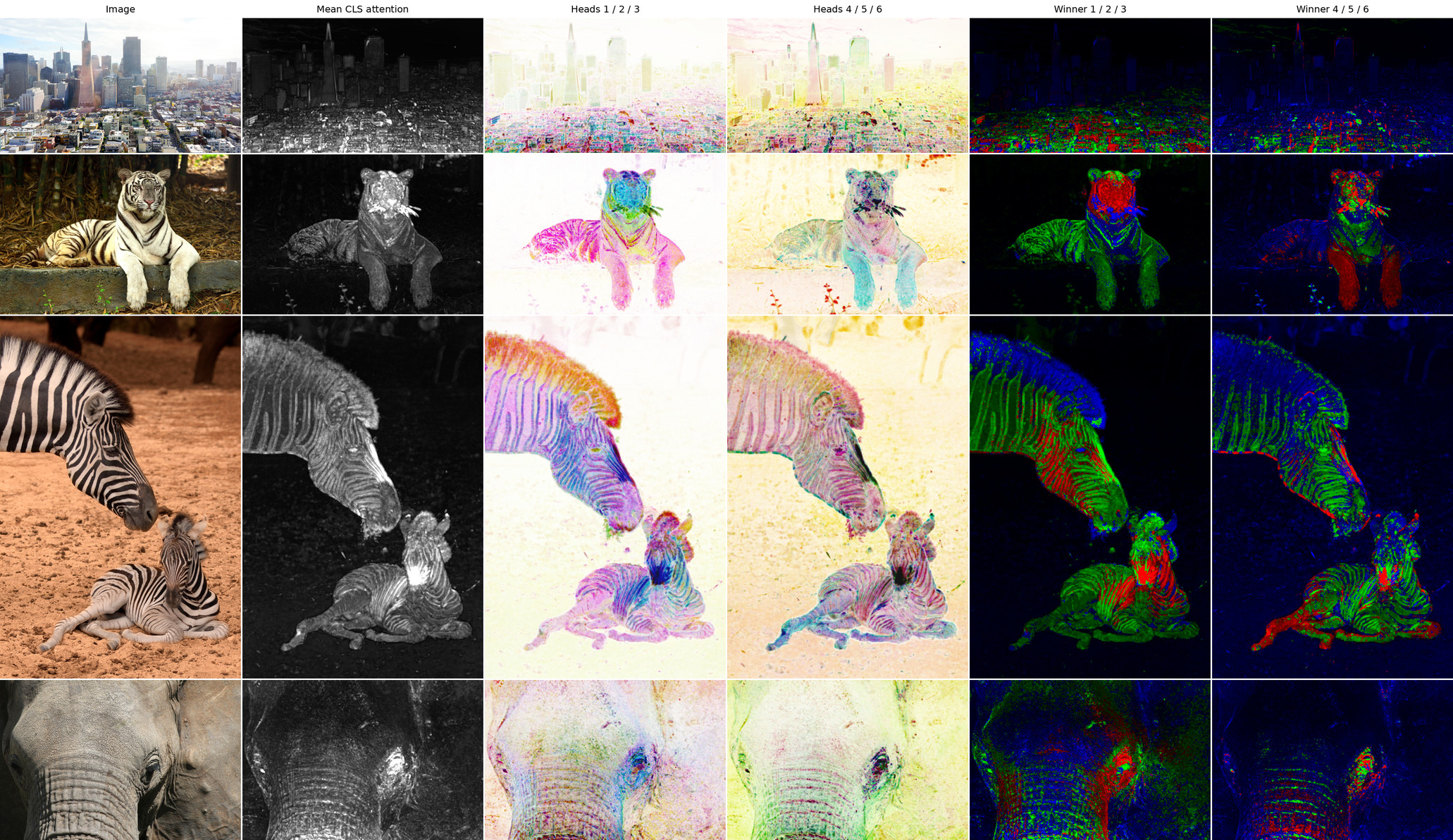}}{\includegraphics[width=\textwidth]{gallery/composed/div2k_heads_dominant_native_1.png}}
\caption{\textbf{Native DIV2K attention.} Frozen DINO ViT-S/16 read by \SF{DINO}. The columns show the image, mean CLS attention, white-background composites of heads 1--3 and 4--6, and the winning head within each triplet (colours: \cref{sec:vizattn}). The rows show the San Francisco skyline, a white tiger, zebras and an African elephant.}
\label{fig:div2k_dominant_1}
\end{figure}
\begin{figure}[H]
\centering
\IfFileExists{appendix_canonical_20261014/div2k_heads_2.jpg}{\includegraphics[width=\textwidth]{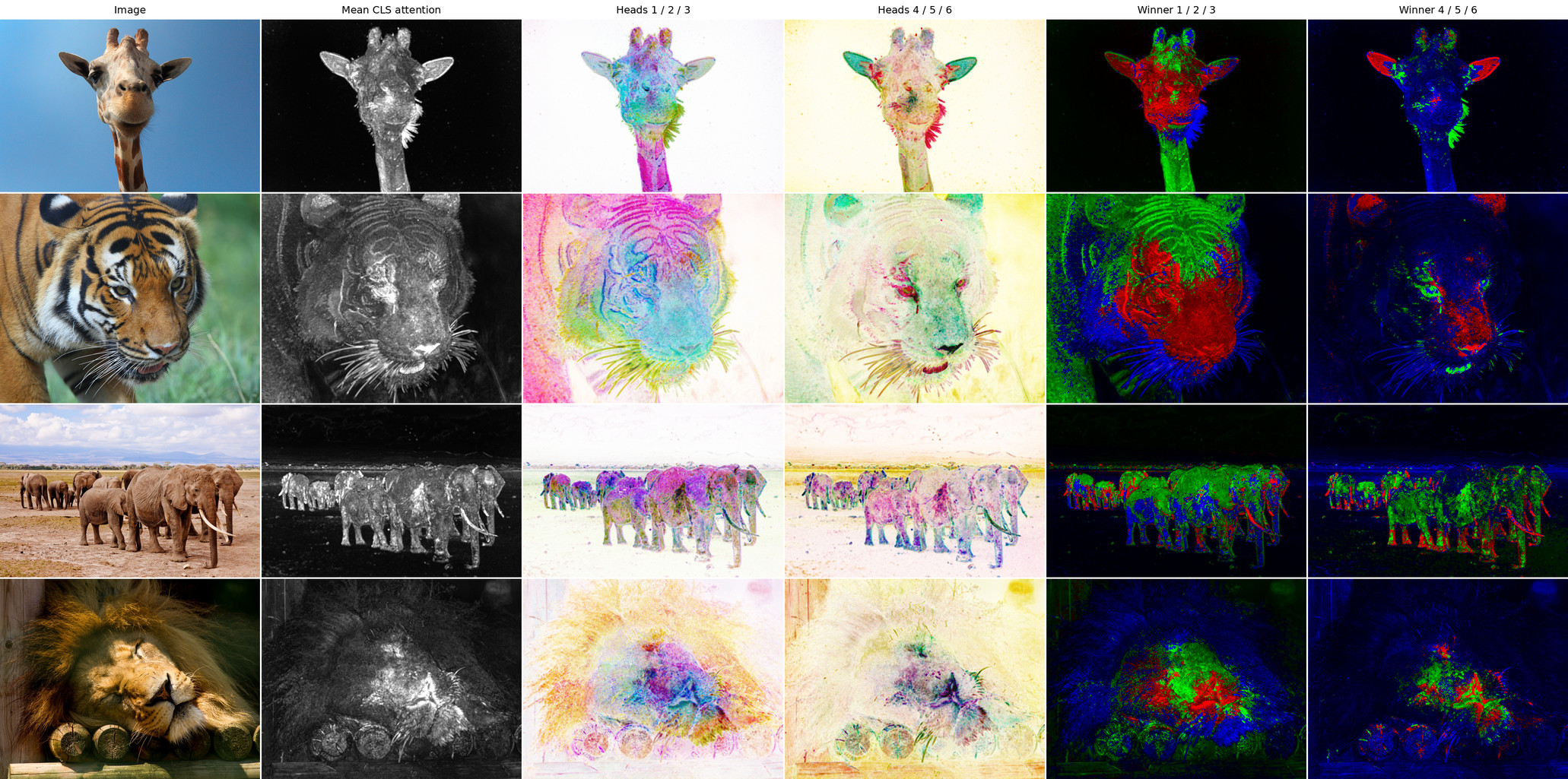}}{\includegraphics[width=\textwidth]{gallery/composed/div2k_heads_dominant_native_2.png}}
\caption{\textbf{Native DIV2K attention.} Frozen DINO ViT-S/16 read by \SF{DINO}; columns as in \cref{fig:div2k_dominant_1}. Rows: giraffe, tiger, elephants and lioness.}
\label{fig:div2k_dominant_2}
\end{figure}

\begin{figure}[H]
\centering
\IfFileExists{appendix_canonical_20261014/div2k_heads_3.jpg}{\includegraphics[width=\textwidth]{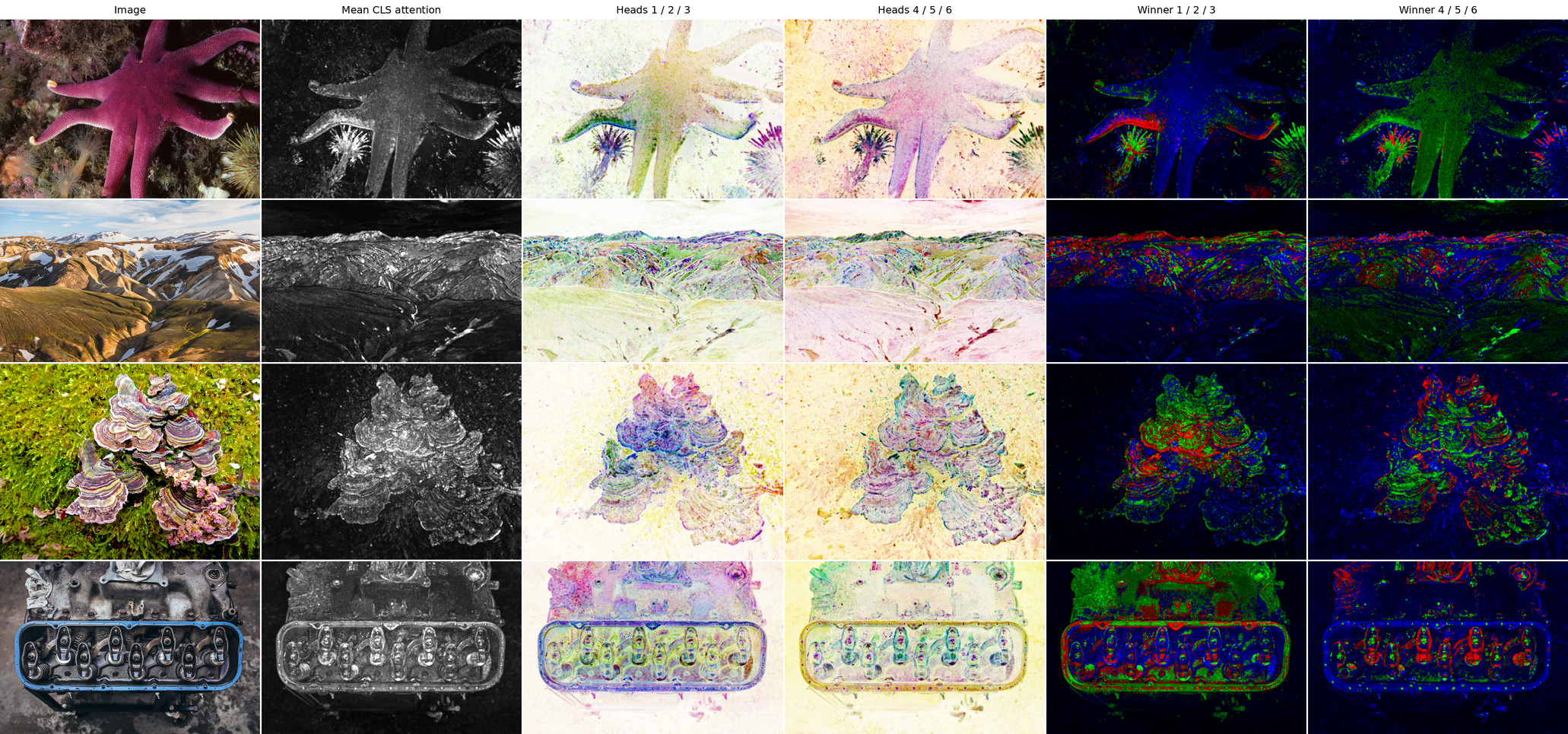}}{\includegraphics[width=\textwidth]{gallery/composed/div2k_heads_dominant_native_3.png}}
\caption{\textbf{Native DIV2K attention.} Frozen DINO ViT-S/16 read by \SF{DINO}; columns as in \cref{fig:div2k_dominant_1}. Rows: starfish, Icelandic landscape, bracket fungi and engine.}
\label{fig:div2k_dominant_3}
\end{figure}
\begin{figure}[H]
\centering
\IfFileExists{appendix_canonical_20261014/div2k_heads_4.jpg}{\includegraphics[width=\textwidth]{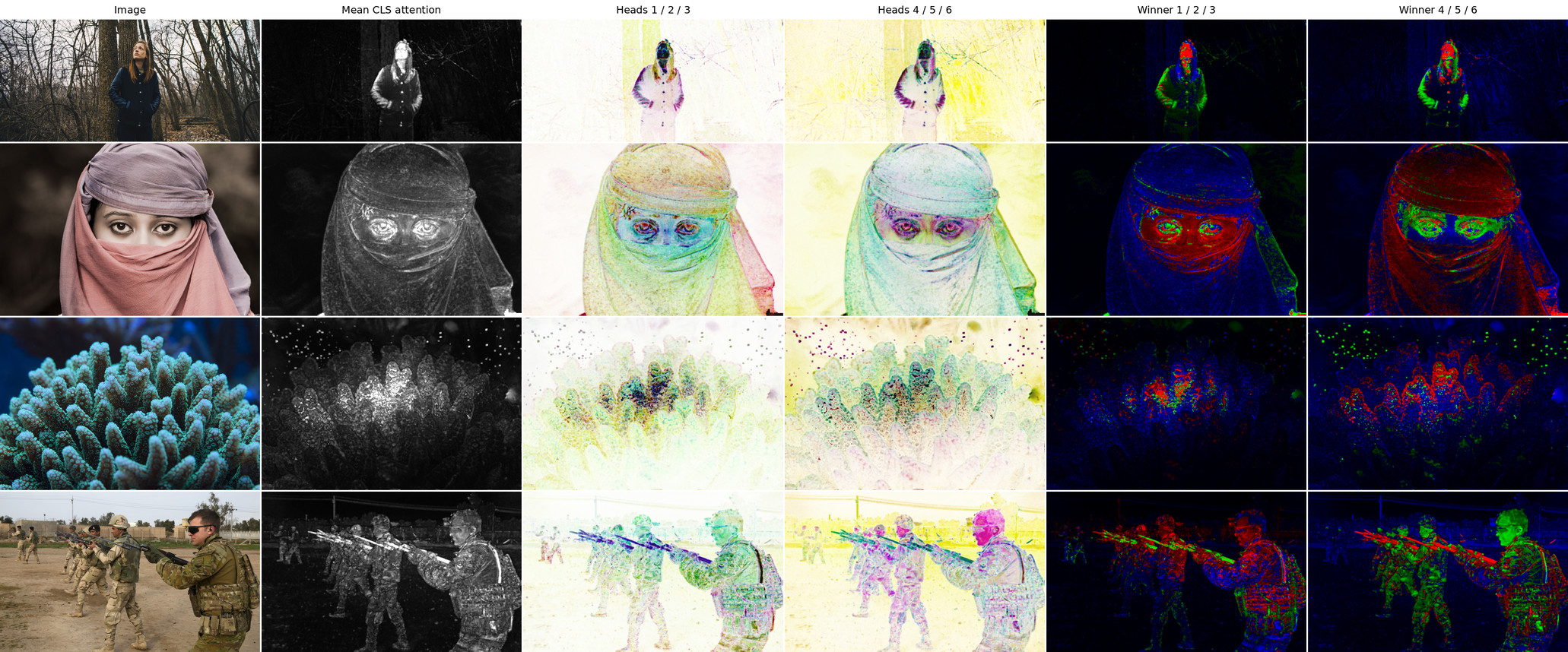}}{\includegraphics[width=\textwidth]{gallery/composed/div2k_heads_dominant_native_4.png}}
\caption{\textbf{Native DIV2K attention.} Frozen DINO ViT-S/16 read by \SF{DINO}; columns as in \cref{fig:div2k_dominant_1}. Rows: woman in forest, veiled portrait, coral reef and marching soldiers.}
\label{fig:div2k_dominant_4}
\end{figure}

\begin{figure}[H]
\centering
\IfFileExists{appendix_canonical_20261014/div2k_heads_5.jpg}{\includegraphics[width=\textwidth]{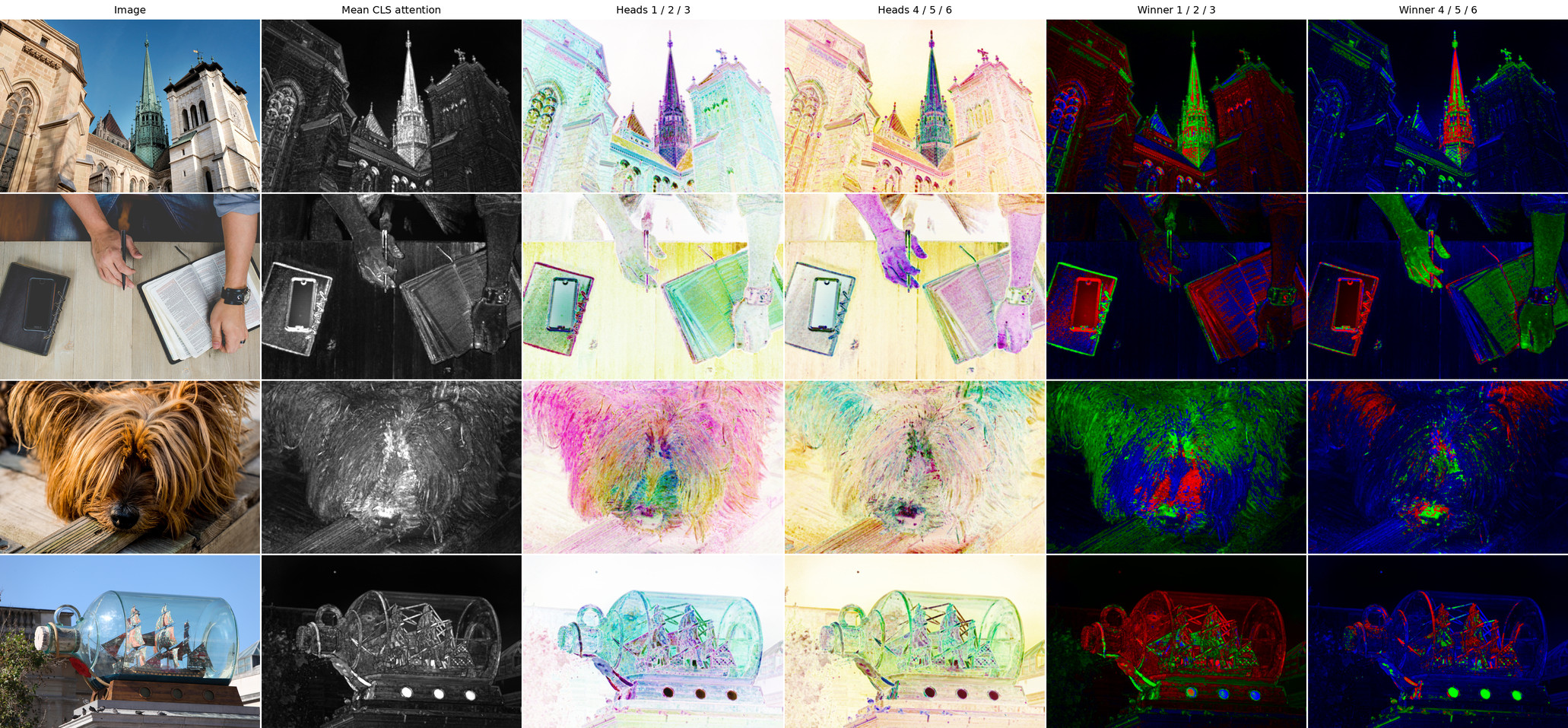}}{\includegraphics[width=\textwidth]{gallery/composed/div2k_heads_dominant_native_5.png}}
\caption{\textbf{Native DIV2K attention.} Frozen DINO ViT-S/16 read by \SF{DINO}; columns as in \cref{fig:div2k_dominant_1}. Rows: cathedral, person writing at a desk, Yorkshire terrier and ship in a bottle.}
\label{fig:div2k_dominant_5}
\end{figure}
\begin{figure}[H]
\centering
\IfFileExists{appendix_canonical_20261014/div2k_heads_6.jpg}{\includegraphics[width=\textwidth]{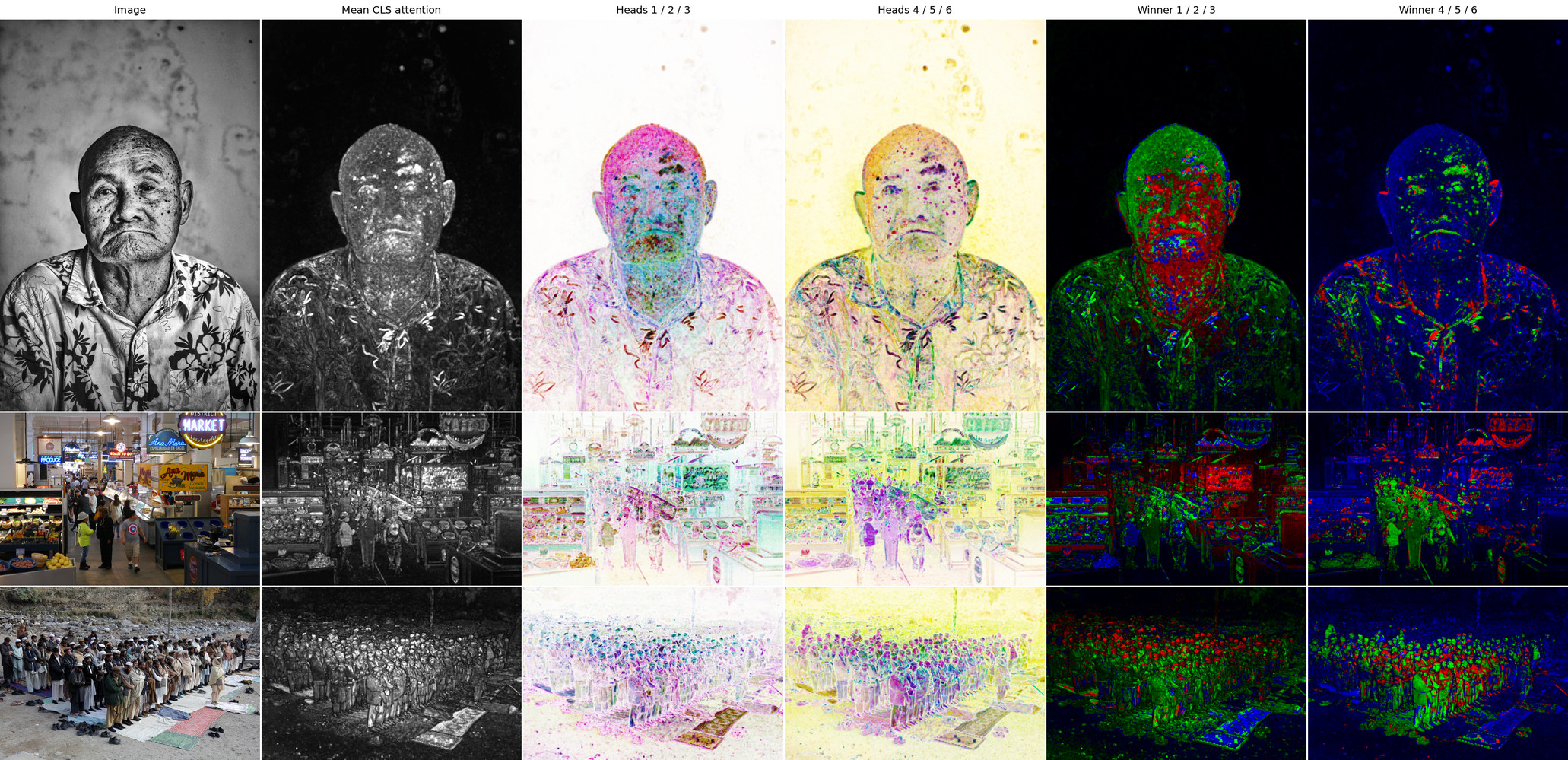}}{\includegraphics[width=\textwidth]{gallery/composed/div2k_heads_dominant_native_6.png}}
\caption{\textbf{Native DIV2K attention.} Frozen DINO ViT-S/16 read by \SF{DINO}; columns as in \cref{fig:div2k_dominant_1}. Rows: elderly man's portrait, indoor market and men praying outdoors.}
\label{fig:div2k_dominant_6}
\end{figure}
\endgroup
\subsection{Four backbones, four readouts}\label{sec:vizattn}
\begin{figure}[H]
\centering
\colheads{\textwidth}{9}{Image&\shortstack{ViT-S\\$k$-means}&\shortstack{\SF{ViT-S}\\$k$-means}&\shortstack{DeiT-Small\\evidence\\regions}&\shortstack{\SF{DeiT-S}\\evidence\\regions}&\shortstack{DINO\\class-token\\attention}&\shortstack{\SF{DINO}\\class-token\\attention}&\shortstack{DINOv2\\feature PCA}&\shortstack{\SF{DINOv2}\\feature PCA}}
\input{cvpr4_revision_20261027/network_figure.tex}
\caption{\textbf{Feature and attention readouts from four frozen backbones.} The tandem (top) and the wine bottle (bottom) are shown unresized ($500\times366$ and $463\times500$, width$\times$height), each with the donor-grid response followed by SeeFine at pixel resolution. ViT-S uses $k$-means ($k=5$) of features. DeiT-Small shows its direct relative-logit attribution over evidence regions from Ward clustering of frozen DINO features. DINO shows class-token attention averaged over heads, and DINOv2 shows PCA colours of final features. Clusters, attention and PCA colours follow structure inside the donor's coarse cells, because SeeFine evaluates the frozen embedding and transformer at every dense location.}
\label{fig:network_same}
\end{figure}

The four frozen backbones in \cref{fig:network_same} expose complementary aspects of the same image, and their colours encode different quantities. PCA colours show directions of feature variation, and cluster colours indicate membership. In the attention galleries of \cref{app:div2k}, colour identifies a CLS head.

\paragraph{Head-colour displays.} Each head's attention map is divided by its own $p_{99.5}$ and clipped to $[0,1]$. White-background composites display $1-0.92(z_R,z_G,z_B)$ for the indicated head triplet. Winner maps assign red, green or blue to the largest normalized head of the triplet, and their brightness follows the triplet sum divided by the $p_{99.5}$ of that sum. \Cref{fig:div2k_desk_heads_main} uses the same convention with four heads, adding yellow for head 4.

\paragraph{Evidence regions.} Regions come from Ward clustering of frozen DINO features and are labelled with partial direct-path shares from DeiT-S. A signed region contribution is divided by the selected class's actual signed relative logit. Shares can be negative or exceed $100\%$, and a negative denominator reverses their sign relative to the numerator. The shares cover the direct attention paths, and the omitted paths form a residual that is reported separately.
\input{cvpr4_revision_20261027/direct_path_definition.tex}

\subsection{Pixel attribution with DAVE}\label{app:dave}
DAVE~\cite{wrobel2026dave} decomposes input gradients to stabilize pixel attribution. We apply it to frozen DeiT-S and to \SF{DeiT-S}, using the same RGB crop and target class. Both produce input-pixel attributions; SeeFine changes the internal feature grid to one spatial token per pixel.

\begin{figure}[H]\centering
{\small\setlength{\tabcolsep}{0pt}\begin{tabular}{@{}*{5}{>{\centering\arraybackslash}p{.2\textwidth}}@{}}
Image & DAVE & DAVE + PP & DAVE on SeeFine & DAVE + PP on SeeFine
\end{tabular}}\par\vspace{2pt}
\includegraphics[width=\textwidth]{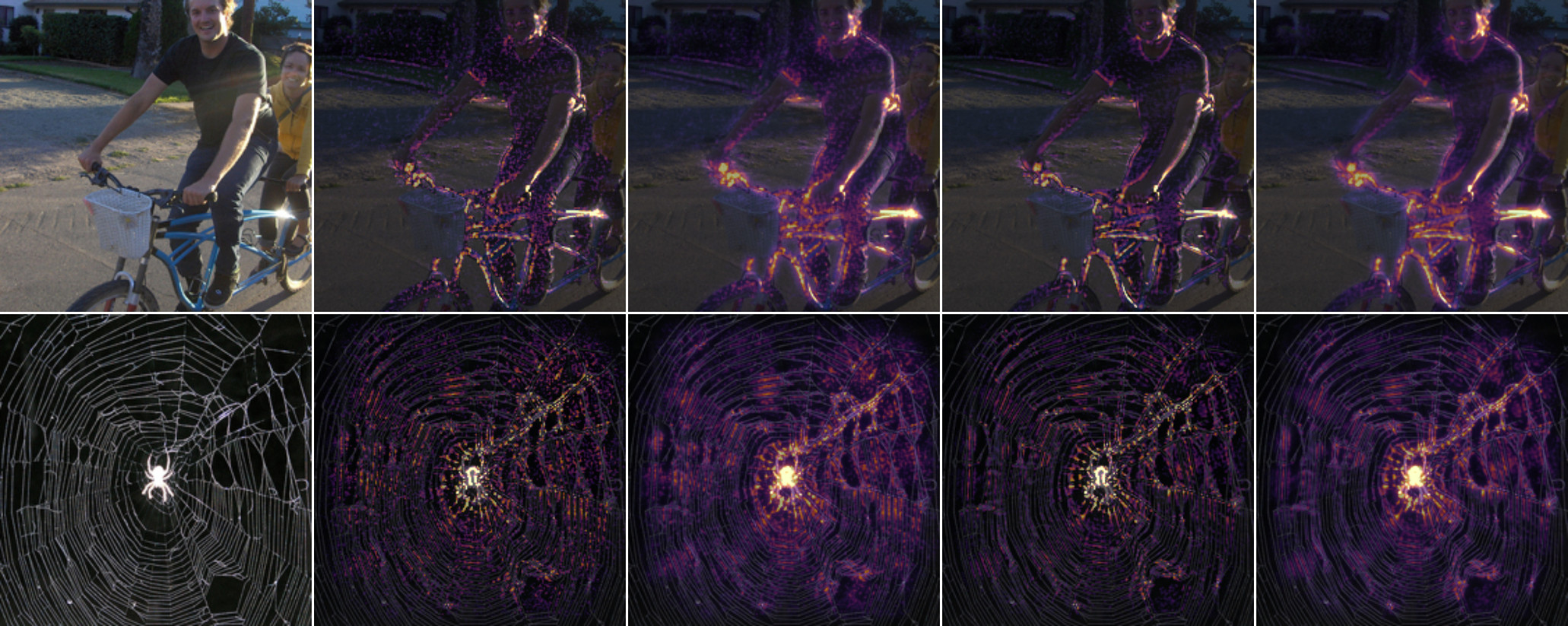}
\caption{\textbf{DAVE on the same photograph and frozen weights.} Columns show the RGB image, raw DAVE and official post-processing on DeiT-S, then the corresponding DAVE maps on \SF{DeiT-S}. Within each row, the four maps use the same $224\times224$ ImageNet crop and DeiT-S top-1 class, with 50 samples per run. Each positive-attribution map is scaled by its own 99.5th percentile and shown as a blend of $40\%$ RGB and $60\%$ heatmap.}\label{fig:canonical_dave_1}
\end{figure}

\begin{figure}[H]\centering
{\small\setlength{\tabcolsep}{0pt}\begin{tabular}{@{}*{5}{>{\centering\arraybackslash}p{.2\textwidth}}@{}}
Image & DAVE & DAVE + PP & DAVE on SeeFine & DAVE + PP on SeeFine
\end{tabular}}\par\vspace{2pt}
\includegraphics[width=\textwidth]{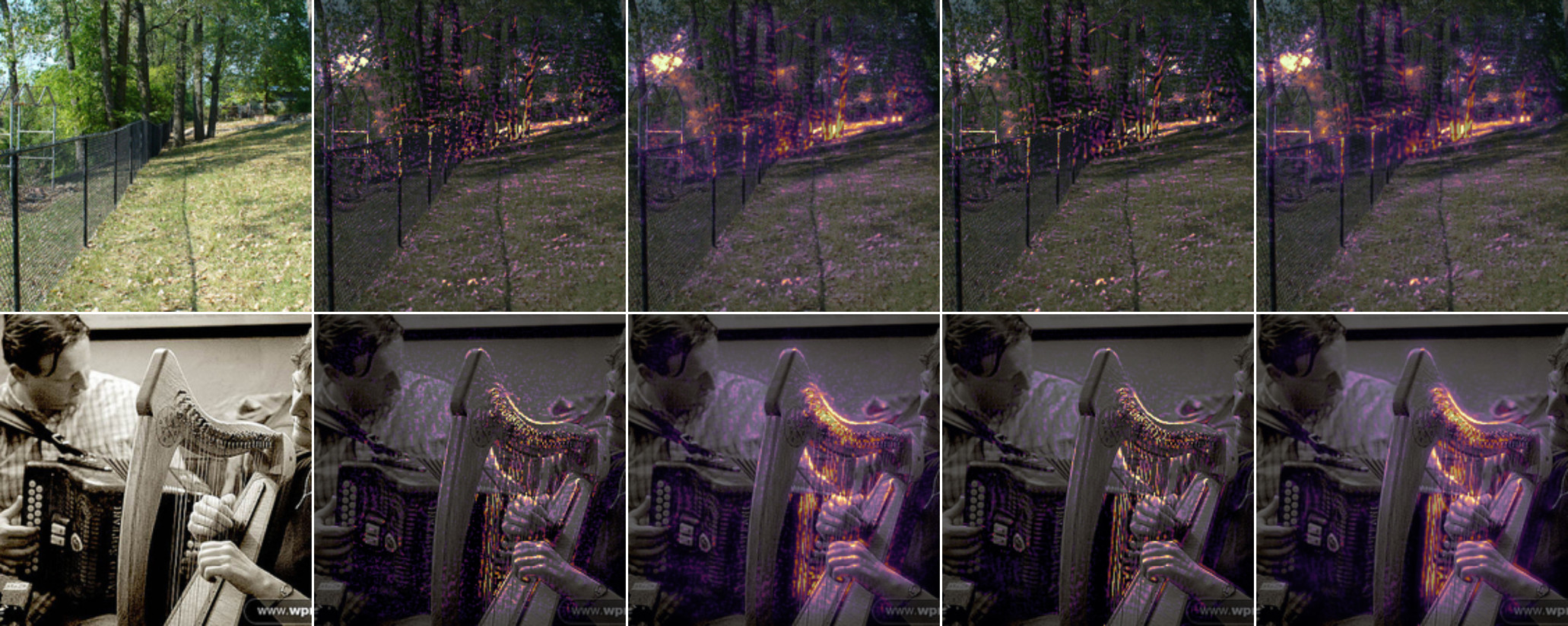}
\caption{\textbf{DAVE on the same photograph and frozen weights, continued.} Two additional ImageNet examples under the protocol and display of \cref{fig:canonical_dave_1}.}\label{fig:canonical_dave_2}
\end{figure}

\clearpage\twocolumn
\section{Runtime and memory}\label{app:runtime}\label{sec:cost}
Phase attention avoids full pixel-to-pixel attention at the million-token scale. For $N=HW$ patch tokens and patch side $P$, it reduces the quadratic patch term from $O(N^2)$ to $O(N^2/P^2)$, while the class query, projections, MLPs and embedding remain linear in $N$. The following forward and backward timings compare the cost of the dense readout with the donor grid.

\paragraph{End-to-end visual encoding.}
Across all 1,449 native VOC images, median visual-encoding time on one GH200 is $1.253$\,s for the recurrent dense call and $0.021$\,s for the ordinary donor, with all readouts of \cref{tab:talk2dinov3_native} timed in one run. Dense recurrence starts from the original initialized CLS/register tokens and requires no donor-state recording. Text encoding, decoding, transfer and rendering are excluded. Each dense visual representation is computed once per image and reused across text queries.

\paragraph{Single-CLS microbenchmark.} \Cref{tab:dino_canonical_runtime} compares the same DINO checkpoint on a $224\times224$ view and the unresized $2040\times1452$ DIV2K photograph 0678. \SF{DINO} takes $29.32$\,ms and $6.33$\,s, respectively, including both corrections. The latter evaluates $2{,}962{,}080$ pixel tokens.
\IfFileExists{cvpr5_revision/runtime_results_v5.tex}{\input{cvpr5_revision/runtime_results_v5.tex}}{}

\input{cvpr5_revision/gradient_cost_v5.tex}

\paragraph{Reusing geometry.} Phase counts, masks, position grids and analytic log shifts are independent of layer activations. They are computed once per image shape and reused throughout the forward pass. Group batching and fused attention reduce overhead without changing the attention rule.
\FloatBarrier

\section{Extended Related Work}\label{app:related}\label{sec:related}

\paragraph{Dense features and feature refinement.} A direct way to obtain dense features from a frozen ViT is to reduce its patch-embedding stride~\cite{amir2022deep}. DINO-Tracker adds a learned residual to half-stride DINOv2 features~\cite{tumanyan2024dinotracker}. FeatUp~\cite{fu2024featup} treats dense-stride inference as an impractical baseline and instead learns an image-guided feature upsampler. LiFT, LoftUp, JAFAR, AnyUp, NAF and FeatSharp likewise learn to refine coarse features~\cite{suri2024lift,huang2025loftup,couairon2025jafar,wimmer2026anyup,chambon2026naf,ranzinger2025featsharp}, and a training-free alternative uses test-time optimisation~\cite{seo2026upsample}. ViT-Up~\cite{wandel2026vitupfaithfulfeatureupsampling} is particularly close in its use of backbone-derived queries and intermediate ViT states: it reuses the patch embedding, then learns coordinate-conditioned cross-attention and sub-token feature modulation through multi-scale feature distillation, and its main configuration also adapts the backbone with LoRA. Docherty et al.~\cite{docherty2024upsampling} obtain dense frozen features by transforming the input, resizing the resulting feature maps, undoing each transform and averaging in source coordinates; shifts and flips can be combined with a reduced patch-embedding stride. This is a close training-free starting point. SeeFine instead integrates phase states inside the transformer through a recurrent shared CLS and analytic key-mass calibration. Distillation of sliding-window inference~\cite{kombol2026spar} and models trained directly on pixel tokens~\cite{nguyen2025image,wang2025scaling} obtain dense features through training. SeeFine starts from the particular checkpoint to be inspected, with no original training dataset assumed available. It keeps every learned weight fixed, evaluates the original patch embedding densely and constructs the dense readout without an additional learned component, backbone adaptation or per-image fitting.

\paragraph{Token mass and attention patterns.} Changing token count also changes attention statistics. FlexiViT, ResFormer and NaViT train for multiple patch sizes or resolutions~\cite{beyer2023flexivit,tian2023resformer,dehghani2023navit}. FixRes studies object-scale shift~\cite{touvron2019fixing}, and Win-Win tunes softmax temperature for high-resolution ViTs~\cite{leroy2024win}. Token merging adds the log mass of merged tokens to their logits~\cite{bolya2023token}. SeeFine applies the same mass-weighting principle to a class token whose relative mass would otherwise shrink among the many spatial keys. This shifts the global-key logits while leaving the spatial-key logits unchanged, whereas a temperature rescales every logit~\cite{chiang2022overcoming,nakanishi2025scalable,anson2025scale,peng2024yarn}. Sparse fixed-spacing attention is familiar in trained architectures~\cite{child2019generating,tu2022maxvit,huang2019interlaced,hassani2022dilated}. SeeFine imposes a related pattern on frozen weights, with phases chosen to match the donor's patch grid. Our calibration makes the global-to-spatial sampling ratio explicit for each query type. A related token-dilution effect has been reported for diffusion transformers~\cite{liu2026tide}.

\paragraph{Shift sensitivity and interpretability.} Small-shift sensitivity in ViTs has been addressed with anti-aliasing, adaptive polyphase sampling and anchoring~\cite{azulay2019why,zhang2019making,chaman2021truly,ding2023reviving,rojasgomez2024making}, and averaging predictions over shifted crops is a standard test-time augmentation~\cite{shanmugam2021better}. SeeFine is related to these polyphase ideas, keeps the weights frozen and couples phases through one recurrent class token. \Cref{app:components_v4} compares it with independently evaluated phases. Our dense readout complements tools such as class-token attention~\cite{caron2021dino}, rollout~\cite{abnar2020quantifying}, relevance propagation~\cite{chefer2021transformer,chefer2021generic}, Grad-CAM and Eigen-CAM~\cite{selvaraju2020gradcam,muhammad2020eigencam}, perturbation maps~\cite{petsiuk2018rise} and token-level decompositions~\cite{gandelsman2024interpreting}, which are ordinarily evaluated on the donor grid. DAVE instead stabilises pixel-level attribution by decomposing input gradients~\cite{wrobel2026dave}, and its attribution method and our denser execution are complementary (\cref{app:dave}). High-norm tokens in large ViTs~\cite{darcet2024vision} also motivate distinguishing feature magnitude from attention, and \cref{app:gallery} explains how to distinguish feature, attention and attribution views.
\clearpage

%% file: cvpr5_revision/appendix_protocol_index.tex
\begin{table*}[t]\centering\footnotesize
\caption{\textbf{Protocol index: which result uses which model and averaging unit.} All backbones are frozen. For NYUv2 we train a separate depth head for each readout; the other studies fit no new predictor weights. Full details and fixed controls appear in the corresponding sections.}
\label{tab:appendix_protocol_index_v4}\setlength{\tabcolsep}{3pt}
\begin{tabular}{@{}p{.18\textwidth}p{.23\textwidth}p{.28\textwidth}p{.25\textwidth}@{}}\toprule
Study & Images / split & Execution and readout & Reported unit\\\midrule
ImageNet & All 50,000 validation images for every classifier; $224^2$ & Canonical single CLS; original classifier & Image-level top-1\\
Talk2DINOv3 / VOC & All 1,449 validation images, native frames & Same text/mask readout; recurrent globals (UP); recorded globals as control (RP) & Pooled 21-label mIoU; paired image-bootstrap contrasts\\
Native five-backbone SPair & All 12,234 test pairs, 88,328 points & Recurrent CLS for four absolute-position backbones; recorded context for DINOv3-L; radius-seven pooling & Mean points within pair, then pairs\\
DINOv3-B / ViT-Up & Same full SPair split; 481 distinct photos & Recurrent globals and recorded-context controls; official 448, native/2$\times$; radius 8 & Point-micro within category, then equal-category macro\\
Affine correspondence & Fixed 100 ImageNet views, 32 points per view & Same frozen DINO, RGB field and matcher & Pixel errors; bootstrap resamples images\\
CUB landmarks & All 5,794 test images; identical $224^2$ crops & Frozen DINO-S/16 CLS attention, ranked area fixed & Image-macro recall; part rows pool visible landmarks\\
CT 3D correspondence & 35 lung scan pairs (3 datasets); 15 abdominal pairs, 13 organs & Frozen TAP-CT-B-3D and 3DINO; donor-pass globals; nearest-neighbour matching & Median landmark error (mm); correct-organ rate\\
BraTS 2D propagation & 4,563 target slices, 150 eligible cases & One labelled support per case; frozen DINO features & Case-macro Dice primary; slice-macro also shown\\
NYUv2 probe & 795 labelled training / 654 test; identical $224^2$ views & Frozen DINOv2; separately trained depth head per readout & SA and scale-and-shift-aligned metrics; one seed\\
\bottomrule\end{tabular}
\end{table*}

%% file: cvpr5_revision/reference_policy_v5.tex
\paragraph{Reference counts.}\label{app:reference_audit}
The reference count sets the nominal global-to-patch attention mass. A fixed reference uses the checkpoint's configured patch count, and an input-specific reference uses the donor's valid patch count on the given image. \Cref{tab:reference_policy_scope_v4} lists both counts for each evaluation setting. Recovery at donor anchors also requires the coordinate and key-support conditions in \cref{app:donor_contract_v4}.

\begin{table}[h]\centering\scriptsize
\caption{\textbf{Reference and donor patch counts by evaluation setting.} The reference count is used in the logit correction. The donor count is the number of patch keys in the ordinary pass on the same RGB input and can vary with native image size.}
\label{tab:reference_policy_scope_v4}
\setlength{\tabcolsep}{3pt}
\begin{tabular}{@{}lrr@{}}\toprule
Evaluation setting & Mass reference & Actual donor count\\\midrule
ImageNet/CUB/spatial, $224$, $P=16$ & 196 & 196\\
DINOv3--ViT-Up, $448$, $P=16$ & 784 & 784\\
NYUv2, $224$, $P=14$ & 256 & 256\\
Native SPair, DINO/ViT/DeiT & 196 & 240--1024\\
Native SPair, DINOv2 & 1369 & 322--1296\\
Native SPair, DINOv3-L & 784 & 209--961\\
Talk2DINOv3 VOC, UP/RP & $N_D$ & $N_D$\\
Talk2DINOv3 gallery figures & 784 & image-dependent\\
BraTS 2D, $240$, $P=16$ & 196 & 225\\
\bottomrule\end{tabular}
\end{table}

Native SPair ranges cover all 481 test photographs. Donors with absolute positional embeddings use reflection padding to reach the next patch multiple along each axis. The DINOv3 donor embeds only complete patches. Native DIV2K figures likewise use a fixed checkpoint reference. Fixed-reference and donor-preserving executions are reported as separate variants.

%% file: cvpr4_revision_20261027/donor_context_algorithm.tex
\begin{algorithm}[H]
\caption{Preservation control: recorded donor-global context}\label{alg:donor_context_v4}
\footnotesize
\begin{algorithmic}[1]
\Require Frozen model $M$ with $K$ global tokens, RGB image $x$, patch side $P$, reference count $\Nref$, key subsets and rotary coordinate domain
\Ensure Pixel-aligned local features; no update to learned parameters
\State Run the unmodified donor on $x$ with its original valid patch support
\State Record incoming global states $C^D_\ell(x)$ before every block $\ell$
\State $Z\gets\operatorname{PatchEmbed}(\operatorname{ReflectRGB}(x);\mathrm{stride}=1)$, with checkpoint normalization
\State Form phase groups $G$ and their real key subsets $J_G$; retain all source pixels as queries
\State Set $\Nref$ to the checkpoint count or to the valid donor count of $x$
\For{each frozen block $\ell$}
  \State Set $C\gets C^D_\ell(x)$, read-only; normalize $[C;Z]$
  \State Compute checkpoint $Q,K,V$, including original Q/K normalization and the chosen rotary coordinate domain
  \For{all phases $G$, independently}
    \State $d_G\gets\log(\max(1,|J_G|)/\Nref)$
    \State $A_G\gets\operatorname{Attn}(Q_G,[K_C;K_{J_G}],[V_C;V_{J_G}])$, adding $d_G$ to each of the $K$ global-key logits
  \EndFor
  \State Restore the source lattice; apply unchanged patch output projection, attention residual and MLP residual
  \State Discard global-query outputs; do not feed dense states back into the recorded donor branch
\EndFor
\State Return the final checkpoint-normalized $Z$
\end{algorithmic}
\end{algorithm}

\paragraph{Global attention used by the segmentation readout.}
For the attention cleaner only, the last incoming donor CLS query is also evaluated against the dense patch keys and all $K=5$ global keys. We add $\log(N_{\rm keys}/\Nref)$ to each global-key logit and mask invalid keys before softmax. This gives the cleaner a dense attention map from the read-only donor query. The VOC evaluation uses the donor patch count as reference, donor-support RoPE and complete-window keys (\cref{tab:talk_policy_v4}), whereas the gallery in \cref{app:talk2dinov3} uses reference 784, source-image RoPE and reflected boundary keys. Full input gradients propagate through the recorded global states to the RGB image.

%% file: cvpr5_revision/donor_anchor_contract_v5.tex
\paragraph{When the dense readout preserves donor anchors.}\label{app:donor_contract_v4}
With read-only donor context, additional query phases need not disturb the original donor phase. The following proposition states the conditions under which the donor's features are preserved at its original patch centres.

\begin{proposition}[Donor-anchor preservation]\label{prop:donor_anchor_v4}
Fix the image geometry and frozen DINOv3 weights. Let $F_D(x)$ be the ordinary donor's normalized patch features, and let $\Pi_A F_S(x)$ select the corresponding original-patch anchors from the dense readout. Assume:
\begin{enumerate}\setlength{\itemsep}{1pt}
\item Dense anchor embeddings use exactly the donor's RGB patch windows, normalization and initial token states.
\item At every block, an anchor patch query reads exactly the donor's spatial keys/values and the incoming donor CLS/register states recorded from the same $x$.
\item Anchor coordinates, Q/K normalization, projections, residuals, MLPs and final token normalization match the donor.
\item The mass reference equals the number $N_D$ of donor patch keys. The original phase contains exactly these $N_D$ keys, so its global-key shift is $\log(N_D/N_D)=0$.
\end{enumerate}
Then, in exact arithmetic,
\begin{equation}
\Pi_A F_S(x)=F_D(x).
\end{equation}
If the recorded donor states retain their dependency on $x$, every identical fixed linear probe $v$ also satisfies
\begin{equation}
\nabla_x\langle v,\Pi_A F_S(x)\rangle
=\nabla_x\langle v,F_D(x)\rangle
\end{equation}
where the derivatives exist.
\end{proposition}

\begin{proof}
The initial anchor states agree by the first condition. Suppose they agree at the input to a block. The spatial keys/values are then identical by induction, and the recorded incoming global states are the donor's own states. The same coordinates, zero anchor-phase correction and checkpoint operations give the same attention output, residual and tokenwise MLP update. Anchor states therefore agree at the next block. Induction and the shared final token normalization establish the feature identity. This identity holds as a function of the same RGB variable, so differentiating the identical scalar probe gives the gradient identity.
\end{proof}

The recorded-context control (RP) realizes these conditions with the actual donor count, donor-support RoPE and complete-window keys. Reflected border-centred windows still receive dense \emph{query} outputs, but do not introduce extra keys into the donor phase. To recover the donor's input gradients, we differentiate through the recorded globals. Detaching them gives the derivative conditional on those states. The feature identity holds at the original patch centres, and the gradient identity holds for fixed linear probes of the features at those centres. The measurements below test both in floating point.

%% file: cvpr5_revision/anchor_agreement_v5.tex
\FloatBarrier
\paragraph{Agreement at donor patch centres.}\label{app:anchor_agreement}
With recorded donor context, SeeFine reproduces DINOv3 features at the donor patch centres to a mean relative error of $1.4\cdot10^{-6}$ and the tested input gradients to within $0.1\%$ (\cref{tab:global_feedback_agreement_v4}), under the conditions of \cref{prop:donor_anchor_v4}. For DeiT-III-S/16, \cref{tab:anchor_agreement_v4} tests how patch-window alignment, positional sampling and the reference count affect agreement under recurrent single-CLS execution. For DINOv3, \cref{tab:global_feedback_agreement_v4} measures donor agreement with recorded and recurrent globals. The two constructions are evaluated separately because frozen weights alone do not imply the same input--output function.

\paragraph{Matched tensors and sampling.}
We fix 24 randomly drawn ImageNet validation images before scoring, without correctness or confidence filtering. Each image is resized to short side 256 and centre-cropped to $224\times224$. A second condition bilinearly interpolates that crop to $448\times448$. Within each condition all executions receive the identical RGB tensor, normalization, frozen checkpoint and classification head. The larger condition adds no new source-image information. The checkpoint reference count remains $N_{\rm ref}=196$ in both conditions.

For patch side $P=16$, the dense locations $(P/2+Pj,P/2+Pk)$ use exactly the original non-overlapping RGB patch windows, with the implementation's even-window anchoring convention. We compare final normalized patch features at these locations, final CLS features and logits. Input gradients target (i) the same donor top-1 class logit, held fixed across variants, and (ii) the same fixed Gaussian linear projection of all aligned final patch features. The gradient of this shared projection with respect to the input is a vector--Jacobian product (VJP). We report cosine similarity and relative $L_2$ error normalized by the donor norm. We also compare the features entering each block to track agreement through depth. All computations use FP32 with TF32 disabled. Parameters stay frozen, and gradients are computed only with respect to the RGB input.

\paragraph{Diagnostic controls.}
The \emph{anchor-only recovery control} first evaluates the patch embedding densely, retains only the original donor phase, adds the original donor positional embeddings, and executes the unchanged full-attention blocks and CLS. It tests indexing and window alignment only. An FP64 version using the original timm blocks on both sides checks numerical recovery independently of FP32 precision. The \emph{anchor-position control} retains every SeeFine phase and the shared CLS but samples positions so that the embeddings at donor anchors exactly match the donor. Comparing this variant with canonical SeeFine isolates the effect of positional sampling. Residual disagreement cannot be attributed solely to the patch windows.

\paragraph{Numerical recovery controls.}
The FP64 recovery control uses the same original blocks and precision on both sides and recovers both forward outputs and input gradients. In FP32, the mean feature discrepancy is below $6\cdot10^{-6}$ and the mean gradient discrepancy below $1.2\cdot10^{-4}$ in relative $L_2$.

\begin{table}[h]\centering\footnotesize
\caption{\textbf{Donor agreement under recurrent DeiT-III execution.} Mean per-image cosine similarity with the frozen DeiT-III-S/16 donor over 24 fixed ImageNet views. The class gradient uses the donor's top-1 logit for every variant, and the feature VJP uses a shared Gaussian probe. ``Input ref.'' sets the reference to the actual donor count (196 or 784). The ``+ aligned positions'' rows match the donor's positional embeddings at the donor patch centres.}
\label{tab:anchor_agreement_v4}
\setlength{\tabcolsep}{3pt}
\begin{tabular}{@{}lrccc@{}}\toprule
Execution & Size & Features & Class grad. & Feature VJP\\\midrule
Anchor-only recovery & 224 & 1.0000 & 1.0000 & 1.0000\\
Canonical, ref. 196 & 224 & 0.9974 & 0.2324 & 0.7404\\
+ aligned positions & 224 & 0.9981 & 0.2337 & 0.8362\\
Anchor-only recovery & 448 & 1.0000 & 1.0000 & 1.0000\\
Fixed ref. 196 & 448 & 0.8278 & 0.1789 & 0.0377\\
+ aligned positions & 448 & 0.8279 & 0.1789 & 0.0323\\
Input ref. 784 & 448 & 0.9992 & 0.3022 & 0.8456\\
+ aligned positions & 448 & 0.9995 & 0.3023 & 0.9252\\
\bottomrule\end{tabular}
\end{table}

At $224\times224$, reference matching leaves the canonical execution unchanged. At $448\times448$, using the donor count $784$ as reference raises anchor-feature cosine from $0.8278$ to $0.9992$, and aligned positions raise it to $0.9995$. The recurrent CLS aggregates all phases, so its class gradients describe the integrated predictor and differ from the donor's.

\paragraph{DINOv3: preservation versus recurrent integration.}
The two global-state variants separate donor recovery from the construction of one integrated model. Both hold the checkpoint, input-specific reference, donor-support RoPE and complete-window keys fixed. The recorded-context control (RP) supplies the donor's incoming CLS/register states and retains their full RGB dependency. Recurrent SeeFine (UP) updates the original CLS and four registers from all dense phases, using the same incoming states within each block; it requires no donor forward.

\paragraph{Feature and gradient probes.}
Forward tests use three prespecified SPair photographs at native and doubled-native dimensions (six cases). At donor patch centres, RP features agree with the donor to a relative $L_2$ error of $1.36\cdot10^{-6}$. UP has mean cosine $0.998998$ and mean relative error $4.04\%$. The backward test uses the same photographs at $224$-pixel centre crops and their bilinear $448$-pixel enlargements. Each case uses three fixed Gaussian projections of local features, giving 18 VJPs. The dense outputs are restricted to donor centres before each projection is applied.

\begin{table}[h]\centering\footnotesize
\caption{\textbf{Preservation and recurrent integration.} Mean feature and input-gradient agreement with frozen DINOv3-B. Feature errors use six cases at native or doubled resolution. Input VJPs use six crop cases at 224 or 448 pixels, with three projections per case shared across readouts. Errors are relative $L_2$ percentages, and cosine uses signed full-RGB gradients.}
\label{tab:global_feedback_agreement_v4}\setlength{\tabcolsep}{3pt}
\begin{tabular}{@{}lrrr@{}}\toprule
Global context & \shortstack{Feature\\error (\%)} & \shortstack{Input VJP\\cosine} & \shortstack{Input VJP\\error (\%)}\\\midrule
Recorded donor states & 0.000136 & 1.000000 & 0.0257\\
Recurrent CLS/registers & 4.04 & 0.963003 & 25.51\\
\bottomrule\end{tabular}\end{table}

\paragraph{Successful donor recovery under the preservation control.}
RP recovers both gradient direction and magnitude for the tested local-feature probes: the mean relative gradient error is $0.0257\%$ and the maximum approximately $0.1\%$. The mean cosine is $0.99999993$ (minimum $0.99999949$), and candidate/donor norm ratios range from $0.999974$ to $1.000126$.

\paragraph{Gradients of the integrated computation.}
With recurrent globals, dense patches influence the shared global states and the spatial tokens in later blocks. The gradients therefore describe the integrated function of a single predictor. UP's local-feature VJPs have mean signed cosine $0.9630$ to the donor and mean relative error $25.51\%$ ($25.35\%$ after fitting a scalar). The six CLS-projection VJPs have mean cosine $0.474$.

\paragraph{Numerical and memory checks.}
On a small input, grouped recurrence agrees with explicit full masked attention: feature relative error is $2.44\cdot10^{-6}$ and input-VJP cosine exceeds $0.99999998$. This verifies the recurrent implementation itself. At $448\times448$, the FP32 probe with MLP checkpointing reaches a peak allocated memory of $59.7$ GiB for RP and $80.1$ GiB for UP on a GH200. \Cref{app:gradient_cost_v4} reports the tested range of full input gradients separately.

\paragraph{Gradient direction and magnitude.}
The primary gradient-agreement metric is the cosine of the raw signed RGB input gradients, before visualization normalization. To distinguish collinearity from equality, we additionally report $\lVert g_{\rm SF}\rVert/\lVert g_D\rVert$ and the least-squares scale $\alpha^*=\langle g_D,g_{\rm SF}\rangle/\lVert g_D\rVert^2$, together with $\lVert g_{\rm SF}-\alpha^*g_D\rVert/\lVert g_{\rm SF}\rVert$. Positive collinearity gives cosine one even when magnitudes differ. We restrict output tokens \emph{before} the scalar reduction and use identical projection coefficients and normalization, so a different dense-token count cannot introduce a trivial sum-versus-mean scaling. These are gradients with respect to the full RGB input.

%% file: cvpr5_revision/component_v5.tex
\FloatBarrier
\subsection{Shared class token and mass correction}\label{app:components_v4}
We isolate mechanisms of the single-CLS construction using one frozen DINO ViT-S/16 checkpoint, a fixed reference of 196, identical dense RGB windows and one positional field. We defined the variants on 100 CUB training images and 36 SPair validation pairs and then evaluated them on the full test splits.

The controls are: ordinary donor (D), calibrated donor (DC), independent CLS per phase without/with correction (I0/I1), and one recurrent shared CLS with neither correction (S0), patch-query correction only (SP), CLS-self correction only (SC), or both (SB). Independent phases receive identical initial CLS states and update each only within its phase. Their CLS attention maps are interleaved as an equal-weight phase ensemble.

At $224\times224$, each phase has 196 keys, giving exact duplicate executions D=DC, I0=I1, S0=SP and SC=SB. \Cref{tab:components_cub_full} therefore lists four rows. Native SPair has variable phase sizes and identifies the patch-query correction. This study interpolates grid features to the pixel lattice before cosine matching, whereas \cref{tab:spair_main} interpolates the cosine maps of grid readouts, so the scores of the two tables are not directly comparable.

\begin{table}[h]\centering\footnotesize
\caption{\textbf{Full CUB test: 5,794 images.} Image-macro visible-landmark recall (\%), using mean-head attention and the same 224-pixel crop and annotations. Columns fix the ranked area at 5/10/20\%.}
\label{tab:components_cub_full}
\setlength{\tabcolsep}{4pt}
\begin{tabular}{@{}lrrr@{}}\toprule
Execution & 5\% & 10\% & 20\%\\\midrule
Donor (D, DC) & 43.31 & 63.49 & 84.11\\
Independent phases (I0, I1) & 49.63 & 70.80 & 89.77\\
Shared CLS, uncorrected (S0, SP) & 49.46 & 70.45 & 89.62\\
Shared CLS, corrected (SC, SB) & 49.52 & 70.77 & 89.80\\
\bottomrule\end{tabular}\end{table}

\begin{table}[h]\centering\footnotesize
\caption{\textbf{Full native SPair component study.} Pair-mean PCK (\%) on all test pairs, using the same frozen DINO ViT-S/16. This averaging differs from the category-macro DINOv3/ViT-Up table.}
\label{tab:components_spair_full}
\setlength{\tabcolsep}{4pt}
\begin{tabular}{@{}lrr@{}}\toprule
Execution & PCK@0.05 & PCK@0.10\\\midrule
D: ordinary donor & 17.71 & 30.36\\
DC: calibrated donor & 17.27 & 29.77\\
I0: independent, no correction & 19.33 & 31.05\\
I1: independent, corrected & 18.62 & 30.37\\
S0: shared, no correction & 19.38 & 31.08\\
SP: shared, patch correction & 18.81 & 30.47\\
SC: shared, CLS correction & 19.27 & 30.93\\
SB: shared, both corrections & 18.63 & 30.37\\
\bottomrule\end{tabular}\end{table}

\paragraph{Results of the full ablation.}
On CUB, shared phases with both corrections (SB) gain $7.28$ points over the donor (paired image-bootstrap 95\% interval $[6.95,7.59]$) and $0.314$ points over the uncorrected shared variant (S0; $[0.146,0.486]$). On native SPair, SB scores $0.758$ PCK@0.05 points below S0 ($[-0.938,-0.578]$). Independent phases with the same corrections (I1) give nearly identical aggregate scores: SB minus I1 is $-0.036$ points on CUB ($[-0.128,0.051]$) and $+0.0023$ points on SPair (paired whole-pair interval $[-0.038,0.044]$), and their SPair predictions differ at 3.12\% of locations. The shared construction integrates the phase states into one predictor and one backward graph, and token-mass calibration determines how each query's key count enters that computation. The SPair intervals describe the fixed split, in which photographs recur across pairs.

\paragraph{Communication, gradients and explanation readouts.}
Each independent phase samples the entire photograph at patch spacing and summarizes that coarse view with its own CLS. The phase states evolve separately inside the transformer, so hidden tokens cannot exchange information between phases. A recurrent shared CLS adds an indirect path $G_a\to\mathrm{CLS}\to G_b$ across successive blocks, and its backward graph contains the corresponding cross-phase paths. This path integrates the phase computations inside one recurrent predictor. Independent phases also yield input gradients, because overlapping RGB windows share source pixels.

\paragraph{Attribution for independent phases.}
Independent phase outputs do not share a single global CLS computation, so attribution methods need an aggregation rule: a scalar target, a way to combine phase-level explanations and a mapping back to source coordinates. For a classifier with frozen head $h$, one possible global target is the mean of phase logits,
\begin{equation}
s_c(x)=\frac{1}{|\mathcal G|}\sum_{g\in\mathcal G}h(\operatorname{LN}(c_g(x)))_c.
\end{equation}
Input gradients and Grad-CAM can differentiate this scalar. Grad-CAM additionally specifies the spatial activation layer, gradient pooling within and across phases and the assembly of the map in source coordinates. Attention rollout specifies the starting mass across CLS tokens, for example a uniform mixture, and gradient-weighted rollout uses the same scalar target. These choices define which ensemble computation the resulting map describes.

In the DINOv3 recorded-context control, every dense phase reads the same globals recorded from the donor on that image, and dense outputs do not update them. Its input gradients are computed through the recorded donor branch (\cref{app:anchor_agreement}).

When a classifier uses only the recorded donor CLS, its score does not depend on the dense activations, and Grad-CAM on them gives zero gradients. The DINOv3 test in \cref{app:anchor_agreement} therefore differentiates matched local-feature projections. In recurrent single-CLS SeeFine the score depends on the dense activations, and an independent-phase classifier can use the ensemble target above.

\paragraph{Computational implications.}
Both executions process all phases. For a mean-logit target, the phase gradients of independent branches can be accumulated separately by linearity, which reduces peak graph memory. A recurrent shared CLS couples the phases in a single backward graph with one internal global readout. Runtime and peak memory depend on how these backward computations are scheduled.

%% file: cvpr5_revision/talk_policy_v5.tex
\begin{table}[t]\centering\footnotesize
\caption{\textbf{Global-context policies on Talk2DINOv3.} Both variants use the same frozen checkpoint and geometry, with $N_D=\lfloor H/16\rfloor\lfloor W/16\rfloor$, and differ only in how the global tokens are supplied.}
\label{tab:talk_policy_v4}\setlength{\tabcolsep}{4pt}
\begin{tabular}{@{}lll@{}}\toprule
Setting & SeeFine (UP) & Recorded control (RP)\\\midrule
Global states & Recurrent & Recorded donor\\
Reference & $N_D$ & $N_D$\\
RoPE domain & Valid donor support & Valid donor support\\
Spatial keys & Complete windows & Complete windows\\
Full VOC mIoU & $66.56$ & $66.60$\\
\bottomrule\end{tabular}\end{table}

%% file: cvpr4_revision_20261027/cleaner_definition.tex
\paragraph{Exact score and cleaner definition.}
Let $f_i$ be the final visual descriptor, $t_k$ a unit projected text ensemble, and $A_{hi}$ the final post-RoPE CLS-to-patch probability for head $h$. With $N$ spatial samples, define
\begin{equation}
 \begin{aligned}
 a_{hi}&=\frac{\max(A_{hi},0)}{\max(\sum_j\max(A_{hj},0),\epsilon)},\\
 d_h&=\sum_i a_{hi}f_i,\\
 q_{kh}&=\operatorname{softmax}_h(\langle t_k,\widehat d_h\rangle).
 \end{aligned}
\end{equation}
A hat denotes unit normalization. The raw class score and head-weighted density are $p_{ki}=\operatorname{sigmoid}(\langle t_k,\widehat f_i\rangle)$ and $D_{ki}=N\sum_h q_{kh}a_{hi}$. Let $D_-,D_+$ and $p_-,p_+$ be the respective minima/maxima across all classes and positions of this image. The range mapping and blend are
\begin{equation}
 \begin{aligned}
 B_{ki}&=p_-+(p_+-p_-)\frac{D_{ki}-D_-}{\max(D_+,D_+-D_-,\epsilon)},\\
 \widetilde p_{ki}&=\frac{p_{ki}+0.2B_{ki}}{1.2},\qquad \epsilon=10^{-12}.
 \end{aligned}
\end{equation}
The denominator follows the released cleaner, and the extrema are taken over all classes and positions of one image. A pixel is background if its highest blended score, $\max_k\widetilde p_{ki}$, is below $0.55$. Otherwise it takes the class with the highest score. The raw control uses $p$ without the blend.

%% file: cvpr5_revision/voc_context_v5.tex
\paragraph{Recurrent versus recorded global tokens.}\label{app:voc_context_v4}
We isolate global-state feedback with the geometry held fixed. UP, the default SeeFine, updates the original CLS and four registers recurrently; RP reads their recorded donor states (\cref{alg:donor_context_v4}). Both use the same frozen checkpoint, reference, rotary domain, keys, text ensemble, threshold and readout (\cref{tab:talk_policy_v4}).
\begin{table}[H]\centering\footnotesize
\caption{\textbf{Recurrent and recorded global tokens on VOC2012.} Pooled 21-label mIoU (\%) on all 1,449 validation images. ``Full'' scores the entire frame, and ``Support'' scores the common rectangle of complete donor patches. Both use the fixed cleaner weight 0.2, and the raw columns use the scores before cleaning.}
\label{tab:voc_current_globals_v4}
\setlength{\tabcolsep}{4pt}
\begin{tabular}{@{}lrrrr@{}}\toprule
& Full & Raw & Support & Raw support\\\midrule
Donor & 61.19 & 53.49 & 61.42 & 53.81\\
Recorded RP & 66.60 & 58.46 & 66.76 & 58.74\\
Recurrent UP & 66.56 & 58.45 & 66.73 & 58.72\\
\bottomrule\end{tabular}\end{table}
Both variants improve full-frame mIoU by more than five points over the donor, and the difference between them is practically negligible: UP$-$RP is $-0.033$ points with the cleaner (paired 95\% image-bootstrap CI $[-0.063,-0.005]$) and $-0.013$ points without it (CI $[-0.042,+0.014]$). The gain of UP persists under the readout controls: $5.37$ points with the cleaner, $4.96$ without it and $5.31$ on common complete-patch support.

%% file: cvpr5_revision/spair_final.tex
\section{Semantic correspondence}\label{app:spair}
SPair-71k~\cite{min2019spair} tests correspondence across object instances. We evaluate all $12{,}234$ test pairs, $88{,}328$ keypoints and 18 categories on the native photographs, without resizing. PCK@$\tau$ accepts a prediction whose error is at most $\tau$ times the larger side of the target box. Pair-mean scores average points within each pair and then over pairs. The comparison with trained upsamplers pools points within each category and averages the category scores, following ViT-Up. Intervals use $2{,}000$ paired bootstrap resamples of whole pairs and describe this fixed split, in which photographs recur across pairs.

\paragraph{Matching.} We sample the source descriptor at the annotated keypoint and predict the target pixel with the highest cosine similarity. Patch-grid readouts interpolate the target cosine map to the pixel lattice, and SeeFine computes it directly on that lattice. The component study in \cref{app:components_v4} interpolates the grid features instead. Target annotations are used only to score predictions.

\paragraph{Local pooling.}\label{app:spair_query} Optionally, unit descriptors $f(u)$ on the image lattice $\Omega$ are averaged around each location $p$ after the frozen forward pass,
\begin{equation}\label{eq:spair_query}
 \begin{aligned}
 \bar f(p)&=\frac{\sum_{u\in\Omega}w_p(u)f(u)}{\sum_{u\in\Omega}w_p(u)},\\
 w_p(u)&=\max\!\Big(0,1-\frac{\lVert u-p\rVert^2}{r^2}\Big),
 \end{aligned}
\end{equation}
and renormalized. We apply it at the source keypoint and at every target pixel, with $r=7$ in the cross-backbone study and $r=8$ for DINOv3-B. Kernel support is truncated at image borders, every target pixel remains a candidate, and no parameters are fitted.

\begin{table*}[t]\centering\footnotesize
\caption{\textbf{SPair-71k with training-free baselines.} Pair-mean PCK@0.05 / PCK@0.10 (\%) on all native-resolution test pairs. All readouts use last-block features after the final LayerNorm (for DVF, applied to the last-block tokens its extractor returns), and Bilinear reproduces the donor result. SeeFine on DINOv3 uses recorded donor context. The lower block adds local pooling with radius seven. ---: not supported by the released code; the HR-Dv2 timm path, needed for ViT-S and DeiT-S, runs only at $224\times224$. HR-Dv2 uses the donor's DINOv2 weights without registers; the separately trained checkpoint with registers gives $29.98/44.01$. \Cref{tab:abl_descriptor,tab:abl_layernorm} give the descriptor and LayerNorm controls.}\label{tab:base_spair}
\setlength{\tabcolsep}{5pt}
\begin{tabular}{@{}lcccccc@{}}\toprule
Backbone & Donor & Nearest & Strided & HR-Dv2 & DVF & SeeFine\\\midrule
DINO ViT-S/16 & 15.74 / 28.34 & 11.90 / 25.53 & 18.90 / \textbf{30.75} & 17.13 / 28.37 & \textbf{18.92} / 30.61 & 18.61 / 30.32\\
DINOv2 ViT-S/14 & 31.89 / 49.03 & 25.52 / 45.89 & 34.96 / 48.38 & 31.33 / 44.47 & --- & \textbf{35.13} / \textbf{50.14}\\
ViT-S/16 (AugReg) & 7.96 / 17.29 & 6.91 / 16.99 & 10.39 / 20.56 & --- & 10.60 / 20.43 & \textbf{11.31} / \textbf{21.50}\\
DeiT-Small & 11.79 / 23.23 & 9.36 / 21.09 & 14.40 / 25.21 & --- & 14.14 / 24.21 & \textbf{15.98} / \textbf{27.32}\\
DINOv3 ViT-L/16 (recorded context) & 32.92 / 51.44 & 24.94 / 48.10 & 40.10 / 55.51 & 38.12 / 54.45 & --- & \textbf{41.37} / \textbf{57.45}\\
\midrule
\multicolumn{7}{@{}l}{+ local pooling, radius seven}\\
DINO ViT-S/16 & 17.90 / 30.57 & 17.00 / 29.76 & \textbf{20.65} / \textbf{32.50} & 17.94 / 29.47 & 20.48 / 32.19 & 20.33 / 32.20\\
DINOv2 ViT-S/14 & 34.37 / 50.24 & 34.11 / 50.17 & 35.52 / 48.51 & 31.74 / 44.98 & --- & \textbf{36.70} / \textbf{51.19}\\
ViT-S/16 (AugReg) & 8.62 / 18.95 & 8.52 / 18.93 & 11.16 / 21.52 & --- & 11.19 / 21.30 & \textbf{12.02} / \textbf{22.52}\\
DeiT-Small & 14.11 / 25.88 & 13.56 / 25.16 & 15.79 / 26.84 & --- & 15.40 / 25.99 & \textbf{17.42} / \textbf{29.31}\\
DINOv3 ViT-L/16 (recorded context) & 35.37 / 52.06 & 35.51 / 52.93 & 39.62 / 54.40 & 37.61 / 54.04 & --- & \textbf{41.39} / \textbf{57.08}\\
\bottomrule\end{tabular}
\end{table*}

\begin{table*}[t]\centering\footnotesize
\caption{\textbf{SPair-71k: baseline minus SeeFine.} PCK@0.05 differences (percentage points), paired 95\% whole-pair bootstrap CI. Pooled columns compare with SeeFine + local pooling. ---: as in \cref{tab:base_spair}. DVF and SeeFine both use last-block features after the final LayerNorm. With DVF's block-9 descriptor applied to every readout, SeeFine exceeds DVF (\cref{tab:abl_descriptor}).}\label{tab:base_spair_delta}
\setlength{\tabcolsep}{3pt}
\begin{tabular}{@{}lccccccc@{}}\toprule
Backbone & Nearest & Strided & HR-Dv2 & DVF & \shortstack{Strided\\+ pooling} & \shortstack{HR-Dv2\\+ pooling} & \shortstack{DVF\\+ pooling}\\\midrule
DINO ViT-S/16 & \shortstack{$-6.71$\\$[-7.05,-6.39]$} & \shortstack{$+0.29$\\$[+0.02,+0.55]$} & \shortstack{$-1.48$\\$[-1.76,-1.21]$} & \shortstack{$+0.32$\\$[+0.05,+0.56]$} & \shortstack{$+0.32$\\$[+0.09,+0.56]$} & \shortstack{$-2.39$\\$[-2.66,-2.12]$} & \shortstack{$+0.15$\\$[-0.07,+0.37]$}\\
DINOv2 ViT-S/14 & \shortstack{$-9.61$\\$[-9.98,-9.26]$} & \shortstack{$-0.17$\\$[-0.49,+0.13]$} & \shortstack{$-3.80$\\$[-4.16,-3.43]$} & --- & \shortstack{$-1.18$\\$[-1.45,-0.91]$} & \shortstack{$-4.96$\\$[-5.30,-4.61]$} & ---\\
ViT-S/16 (AugReg) & \shortstack{$-4.40$\\$[-4.66,-4.12]$} & \shortstack{$-0.92$\\$[-1.17,-0.65]$} & --- & \shortstack{$-0.71$\\$[-0.98,-0.46]$} & \shortstack{$-0.86$\\$[-1.11,-0.61]$} & --- & \shortstack{$-0.83$\\$[-1.08,-0.58]$}\\
DeiT-Small & \shortstack{$-6.63$\\$[-6.93,-6.32]$} & \shortstack{$-1.58$\\$[-1.86,-1.30]$} & --- & \shortstack{$-1.84$\\$[-2.11,-1.56]$} & \shortstack{$-1.62$\\$[-1.88,-1.35]$} & --- & \shortstack{$-2.02$\\$[-2.27,-1.75]$}\\
DINOv3 ViT-L/16 & \shortstack{$-16.43$\\$[-16.83,-16.02]$} & \shortstack{$-1.27$\\$[-1.59,-0.96]$} & \shortstack{$-3.25$\\$[-3.61,-2.90]$} & --- & \shortstack{$-1.77$\\$[-2.06,-1.49]$} & \shortstack{$-3.78$\\$[-4.12,-3.43]$} & ---\\
\bottomrule\end{tabular}
\end{table*}
\subsection{Cross-backbone results}
SeeFine improves on every frozen donor at both thresholds (\cref{tab:spair_multibackbone_gains}). The four backbones with absolute positional embeddings use recurrent SeeFine, and the DINOv3-L column uses recorded context with a fixed reference of 784. \Cref{tab:base_spair,tab:base_spair_delta} add the training-free baselines and local pooling. SeeFine exceeds Strided by $0.92$ to $1.58$ PCK@0.05 points on ViT-S, DeiT-S and DINOv3 and is on par with it on DINOv2 (Strided minus SeeFine: $-0.17$, not significant). On DINO, Strided scores $0.29$ points higher at about one fifteenth of the extraction cost. HR-Dv2 stays below SeeFine wherever it runs. With the same last-block features, DVF exceeds SeeFine by $0.32$ PCK@0.05 points on DINO (95\% CI $[0.05,0.56]$) and scores $0.71$ and $1.84$ points lower on ViT-S and DeiT-S, respectively. Over all keypoints, \SF{DINOv2} is closer to the annotation than DINOv2 in $53.4\%$ of cases, and in $57.8\%$ with local pooling (\cref{tab:spair_winrates}).

\paragraph{The DVF descriptor.}\label{app:abl_descriptor} DVF~\cite{amir2022deep} matches block-9 keys with logarithmic binning, whereas \cref{tab:base_spair} reads every readout, DVF included, on last-block features. The descriptor is independent of how resolution is increased, so \cref{tab:abl_descriptor} applies it to every readout. For SeeFine, the bins sit at the offsets of the stride-4 grid, and on DINOv3 we use block 18 of 24 without a layer search. With the same descriptor, SeeFine is the best readout on all five backbones and exceeds DVF by $0.76$ ($[0.57,0.95]$) on DINO, $1.51$ ($[1.25,1.77]$) on ViT-S and $3.12$ ($[2.84,3.40]$) on DeiT-S. The descriptor helps DINO, ViT-S and DeiT-S, raising ViT-S from $11.31$ to $23.81$, but lowers DINOv2 and DINOv3 below their last-block scores ($7.29$ versus $35.13$ and $31.65$ versus $41.37$).

\paragraph{The final LayerNorm.}\label{app:abl_layernorm} The SPair readouts use last-block features after the model's final LayerNorm. \Cref{tab:abl_layernorm} repeats the comparison on the same last-block tokens before the LayerNorm; for DVF, these are the tokens returned by its extractor. Removing the LayerNorm lowers PCK@0.05 in 24 of 26 readout--backbone combinations and raises the other two by at most $0.10$ points. SeeFine ranks highest with either setting on DINOv2, ViT-S, DeiT-S and DINOv3. On DINO, Strided, DVF and SeeFine stay within $0.31$ points of one another, and their order changes. BraTS uses the tokens before the LayerNorm for every readout (\cref{app:brats-slices}). With it, parts of the constant black background become false positives, and case-macro Dice falls from $75.98\%$ to $63.77\%$ for SeeFine and from $74.20\%$ to $67.86\%$ for DVF.

\input{cvpr5_revision/dvf_scope.tex}

\begin{table}[t]\centering\footnotesize
\caption{\textbf{Cross-backbone SPair-71k gains.} SeeFine minus its frozen donor in pair-mean PCK (percentage points), computed from the rounded scores in \cref{tab:base_spair}. All rows use the full test split. RP denotes recorded donor context. Paired baseline contrasts are reported in \cref{tab:base_spair_delta}.}
\label{tab:spair_multibackbone_gains}
\setlength{\tabcolsep}{4pt}
\begin{tabular}{@{}lrr@{}}\toprule
Backbone & $\Delta$@0.05 & $\Delta$@0.10\\\midrule
DINO ViT-S/16 & $+2.87$ & $+1.98$\\
DINOv2 ViT-S/14 & $+3.24$ & $+1.11$\\
ViT-S/16 (AugReg) & $+3.35$ & $+4.21$\\
DeiT-Small & $+4.19$ & $+4.09$\\
DINOv3 ViT-L/16 (RP) & $+8.45$ & $+6.01$\\
\bottomrule\end{tabular}
\end{table}

\begin{table}[H]\centering\footnotesize
\caption{\textbf{Pointwise localization error relative to DINOv2.} Percent of all SPair-71k test correspondences with smaller, equal or larger Euclidean error than the frozen donor. All correspondences, including ties, enter the denominator.}\label{tab:spair_winrates}
\setlength{\tabcolsep}{4pt}
\begin{tabular}{@{}lrrr@{}}\toprule
Readout & Closer & Tie & Farther\\\midrule
\SF{DINOv2} & 53.41 & 0.91 & 45.69\\
\quad + local pooling & 57.81 & 0.84 & 41.34\\
\bottomrule\end{tabular}
\par\smallskip\noindent\scriptsize
95\% pair-bootstrap intervals for ``Closer'': $[53.06,53.76]$ and $[57.47,58.15]\%$. Local pooling also reduces unpooled SeeFine's error on $55.19\%$ of points.
\end{table}

\begin{table}[H]\centering\scriptsize
\caption{\textbf{Effect of the descriptor on SPair-71k.} Pair-mean PCK@0.05 (\%). The last column uses last-block features, all others the DVF descriptor. }\label{tab:abl_descriptor}
\setlength{\tabcolsep}{3pt}
\begin{tabular}{@{}lrrrrr@{}}\toprule
Backbone & Donor & Strided & DVF & SeeFine & SeeFine, last block\\\midrule
DINO ViT-S/16 & 13.37 & 20.06 & 21.50 & \textbf{22.26} & 18.61\\
ViT-S/16 (AugReg) & 14.62 & 21.63 & 22.29 & \textbf{23.81} & 11.31\\
DeiT-Small & 11.18 & 16.94 & 17.72 & \textbf{20.84} & 15.98\\
DINOv2 ViT-S/14 & 5.02 & 4.78 & --- & 7.29 & \textbf{35.13}\\
DINOv3 ViT-L/16 & 16.83 & 24.49 & --- & 31.65 & \textbf{41.37}\\
\bottomrule\end{tabular}
\end{table}

\begin{table*}[t]\centering\footnotesize
\caption{\textbf{Effect of the final LayerNorm on SPair-71k.} Pair-mean PCK@0.05 (\%) on all test pairs without local pooling. Each cell gives the score after / before the final LayerNorm. The ``after'' values are those of \cref{tab:base_spair}; for DVF, ``before'' uses the last-block tokens returned by its extractor. ---: as in \cref{tab:base_spair}.}\label{tab:abl_layernorm}
\setlength{\tabcolsep}{5pt}
\begin{tabular}{@{}lcccccc@{}}\toprule
Backbone & Donor & Nearest & Strided & HR-Dv2 & DVF & SeeFine\\\midrule
DINO ViT-S/16 & 15.74 / 15.12 & 11.90 / 11.67 & 18.90 / 18.53 & 17.13 / 17.23 & 18.92 / 18.79 & 18.61 / 18.63\\
DINOv2 ViT-S/14 & 31.89 / 31.03 & 25.52 / 24.96 & 34.96 / 34.61 & 31.33 / 29.67 & --- & 35.13 / 34.73\\
ViT-S/16 (AugReg) & 7.96 / 6.83 & 6.91 / 6.12 & 10.39 / 9.18 & --- & 10.60 / 9.37 & 11.31 / 10.43\\
DeiT-Small & 11.79 / 10.85 & 9.36 / 8.94 & 14.40 / 13.90 & --- & 14.14 / 13.79 & 15.98 / 15.71\\
DINOv3 ViT-L/16 (recorded context) & 32.92 / 29.63 & 24.94 / 23.20 & 40.10 / 38.53 & 38.12 / 35.09 & --- & 41.37 / 39.56\\
\bottomrule\end{tabular}
\end{table*}

\subsection{Comparison with trained upsamplers}\label{app:vitup}\label{app:native_scale_v4}\label{app:native_controls_v4}
\Cref{tab:vitup_comparison} extends the DINOv3-B comparison of \cref{tab:vitup_main} with all controls, including recorded context and local pooling.

\paragraph{ViT-Up protocol.} The released ViT-Up~\cite{wandel2026vitupfaithfulfeatureupsampling} keeps its authors' $448\times448$ setting: square padding, antialiased resizing, floor source-cell lookup and cosine nearest-neighbour target search, with BF16 inference and FP32 search. Its decoder was trained with teacher inputs of $224$, $448$ and $896$ pixels. Our run of the released checkpoint matches the published scores to within $0.16$ points.

\paragraph{Native and doubled inputs.} The native and doubled-native blocks keep each photograph's aspect ratio and use no padding at the input stage. The doubled input is a bilinear $2W\times2H$ enlargement of the same RGB image and adds no information. Donor features are interpolated from their patch centres onto the pixel lattice, and all fields use the same matcher. SeeFine uses the actual donor patch count $N_D=\lfloor H/16\rfloor\lfloor W/16\rfloor$ as reference, donor-support rotary coordinates and complete RGB windows as keys. Local pooling uses a radius of eight input pixels.

\paragraph{Results.} At $448\times448$ ViT-Up leads, $41.84$ versus $40.00$ PCK@0.05. On native and doubled inputs SeeFine scores $56.95/41.48/6.37$ and $58.04/44.10/10.60$ at PCK@0.10/0.05/0.01. On the same doubled input it gains $3.26/3.00/2.19$ points over the donor, with category-stratified 95\% intervals $[3.05,3.47]$, $[2.76,3.24]$ and $[1.99,2.41]$. The recorded-context control RP gives $58.07/44.07/10.61$ at the same thresholds. Its difference from the recurrent model at PCK@0.05 is $-0.03$ points (95\% interval $[-0.11,+0.05]$).

\paragraph{AnyUp and sparser sampling.} AnyUp~\cite{wimmer2026anyup} is run with its released multi-backbone checkpoint and official implementation. It receives the ImageNet-normalized RGB rectangle and the raw $\lfloor H/16\rfloor\times\lfloor W/16\rfloor$ donor grid, passed with the complete $H\times W$ guide without cropping trailing strips. Stride-$S$ controls sample source positions $0,S,2S,\ldots$ with the recorded-context policy and interpolate their descriptors onto the pixel lattice.

\paragraph{Cost.} On native images, moving from stride 1 to stride 4 with recorded context reduces median extraction time from $0.370$ to $0.039$\,s and peak allocated memory from $10.42$ to $1.62$\,GiB, while PCK@0.05 stays at $41.52$, $41.41$ and $41.50$ for strides 1, 2 and 4. \Cref{tab:current_operating_cost} reports feature-extraction time and peak allocated GPU memory on 24 fixed validation photographs (one GH200, one warmup and three synchronized repeats per image, one configuration per process). Timing runs from RGB on the GPU to an FP32 descriptor field at the input lattice and includes the donor pass in the recorded-context rows. Image decoding, matching and model loading are outside the timed region.

\input{cvpr5_revision/comparison_full_table_v5.tex}

\begin{table}[t]\centering\footnotesize
\caption{\textbf{Quality and feature-extraction cost on SPair-71k.} Category-macro PCK@0.05 on the full test split. Median extraction time and peak allocated GPU memory per image are measured on 24 fixed validation photographs.}\label{tab:current_operating_cost}
\setlength{\tabcolsep}{3pt}
\begin{tabular}{@{}llrrr@{}}\toprule
Method & Input & PCK@0.05 & Time (s) & Mem.\ (GiB)\\\midrule
ViT-Up (released) & $448\times448$ & 41.84 & 0.320 & 1.34\\
SeeFine & $448\times448$ & 40.00 & 0.501 & 9.52\\
\midrule
DINOv3-B & $W\times H$ & 35.84 & 0.011 & 1.59\\
\quad + local pooling & $W\times H$ & 35.16 & 0.032 & 3.42\\
SeeFine & $W\times H$ & 41.48 & 0.439 & 10.41\\
\quad + local pooling & $W\times H$ & 41.25 & 0.458 & 10.41\\
\midrule
DINOv3-B & $2W\times2H$ & 41.09 & 0.019 & 5.29\\
SeeFine & $2W\times2H$ & 44.10 & 2.248 & 40.00\\
\quad + local pooling & $2W\times2H$ & 43.53 & 2.340 & 40.00\\
\midrule
Recorded control RP & $W\times H$ & 41.52 & 0.370 & 10.42\\
\quad + local pooling & $W\times H$ & 41.45 & 0.388 & 10.42\\
RP, stride 2 & $W\times H$ & 41.41 & 0.107 & 2.90\\
RP, stride 4 & $W\times H$ & 41.50 & 0.039 & 1.62\\
AnyUp & $W\times H$ & 19.04 & 1.600 & 1.50\\
\bottomrule\end{tabular}\end{table}

\subsection{Spatial precision under known transformations}\label{app:spatial}
We fix 100 ImageNet validation indices before scoring (random permutation, seed 1310). Each source view is the 224-pixel centre of the same 256-pixel gallery canvas. Frozen DINO ViT-S/16 and \SF{DINO} receive the same RGB views, and no checkpoint or readout is trained.

\paragraph{Known affine correspondence.} A $3^\circ$ rotation, scale $1.05$ and translation $(+4,-3)$ define a known source-to-target transform. The target RGB view is sampled bilinearly with reflection outside its canvas. We sample 32 source points per image uniformly in $[40,184]^2$, keeping their transformed locations inside the target. We normalize final patch features, interpolate donor features at their token centres and use the same cosine-argmax matcher over the target pixel lattice. Because both views show the same image under a known transform, the error measures spatial localization directly.

SeeFine reduces median error from $4.33$ to $0.80$ pixels and raises the fraction of matches within four pixels from $45.75$ to $96.12\%$. A paired bootstrap resamples whole images, preserving their 32 correlated points (2000 resamples, seed 1310). The 95\% intervals are $[3.34,3.73]$ pixels for the median-error reduction and $[47.94,52.81]$ percentage points for the gain in matches within four pixels. HR-Dv2 has a median error of $0.85$ pixels, $0.05$ above SeeFine (95\% CI $[0.02,0.09]$), but places more matches within four pixels ($97.09\%$). Median errors are $1.03$ pixels for DVF and $1.60$ for Strided, and Nearest is worse than the donor on every measure.

\paragraph{Translation consistency.} We crop each shifted 224-pixel view independently from the same 256-pixel canvas and encode it with the same model. We then undo the known horizontal shift, compare normalized features on the common domain, excluding a 16-pixel boundary band, and report image-mean cosine similarity and relative $L_2$ discrepancy at shifts 1, 4 and 8 pixels. At all three tested shifts, SeeFine has higher cosine similarity and lower relative $L_2$ discrepancy than the donor (\cref{tab:canonical_spatial}). SeeFine has the lowest relative $L_2$ discrepancy at a one-pixel shift. DVF and HR-Dv2 have lower discrepancies at four- and eight-pixel shifts.
\input{cvpr5_revision/spatial_results_v5.tex}

\begin{figure*}[p]\centering
\includegraphics[width=\textwidth,height=.84\textheight,keepaspectratio]{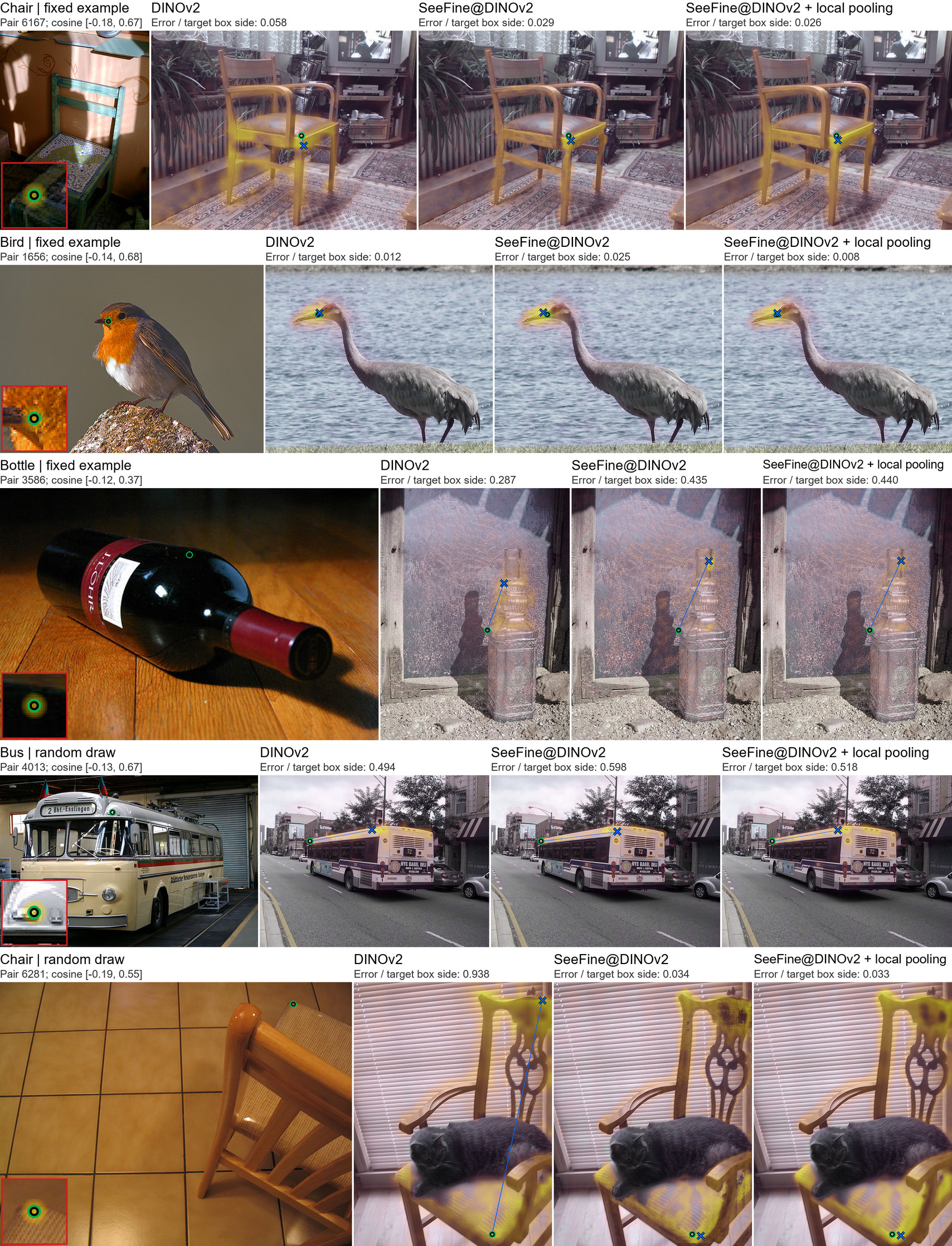}
\caption{\textbf{SPair-71k: diagnostic examples and random draws.} Frozen DINOv2 ViT-S/14 on uncropped native photographs. From left to right, the panels show the source query with a radius-seven kernel inset, the DINOv2 target cosine map, unpooled \SF{DINOv2} and \SF{DINOv2} with local pooling of both source and target features. Green circles mark annotations, and blue crosses mark predictions. Each row shares one cosine scale and divides the error by the longer side of the target box. The first three rows are diagnostic examples, and the last two are random draws.}\label{fig:spair_query_gains}
\end{figure*}

\begin{figure*}[p]\centering
\includegraphics[width=\textwidth,height=.84\textheight,keepaspectratio]{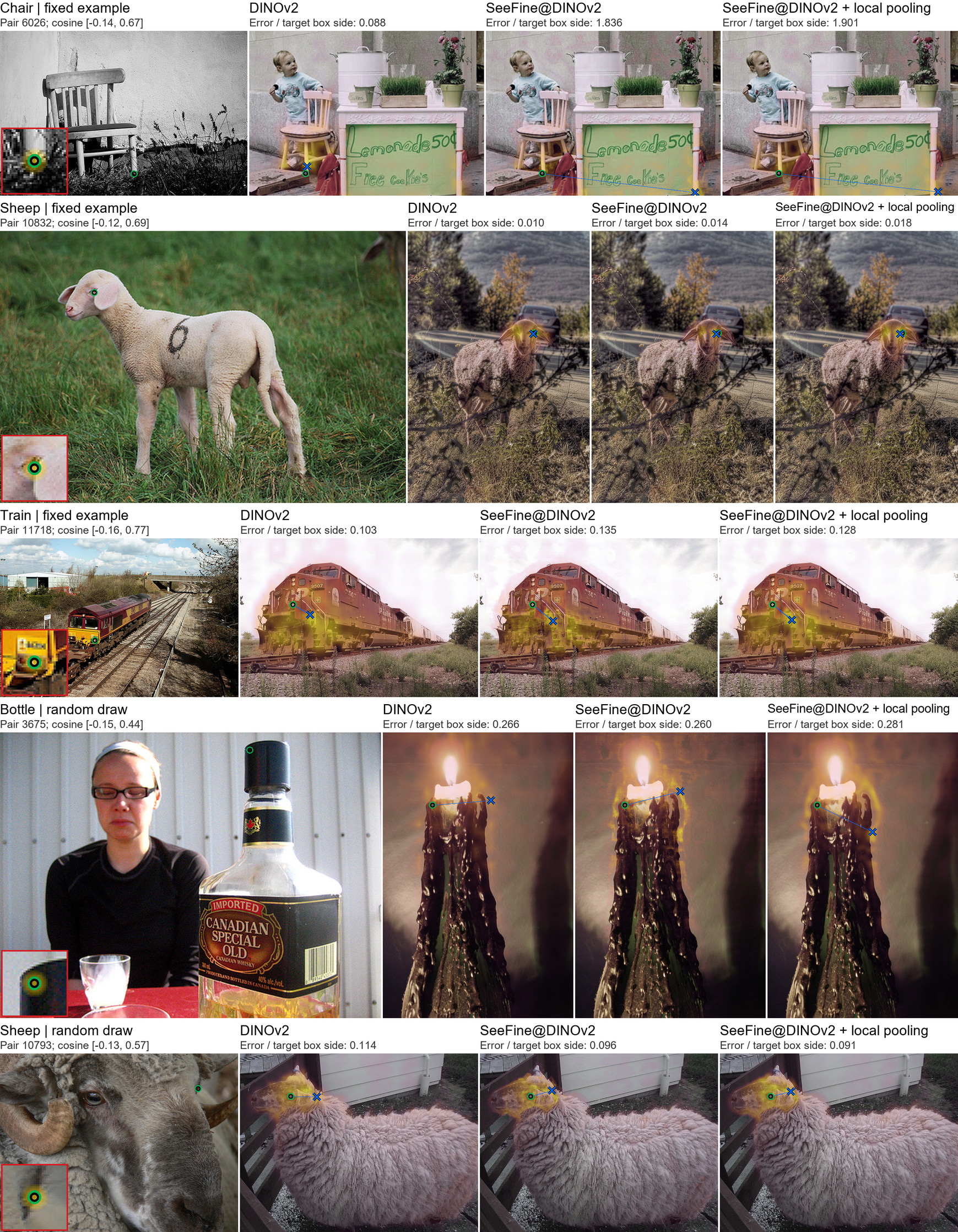}
\caption{\textbf{SPair-71k: further diagnostic examples and random draws.} Five additional pairs with the checkpoint, readouts and display of \cref{fig:spair_query_gains}. The first three rows are diagnostic examples, and the last two are random draws. All ten examples use distinct pairs. In the first row, both SeeFine readouts place the match far from the annotation. Normalized distances can exceed one when the target box is much smaller than the photograph. Weak heatmaps retain the common per-row cosine scale.}\label{fig:spair_query_limits}
\end{figure*}

\paragraph{Examples.} \Cref{fig:spair_query_gains,fig:spair_query_limits} show six diagnostic pairs and four randomly drawn pairs with frozen DINOv2, including failures. The example in the main paper illustrates successive improvements from the donor to SeeFine and to pooling. \Cref{tab:spair_winrates} reports how often SeeFine is closer to the annotation than the donor across the full split.

%% file: cvpr5_revision/dvf_scope.tex
\par{
\paragraph{Compatibility of the released DVF extractor.}\label{app:dvf_scope}
We run the released extractor~\cite{amir2022deep} unchanged on compatible backbones. It has four requirements. (i) The checkpoint must load into the original DINO ViT architecture~\cite{caron2021dino}. (ii) The stride $s$ must divide the patch size $P$. (iii) The backbone uses a learned absolute position table, which the extractor interpolates to the denser grid. (iv) The CLS token is the sole prefix token. The extractor drops it and reshapes the remaining tokens into the patch grid. DINO ViT-S/16, ViT-S/16 (AugReg) and DeiT-S meet these requirements. The extractor loads DeiT-S through the same path as the supervised ViT-S.

\paragraph{DINOv2.} The ViT-S/14 checkpoint~\cite{oquab2023dinov2} has a learned $37\times37$ position table and a single CLS prefix token, satisfying (iii) and (iv). Loading it into the original DINO ViT class would discard the 24 LayerScale tensors that scale the attention and MLP outputs across its 12 blocks. With $P=14$, the divisor requirement limits denser sampling to strides 1, 2 and 7, so the extractor rejects the released stride-4 setting on this backbone. LayerScale also prevents the released extractor from loading the DeiT-III checkpoints in \cref{tab:donors}.

\paragraph{DINOv3.} The ViT-L/16 checkpoint~\cite{simeoni2025dinov3} meets the stride requirement because 4 divides 16. Its blocks combine rotary attention with LayerScale, which requires changes to the original DINO ViT class. Rotary positions are computed from token coordinates, leaving the extractor without a learned position table to interpolate. The extractor would also retain the four register tokens~\cite{darcet2024vision} when reshaping the patch tokens. Supporting DINOv3-L, DINOv3-B (\cref{tab:vitup_main}) and the Talk2DINOv3 encoder (\cref{tab:talk2dinov3_native}) would require adapting the extractor for these architectural differences. The HR-Dv2 release already includes a DINOv3 execution path for its shifted and flipped passes.

\paragraph{Separate controls for dense execution and descriptor choice.} On DINOv2 and DINOv3, Strided tests overlapping patches with full attention using last-block features. \Cref{tab:abl_descriptor} tests DVF's block-9 descriptor by applying it to every readout. At stride 4, Strided differs from the released DVF extractor by at most $0.20$ PCK@0.05 points across the three backbones supported by both implementations (\cref{tab:dvf_scope}). The paired differences (Strided minus DVF) are $-0.02$ ($[-0.10,+0.06]$) on DINO, $-0.20$ ($[-0.29,-0.10]$) on ViT-S and $-0.03$ ($[-0.11,+0.05]$) on DeiT-S. Strided supports strides that do not divide the patch size, including stride 4 on DINOv2. It resamples learned position tables and uses DINOv3's rotary positions on the denser grid. SeeFine exceeds the stride-4 Strided baseline by $2.21$ PCK@0.05 points on DINOv2 ($[1.87,2.53]$) and $1.12$ points on DINOv3 ($[0.80,1.46]$).
}\par

\begin{table}[t]\centering\footnotesize\par{
\caption{\textbf{Strided closely matches released DVF at stride 4.} SPair-71k pair-mean PCK@0.05 (\%) on all test pairs, using last-block features after the final LayerNorm (\cref{tab:base_spair}). The released DVF extractor uses stride 4. Strided is evaluated at stride 4 and at its default stride (6 for $P=14$, 7 for $P=16$). SeeFine uses recorded donor context on DINOv3.}\label{tab:dvf_scope}
\setlength{\tabcolsep}{3pt}
\begin{tabular}{@{}lrrrr@{}}\toprule
Backbone & DVF & Strided, $s=4$ & Strided, $s=6/7$ & SeeFine\\\midrule
DINO ViT-S/16 & \textbf{18.92} & 18.90 & 18.90 & 18.61\\
ViT-S/16 (AugReg) & 10.60 & 10.40 & 10.39 & \textbf{11.31}\\
DeiT-Small & 14.14 & 14.11 & 14.40 & \textbf{15.98}\\
DINOv2 ViT-S/14 & --- & 32.93 & 34.96 & \textbf{35.13}\\
DINOv3 ViT-L/16 & --- & 40.25 & 40.10 & \textbf{41.37}\\
\bottomrule\end{tabular}
}\par\end{table}

%% file: cvpr5_revision/comparison_full_table_v5.tex
\begin{table*}[t]\centering\footnotesize
\caption{\textbf{Released learned upsampling, integrated SeeFine and its preservation control.} SPair-71k category-macro PCK (\%) on all test pairs across the 18 categories. The $448\times448$ block follows ViT-Up's official protocol, and the native and $2\times$ blocks use uncropped rectangles; comparisons across blocks differ in input and compute. ViT-Up adds LoRA and a decoder, and AnyUp adds a trained upsampler. Nearest, Strided, HR-Dv2 and the SeeFine variants keep DINOv3-B frozen. RP reads recorded donor globals, and the other SeeFine rows update the globals recurrently. Bold marks the highest score within each block.}\label{tab:vitup_comparison}
\setlength{\tabcolsep}{5pt}\begin{tabular}{@{}llrrr@{}}\toprule
Method & Additional learned components & PCK@0.10 & PCK@0.05 & PCK@0.01\\\midrule
\multicolumn{5}{l}{\emph{Published ViT-Up Table IV: official $448\times448$ protocol}}\\
DINOv3-B, bilinear & None & 50.01 & 34.03 & 3.89\\
ViT-Up & LoRA + decoder & \textbf{58.10} & \textbf{42.00} & \textbf{7.93}\\
\midrule
\multicolumn{5}{l}{\emph{Our measurements: unchanged official $448\times448$ input and matching protocol}}\\
DINOv3-B, bilinear & None & 50.03 & 34.06 & 3.90\\
DINOv3-B, nearest & None & 45.69 & 22.64 & 1.09\\
DINOv3-B, Strided (stride 7) & None & 51.90 & 37.96 & 6.55\\
DINOv3-B, HR-Dv2 & None & 52.45 & 36.26 & 5.17\\
ViT-Up, released checkpoint & LoRA + decoder & \textbf{57.97} & \textbf{41.84} & \textbf{7.93}\\
\SF{DINOv3-B}, recurrent & None & 55.92 & 40.00 & 5.43\\
\quad + pooling ($r=8$) & None & 54.58 & 39.89 & 6.69\\
Recorded-context control RP & None & 55.92 & 40.01 & 5.32\\
\quad + pooling ($r=8$) & None & 54.67 & 39.93 & 6.66\\
\midrule
\multicolumn{5}{l}{\emph{Native rectangles: original $W\times H$}}\\
DINOv3-B & None & 51.72 & 35.84 & 4.62\\
\quad + pooling ($r=8$) & None & 51.05 & 35.16 & 4.62\\
DINOv3-B, nearest & None & 48.24 & 25.27 & 1.34\\
DINOv3-B, Strided (stride 7) & None & 52.05 & 38.68 & 7.46\\
DINOv3-B, HR-Dv2 & None & 52.86 & 36.79 & 5.46\\
\SF{DINOv3-B}, recurrent & None & 56.95 & 41.48 & 6.37\\
\quad + pooling ($r=8$) & None & 55.76 & 41.25 & 7.65\\
Recorded-context control RP & None & \textbf{57.16} & \textbf{41.52} & 6.28\\
\quad + pooling ($r=8$) & None & 56.04 & 41.45 & \textbf{7.66}\\
Recorded context, stride 2 & None & 57.10 & 41.41 & 6.33\\
Recorded context, stride 4 & None & 56.92 & 41.50 & 6.60\\
AnyUp, multi-backbone & Released upsampler & 36.51 & 19.04 & 1.92\\
\midrule
\multicolumn{5}{l}{\emph{Doubled-native rectangles: $2W\times2H$}}\\
DINOv3-B & None & 54.78 & 41.09 & 8.41\\
\quad + pooling ($r=8$) & None & 54.29 & 40.72 & 8.50\\
DINOv3-B, nearest & None & 54.92 & 37.85 & 4.13\\
DINOv3-B, Strided (stride 7) & None & 57.08 & 41.95 & 9.91\\
DINOv3-B, HR-Dv2 & None & 49.31 & 33.30 & 6.02\\
\SF{DINOv3-B}, recurrent & None & 58.04 & \textbf{44.10} & 10.60\\
\quad + pooling ($r=8$) & None & 57.14 & 43.53 & 11.18\\
Recorded-context control RP & None & \textbf{58.07} & 44.07 & 10.61\\
\quad + pooling ($r=8$) & None & 57.22 & 43.60 & \textbf{11.22}\\
\bottomrule\end{tabular}\end{table*}

%% file: cvpr5_revision/spatial_results_v5.tex
\begin{table*}[t]\centering\footnotesize
\caption{\textbf{The frozen DINO representation becomes more spatially precise.} All readouts use frozen DINO ViT-S/16 and the same 100 fixed $224\times224$ ImageNet views in one evaluation run. Affine correspondence uses a common matcher on 3,200 points. Translation consistency compares cosine similarity and relative $L_2$ error on the common interior after undoing each shift. Bilinear reproduces the donor result, and DVF uses stride-four last-block features. $\downarrow$/$\uparrow$: lower/higher is better; bold: best per column.}\label{tab:canonical_spatial}
\setlength{\tabcolsep}{3pt}
\begin{tabular}{@{}lrrrrrrrr@{}}\toprule
& \multicolumn{2}{c}{Affine correspondence} & \multicolumn{3}{c}{Translation, cosine $\uparrow$} & \multicolumn{3}{c}{Translation, relative $L_2$ $\downarrow$}\\
\cmidrule(lr){2-3}\cmidrule(lr){4-6}\cmidrule(l){7-9}
Readout & Median (px) $\downarrow$ & $\le$4 px (\%) $\uparrow$ & 1 px & 4 px & 8 px & 1 px & 4 px & 8 px\\\midrule
DINO ViT-S/16 (= Bilinear) & 4.33 & 45.75 & 0.9763 & 0.9087 & 0.8622 & 0.2025 & 0.4013 & 0.4952\\
Nearest & 13.68 & 4.44 & 0.9417 & 0.8036 & 0.7228 & 0.2796 & 0.5694 & 0.7006\\
Strided & 1.60 & 89.44 & 0.9818 & 0.9568 & 0.9775 & 0.1733 & 0.2699 & 0.1976\\
DVF & 1.03 & 95.44 & 0.9878 & \textbf{0.9987} & \textbf{0.9958} & 0.1399 & \textbf{0.0451} & \textbf{0.0831}\\
HR-Dv2 & 0.85 & \textbf{97.09} & \textbf{0.9984} & 0.9979 & 0.9942 & 0.0521 & 0.0548 & 0.0970\\
\midrule
\SF{DINO} & \textbf{0.80} & 96.12 & 0.9982 & 0.9926 & 0.9875 & \textbf{0.0425} & 0.0989 & 0.1373\\
\bottomrule\end{tabular}\end{table*}

%% file: cvpr5_revision/nyu_results_v5.tex
\begin{table}[H]\centering\footnotesize
\caption{\textbf{NYUv2 depth from frozen features.} Frozen DINOv2 features at $224\times224$. Each readout has a separately trained head with the same 795 training images, 654 test images, training recipe and seed. SA denotes scale-aware and SI scale-and-shift-aligned evaluation. Lower RMSE and higher $d_1$ are better. Bilinear equals the donor, and Nearest assigns each location the feature of its nearest patch. }\label{tab:canonical_nyu}
\setlength{\tabcolsep}{4pt}
\begin{tabular}{@{}lrrrr@{}}\toprule
& \multicolumn{2}{c}{SA} & \multicolumn{2}{c}{SI}\\
Readout & RMSE & $d_1$ (\%) & RMSE & $d_1$ (\%)\\\midrule
DINOv2 (= Bilinear) & 0.5456 & 76.86 & \textbf{0.4137} & 87.61\\
Nearest & 0.5779 & 74.78 & 0.4532 & 85.10\\
Strided & 0.5486 & 76.08 & 0.4151 & 87.29\\
HR-Dv2 & 0.5607 & 74.89 & 0.4251 & 86.58\\
\midrule
\SF{DINOv2} & \textbf{0.5388} & \textbf{77.57} & 0.4175 & \textbf{87.78}\\
\bottomrule\end{tabular}
\end{table}

%% file: cvpr4_revision_20261027/network_figure.tex
{\setlength{\tabcolsep}{1pt}\begin{tabular}{@{}*{9}{c}@{}}
\includegraphics[width=.105\textwidth]{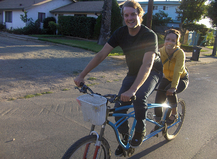}&\includegraphics[width=.105\textwidth]{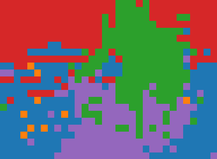}&\includegraphics[width=.105\textwidth]{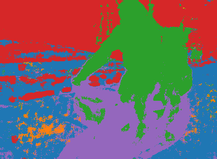}&\includegraphics[width=.105\textwidth]{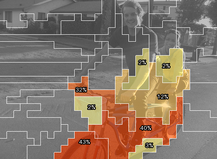}&\includegraphics[width=.105\textwidth]{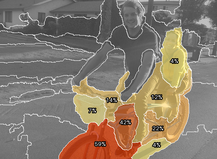}&\includegraphics[width=.105\textwidth]{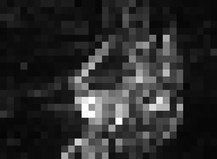}&\includegraphics[width=.105\textwidth]{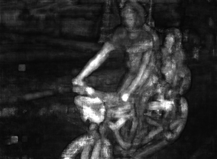}&\includegraphics[width=.105\textwidth]{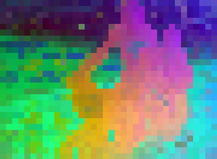}&\includegraphics[width=.105\textwidth]{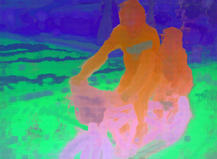}\\
\includegraphics[width=.105\textwidth]{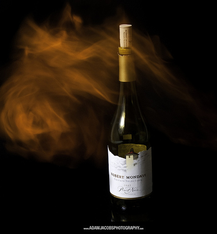}&\includegraphics[width=.105\textwidth]{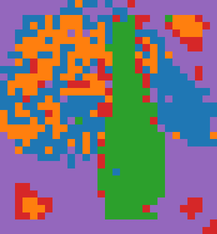}&\includegraphics[width=.105\textwidth]{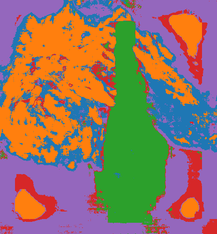}&\includegraphics[width=.105\textwidth]{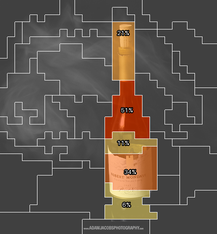}&\includegraphics[width=.105\textwidth]{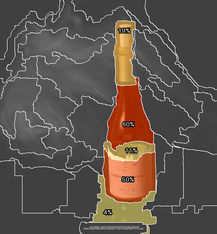}&\includegraphics[width=.105\textwidth]{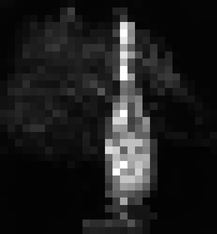}&\includegraphics[width=.105\textwidth]{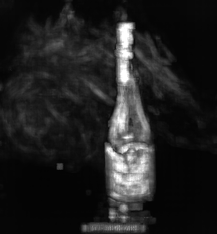}&\includegraphics[width=.105\textwidth]{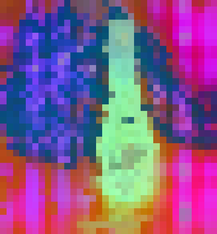}&\includegraphics[width=.105\textwidth]{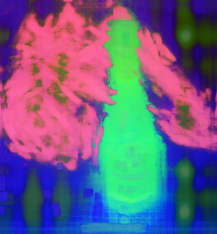}\\
\end{tabular}}

%% file: cvpr4_revision_20261027/direct_path_definition.tex
\paragraph{Exact partial direct-path quantity.}
For classifier logits $\ell\in\mathbb R^C$, the relative-logit denominator is $r_c=\ell_c-C^{-1}\sum_k\ell_k$. Let $z$ be the CLS before final LayerNorm, $\sigma=\sqrt{\operatorname{Var}(z)+\epsilon_{\rm LN}}$, and $\gamma,\beta$ its affine parameters. With classifier rows $W_k$ and biases $b_k$, define
\begin{equation}
 \begin{aligned}
 u_c&=(W_c-\overline W)\odot\gamma/\sigma,\\
 w_c&=u_c-\overline{u_c}\,\mathbf1,\\
 r_c&=w_c^\top z+(W_c-\overline W)^\top\beta+b_c-\overline b.
 \end{aligned}
\end{equation}
A bar over a classifier row or bias denotes a mean over classes, and the bar over $u_c$ denotes its mean across coordinates. The affine identity holds at the observed forward state, where the LayerNorm denominator is fixed at its computed value. For the actual shifted CLS attention probability $\alpha^\ell_{hi}$, patch value $v^\ell_{hi}$, attention output projection $W_O^\ell$ and LayerScale vector $\lambda^\ell$ (ones if absent), the displayed patch numerator is
\begin{equation}
 \begin{aligned}
 \eta^\ell_{ch}&=[W_O^{\ell\top}(\lambda^\ell\odot w_c)]_h,\\
 n_i&=\sum_{\ell,h}\alpha^\ell_{hi}\langle v^\ell_{hi},\eta^\ell_{ch}\rangle.
 \end{aligned}
\end{equation}
For a region $R$, its signed share is $100\sum_{i\in R}n_i/r_c$. Shares are undefined for $|r_c|\leq10^{-6}$ and are then not displayed. A negative $r_c$ reverses their sign. The unassigned residual $r_c-\sum_i n_i$ is retained numerically and never redistributed over pixels. It contains the initial CLS, self paths, biases, CLS MLP updates and final affine terms, including finite-precision residuals. The maps visualize the subset of direct attention paths included in the sum above.

%% file: cvpr5_revision/runtime_results_v5.tex
\begin{table}[H]\centering\small
\caption{\textbf{Forward cost of \SF{DINO}.} One GH200, batch one, eager PyTorch, BF16 autocast with FP32 position and residual states. We report the median of 10 synchronized forward passes after 3 warm-up passes. Peak allocated memory includes the loaded model and retained outputs, and timing excludes decoding, transfer and rendering.}\label{tab:dino_canonical_runtime}
\setlength{\tabcolsep}{3pt}
\begin{tabular}{@{}llrr@{}}\toprule
RGB view & Readout & ms/image & Peak GiB\\\midrule
$224\times224$ & DINO & 14.03 & 0.20\\
$224\times224$ & \SF{DINO} & 29.32 & 1.06\\
$2040\times1452$ & DINO & 15.29 & 0.35\\
$2040\times1452$ & \SF{DINO} & 6330.89 & 51.34\\
\bottomrule\end{tabular}\end{table}

\begin{figure}[t]\centering
\input{appendix_canonical_20261014/runtime_plot.tex}
\caption{\textbf{Single-image forward latency.} Same GH200 benchmark as \cref{tab:dino_canonical_runtime}, with a logarithmic time axis. On the native $2040\times1452$ photograph, SeeFine evaluates $2.96$ million spatial tokens, and its runtime reflects this larger token count.}\label{fig:canonical_runtime}
\end{figure}
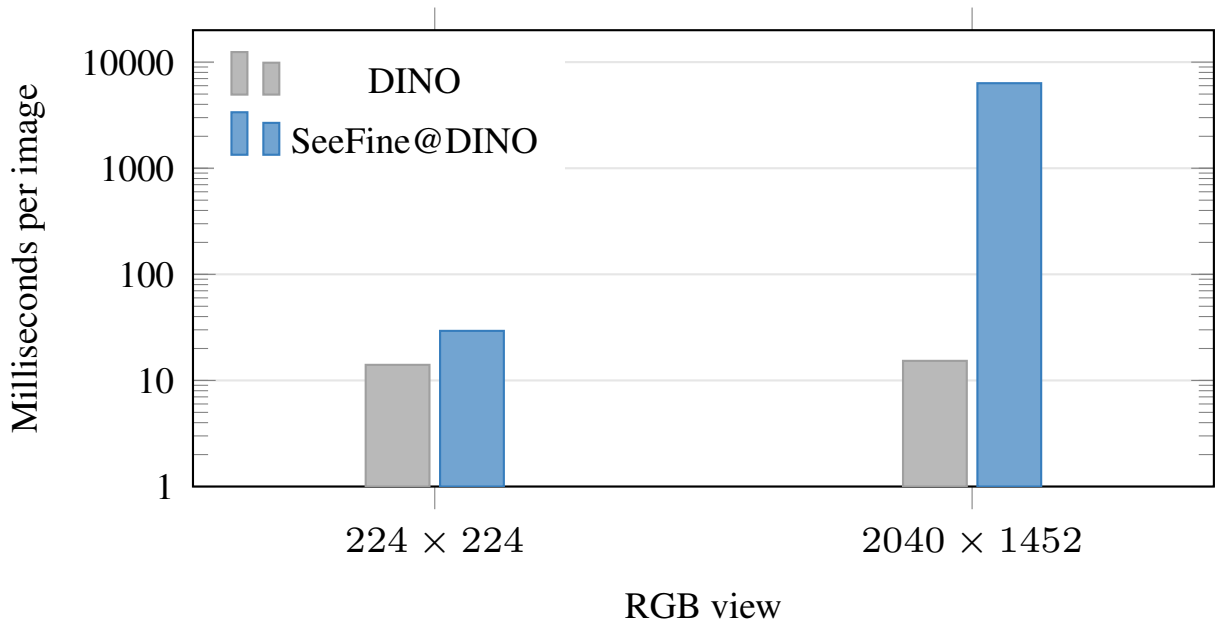

%% file: appendix_canonical_20261014/runtime_plot.tex
\begin{tikzpicture}
\begin{axis}[width=\columnwidth,height=4.6cm,ybar=2pt,bar width=12pt,
ymode=log,ymin=1,ymax=20000,log origin=infty,
symbolic x coords={small,native},xtick=data,
xticklabels={$224\times224$,$2040\times1452$},
ylabel={Milliseconds per image},xlabel={RGB view},
ytick={1,10,100,1000,10000},yticklabels={1,10,100,1000,10000},
font=\scriptsize,ymajorgrids=true,grid style={gray!20},
legend style={at={(.02,.98)},anchor=north west,draw=none,font=\scriptsize},
enlarge x limits=.45]
\addplot[fill=gray!55,draw=gray!75] coordinates {(small,14.026) (native,15.294)};
\addplot[fill=cvprblue!70,draw=cvprblue] coordinates {(small,29.320) (native,6330.889)};
\legend{DINO,SeeFine@DINO}
\end{axis}\end{tikzpicture}

%% file: cvpr5_revision/gradient_cost_v5.tex
\paragraph{Full RGB input-gradient feasibility.}\label{app:gradient_cost_v4}
These measurements use the recurrent single-CLS DeiT-S classifier. We time one forward or one forward/backward to a fixed class logit on the same RGB photograph, with all weights frozen. Each entry is the median of three synchronized runs after one warmup on one GH200. Peak memory is allocated GPU memory, capped at $85\%$ of a $95.12$-GiB device. We benchmark native-resolution execution on two fixed photographs. The $224$-pixel crop is also reported.

\begin{table}[H]\centering\small
\caption{\textbf{Forward and full RGB-gradient cost.} B16 uses a BF16 encoder and FP32 head, F32 is an FP32 forward pass, and F+B adds the FP32 backward pass. OOM denotes a measured allocation failure under the stated cap.}
\label{tab:gradient_cost_v4}\setlength{\tabcolsep}{3pt}
\begin{tabular}{@{}lllrr@{}}\toprule
RGB size & Model & Operation & Seconds & GiB\\\midrule
$224\times224$ & Donor & B16 & 0.014 & 0.12\\
$224\times224$ & Donor & F32 & 0.014 & 0.12\\
$224\times224$ & Donor & F+B & 0.037 & 0.19\\
$224\times224$ & SeeFine & B16 & 0.029 & 1.02\\
$224\times224$ & SeeFine & F32 & 0.083 & 1.66\\
$224\times224$ & SeeFine & F+B & 0.184 & 12.21\\
$500\times319$ & Donor & B16 & 0.012 & 0.16\\
$500\times319$ & Donor & F32 & 0.026 & 0.16\\
$500\times319$ & Donor & F+B & 0.043 & 0.28\\
$500\times319$ & SeeFine & B16 & 0.101 & 2.91\\
$500\times319$ & SeeFine & F32 & 0.307 & 4.99\\
$500\times319$ & SeeFine & F+B & 0.707 & 38.71\\
$2040\times1356$ & Donor & B16 & 0.022 & 0.32\\
$2040\times1356$ & Donor & F32 & 0.086 & 0.41\\
$2040\times1356$ & Donor & F+B & 0.249 & 2.59\\
$2040\times1356$ & SeeFine & B16 & 5.972 & 47.85\\
$2040\times1356$ & SeeFine & F32 & \multicolumn{2}{c}{OOM}\\
$2040\times1356$ & SeeFine & F+B & \multicolumn{2}{c}{OOM}\\
\bottomrule\end{tabular}\end{table}
We measure the overhead of differentiation by comparing FP32 forward and forward-plus-backward runs at the sizes where both fit. At $500\times319$, SeeFine takes $0.307$\,s and $4.99$\,GiB for the forward pass and $0.707$\,s and $38.71$\,GiB with the full RGB backward pass. At $2040\times1356$, the BF16 forward fits within the cap, while both FP32 runs exceed it. The 50-sample DAVE examples use $224\times224$ crops (\cref{app:dave}).

%% file: references.bib
@inproceedings{dosovitskiy2021vit,
  title = {An Image is Worth 16x16 Words: Transformers for Image Recognition at Scale},
  author = {Dosovitskiy, A. and Beyer, Lucas and Kolesnikov, Alexander and Weissenborn, Dirk and Zhai, Xiao-Hua and Unterthiner, T. and Dehghani, Mostafa and Minderer, M. and Heigold, G. and Gelly, S. and others},
  booktitle = {International Conference on Learning Representations},
  year = {2021},
  doi = {10.48550/arXiv.2010.11929},
}

@inproceedings{dehghani2023navit,
  title = {Patch n' Pack: {NaViT}, a Vision Transformer for any Aspect Ratio and Resolution},
  author = {Dehghani, Mostafa and Mustafa, Basil and Djolonga, J. and Heek, Jonathan and Minderer, M. and Caron, Mathilde and Steiner, A. and Puigcerver, J. and Geirhos, Robert and Alabdulmohsin, Ibrahim M. and others},
  booktitle = {Advances in Neural Information Processing Systems 36},
  year = {2023},
  pages = {2252-2274},
  doi = {10.52202/075280-0106},
  publisher = {Neural Information Processing Systems Foundation, Inc. (NeurIPS)},
}

@inproceedings{beyer2023flexivit,
  title = {{FlexiViT}: One Model for All Patch Sizes},
  author = {Beyer, Lucas and Izmailov, Pavel and Kolesnikov, Alexander and Caron, Mathilde and Kornblith, Simon and Zhai, Xiaohua and Minderer, Matthias and Tschannen, Michael and Alabdulmohsin, Ibrahim and Pavetic, Filip},
  booktitle = {Computer Vision and Pattern Recognition},
  year = {2023},
  pages = {14496-14506},
  doi = {10.1109/CVPR52729.2023.01393},
  publisher = {IEEE},
}

@inproceedings{caron2021dino,
  title = {Emerging Properties in Self-Supervised Vision Transformers},
  author = {Caron, Mathilde and Touvron, Hugo and Misra, Ishan and Jégou, Hervé and Mairal, Julien and Bojanowski, P. and Joulin, A.},
  booktitle = {IEEE International Conference on Computer Vision},
  year = {2021},
  pages = {9630-9640},
  doi = {10.1109/ICCV48922.2021.00951},
  publisher = {IEEE},
}

@inproceedings{amir2022deep,
  title = {{Deep ViT Features as Dense Visual Descriptors}},
  author = {Shir Amir and Yossi Gandelsman and Shai Bagon and Tali Dekel},
  booktitle = {ECCV Workshops (What is Motion For?)},
  year = {2022},
  eprint = {2112.05814},
  archiveprefix = {arXiv},
}

@inproceedings{tumanyan2024dinotracker,
  title = {{DINO-Tracker: Taming DINO for Self-Supervised Point Tracking in a Single Video}},
  author = {Narek Tumanyan and Assaf Singer and Shai Bagon and Tali Dekel},
  booktitle = {European Conference on Computer Vision (ECCV)},
  year = {2024},
  eprint = {2403.14548},
  archiveprefix = {arXiv},
}

@inproceedings{gandelsman2024interpreting,
  title = {{Interpreting CLIP's Image Representation via Text-Based Decomposition}},
  author = {Yossi Gandelsman and Alexei A. Efros and Jacob Steinhardt},
  booktitle = {International Conference on Learning Representations},
  year = {2024},
  eprint = {2310.05916},
  archiveprefix = {arXiv},
  doi = {10.48550/arXiv.2310.05916},
}

@inproceedings{fu2024featup,
  title = {{FeatUp: A Model-Agnostic Framework for Features at Any Resolution}},
  author = {Stephanie Fu and Mark Hamilton and Laura Brandt and Axel Feldman and Zhoutong Zhang and William T. Freeman},
  booktitle = {International Conference on Learning Representations (ICLR)},
  year = {2024},
  eprint = {2403.10516},
  archiveprefix = {arXiv},
}

@inproceedings{wang2025scaling,
  title = {{Scaling Laws in Patchification: An Image Is Worth 50,176 Tokens And More}},
  author = {Feng Wang and Yaodong Yu and Guoyizhe Wei and Wei Shao and Yuyin Zhou and Alan Yuille and Cihang Xie},
  booktitle = {International Conference on Machine Learning},
  year = {2025},
  eprint = {2502.03738},
  archiveprefix = {arXiv},
  doi = {10.48550/arXiv.2502.03738},
}

@inproceedings{chefer2021transformer,
  title = {{Transformer Interpretability Beyond Attention Visualization}},
  author = {Hila Chefer and Shir Gur and Lior Wolf},
  booktitle = {Computer Vision and Pattern Recognition},
  year = {2021},
  eprint = {2012.09838},
  archiveprefix = {arXiv},
  pages = {782-791},
  doi = {10.1109/CVPR46437.2021.00084},
  publisher = {IEEE},
}

@inproceedings{abnar2020quantifying,
  title = {{Quantifying Attention Flow in Transformers}},
  author = {Samira Abnar and Willem Zuidema},
  booktitle = {Proceedings of the 58th Annual Meeting of the Association for Computational Linguistics},
  year = {2020},
  doi = {10.18653/v1/2020.acl-main.385},
  eprint = {2005.00928},
  archiveprefix = {arXiv},
  pages = {4190-4197},
  publisher = {Association for Computational Linguistics},
}

@inproceedings{wimmer2026anyup,
  title = {{AnyUp: Universal Feature Upsampling}},
  author = {Thomas Wimmer and Prune Truong and Marie-Julie Rakotosaona and Michael Oechsle and Federico Tombari and Bernt Schiele and Jan Eric Lenssen},
  booktitle = {International Conference on Learning Representations},
  year = {2026},
  eprint = {2510.12764},
  archiveprefix = {arXiv},
  doi = {10.48550/arXiv.2510.12764},
}

@article{kopf2007joint,
  title = {{Joint bilateral upsampling}},
  author = {Johannes Kopf and Michael F. Cohen and Dani Lischinski and Matt Uyttendaele},
  journal = {ACM Transactions on Graphics},
  volume = {26},
  number = {3},
  pages = {96},
  year = {2007},
  doi = {10.1145/1276377.1276497},
  note = {SIGGRAPH 2007},
}

@inproceedings{huang2025loftup,
  title = {{LoftUp: Learning a Coordinate-Based Feature Upsampler for Vision Foundation Models}},
  author = {Haiwen Huang and Anpei Chen and Volodymyr Havrylov and Andreas Geiger and Dan Zhang},
  booktitle = {IEEE International Conference on Computer Vision},
  year = {2025},
  eprint = {2504.14032},
  archiveprefix = {arXiv},
  pages = {9913-9923},
  doi = {10.1109/ICCV51701.2025.00924},
  publisher = {IEEE},
}

@inproceedings{couairon2025jafar,
  title = {{JAFAR: Jack up Any Feature at Any Resolution}},
  author = {Paul Couairon and Loick Chambon and Louis Serrano and Jean-Emmanuel Haugeard and Matthieu Cord and Nicolas Thome},
  booktitle = {Advances in Neural Information Processing Systems 38},
  year = {2025},
  eprint = {2506.11136},
  archiveprefix = {arXiv},
  pages = {110050-110077},
  doi = {10.52202/085713-3316},
  publisher = {Neural Information Processing Systems Foundation, Inc. (NeurIPS)},
}

@inproceedings{tian2023resformer,
  title = {{ResFormer: Scaling ViTs with Multi-Resolution Training}},
  author = {Tian, Rui and Wu, Zu-Xuan and Dai, Qiuju and Hu, Hang-Rui and Qiao, Y. and Jiang, Yu-Gang},
  booktitle = {Computer Vision and Pattern Recognition},
  year = {2023},
  eprint = {2212.00776},
  archiveprefix = {arXiv},
  pages = {22721-22731},
  doi = {10.1109/CVPR52729.2023.02176},
  publisher = {IEEE},
}

@misc{ding2023reviving,
  title = {{Reviving Shift Equivariance in Vision Transformers}},
  author = {Peijian Ding and Davit Soselia and Thomas Armstrong and Jiahao Su and Furong Huang},
  year = {2023},
  eprint = {2306.07470},
  archiveprefix = {arXiv},
  journal = {arXiv.org},
  doi = {10.48550/arXiv.2306.07470},
}

@inproceedings{rojasgomez2024making,
  title = {{Making Vision Transformers Truly Shift-Equivariant}},
  author = {Renan A. Rojas-Gomez and Teck-Yian Lim and Minh N. Do and Raymond A. Yeh},
  booktitle = {Computer Vision and Pattern Recognition},
  year = {2024},
  eprint = {2305.16316},
  archiveprefix = {arXiv},
  pages = {5568-5577},
  doi = {10.1109/CVPR52733.2024.00532},
  publisher = {IEEE},
}

@inproceedings{zhang2019making,
  title = {{Making Convolutional Networks Shift-Invariant Again}},
  author = {Richard Zhang},
  booktitle = {International Conference on Machine Learning},
  year = {2019},
  eprint = {1904.11486},
  archiveprefix = {arXiv},
  doi = {10.48550/arXiv.1904.11486},
}

@article{azulay2019why,
  title = {{Why Do Deep Convolutional Networks Generalize So Poorly to Small Image Transformations?}},
  author = {Aharon Azulay and Yair Weiss},
  journal = {Journal of Machine Learning Research},
  volume = {20},
  number = {184},
  pages = {1--25},
  year = {2019},
}

@inproceedings{tu2022maxvit,
  title = {{MaxViT}: Multi-Axis Vision Transformer},
  author = {Tu, Zhengzhong and Talebi, Hossein and Zhang, Han and Yang, Feng and Milanfar, Peyman and Bovik, Alan and Li, Yinxiao},
  booktitle = {European Conference on Computer Vision (ECCV)},
  year = {2022},
}

@article{oquab2023dinov2,
  title = {{DINOv2: Learning Robust Visual Features without Supervision}},
  author = {Oquab, M. and Darcet, Timothée and Moutakanni, Théo and Vo, Huy V. and Szafraniec, Marc and Khalidov, Vasil and Fernandez, Pierre and Haziza, Daniel and Massa, Francisco and El-Nouby, Alaaeldin and others},
  journal = {Trans. Mach. Learn. Res.},
  year = {2024},
  doi = {10.48550/arXiv.2304.07193},
}

@inproceedings{suri2024lift,
  title = {{LiFT}: A Surprisingly Simple Lightweight Feature Transform for Dense {ViT} Descriptors},
  author = {Suri, Saksham and Walmer, Matthew and Gupta, Kamal and Shrivastava, Abhinav},
  booktitle = {European Conference on Computer Vision (ECCV)},
  year = {2024},
  note = {arXiv:2403.14625},
}

@inproceedings{chambon2026naf,
  title = {{NAF}: Zero-Shot Feature Upsampling via Neighborhood Attention Filtering},
  author = {Chambon, Lo{\"i}ck and Couairon, Paul and Zablocki, Eloi and Boulch, Alexandre and Thome, Nicolas and Cord, Matthieu},
  booktitle = {Computer Vision and Pattern Recognition},
  year = {2026},
  note = {arXiv:2511.18452},
  doi = {10.48550/arXiv.2511.18452},
}

@inproceedings{ranzinger2025featsharp,
  title = {{FeatSharp}: Your Vision Model Features, Sharper},
  author = {Ranzinger, Mike and Heinrich, Greg and Molchanov, Pavlo and Kautz, Jan and Catanzaro, Bryan and Tao, Andrew},
  booktitle = {International Conference on Machine Learning (ICML)},
  year = {2025},
  note = {arXiv:2502.16025},
}

@inproceedings{seo2026upsample,
  title = {Upsample Anything: A Simple and Hard to Beat Baseline for Feature Upsampling},
  author = {Seo, Minseok and Hamilton, Mark and Kim, Changick},
  booktitle = {Computer Vision and Pattern Recognition},
  year = {2026},
  note = {arXiv:2511.16301},
  doi = {10.48550/arXiv.2511.16301},
}

@article{docherty2024upsampling,
  title={Upsampling DINOv2 features for unsupervised vision tasks and weakly supervised materials segmentation},
  author={Docherty, Ronan and Vamvakeros, Antonis and Cooper, Samuel J},
  journal={Advanced Intelligent Systems},
  volume={8},
  number={4},
  pages={e202501094},
  year={2026},
  publisher={Wiley Online Library}
}

@inproceedings{darcet2024vision,
  title = {Vision Transformers Need Registers},
  author = {Darcet, Timoth{\'e}e and Oquab, Maxime and Mairal, Julien and Bojanowski, Piotr},
  booktitle = {International Conference on Learning Representations},
  year = {2024},
  note = {arXiv:2309.16588},
  doi = {10.48550/arXiv.2309.16588},
}

@article{simeoni2025dinov3,
  title = {{DINOv3}},
  author = {Siméoni, Oriane and Vo, Huy V. and Seitzer, Maximilian and Baldassarre, Federico and Oquab, Maxime and Jose, Cijo and Khalidov, Vasil and Szafraniec, Marc and Yi, Seungeun and Ramamonjisoa, Michael and others},
  journal = {arXiv},
  year = {2025},
  note = {arXiv:2508.10104; a TMLR version is listed by mlanthology (2026)},
}

@inproceedings{touvron2019fixing,
  title = {Fixing the train-test resolution discrepancy},
  author = {Touvron, Hugo and Vedaldi, A. and Douze, Matthijs and Jégou, H.},
  booktitle = {Neural Information Processing Systems},
  year = {2019},
  note = {arXiv:1906.06423},
  doi = {10.48550/arXiv.1906.06423},
}

@inproceedings{leroy2024win,
  title = {Win-Win: Training High-Resolution Vision Transformers from Two Windows},
  author = {Leroy, Vincent and Revaud, Jerome and Lucas, Thomas and Weinzaepfel, Philippe},
  booktitle = {International Conference on Learning Representations},
  year = {2024},
  note = {arXiv:2310.00632},
  doi = {10.48550/arXiv.2310.00632},
}

@inproceedings{nguyen2025image,
  title = {An Image is Worth More Than 16x16 Patches: Exploring Transformers on Individual Pixels},
  author = {Nguyen, Duy-Kien and Assran, Mahmoud and Jain, Unnat and Oswald, Martin R. and Snoek, Cees G. M. and Chen, Xinlei},
  booktitle = {International Conference on Learning Representations (ICLR)},
  year = {2025},
  note = {arXiv:2406.09415},
}

@inproceedings{chiang2022overcoming,
  title = {Overcoming a Theoretical Limitation of Self-Attention},
  author = {Chiang, David and Cholak, Peter},
  booktitle = {Proceedings of the 60th Annual Meeting of the Association for Computational Linguistics (Volume 1: Long Papers)},
  year = {2022},
  note = {arXiv:2202.12172; Sec. 5.3 "Log-length scaled attention"},
  pages = {7654-7664},
  doi = {10.18653/v1/2022.acl-long.527},
  publisher = {Association for Computational Linguistics},
}

@article{nakanishi2025scalable,
  title = {Scalable-Softmax Is Superior for Attention},
  author = {Nakanishi, Ken M.},
  journal = {arXiv.org},
  year = {2025},
  note = {arXiv:2501.19399},
  doi = {10.48550/arXiv.2501.19399},
}

@inproceedings{anson2025scale,
  title = {Scale-invariant Attention},
  author = {Anson, Ben and Wang, Xi and Aitchison, Laurence},
  booktitle = {Advances in Neural Information Processing Systems 38},
  year = {2025},
  note = {arXiv:2505.17083},
  pages = {16619-16651},
  doi = {10.52202/085713-0500},
  publisher = {Neural Information Processing Systems Foundation, Inc. (NeurIPS)},
}

@inproceedings{peng2024yarn,
  title = {{YaRN}: Efficient Context Window Extension of Large Language Models},
  author = {Peng, Bowen and Quesnelle, Jeffrey and Fan, Honglu and Shippole, Enrico},
  booktitle = {International Conference on Learning Representations},
  year = {2024},
  note = {arXiv:2309.00071},
  doi = {10.48550/arXiv.2309.00071},
}

@article{liu2026tide,
  title = {{TIDE}: Text-Informed Dynamic Extrapolation with Step-Aware Temperature Control for Diffusion Transformers},
  author = {Liu, Yihua and Ye, Fanjiang and Lin, Bowen and Fang, Rongyu and Zhang, Chengming},
  journal = {arXiv.org},
  year = {2026},
  note = {arXiv:2603.08928; attention dilution at higher resolution, text-token anchoring},
  doi = {10.48550/arXiv.2603.08928},
}

@inproceedings{bolya2023token,
  title = {Token Merging: Your {ViT} But Faster},
  author = {Bolya, Daniel and Fu, Cheng-Yang and Dai, Xiaoliang and Zhang, Peizhao and Feichtenhofer, Christoph and Hoffman, Judy},
  booktitle = {International Conference on Learning Representations},
  year = {2023},
  note = {Oral. arXiv:2210.09461; proportional attention: log token size added to the attention logits},
  doi = {10.48550/arXiv.2210.09461},
}

@inproceedings{chaman2021truly,
  title = {Truly Shift-Invariant Convolutional Neural Networks},
  author = {Chaman, Anadi and Dokmanić, Ivan},
  booktitle = {Computer Vision and Pattern Recognition},
  year = {2021},
  note = {arXiv:2011.14214},
  pages = {3772-3782},
  doi = {10.1109/CVPR46437.2021.00377},
  publisher = {IEEE},
}

@inproceedings{shanmugam2021better,
  title = {Better Aggregation in Test-Time Augmentation},
  author = {Shanmugam, Divya and Blalock, Davis and Balakrishnan, Guha and Guttag, John},
  booktitle = {IEEE International Conference on Computer Vision},
  year = {2021},
  note = {arXiv:2011.11156},
  pages = {1194-1203},
  doi = {10.1109/ICCV48922.2021.00125},
  publisher = {IEEE},
}

@inproceedings{sermanet2014overfeat,
  title = {{OverFeat}: Integrated Recognition, Localization and Detection using Convolutional Networks},
  author = {Sermanet, Pierre and Eigen, David and Zhang, Xiang and Mathieu, Michael and Fergus, Rob and LeCun, Yann},
  booktitle = {International Conference on Learning Representations},
  year = {2014},
  note = {arXiv:1312.6229},
  doi = {10.48550/arXiv.1312.6229},
}

@inproceedings{long2015fully,
  title={Fully convolutional networks for semantic segmentation},
  author={Long, Jonathan and Shelhamer, Evan and Darrell, Trevor},
  booktitle={Proceedings of the IEEE conference on computer vision and pattern recognition},
  pages={3431--3440},
  year={2015}
}

@article{child2019generating,
  title = {Generating Long Sequences with Sparse Transformers},
  author = {Child, Rewon and Gray, Scott and Radford, Alec and Sutskever, Ilya},
  journal = {arXiv.org},
  year = {2019},
  note = {arXiv:1904.10509},
  doi = {10.48550/arXiv.1904.10509},
}

@article{huang2019interlaced,
  title = {Interlaced Sparse Self-Attention for Semantic Segmentation},
  author = {Huang, Lang and Yuan, Yuhui and Guo, Jianyuan and Zhang, Chao and Chen, Xilin and Wang, Jingdong},
  journal = {arXiv.org},
  year = {2019},
  note = {arXiv:1907.12273},
  doi = {10.48550/arXiv.1907.12273},
}

@article{hassani2022dilated,
  title={Dilated neighborhood attention transformer},
  author={Hassani, Ali and Shi, Humphrey},
  journal={arXiv preprint arXiv:2209.15001},
  year={2022}
}

@inproceedings{chefer2021generic,
  title = {Generic Attention-model Explainability for Interpreting Bi-Modal and Encoder-Decoder Transformers},
  author = {Chefer, Hila and Gur, Shir and Wolf, Lior},
  booktitle = {IEEE International Conference on Computer Vision},
  year = {2021},
  eprint = {2103.15679},
  archiveprefix = {arXiv},
  pages = {387-396},
  doi = {10.1109/ICCV48922.2021.00045},
  publisher = {IEEE},
}

@article{selvaraju2020gradcam,
  title = {Grad-{CAM}: Visual Explanations from Deep Networks via Gradient-based Localization},
  author = {Selvaraju, Ramprasaath R. and Cogswell, Michael and Das, Abhishek and Vedantam, Ramakrishna and Parikh, Devi and Batra, Dhruv},
  journal = {International Journal of Computer Vision},
  year = {2020},
  note = {arXiv:1610.02391; IJCV version of the ICCV 2017 paper (per arXiv comment)},
}

@inproceedings{muhammad2020eigencam,
  title={Eigen-cam: Class activation map using principal components},
  author={Muhammad, Mohammed Bany and Yeasin, Mohammed},
  booktitle={2020 international joint conference on neural networks (IJCNN)},
  pages={1--7},
  year={2020},
  organization={IEEE}
}

@inproceedings{petsiuk2018rise,
  title = {{RISE}: Randomized Input Sampling for Explanation of Black-box Models},
  author = {Petsiuk, Vitali and Das, Abir and Saenko, Kate},
  booktitle = {British Machine Vision Conference},
  year = {2018},
  eprint = {1806.07421},
  archiveprefix = {arXiv},
  doi = {10.48550/arXiv.1806.07421},
}

@article{min2019spair,
  title = {{SPair-71k}: A Large-scale Benchmark for Semantic Correspondence},
  author = {Min, Juhong and Lee, Jongmin and Ponce, Jean and Cho, Minsu},
  journal = {arXiv.org},
  year = {2019},
  note = {arXiv:1908.10543; extension of Hyperpixel Flow (ICCV 2019)},
  doi = {10.48550/arXiv.1908.10543},
}

@inproceedings{kombol2026spar,
  title = {{SPAR}: Single-Pass Any-Resolution {ViT} for Open-vocabulary Segmentation},
  author = {Kombol, Naomi and Martinović, Ivan and Šegvić, Siniša and Tolias, Giorgos},
  booktitle = {Computer Vision and Pattern Recognition},
  year = {2026},
  note = {arXiv:2604.02252},
  doi = {10.48550/arXiv.2604.02252},
}

@inproceedings{wrobel2026dave,
  title = {{DAVE}: Distribution-aware Attribution via {ViT} Gradient Decomposition},
  author = {Wr{\'o}bel, Adam and Gairola, Siddhartha and Tabor, Jacek and Schiele, Bernt and Zieli{\'n}ski, Bartosz and Rymarczyk, Dawid},
  booktitle = {International Conference on Machine Learning (ICML)},
  year = {2026},
  note = {arXiv:2602.06613},
}

@inproceedings{giambellutti2025talk2dino,
  title = {Talking to {DINO}: Bridging Self-Supervised Vision Backbones with Language for Open-Vocabulary Segmentation},
  author = {Barsellotti, Luca and Bianchi, Lorenzo and Messina, Nicola and Carrara, Fabio and Cornia, Marcella and Baraldi, Lorenzo and Falchi, Fabrizio and Cucchiara, Rita},
  booktitle = {IEEE International Conference on Computer Vision},
  year = {2025},
  note = {arXiv:2411.19331},
  url = {https://arxiv.org/abs/2411.19331},
  pages = {22025-22035},
  doi = {10.1109/ICCV51701.2025.02045},
  publisher = {IEEE},
}

@techreport{wah2011cub,
  author = {Wah, Catherine and Branson, Steve and Welinder, Peter and Perona, Pietro and Belongie, Serge},
  title = {The {Caltech-UCSD} Birds-200-2011 Dataset},
  institution = {California Institute of Technology},
  number = {CNS-TR-2011-001},
  year = {2011},
}

@article{menze2015brats,
  title = {The Multimodal Brain Tumor Image Segmentation Benchmark (BRATS)},
  author = {Menze, Bjoern H and Jakab, A. and Bauer, S. and Kalpathy-Cramer, Jayashree and Farahani, K. and Kirby, J. and Burren, Y. and Porz, N. and Slotboom, J. and Wiest, R. and others},
  journal = {IEEE Transactions on Medical Imaging},
  volume = {34},
  number = {10},
  pages = {1993--2024},
  year = {2015},
  doi = {10.1109/TMI.2014.2377694},
  publisher = {Institute of Electrical and Electronics Engineers (IEEE)},
}

@article{bakas2017tcga,
  title = {Advancing The Cancer Genome Atlas glioma MRI collections with expert segmentation labels and radiomic features},
  author = {Bakas, S. and Akbari, H. and Sotiras, Aristeidis and Bilello, M. and Rozycki, M. and Kirby, J. and Freymann, J. and Farahani, K. and Davatzikos, C.},
  journal = {Scientific Data},
  volume = {4},
  pages = {170117},
  year = {2017},
  doi = {10.1038/sdata.2017.117},
  number = {1},
  publisher = {Springer Science and Business Media LLC},
}

@misc{balakrishnan2023brats2d,
  author = {Balakrishnan},
  title = {Brain 2D MRI Images and Mask},
  howpublished = {Kaggle},
  year = {2023},
  url = {https://www.kaggle.com/datasets/balakrishcodes/brain-2d-mri-imgs-and-mask},
  note = {BraTS-2019 FLAIR slices and tumour masks; accessed 2026-10-23},
}

@misc{wandel2026vitupfaithfulfeatureupsampling,
  title = {{ViT-Up: Faithful Feature Upsampling for Vision Transformers}},
  author = {Krispin Wandel and Jingchuan Wang and Hesheng Wang},
  year = {2026},
  eprint = {2606.14024},
  archiveprefix = {arXiv},
  primaryclass = {cs.CV},
  url = {https://arxiv.org/abs/2606.14024},
  journal = {arXiv.org},
  doi = {10.48550/arXiv.2606.14024},
}

@article{veenboer2025tapct,
  title = {{TAP-CT}: {3D} Task-Agnostic Pretraining of Computed Tomography Foundation Models},
  author = {Veenboer, T. and Yiasemis, George and Marcus, E. and Veldhuizen, Vivien van and Snoek, Cees G. M. and Teuwen, Jonas and Lipman, K. G. Groot},
  journal = {arXiv.org},
  year = {2025},
  doi = {10.48550/arXiv.2512.00872},
}

@inproceedings{falta2023lung250m,
  title = {{Lung250M-4B}: A Combined {3D} Dataset for {CT}- and Point Cloud-Based Intra-Patient Lung Registration},
  author = {Falta, Fenja and Gro{\ss}br{\"o}hmer, Christoph and Hering, Alessa and Bigalke, Alexander and Heinrich, Mattias P.},
  booktitle = {Advances in Neural Information Processing Systems (Datasets and Benchmarks Track)},
  year = {2023},
}

@article{castillo2009dirlab,
  title = {A framework for evaluation of deformable image registration spatial accuracy using large landmark point sets},
  author = {Castillo, Richard and Castillo, Edward and Guerra, Rudy and Johnson, Valen E. and McPhail, Travis and Garg, Amit K. and Guerrero, Thomas},
  journal = {Physics in Medicine and Biology},
  volume = {54},
  number = {7},
  pages = {1849--1870},
  year = {2009},
  doi = {10.1088/0031-9155/54/7/001},
  publisher = {IOP Publishing},
}

@article{castillo2013copdgene,
  title = {A reference dataset for deformable image registration spatial accuracy evaluation using the {COPDgene} study archive},
  author = {Castillo, Richard and Castillo, Edward and Fuentes, David and Ahmad, Moiz and Wood, Abbie M. and Ludwig, Michelle S. and Guerrero, Thomas},
  journal = {Physics in Medicine and Biology},
  volume = {58},
  number = {9},
  pages = {2861--2877},
  year = {2013},
  doi = {10.1088/0031-9155/58/9/2861},
  publisher = {IOP Publishing},
}

@inproceedings{heinrich2013mind,
  title = {Towards Realtime Multimodal Fusion for Image-Guided Interventions Using Self-Similarities},
  author = {Heinrich, M. and Jenkinson, M. and Papież, B. and Brady, M. and Schnabel, J.},
  booktitle = {International Conference on Medical Image Computing and Computer-Assisted Intervention},
  year = {2013},
  pages = {187-194},
  doi = {10.1007/978-3-642-40811-3_24},
  publisher = {Springer Berlin Heidelberg},
}

@article{xu2025threedino,
  title = {A generalizable {3D} framework and model for self-supervised learning in medical imaging},
  author = {Xu, Tony and Hosseini, Sepehr and Anderson, Chris and Rinaldi, Anthony and Krishnan, Rahul G. and Martel, Anne L. and Goubran, Maged},
  journal = {npj Digital Medicine},
  year = {2025},
  note = {Verify author list and volume before submission},
  volume = {8},
  number = {1},
  doi = {10.1038/s41746-025-02035-w},
  publisher = {Springer Science and Business Media LLC},
}

@article{xu2016abdomen,
  title = {Evaluation of Six Registration Methods for the Human Abdomen on Clinically Acquired {CT}},
  author = {Xu, Zhoubing and Lee, Christopher P. and Heinrich, Mattias P. and Modat, Marc and Rueckert, Daniel and Ourselin, Sebastien and Abramson, Richard G. and Landman, Bennett A.},
  journal = {IEEE Transactions on Biomedical Engineering},
  volume = {63},
  number = {8},
  pages = {1563--1572},
  year = {2016},
  doi = {10.1109/TBME.2016.2574816},
  publisher = {Institute of Electrical and Electronics Engineers (IEEE)},
}

@article{hering2023learn2reg,
  title = {{Learn2Reg}: Comprehensive Multi-Task Medical Image Registration Challenge, Dataset and Evaluation in the Era of Deep Learning},
  author = {Hering, Alessa and Hansen, Lasse and Mok, Tony C. W. and Chung, Albert C. S. and Siebert, Hanna and H{\"a}ger, Stephanie and Lange, Annkristin and Kuckertz, Sven and Heldmann, Stefan and Shao, Wei and others},
  journal = {IEEE Transactions on Medical Imaging},
  volume = {42},
  number = {3},
  pages = {697--712},
  year = {2023},
}
